\documentclass[10pt,twocolumn,letterpaper]{article}

\usepackage{iccv}
\usepackage{times}
\usepackage{epsfig}
\usepackage{graphicx}
\usepackage{amsmath}
\usepackage{amssymb}

\usepackage[pagebackref=true,breaklinks=true,letterpaper=true,colorlinks,bookmarks=false]{hyperref}

\usepackage{wrapfig}

\usepackage[linesnumbered,ruled]{algorithm2e}

\usepackage{subfig}
\usepackage{tabularx}
\usepackage{afterpage}
\usepackage{floatflt}
\usepackage{array}
\usepackage{arydshln}
\usepackage{verbatim}
\usepackage{makecell}
\usepackage[english]{babel}
\usepackage [autostyle, english = american]{csquotes}
\MakeOuterQuote{"}

\usepackage{multirow}
\usepackage{bbm}
\usepackage{nicefrac}

\iccvfinalcopy % *** Uncomment this line for the final submission

\def\iccvPaperID{6410} % *** Enter the ICCV Paper ID here

\def\Pr{\mathrm{P}}

\def\R{\mathbb{R}}

\newcommand{\norm}[1]{\left\|#1\right\|}

\def\Pr{\mathrm{P}}

\newcommand{\Id}{\mathbbm{1}}

\def\argmax{\mathop{\rm arg\,max}\limits}%    a math operator.
\def\minop{\mathop{\rm min}\limits}
\def\maxop{\mathop{\rm max}\limits}

\def\min{\mathop{\rm min}\nolimits}
\def\max{\mathop{\rm max}\nolimits}

\newcolumntype{C}[1]{>{\centering\arraybackslash}p{#1}}
\newcolumntype{L}[1]{>{\raggedright\arraybackslash}p{#1}}
\newcolumntype{R}[1]{>{\raggedleft\arraybackslash}p{#1}}

\ificcvfinal\fi

\newif\ifapp
\apptrue
\begin{document}

%%%%%%%%% TITLE
\title{Sparse and Imperceivable Adversarial Attacks}

\author{Francesco Croce \\ University of T{\"u}bingen \and
	Matthias Hein \\ University of T{\"u}bingen
}

\maketitle

\begin{abstract}
	Neural networks have been proven to be vulnerable to a variety of adversarial attacks. From a safety perspective, highly sparse adversarial attacks are particularly dangerous. On the other hand the pixelwise perturbations of sparse attacks are typically large and thus can be potentially detected. We propose a new black-box technique to craft adversarial examples aiming at minimizing $l_0$-distance to the original image. Extensive experiments show that our attack is better or competitive to the state of the art. Moreover, we can integrate additional bounds on the componentwise perturbation. Allowing pixels to change only in region of high variation and avoiding changes along axis-aligned edges makes our adversarial examples almost non-perceivable. Moreover, we adapt the Projected Gradient Descent attack to the $l_0$-norm integrating componentwise constraints. This allows us to do adversarial training to enhance the robustness of classifiers against sparse and imperceivable adversarial manipulations.
	
\end{abstract}

\section{Introduction}

State-of-the-art neural networks are not robust \cite{BigEtAl13,SzeEtAl2014,GooShlSze2015}, in the sense that a very small adversarial change of a correctly classified input leads to a wrong decision. While \cite{SzeEtAl2014,GooShlSze2015} have brought up this problem in object recognition tasks, the problem itself has been discussed for some time in the area of email spam classification \cite{DalEtAl2004,LowMee2005}. This non-robust behavior of neural networks is a problem when such classifiers are used for decision making in safety-critical systems e.g. in  autonomous driving or medical diagnosis systems. Thus it is important to be aware of the possible vulnerabilities as they can lead to fatal failures beyond the eminent security issue \cite{LiuEtAl2016}.

\begin{figure}[ht]\centering
	\includegraphics[width=0.45\columnwidth]{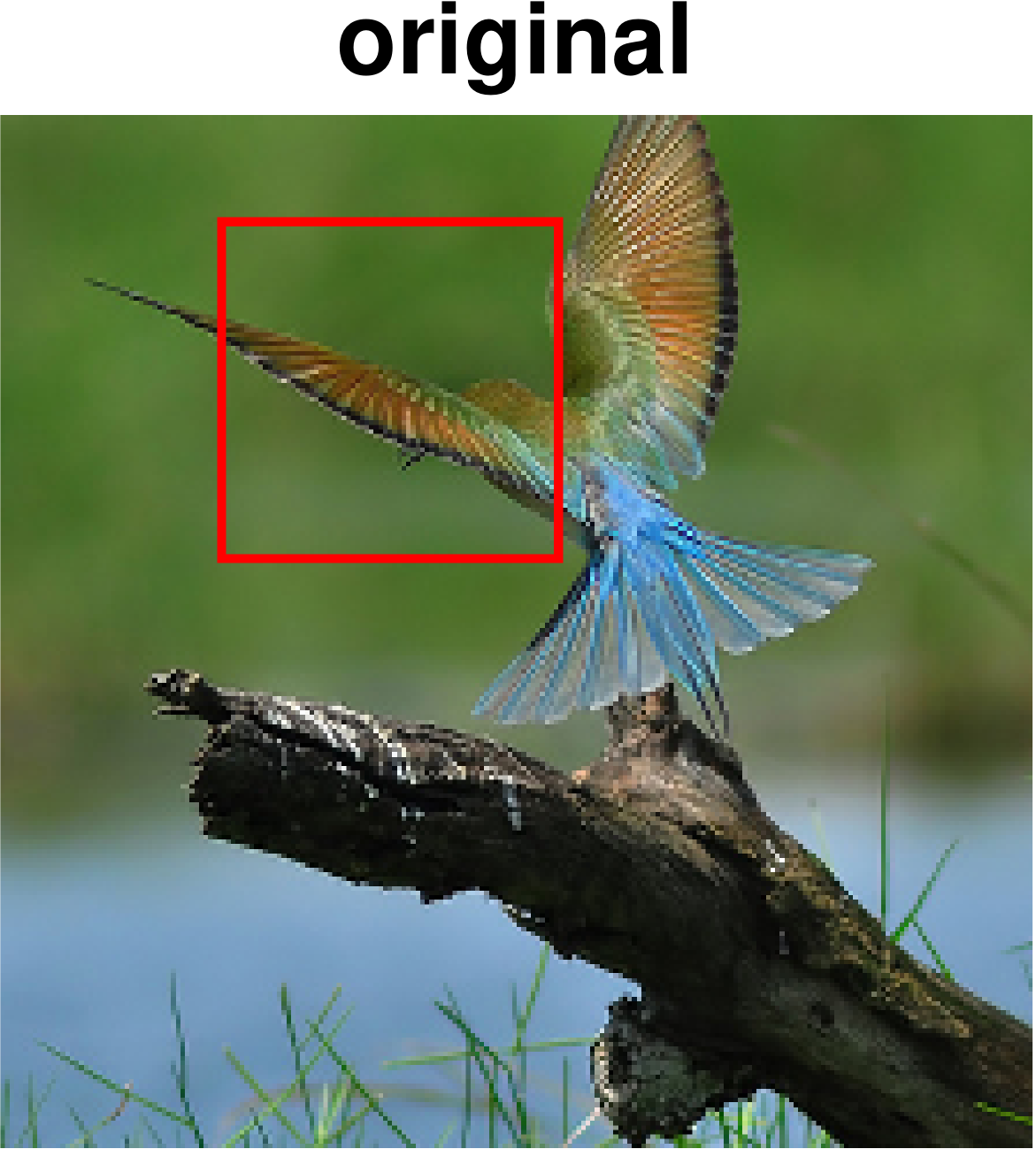}
	\includegraphics[width=0.45\columnwidth]{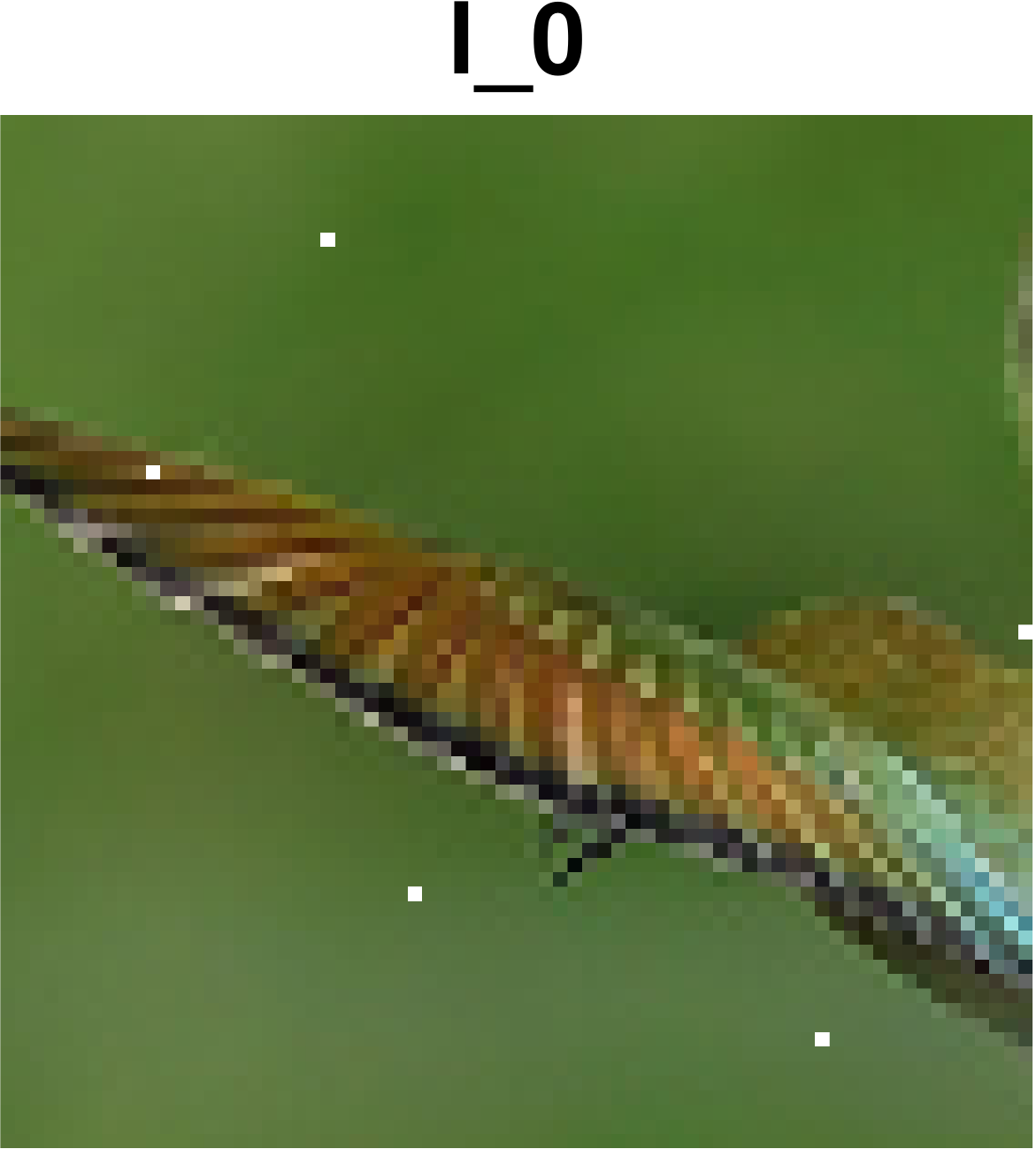}\\
	\vspace{3mm}
	\includegraphics[width=0.45\columnwidth]{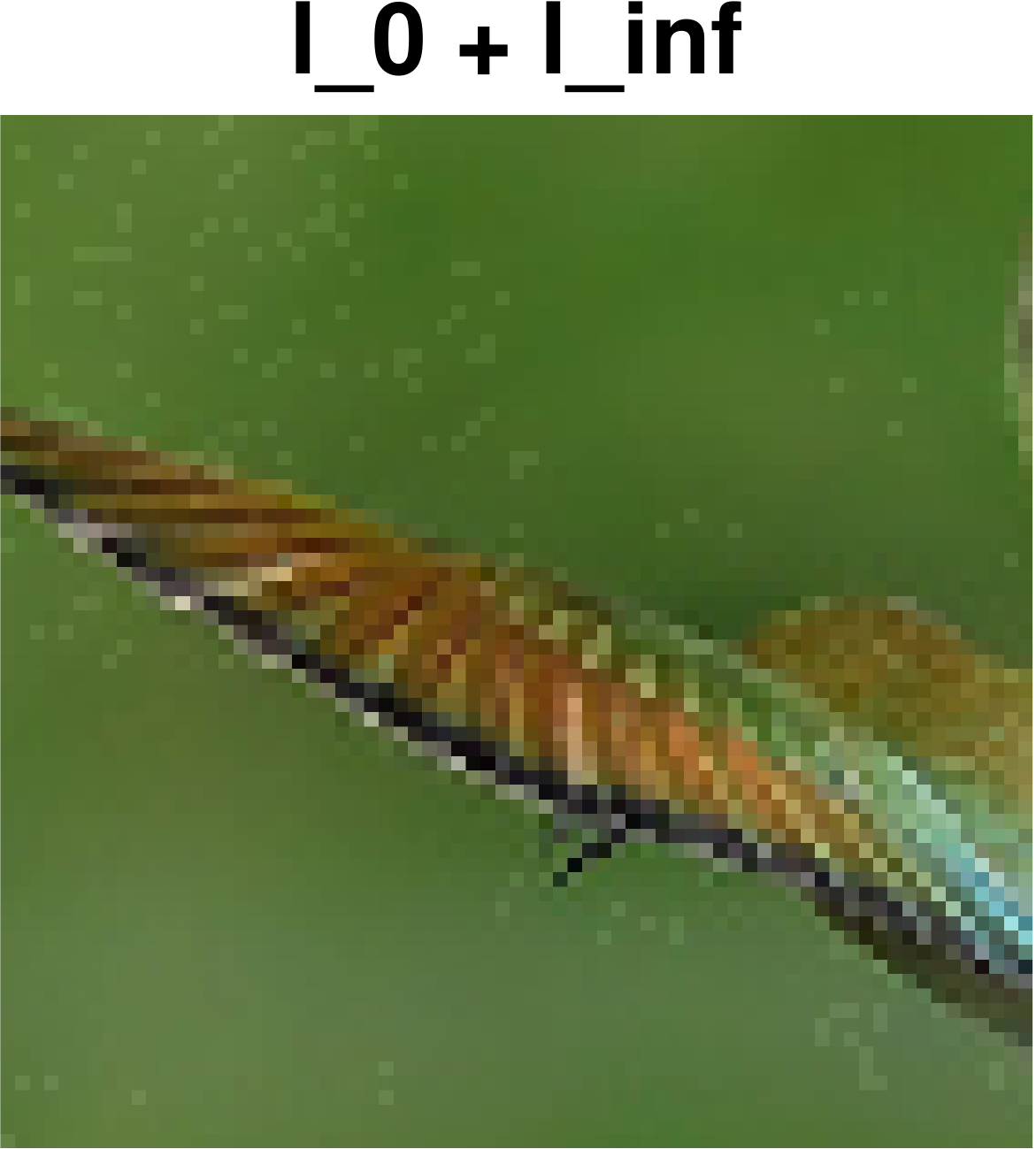}
	\includegraphics[width=0.45\columnwidth]{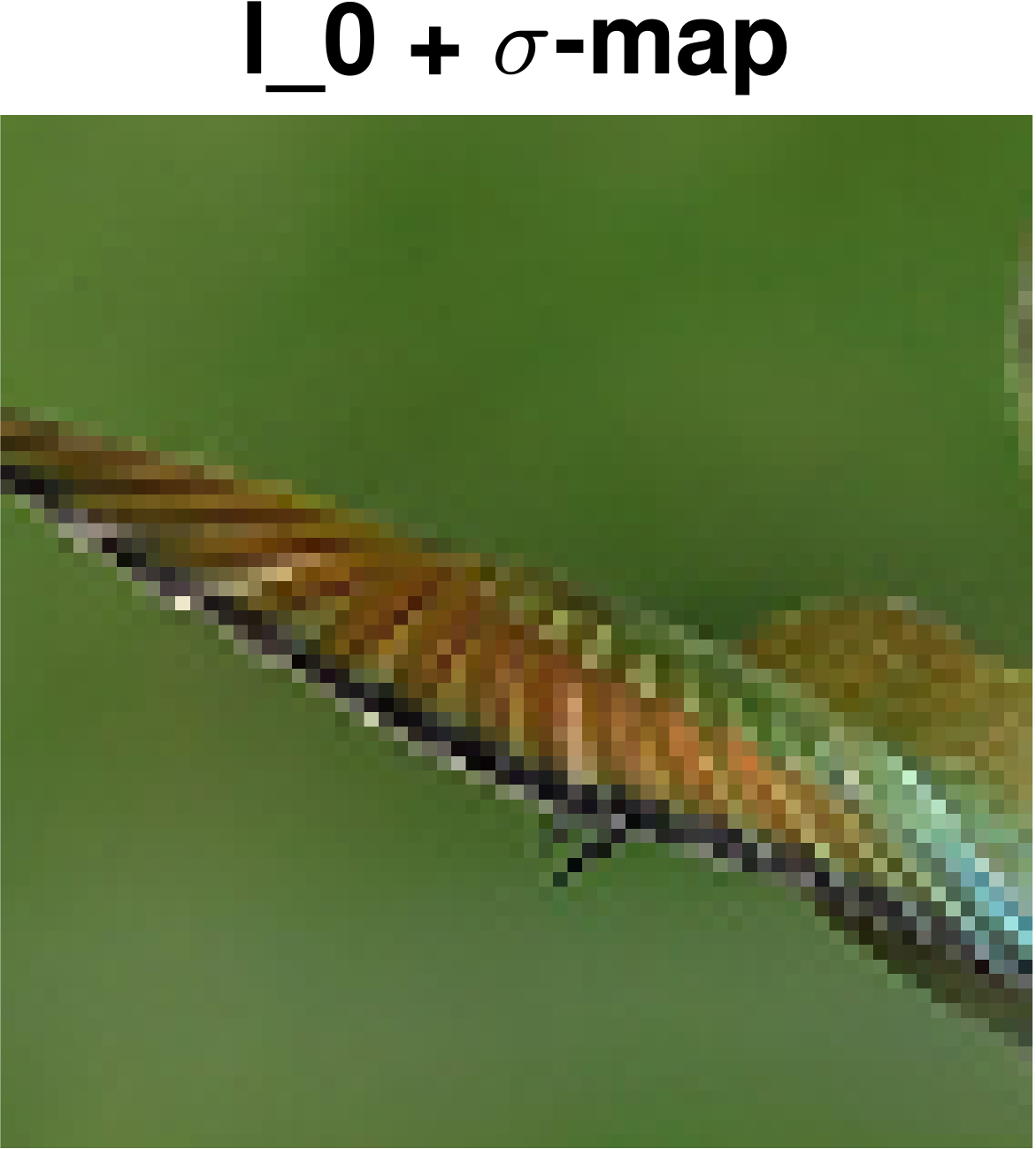}
\caption{\label{fig:teaser}
Upper left: original image with box for zoom,  Upper Right: our $l_0$-attack, very few pixels, only $0.04\%$, are changed, but the modified pixels are clearly visible,
Lower left: the result of the $l_0+l_\infty$-attack as proposed in \cite{ModEtAl19}, the modifications are sparse, $2.7\%$ of the pixels are changed, but clearly visible, Lower right: 
our sparse, $2.7\%$ of the pixels are changed, but imperceivable attack ($l_0+\sigma$-map).}
\end{figure}

Recent research on attacks can be divided into white-box attacks \cite{MooFawFro2016,CarWag2016,CarWag2017,MadEtAl2018,CroHei18}, where one has access to the model at attack time, and black box attacks \cite{CheEtAl17,BreRauBet18,IlyEtAl18,BhaEtAl18} where one can just query the output of the classifier or the confidence scores of all classes. Typically the attacks
try to find points on or close to the decision boundary, where the distance is measured in the pixels space, most often wrt the $l_\infty$- and $l_2$-norm \cite{MooFawFro2016,CarWag2016,CarWag2017,CroHei18}, or one tries to maximize the loss resp. minimize the confidence in the correct class in some $\epsilon$-ball around the original image \cite{MadEtAl2018}. Non pixelwise attacks exploiting geometric transformations have been proposed in \cite{KanMooFro18,XiaEtAl18}. While it has been argued that adversarial changes will not happen in practical scenarios, this argument has been refuted in \cite{KurGooBen2016a, EykEtAl2018}. 
Adversarial attacks during training have been early on proposed as a potential defense \cite{SzeEtAl2014,GooShlSze2015}, now known as adversarial training. In the form proposed in
\cite{MadEtAl2018} this is one of the few defenses which could not be broken easily \cite{CarWag2017,AthEtAl2018}.
%Another direction to achieve robustness are provably robust models \cite{TjeTed2017,HeiAnd2017,RagSteLia2018,WonKol2018,MirGehVec2018,WengEtAl2018,CroEtAl2018} which however up to now do not scale well to large networks.

In this paper we are dealing specifically with sparse adversarial attacks, that is we want to modify the smallest amount of pixels in order to change the decision. There are currently white-box attacks based on variants of gradient based methods integrating the $l_0$-constraint \cite{CarWag2016,ModEtAl19} or mainly black-box attacks which use either local
search or evolutionary algorithms \cite{NarKas16, SuVarKou19,SchEtAl19}. The paper has the following methodological contributions:
1) we suggest a novel black-box attack based on local search which outperforms all existing $l_0$-attacks, 2) we present closed form expressions or simple algorithms for the projections onto
the $l_0$-ball (or intersection of $l_0$-ball and componentwise constraints) in order to extend the PGD attack of \cite{MadEtAl2018} to the considered scenario, 3) since sparse attacks are often clearly visible and thus, at least in some cases, easy to detect (see upper right image of Figure \ref{fig:teaser}), we combine the sparsity constraint ($l_0$-ball) with componentwise constraints, and we extend the two $l_0$-attacks mentioned above to produce sparse and imperceivable adversarial perturbations.\\
Compared to \cite{ModEtAl19} who introduce global componentwise constraints (see lower left image of Figure \ref{fig:teaser}) we propose to use locally adaptive componentwise constraints. These local constraints ensure that the change is typically not visible, that is we neither change color too much, nor we change pixels along edges aligned with the coordinate axis (see the Appendix for a visualization) or in regions which have uniform color (see lower right image of Figure \ref{fig:teaser}). This is in line with, and significantly improves upon, \cite{LuoEtAl18}, who suggest to perturb pixels in regions of high variance to have less recognizable modifications. In fact the often employed $l_\infty$-attacks which modify each pixel only slightly but have to manipulate all pixels seem not to model perturbations which could actually occur. We think that our sparse and imperceivable attacks could happen in practice and correspond to modifications which do not change the semantics
of the images even on very small scales. The good news of our paper is that the success rate of such attacks (50-70\% success rate for standard models) is smaller than that of the commonly used ones - nevertheless we find it disturbing that such manipulations are possible at all. Thus we also test if adversarial training can reduce the success rate of such attacks. We find that adversarial training wrt $l_2$ partially decreases the effectiveness of $l_0$-attacks, while adversarial training wrt either $l_2$ or $l_\infty$ helps to be more robust against sparse and imperceivable attacks. Finally, we introduce adversarial training aiming specifically at robustness wrt both our attack models.

%%%%%%%%% BODY TEXT
\section{Sparse and imperceivable adversarial attacks}\label{sec:attack-models}
Let $f:\R^d \longrightarrow \R^K$ be a multi-class classifier, where $d$ is the input dimension and $K$ the number of classes. A test point $x\in \R^d$ is classified as $c = \argmax_{r=1,\ldots, K} f_r(x)$. The minimal adversarial perturbation $y^*$ of $x\in\R^d$ with respect to a distance function $\gamma:\R^d \longrightarrow \R_+$ is given  as the solution $y^*\in\R^d$ of the optimization problem 
\begin{equation} \begin{split} \minop_{y\in\R^d} \; &\; \gamma(y-x) \\
\textrm{s.th.} \; & \;\argmax_{r=1,\ldots, K} f_r(y) \neq \argmax_{r=1,\ldots, K} f_r(x),\\ & \; y \in C, \end{split} \label{eq:adv_opt}
\end{equation} 
where $C$ is a set of constraints valid inputs need to satisfy (e.g. images are scaled to be in $[0,1]^d$). 
Said otherwise: $y^*$ is the closest point to $x$ wrt the distance function $\gamma$ which is classified differently from $x$.\\

\subsection{Sparse $l_0$-attack}
In an $l_0$-attack one is interested in finding the smallest number of pixels which need to be changed so that the decision changes. We write in the following gray-scale images $x$
with $d$ pixels as vectors in $[0,1]^d$ and color images $x$ with $d$ pixels as matrices $x$ in $[0,1]^{d \times 3}$, and $x_i$ denotes the $i$-th pixel with the three color
channels in RGB. The corresponding distance function $\gamma$ is thus given for gray-scale images as the standard $l_0$-norm
\begin{align}\label{eq:l0metric-gray}
 \gamma(y-x)=\sum_{i=1}^d \Id_{|y_i-x_i|\neq 0},
\end{align}
and for color images as 
\begin{align}\label{eq:l0metric-color}
 \gamma(y-x)=\sum_{i=1}^d \maxop_{j=1,\ldots,3} \Id_{|y_{ij}-x_{ij}|\neq 0},
\end{align}
where the inner maximization checks if any color channel $j$ of the pixel $i$ is changed. %This can be seen as a kind of group sparsity norm.
From a practical point of view the $l_0$-attack tests basically how vulnerable the model is to failure of pixels or large localized changes on an object e.g. a
sticker on a refrigerator or dirt/dust on a windshield.

\subsection{Sparse and Imperceivable attack}
The problem of $l_0$-attacks is that they are completely unconstrained in the way how they change each pixel. Thus the perturbed pixels have usually completely different color than the surrounding ones and thus are easily visible. On the other hand $l_\infty$-attacks, using the distance function
\[ \gamma(y-x)=\maxop_{i=1,\ldots,d}\, \maxop_{j=1,\ldots,3} |y_{ij}-x_{ij}|,\]
are known to result in very small changes per pixel but have to modify every pixel and color channel. This seems to be a quite unrealistic perturbation model from a practical point of view.
A much more realistic attack model which could happen in a practical scenario is when the changes are sparse but also imperceivable. In order to achieve this we come up
with additional constraints on the allowed channelwise change. In \cite{ModEtAl19} they suggest to have global bounds, for some fixed $\delta>0$, in the form
\[ x_{ij} - \delta \leq y_{ij} \leq x_{ij}+\delta,\]
which should ensure that the changes are not visible (we call an attack with $l_0$-norm and these global componentwise bounds an $l_0 +l_\infty$-attack in the following). However, these global bounds are completely agnostic of the image and thus $\delta$ has to be really small so that the changes are not
visible even in regions of homogeneous color, e.g. sky, where almost any variation is easily spotted. We suggest image-specific local bounds taking into account the image structure. We have two specific goals:
\begin{itemize}
\item[1)] We do not want to make changes along edges which are aligned with the coordinate axis as they can be easily spotted and detected.
\item[2)] We do not want to change the color too much and rather just adjust its intensity and keep approximately also its saturation level.
\end{itemize}
In order to achieve this we compute the standard deviation of each color channel in $x$- and $y$-axis directions with the two immediate neighboring pixels and the original pixel. We denote
the corresponding values as $\sigma^{(x)}_{ij}$ and $\sigma^{(y)}_{ij}$ and define $\sigma_{ij}=\sqrt{\min\{\sigma^{(x)}_{ij},\sigma^{(y)}_{ij}\}}$.
%We apply the square root in the definition of $\sigma$ to decrease the relative difference between large and small values (notice that $\sigma_{ij}\in[0,1]$) and have a wider space to search for adversarial examples.
Since $\sigma^{(x)}_{ij},\sigma^{(y)}_{ij}\in [0,1]$ the square root increases more significantly, in relative value, smaller $\min\{\sigma^{(x)}_{ij},\sigma^{(y)}_{ij}\}$. In this way we both enlarge the space of the possible adversarial examples and prevent perturbations in areas of zero variance.
In fact we allow the changed image $y$ just to have values given by
\begin{equation} y_{ij} = (1+\lambda_i \sigma_{ij})x_{ij}, \; \textrm{ with } -\kappa \leq \lambda_i \leq \kappa,\label{eq:sigma_constr}\end{equation}
where $\kappa>0$.
Additionally, we enforce box constraints $y \in [0,1]^{d \times 3}$.
%where $\lambda_i$ is the parameter which we constrain to satisfy, $-\kappa \leq \lambda_i \leq \kappa$ (in the experiments we always use $\kappa=0.2$). 
Note that the parameter $\lambda_i$ corresponds to a change in intensity of pixel $i$ by maximally plus/minus  $\kappa \sum_{j=1}^3 \sigma_{ij}x_{ij}$ as
\[ \sum_{j=1}^3 y_{ij} = \sum_{j=1}^3 x_{ij} + \lambda_i \sum_{j=1}^3 \sigma_{ij}x_{ij}.\]
Thus we are just changing intensity of the pixel instead of the actual color. Moreover, note that this change also preserves the saturation of the color value\footnote{In the HSV color space the saturation of 
a color is defined as $1-\frac{\min\{R,G,B\}}{\max\{R,G,B\}}$, where $R,G,B$ are the red/green/blue color channels in RGB color space.} if the $\sigma_{ij}$ are equal for $j=1,\ldots,3$. Thus we fulfill the second requirement from above. Moreover, the first requirement is satisfied as $\sigma_{ij}=\sqrt{\min\{\sigma^{(x)}_{ij},\sigma^{(y)}_{ij}\}}$,
meaning that if along one of the coordinates there is no change in all color channels then the pixel cannot be modified at all. Thus pixels along a coordinate-aligned edge
showing no change in color will not be changed. %Interestingly, even though this model seems to be rather complicated we can later one give a simple algorithm for the projection
%onto this set which allows to generalize PGD attacks \cite{MadEtAl2018} to this attack model. 
The attack model of sparse and imperceivable attacks will be abbreviated as
$l_0+\sigma$-map.
For gray-scale images $x \in [0,1]^d$ we use instead
%\[ y_{ij} = x_{ij} +\lambda_i \sigma_{ij},\]
\begin{equation} y_{i} = x_{i} + \lambda_i \sigma_{i}, \; \textrm{ with } -\kappa \leq \lambda_i \leq \kappa.\label{eq:sigma_constr_gray}\end{equation}
as there the approximate preservation of color saturation is not needed.

\section{Algorithms for sparse (and imperceivable) attacks}\label{sec:attacks}
In this paper we propose two methods to generate $l_0$-, $l_0+l_\infty$ and $l_0+\sigma$-attacks. The first one is a randomized black-box attack based on the logits (the output of the neural network before the softmax layer) of the classifier. The second is a generalization of projected gradient descent (PGD) on the loss of the correct label \cite{MadEtAl2018} to our different attack models. For each attack model we will derive algorithms for the projection onto the  corresponding sets.

\subsection{Score-based sparse (and imperceivable) attack}\label{sac:our_attack}
Most of the existing black-box $l_0$-attacks either start with perturbing a small set of pixels and then enlarge this set until they find an adversarial example \cite{PapEtAl15,NarKas16} or, given a successful adversarial manipulation, try to progressively reduce the number of pixels exploited to change the classification \cite{CarWag2016,SchEtAl19}. Instead we introduce a flexible attack scheme where at the beginning one checks pixelwise targeted attacks and then sorts them according to the resulting gap in the classifier outputs. Then we introduce a probability
distribution on the sorted list and sample one-pixel changes to generate attacks where more pixels are manipulated simultaneously. The distribution we use is biased towards the one-pixel perturbations which
produce, when applied individually, already large changes in the classifier output. In this non-iterative scheme there is thus no danger to get stuck in suboptimal points. Moreover, while the attack has to test many points, its non-iterative nature allows to check the perturbed points %for a change in decision
in large batches which is thus much faster than an evolutionary attack. Even if the scheme is simple it outperforms all existing methods including white-box attacks.

\paragraph{One-pixel modifications}

In the first step we check all one pixel modifications of the original image $x \in [0,1]^{d \times 3}$ (color) or $x \in [0,1]^d$ (gray-scale). The tested modifications depend on the attack model.
\begin{enumerate}
\item \textbf{$l_0$-attack:} for each pixel $i$ we generate $8=2^3$ images changing the original color value to one of the $8$ corners of the RGB color cube. Thus we name our method \textbf{CornerSearch}. This results in a set of 
      $8d$ images, all one pixel modifications of the original image $x$, which we denote by  $(z^{(j)})_{j=1}^{8d}$. For gray-scale images one just checks the extreme gray-scale values 
      (black and white) and gets $(z^{(j)})_{j=1}^{2d}$.
\item \textbf{$l_0+l_\infty$-attack:} for each pixel $i$ we generate $8$ images changing the original color value of $(x_{ij})_{j=1}^3$ by the corners of the cube
             $[-\epsilon,\epsilon]^3$ resulting again in $(z^{(j)})_{j=1}^{8d}$ images. For gray-scale we use $x_{i}\pm \epsilon$ resulting in total in $(z^{(j)})_{j=1}^{2d}$ images. If necessary we clip to satisfy the constraint $z^{(j)}\in[0,1]^{d\times 3}$ or $z^{(j)}\in[0,1]^{d}$.
\item \textbf{$l_0+\sigma$-map attack:} for color images we generate for each pixel $i$ two  images by setting
      \[  y_{ij} = (1\pm \kappa \sigma_{ij})x_{ij}, \quad j=1,\ldots,3,\]
      where $\kappa$ and $\sigma_{ij}$ are as defined in Section \ref{sec:attack-models}. For gray scale images $x \in [0,1]^d$ we use 
      \[ y_{i} = x_{i} \pm \kappa \sigma_{i}.\]
      Finally, we clip $y_{ij}$ and $y_i$ to $[0,1]$. Thus, this results in $(z^{(j)})_{j=1}^{2d}$ images. We call it \textbf{$\sigma$-CornerSearch}.
\end{enumerate}
After the generation of all the images we get the classifier output $f(z^{(j)})_{j=1}^{M}$ for each of them, where $M$ is the total number of generated images, either $M=2d$ or $M=8d$. Then, separately for each class $r\neq c$, where $c=\argmax_{r=1,\ldots,K} f_r(x)$, we sort the values of
\[ f_r(z^{(j)})- f_c(z^{(j)})\]
in decreasing order $\pi^{(r)}$. That means for all $1\leq s \leq M-1$
\[ f_r(z^{(\pi^{(r)}_s)})- f_c(z^{(\pi^{(r)}_s)}) \geq  f_r(z^{(\pi^{(r)}_{s+1})})- f_c(z^{(\pi^{(r)}_{s+1})}).\]
%If $f_r(z^{(\pi^{(r)}_1)})-f_c(z^{(\pi^{(r)}_1)})> 0$, then the decision has changed by only modifying one pixel. In this case the algorithm stops immediately
We introduce also an order $\pi^{(c)}$, sorting in decreasing order the quantities
\[ \max_{r \neq c} f_r(z^{(j)})- f_c(z^{(j)}).\]
The idea behind generating these one-pixel perturbations is to identify the pixels which push most the decision towards a particular class $r$ or in case of the set $\pi^{(c)}$ towards
an unspecific change.
If $f_r(z^{(\pi^{(r)}_1)})-f_c(z^{(\pi^{(r)}_1)})> 0$ for some $r$, then the decision has changed by only modifying one pixel. In this case the algorithm stops immediately. Otherwise, one could try to iteratively select the most effective change and repeat the one-pixel perturbations. However, this is overly expensive
and again suffers if suboptimal pixel modifications are chosen in the initial steps of the iterative scheme. Thus we suggest in the next paragraph a sampling scheme based on the obtained orderings, where one randomly
selects $k$ one-pixel modifications to combine in order to produce a multi-pixels attack.

\paragraph{Multi-pixels modifications} \label{par:multi-pixels} 
Most of the times the modifications of one pixel are not sufficient to change the decision.
%Let $N$ be the number of pixels which we assume the most influential and
Suppose we want to generate
a candidate for a targeted adversarial sample towards class $r$ by changing at most $k$ pixels, choosing among the first $N$ one-pixel perturbations according to the ordering $\pi^{(r)}$. We do this by sampling $k$ indices $(s_1,\ldots,s_k)$ in $\{1,\ldots,N\}$ from the probability distribution on $\{1,\ldots,N\}$ defined as
\begin{equation}P(Z =i) = \frac{2N-2i + 1}{N^2}, \quad i=1,\ldots, N. \label{eq:sampling} \end{equation} 
The candidate image $y^{(r)}$ is generated by applying all the $k$ one-pixel changes defined in the images $z^{(\pi^{(r)}_{s_1})},\ldots,z^{(\pi^{(r)}_{s_k})}$ to the original image $x$.
Please note that we only sample from the top $N$ one-pixel changes found in the previous paragraph and that the distribution on $\{1,\ldots,N\}$ is biased towards sampling
on the top of the list e.g. $P(Z=1)=\frac{2N-1}{N^2}$ is $2N-1$ larger than $\Pr(Z=N)=\frac{1}{N^2}$. This bias ensures that we are mainly accumulating one-pixel changes which have led
individually already to a larger change of the decision towards the target class $r$. We produce candidate images $y^{(1)},\ldots,y^{(K)}$ for all $K$ classes, having $K-1$ candidate images targeted towards changes in a particular class and one image where the attack is untargeted (for $r=c$). In total we repeat this process $N_{iter}$ times. The big advantage of the sampling
scheme compared to an iterative scheme is that all these images can be fed into the classifier in batches in parallel which compared to a sequential processing is significantly faster.
Moreover, it does not depend on previous steps and thus cannot get stuck in some suboptimal regions.
As shown in the experiments this relatively simple sampling scheme performs better than sophisticated evolutionary algorithms (black-box attacks) and even white-box attacks.

Since we want to find adversarial examples differing from $x$ in as few pixels as possible, we generate the batches $y^{(1)},\ldots,y^{(K)}$ of candidate images as described above, gradually increasing $k$, up to a threshold $k_\textrm{max}$, until we get a classification different from the original class $c$. Algorithm \ref{alg:algorithm-label} summarizes the main steps.% with $N_\textrm{iter}$ is the number of iterations we perform for each values of $\epsilon_0$.

\begin{algorithm}[t]
	\SetAlgoNoLine
	\caption{CornerSearch}
	\label{alg:algorithm-label}
	\SetKwInOut{Input}{Input}
	\SetKwInOut{Output}{Output}
	\Input{$x$ original image classified as class $c$, $K$ number of classes, $N, k_\textrm{max}, N_\textrm{iter}$   }
	\Output{$y$ adversarial example}
	$y\gets \emptyset$\\
	create \textit{one-pixels modifications} $(z^{(i)})_{i=1}^M$ \\
	\If{exists $u \in(z^{(i)})_{i=1}^M$ classified not as $c$}{$y\gets u$, \quad \Return}
	compute orderings $\pi^{(1)},\ldots,\pi^{(K)}$,\\ %sets $I^{(i)}$ for $i=1,\ldots,K$\\
	$k\gets 2$\\
	\While{$k\leq k_\textrm{max}$}{
	%\For{$j=1,\ldots,N_\textrm{iter}$}{
		%create $y^{(1)}, \ldots, y^{(K)}$ as explained in "Multi-pixels modifications" paragraph}
		%\If{$\exists u \in (y^{(i)})_{i=1}^K$ classified not as $c$ }{$y\gets u$, \quad \Return}
	\For{$r=1,\ldots,K$}{
		create the set $Y^{(r)}$ of $N_\textrm{iter}$ "$k$-pixels modifications" towards class $r$ (see paragraph above)\\
	\If{$\exists u \in Y^{{(r)}}$ classified not as $c$}{$y\gets u$, \quad \Return}}
	$k \gets k +1$\\
	%Return: no adversarial sample found with less than $k_{\max}$ pixels changed
	}
\end{algorithm}

\section{PGD for sparse and imperceivable attacks}\label{sec:pgd_l0}
The projected gradient descent (PGD) attack of Madry et al \cite{MadEtAl2018} is not aiming at finding the smallest adversarial perturbation but instead argues from the viewpoint of robust optimization about
maximizing the loss
\[ \maxop_{z \in C(x)} L(c,f(z)),\]
where $L:\{1,\ldots,K\} \times \R^K \rightarrow \R_+$ is usually chosen to be the cross-entropy loss, $c$ is the correct label of the point $x$ and the set $C(x) \subset [0,1]^{d\times 3}$ (color images with $d$ pixels) or $C(x)\subset [0,1]^d$ (gray-scale images). The interpretation in terms of robust optimization \cite{MadEtAl2018} has led to a now well-accepted way of adversarial training with the goal of getting robust wrt a fixed set of perturbations. The usage of PGD attacks during training is the de facto standard for adversarial training, which we will also use later on in Section \ref{sec:experiments}. Commonly used as the set of allowed perturbations is the $l_\infty$-ball: $C(x)=\{z\,|\,\norm{z-x}_\infty \leq \epsilon, \, z \in [0,1]^d \}$ as the projection can be done analytically. 

In order to extend PGD to $l_0$- and $l_0+l_\infty$-attacks, we first have to capture the sets allowed in our attack models in Section \ref{sec:attack-models} and then find fast algorithms for the projections onto these sets. Once this is available PGD is ready to be used as an attack and for adversarial training.
%We leave the projection onto the set of sparse and imperceivable attacks as an open problem.
In the Appendix we also show how to project onto the intersection of the $l_0$-ball and the componentwise constraints given by the $\sigma$-map, for both color and gray-scale images. Thanks to this, we can introduce an $l_0+\sigma$-map version of PGD, called $\sigma$-PGD, able to produce the sparse and imperceivable perturbations we have introduced.

\subsection{Projection onto the $l_0$-ball and $l_0$+$l_\infty$-ball}\label{sec:proj_1}
Given an original color image $x \in [0,1]^{d \times 3}$ we want to project a given point $y \in \R^{d \times 3}$ onto the set
\begin{align*} C(x)=\big\{ z \in \R^{d\times 3}\,\big|\, & \sum_{i=1}^d \maxop_{j=1,2,3} \Id_{|z_{ij}-x_{ij}|>0} \leq k,\\
& l_{ij} \leq z_{ij}\leq u_{ij}\big\}.
\end{align*}
We can write the projection problem onto $C(x)$ as
\begin{align*}
\minop_{z \in \R^{d\times 3}}\quad &  \sum_{i=1}^d \sum_{j=1}^3 (y_{ij} - z_{ij})^2\\
\textrm{s. th.} \quad & l_{ij} \leq z_{ij}\leq u_{ij}, \quad i=1,\ldots,d, \; j=1,\ldots,3 \\
                \quad & \sum_{i=1}^d \maxop_{j=1,2,3}\Id_{|z_{ij}-x_{ij}|>0} >0\leq k
\end{align*}
Ignoring the combinatorial constraint, we first solve for each pixel $i$ the problem
\begin{align*}
\minop_{z_i \in \R^{3}}\quad &  \sum_{i=1}^d \sum_{j=1}^3 (y_{ij} - z_{ij})^2\\
\textrm{s. th.} \quad & l_{ij} \leq z_{ij}\leq u_{ij}, \quad i=1,\ldots,d, \; j=1,\ldots,3 
\end{align*}
The solution is given by $z_{ij}^*=\max\{l_{ij},\min\{y_{ij},u_{ij}\}\}$. We note that each pixel
can be optimized independently from the other pixels. Thus we sort in decreasing order $\pi$ the gains
\[ \phi_i:=\sum_{j=1}^3 (y_{ij}-x_{ij})^2 - \sum_{j=1}^3 (y_{ij}-z_{ij}^*)^2.\]
achieved by each pixel $i$.
%modify the $k$ pixels which have the largest gain.
Thus the final solution differs from $x$ in the $k$ pixels (or less if there are less than $k$ pixels with positive $\phi_i$) which have the largest gain and is given by
\[ z_{\pi_i j} =\begin{cases} z^*_{\pi_i j} &\textrm{ for } i=1,\ldots,k, \; j=1,\ldots,3,\\ x_{\pi_i j} &\textrm{ else. }\end{cases}.\]

Using $l_{ij}=0$ and $u_{ij}=1$ we recover the projection onto the intersection of $l_0$-ball and $[0,1]^{d\times3}$.
%\[ C(x)=\{ z \in \R^{d\times 3}\,|\, \sum_{i=1}^d \maxop_{j=1,2,3} \Id_{|z_{ij}- x_{ij}|>0} \leq k,\; 0 \leq z_{ij}\leq 1\}.\]
For $l_0+l_\infty$ note that the two constraints
\[ 0\leq z_{ij} \leq 1, \quad -\epsilon \leq z_{ij}-x_{ij} \leq \epsilon,\]
are equivalent to:
\[ \max\{0,-\epsilon+x_{ij}\} \leq z_{ij} \leq \min\{1,x_{ij}+\epsilon\}.\]
Thus by using 
\[ l_{ij}=\max\{0,-\epsilon+x_{ij}\}, \quad u_{ij}=\min\{1,x_{ij}+\epsilon\},\]
the set $C(x)$ is equal to the intersection of the $l_0$-ball of radius $k$, the $l_\infty$-ball of radius $\epsilon$ around $x$ and $[0,1]^{d \times 3}$,

\section{Experiments}\label{sec:experiments}
In the experimental section, we evaluate the effectiveness of our score-based $l_0$-attack CornerSearch and our white-box attack $\textrm{PGD}_0$. Moreover,
we give illustrative examples of our sparse and imperceivable $l_0+\sigma$-map attacks $\sigma$-CornerSearch and $\sigma$-PGD (the latter in the Appendix). Finally, we test
adversarial training wrt various norms as a defense against our $l_0$- and $l_0+\sigma$-map-attacks. The code is available at \url{https://github.com/fra31/sparse-imperceivable-attacks}.

\begin{table*}[t]
	\centering 
	\begin{tabular}{L{15mm} L{22mm} *{6}{C{17mm}} }
		& & LocSearchAdv & PA 10x & CW & SparseFool & JSMA & CornerSearch\\[4pt]
		\hline 
		& \textit{black-box} & Yes & Yes & No & No & No & Yes\\
		
		\hline
		\multirow{3}{*}{MNIST} & \textit{success rate} & 91.39\% & 92.35\% & 87.9\% & 100\% & 99.6\% & 97.38\% \\
		& \textit{mean (pixels)} & 17.56 & 8.82 & 46.04 & 19.44 &83.92 & 9.21 \\
		& \textit{median (pixels)} & - & 8 & 44 & 12 & 46 & 7 \\
		\hline
		\multirow{3}{*}{CIFAR-10} & \textit{success rate} & 97.32\% & 100\% & 100\% & 100\% & 100\% & 99.56\% \\
		& \textit{mean (pixels)} & 38.4 & 4.63 &16.55 & 16.10& 54.5 & 2.75 \\
		& \textit{median (pixels)} & - & 3 & 11 & 12 &47 & 2 \\
		\hline
		
	\end{tabular}
	\caption{\textbf{Comparison of different $l_0$-attacks.} While SparseFool is always successful it requires significantly more pixels
		to be changed. Our method CornerSearch requires out of all attacks the least \textit{median} amount of pixels to be changed.}
	\label{tab:comp_attacks}
\end{table*}

\begin{table}[h]
	\centering 
	\begin{tabular}{L{22mm} C{20mm} C{24mm}}
		%\multicolumn{4}{c}{\textbf{sparse attacks on ImageNet}}\\[4pt]
		& SparseFool & CornerSearch\\[4pt]
		\hline 
		\textit{black-box} & No & Yes\\
		
		\hline
		\textit{success rate}  & 100\% & 93.26\% \\
		\textit{mean (pixels)} &  143.2 & 106.7 \\
		\textit{median (pixels)} &  101 & 50 \\
		\hline
		
	\end{tabular}
	\caption{\textbf{$l_0$-attacks on Restricted ImageNet.} We attack the 89 correctly classified points out of 100 points from the validation set with SparseFool \cite{ModEtAl19}
		and our algorithm CornerSearch. Due to the limit on the allowed number of pixel changes, CornerSearch is not always successful, but requires many less pixels to be changed.
		%Considering the points for which the attack succeeds in producing an adversarial manipulation, we report the \textit{mean} and \textit{median} number of pixels modified.
	}
	\label{tab:comp_attacks_IN}
\end{table}
\begin{figure}[h]
	\centering
	\includegraphics[scale=0.28]{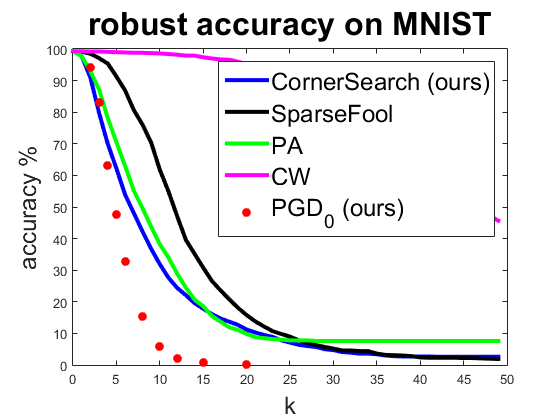}\\
	\includegraphics[scale=0.28]{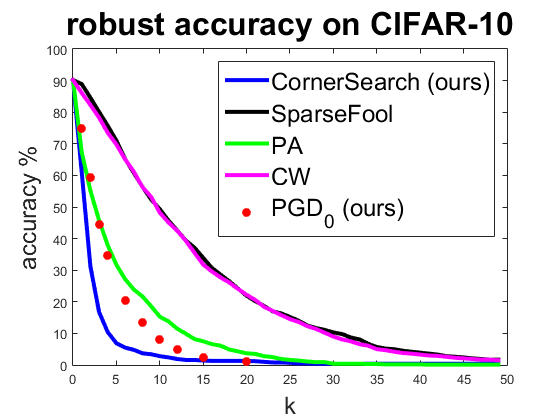}
	\caption{\textbf{Evaluation of $\textrm{PGD}_0$.} We compute for 1000 test points the robust accuracy of the classifier when the attack is allowed to perturb at most $k$ pixels. We can see that $\textrm{PGD}_0$ (red dots) outperforms SparseFool, thus being the best ``cheap'' attack, and it is even the best one on MNIST for $k\geq 4$.}
	\label{fig:pgd}
\end{figure}

\subsection{Evaluation of $l_0$-attacks}\label{sec:experiments_l0_attacks}

We compare CornerSearch with state-of-the-art attacks for sparse adversarial perturbations: LocSearchAdv \cite{NarKas16}, Pointwise Attack (PA) \cite{SchEtAl19}, Carlini-Wagner $l_0$-attack (CW) \cite{CarWag2016}, SparseFool (SF) \cite{ModEtAl19}, JSMA \cite{PapEtAl15}. The first two operate in a black-box scenario, exploiting only the classifier output, like our method, while the latter three require access to the network itself (white-box attacks). Note that SparseFool is actually an $l_1$-attack, that means it uses the $l_1$-norm as distance
measure in \eqref{eq:adv_opt} in order to avoid the combinatorial problem arising from the usage of the $l_0$-norm. However, SparseFool can produce sparse attacks and in \cite{ModEtAl19} has been shown to outperform $l_0$-attacks in terms of sparsity. We use the implementation of the Pointwise Attack in \cite{foolbox} with 10 restarts as done in \cite{SchEtAl19}, CW and JSMA from \cite{Cleverhans2017}, while we reimplemented SparseFool. Since neither the code nor the models used in \cite{NarKas16} are available (the results for LocSearchAdv are taken from \cite{NarKas16}), we decided to compare the performance of the different attacks on one of the architectures reported in \cite{NarKas16}, the Network in Network \cite{LinEtAl14} with batch normalization, retrained on MNIST and CIFAR-10.\\
We run the attacks on the first 1000 points of the corresponding test sets. We use CornerSearch with $k_\textrm{max}=50$, $N=100$ and $N_\textrm{iter}=1000$. In Table \ref{tab:comp_attacks} we report the \textit{success rate} of each method, that is the fraction of correctly classified points which can be successfully attacked, \textit{mean} and \textit{median} number of pixels that every attack needs to modify to change the decision. Please recall that MNIST consists of images with 784 pixels and CIFAR-10 with 1024 pixels. Although CornerSearch does not find an adversarial example for each test point, since we fix the maximum number of pixels that can be modified, both the average and median number of changed pixels are lower than those of the other methods, that is less pixels need to be perturbed by our method to change the decision (with the only exception of the mean on MNIST, where anyway CornerSearch has higher success rate and lower median than PA). On MNIST CornerSearch requires for at least $50\%$ of all
test images $0.89\%$ of the pixels to be changed and for CIFAR-10 it is even just $0.2\%$.\\

\begin{figure*}
	\centering
	\begin{tabular}{c c  c c| c c c c}
		\includegraphics[width=0.2\columnwidth]{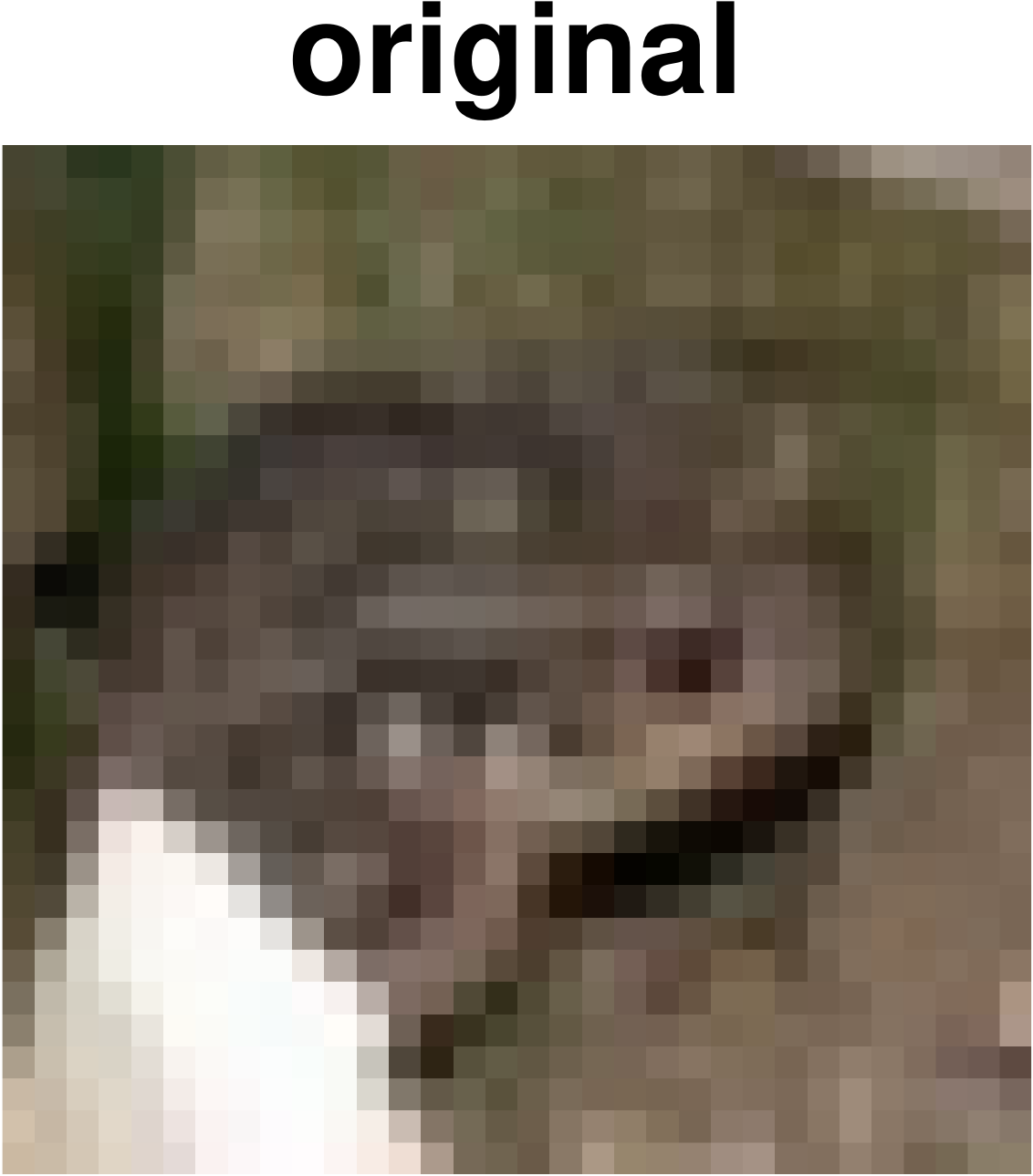}&
		\includegraphics[width=0.2\columnwidth]{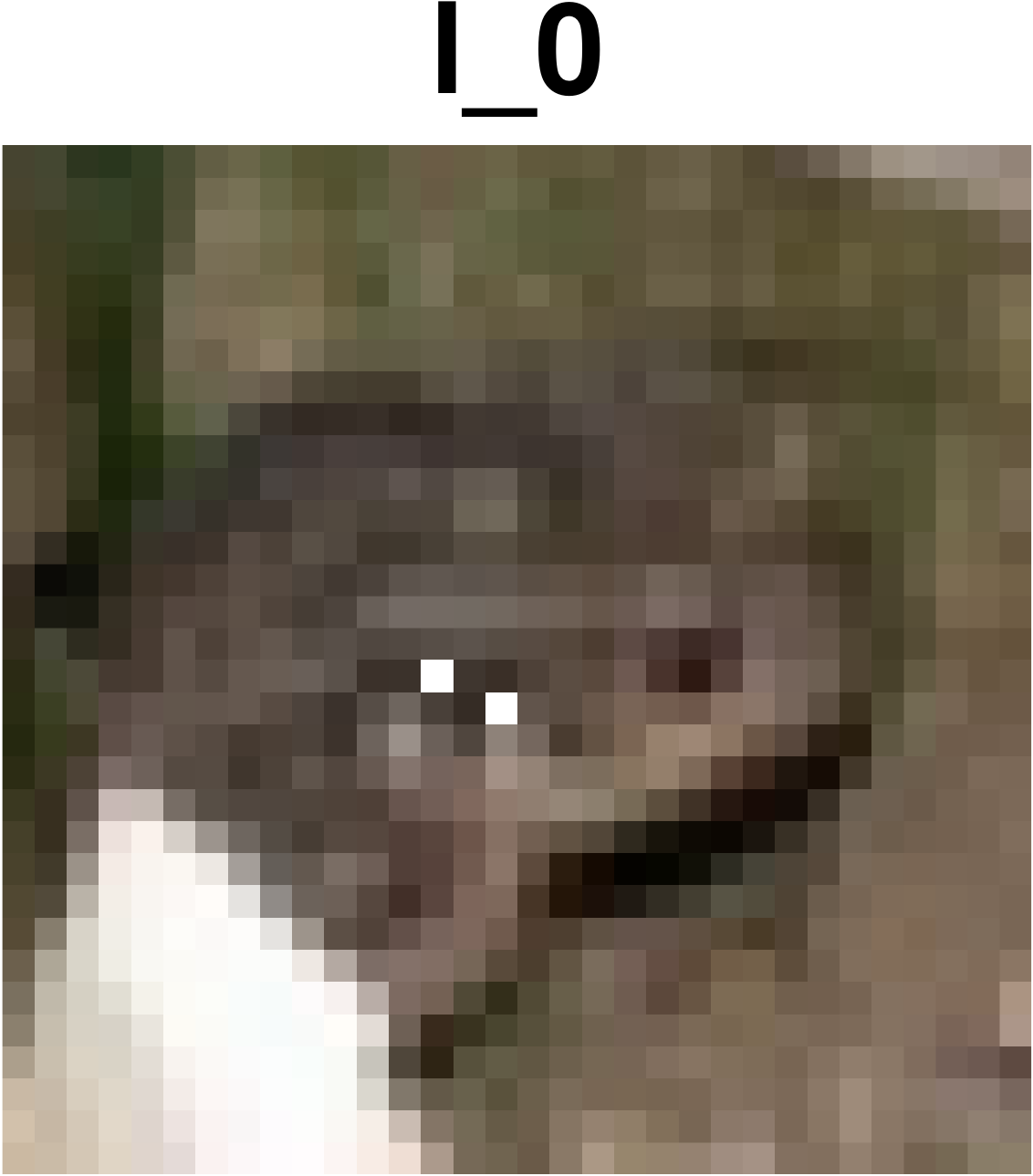}&
		\includegraphics[width=0.2\columnwidth]{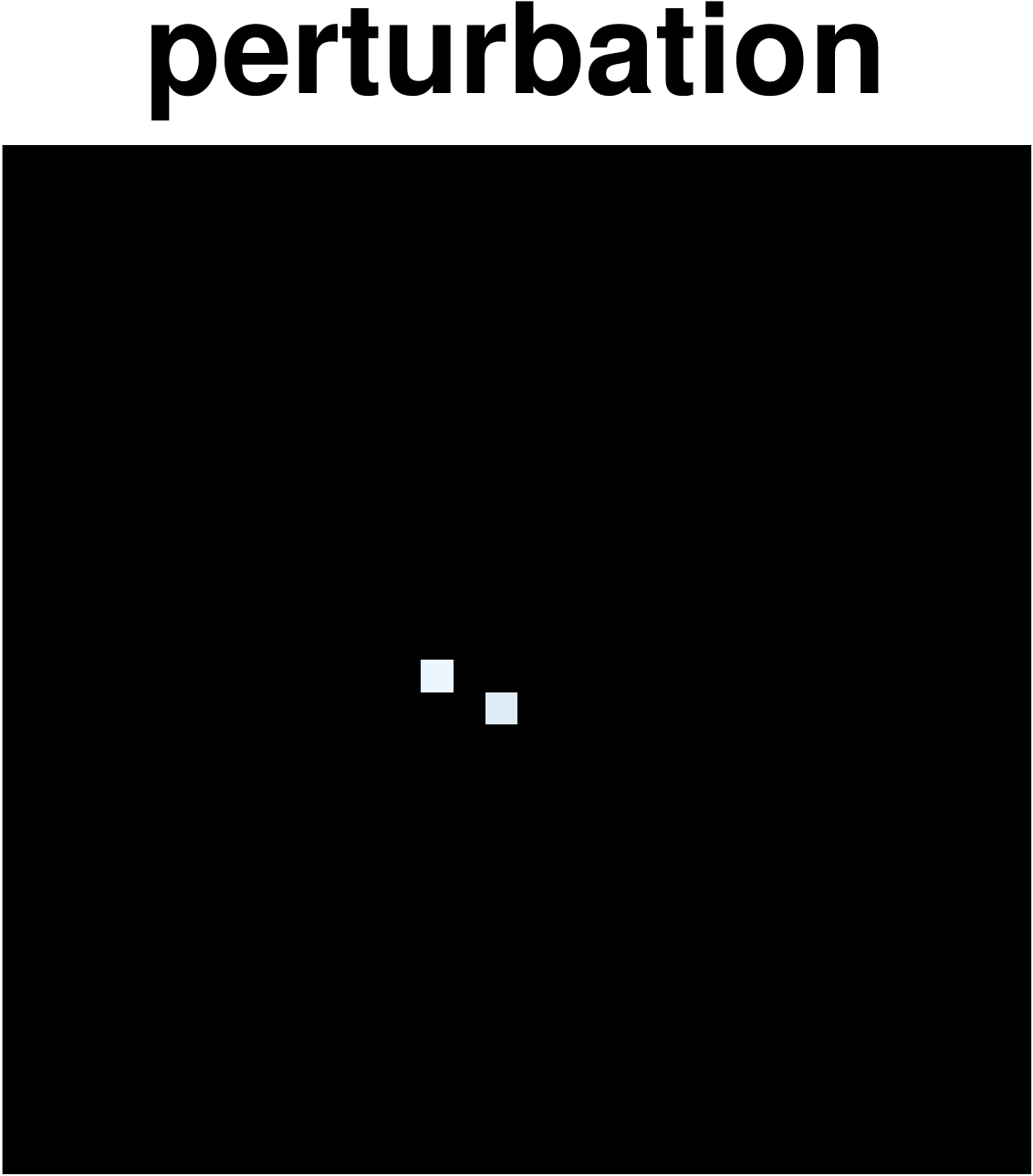}&
		\includegraphics[width=0.2\columnwidth]{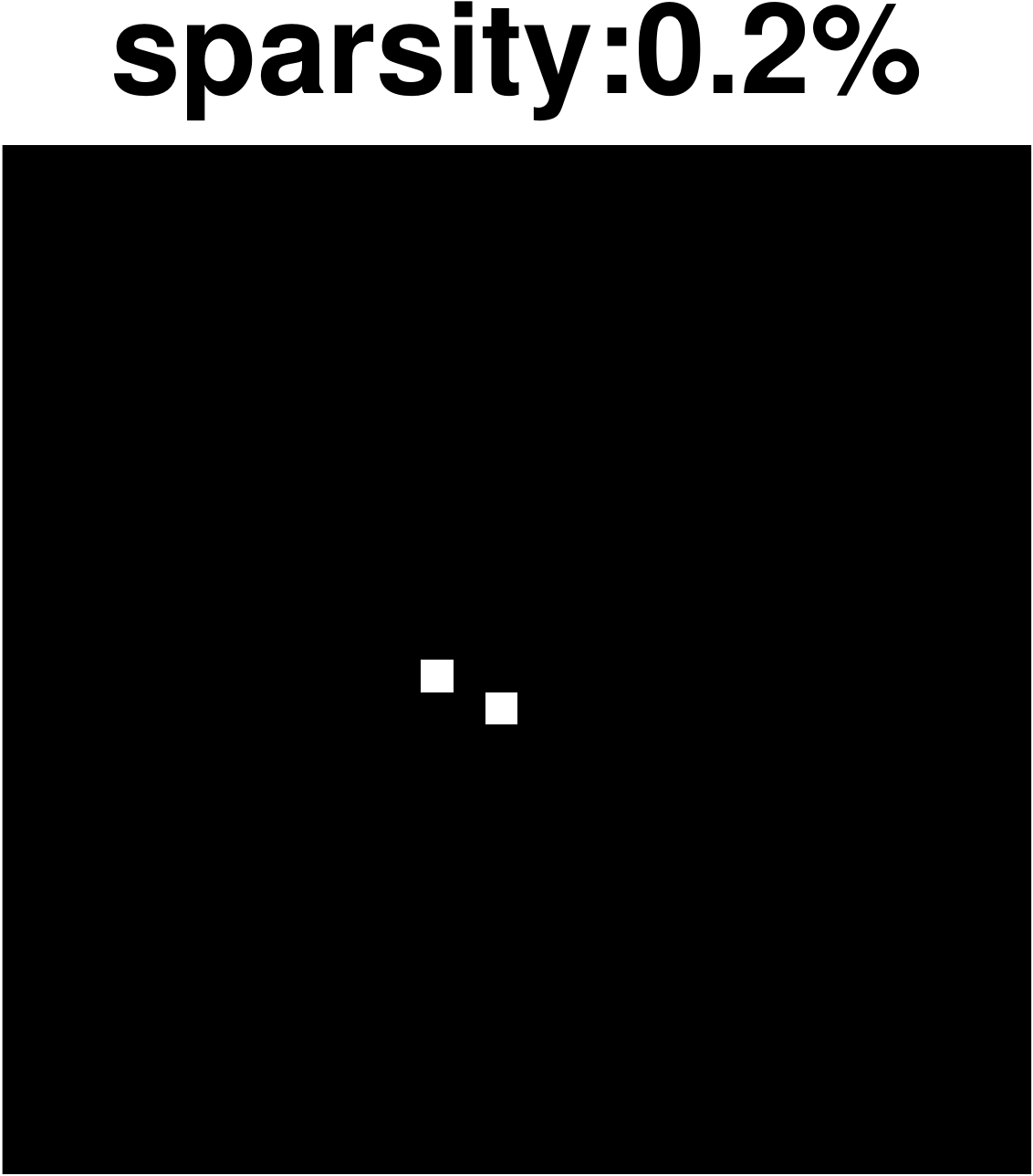}&
		
		\includegraphics[width=0.2\columnwidth]{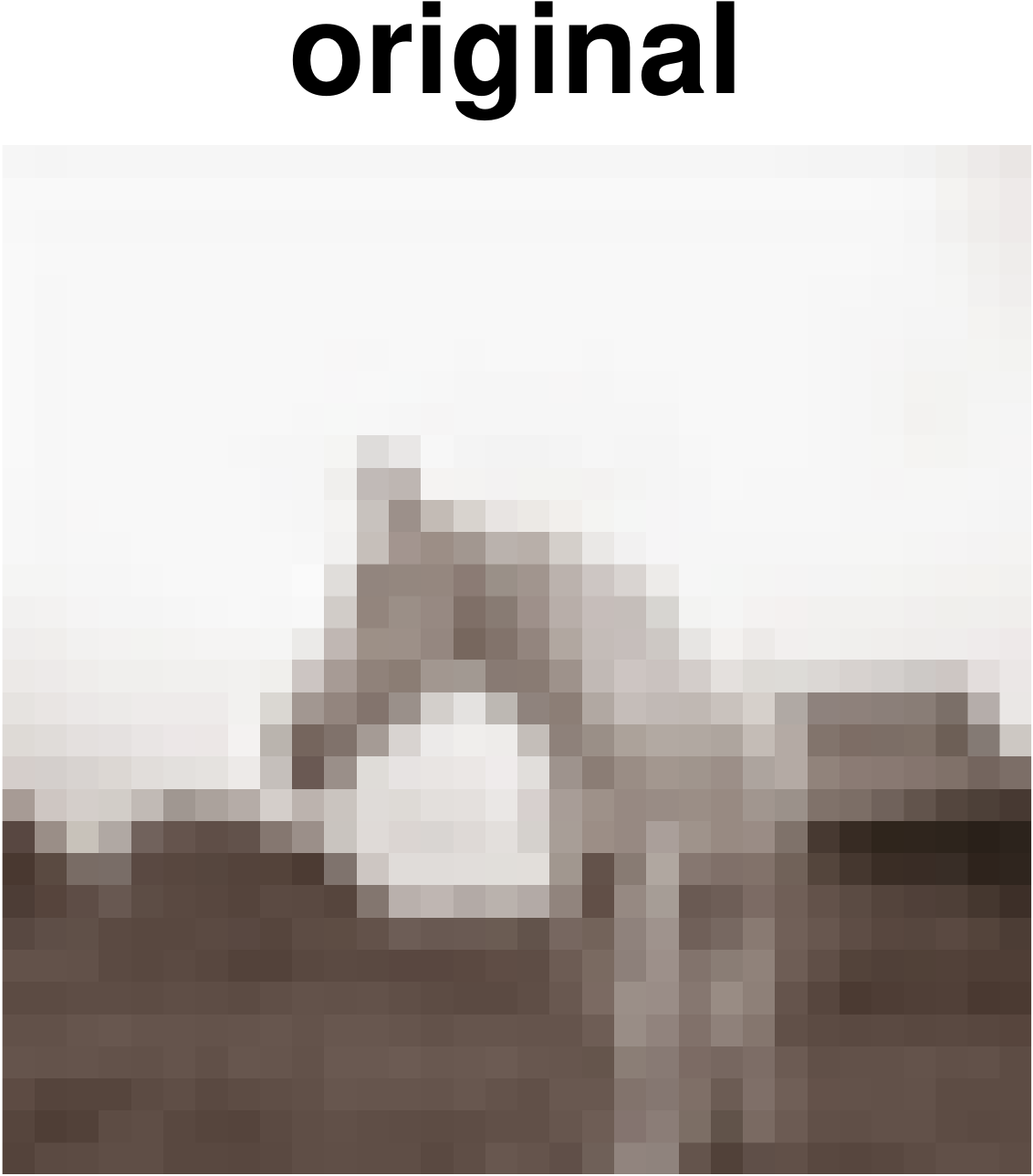}&
		\includegraphics[width=0.2\columnwidth]{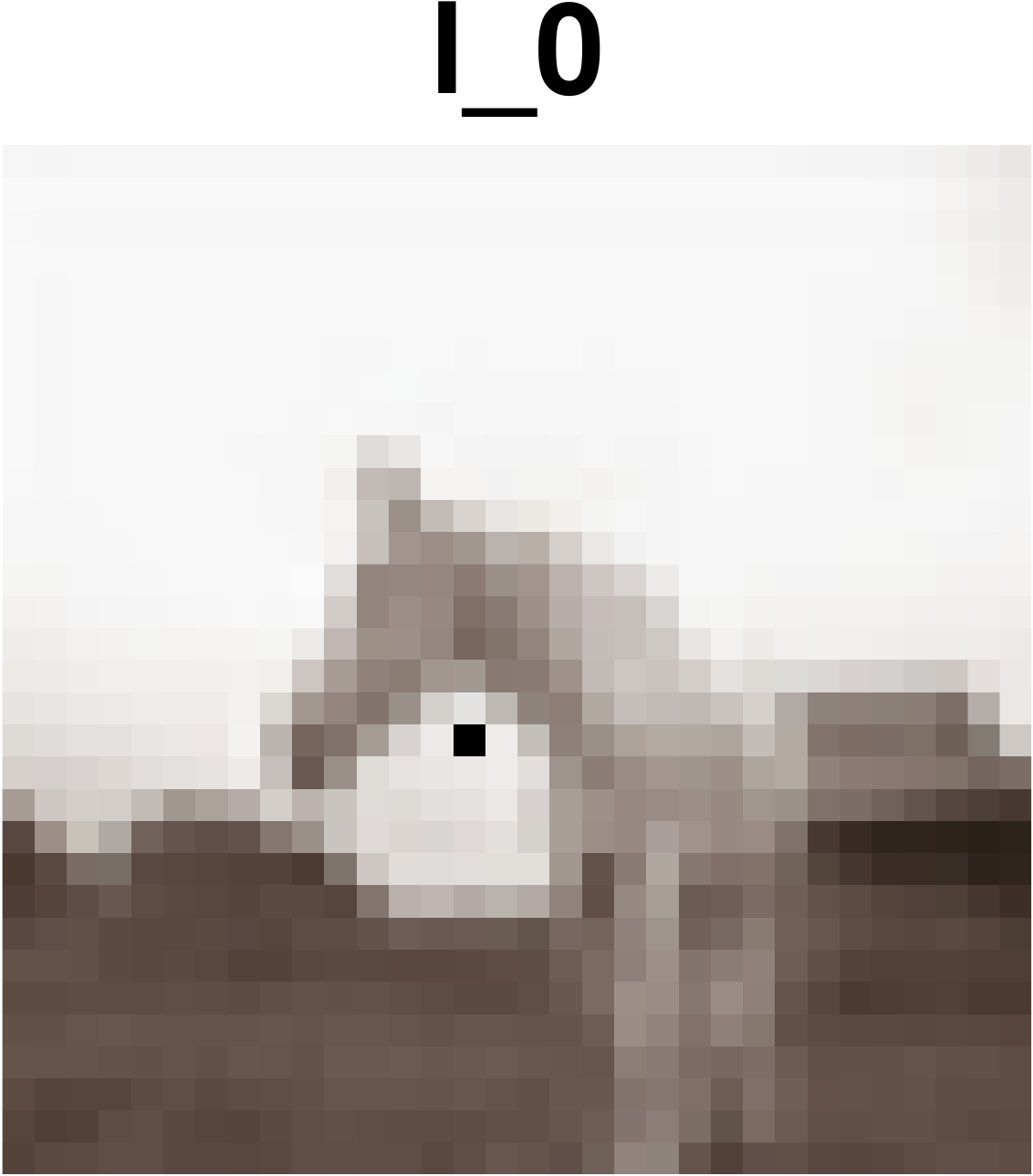}&
		\includegraphics[width=0.2\columnwidth]{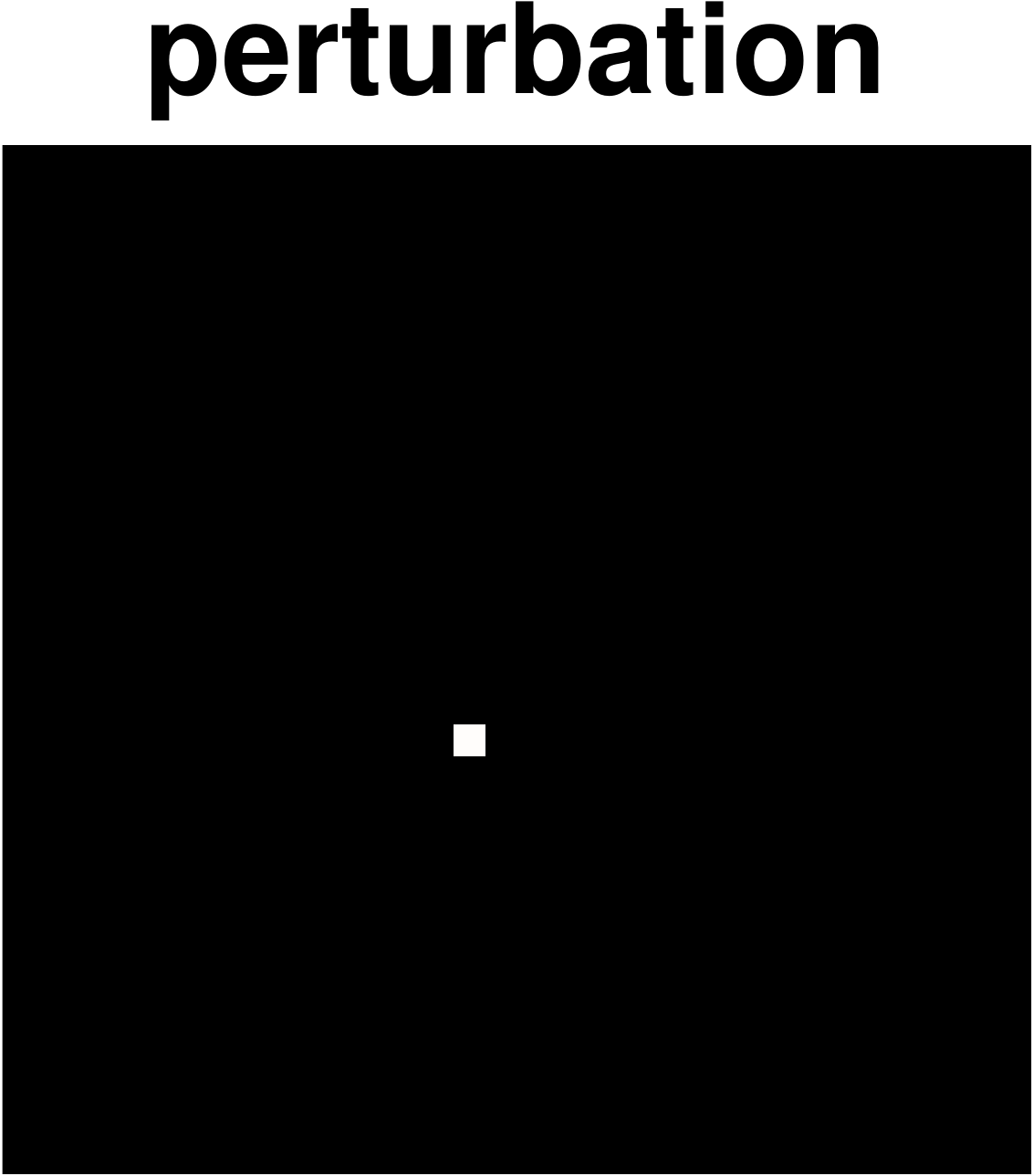}&
		\includegraphics[width=0.2\columnwidth]{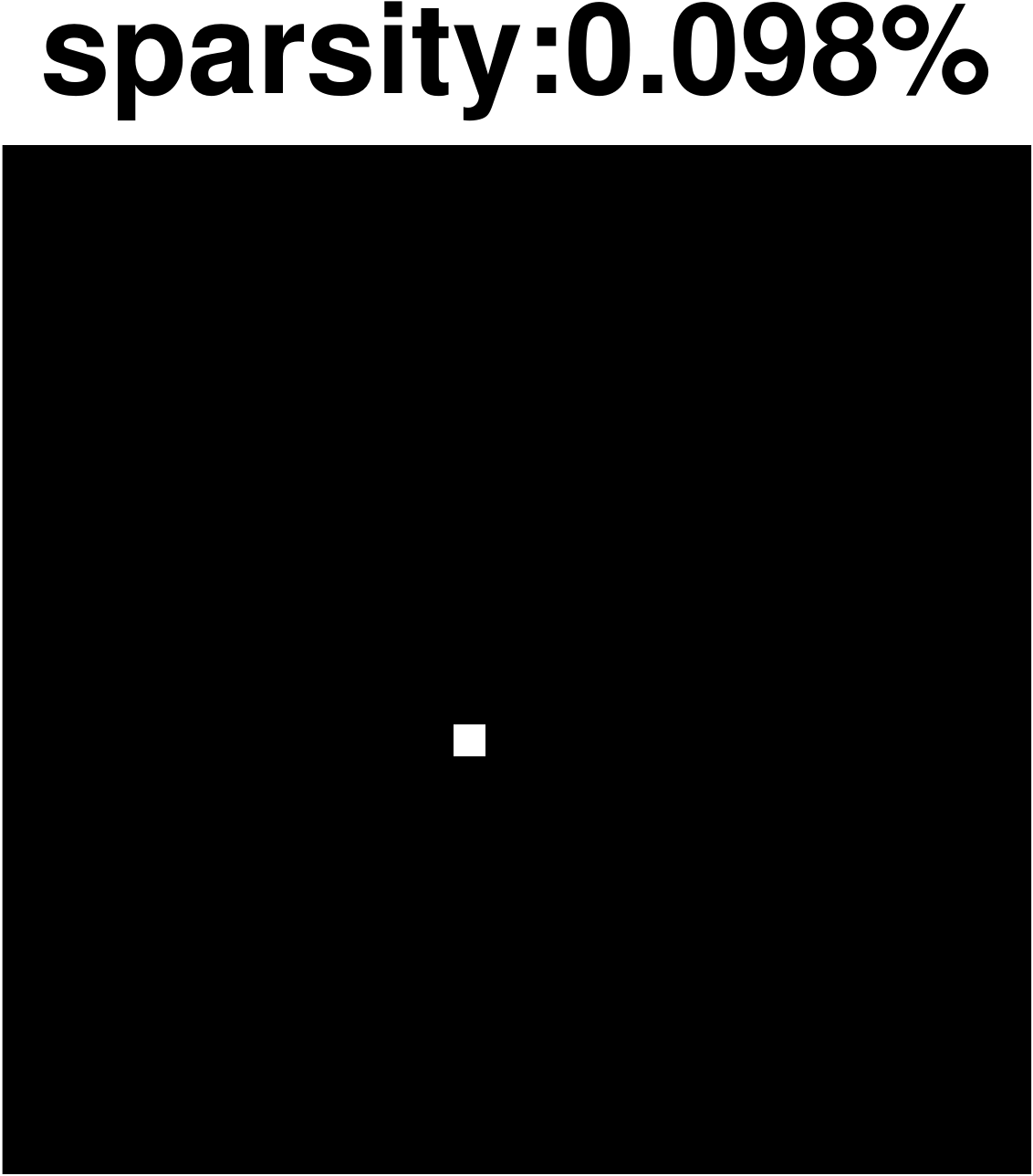}\\
		
		& \includegraphics[width=0.2\columnwidth]{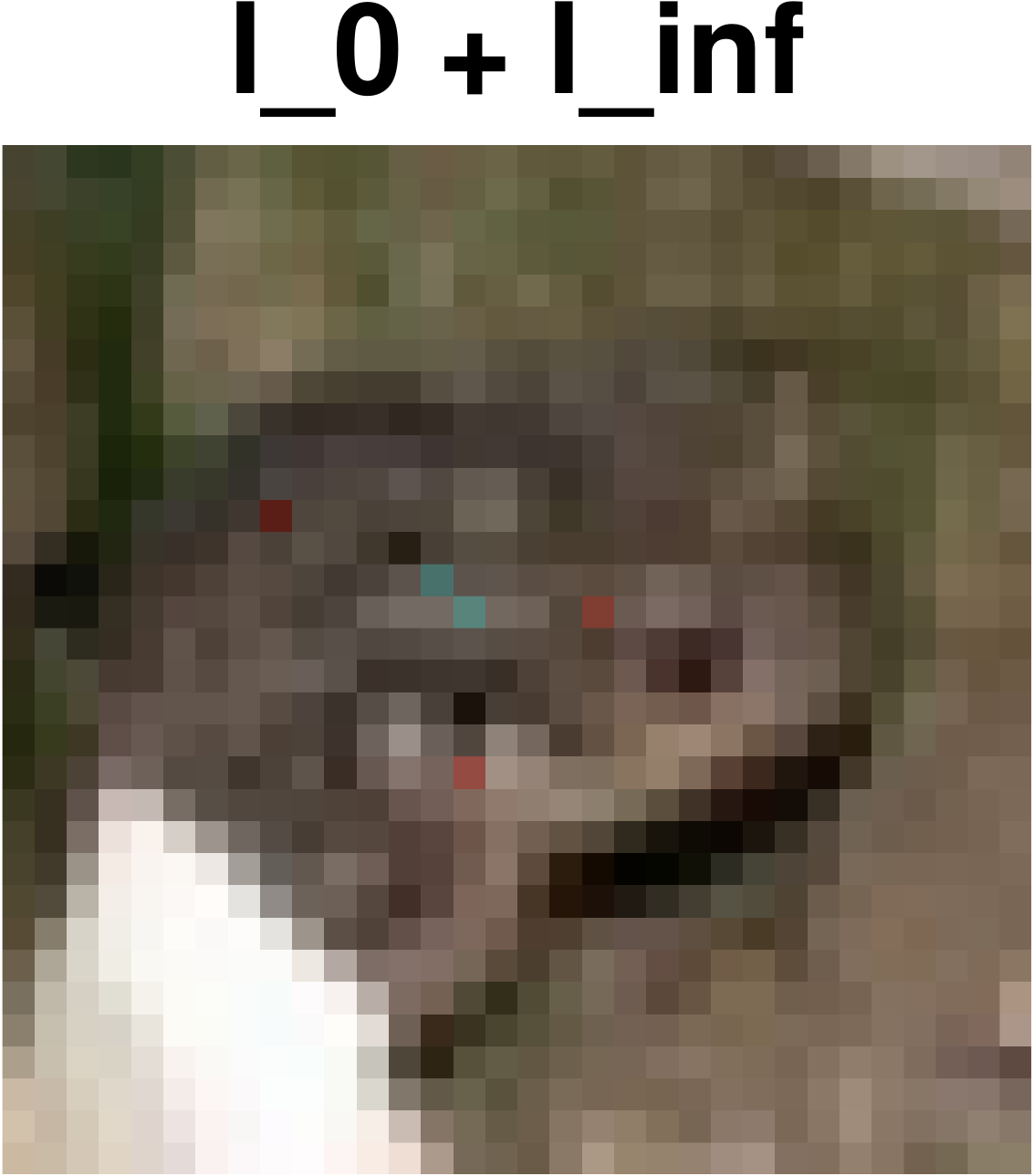}&
		\includegraphics[width=0.2\columnwidth]{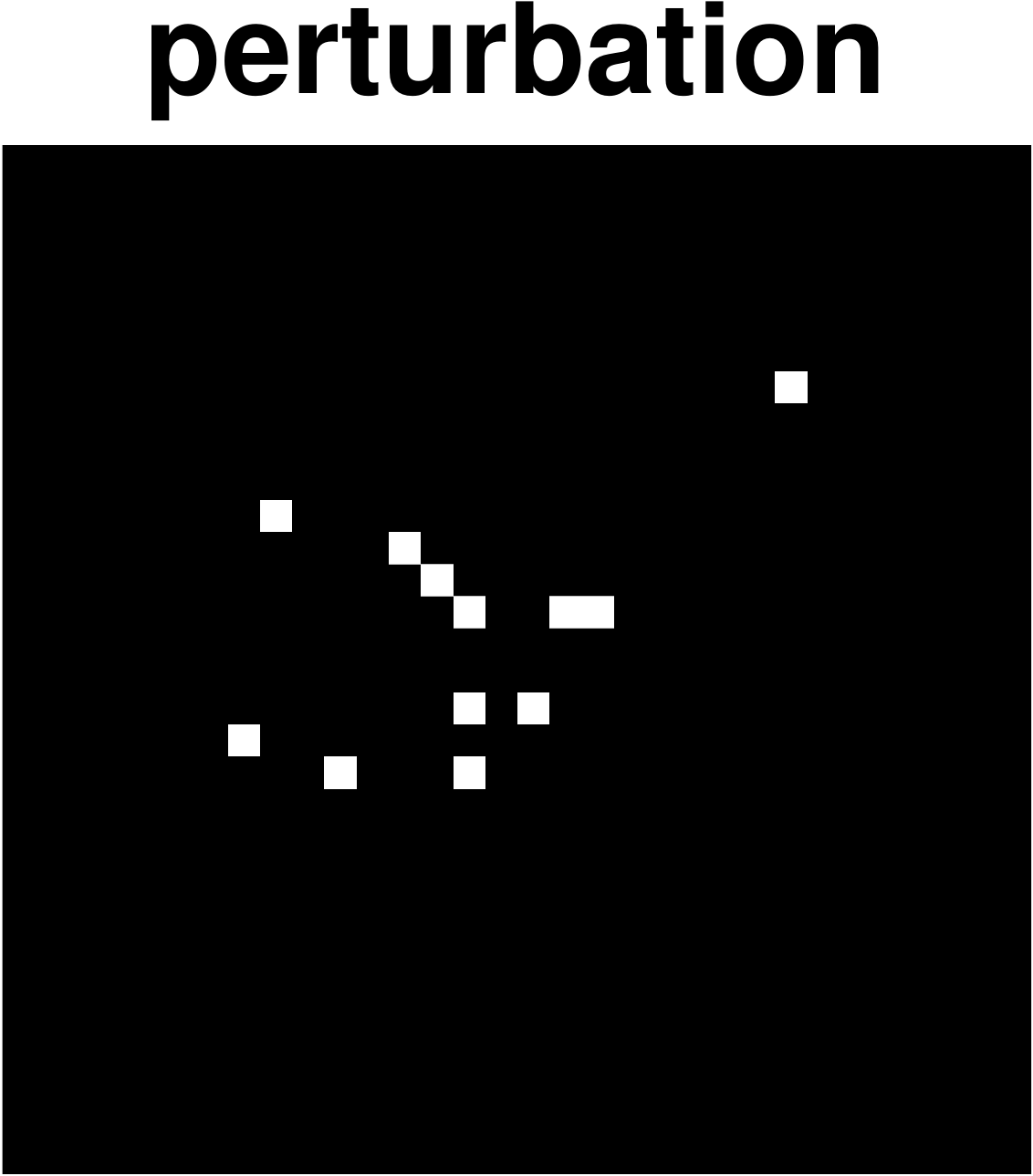}&
		\includegraphics[width=0.2\columnwidth]{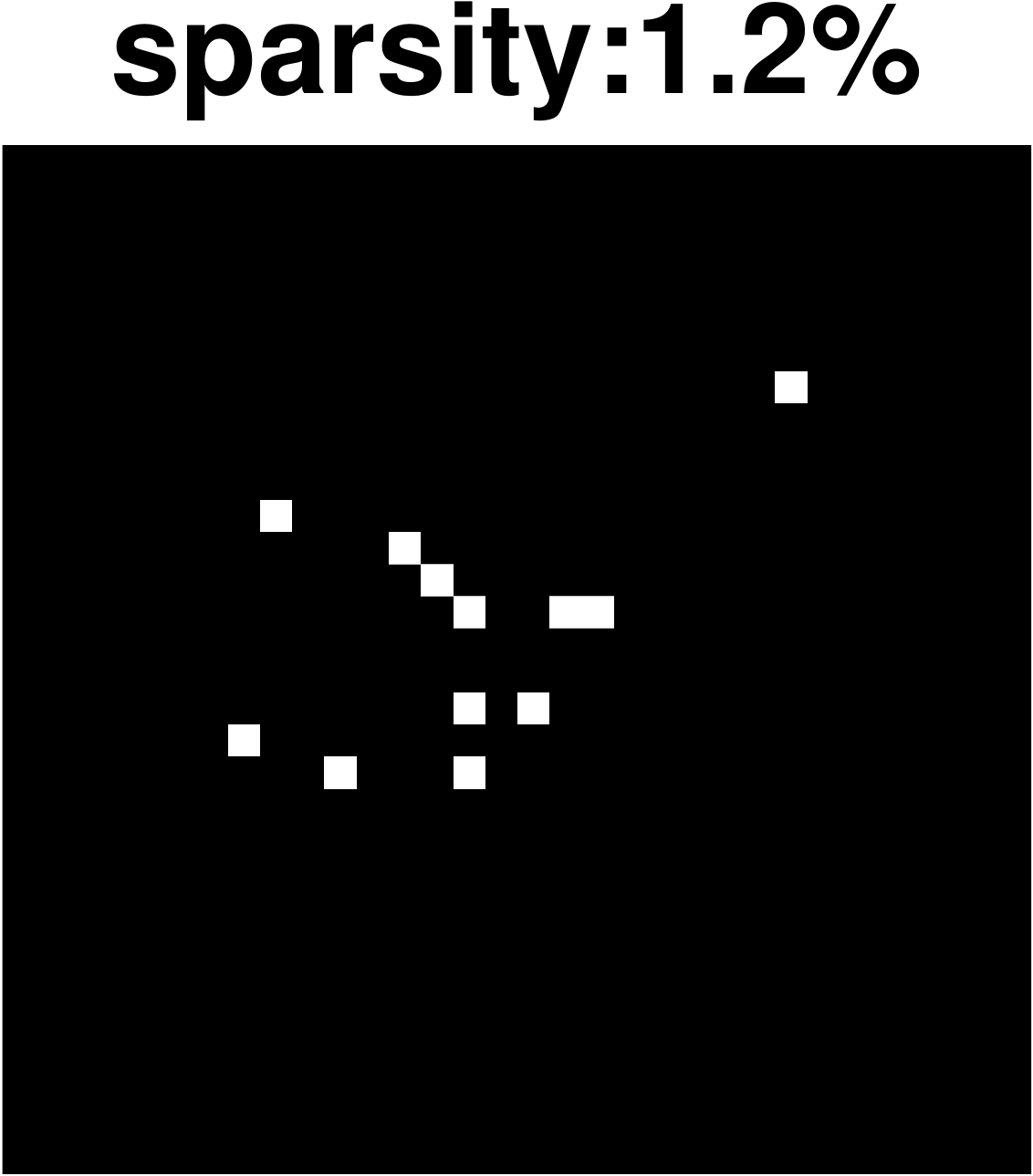}&
		&
		\includegraphics[width=0.2\columnwidth]{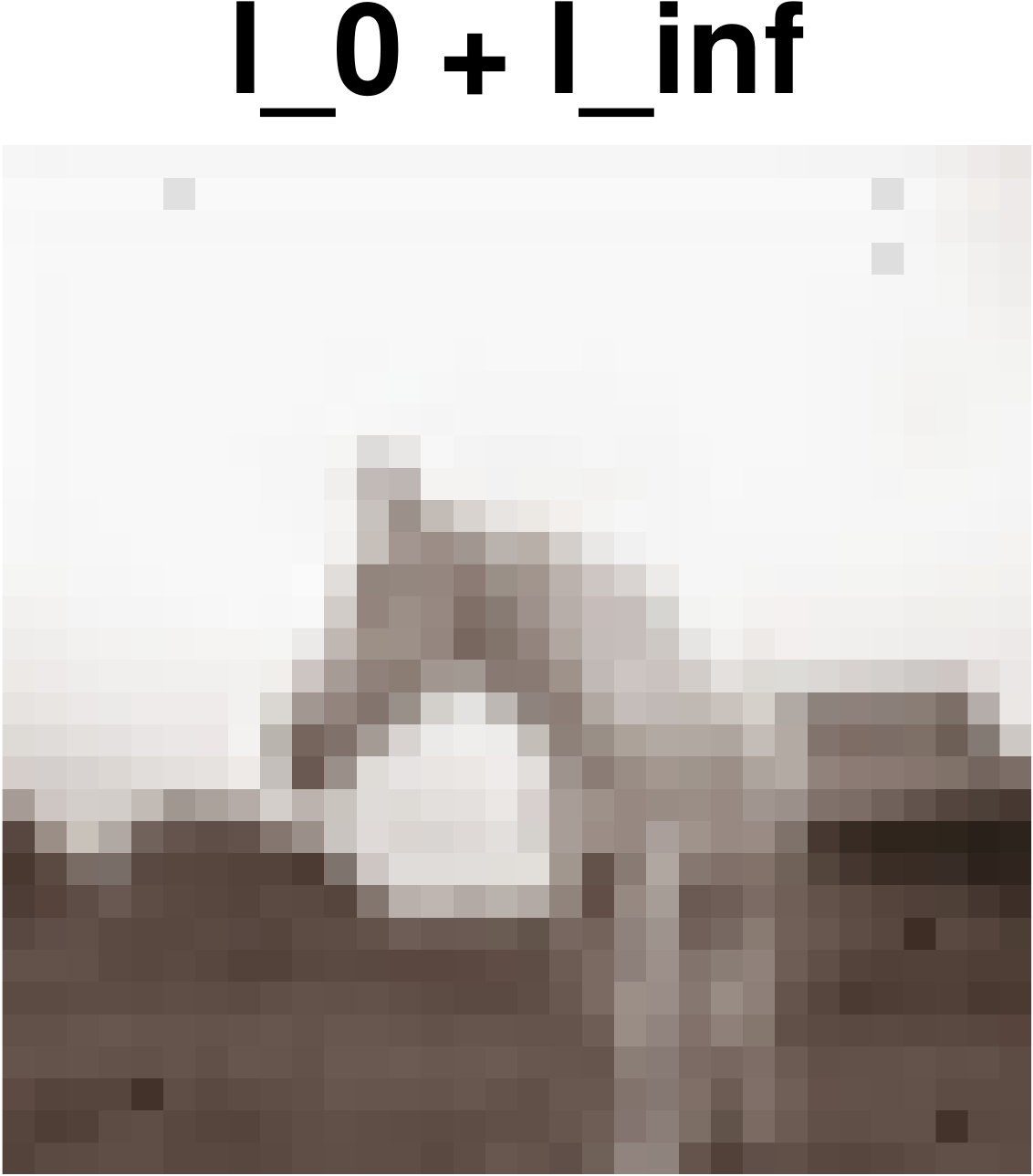}&
		\includegraphics[width=0.2\columnwidth]{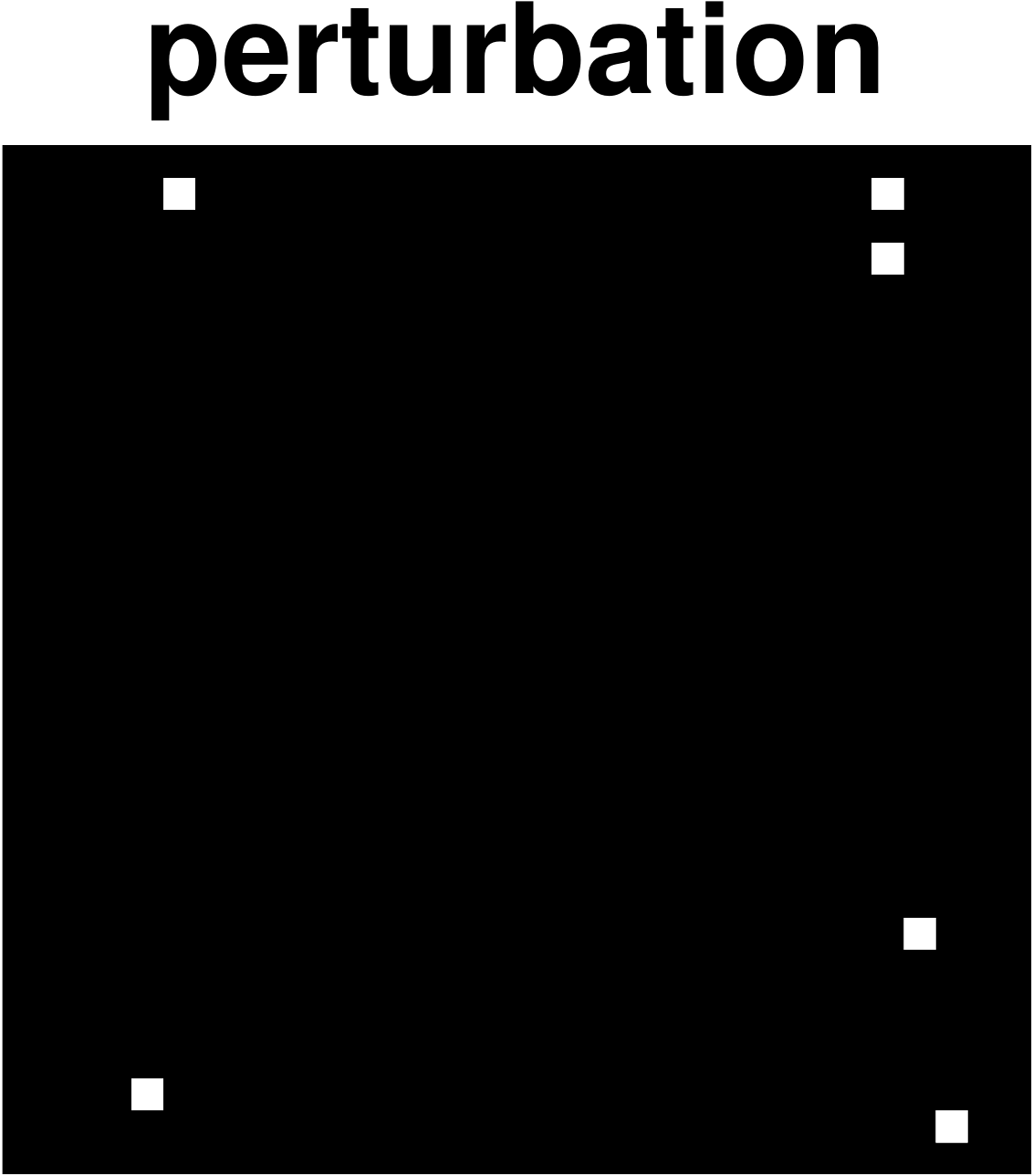}&
		\includegraphics[width=0.2\columnwidth]{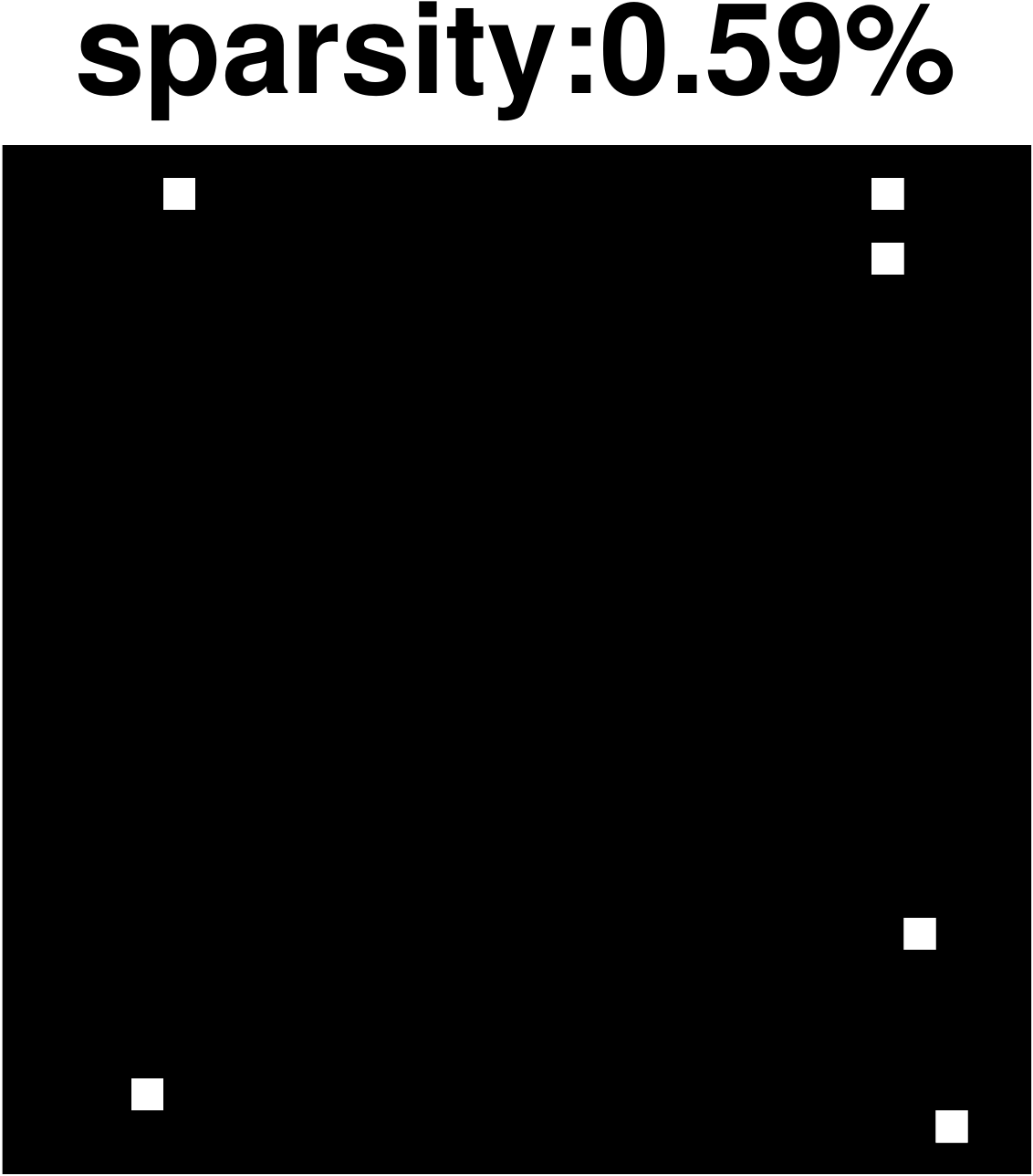}\\
		
		\includegraphics[width=0.2\columnwidth]{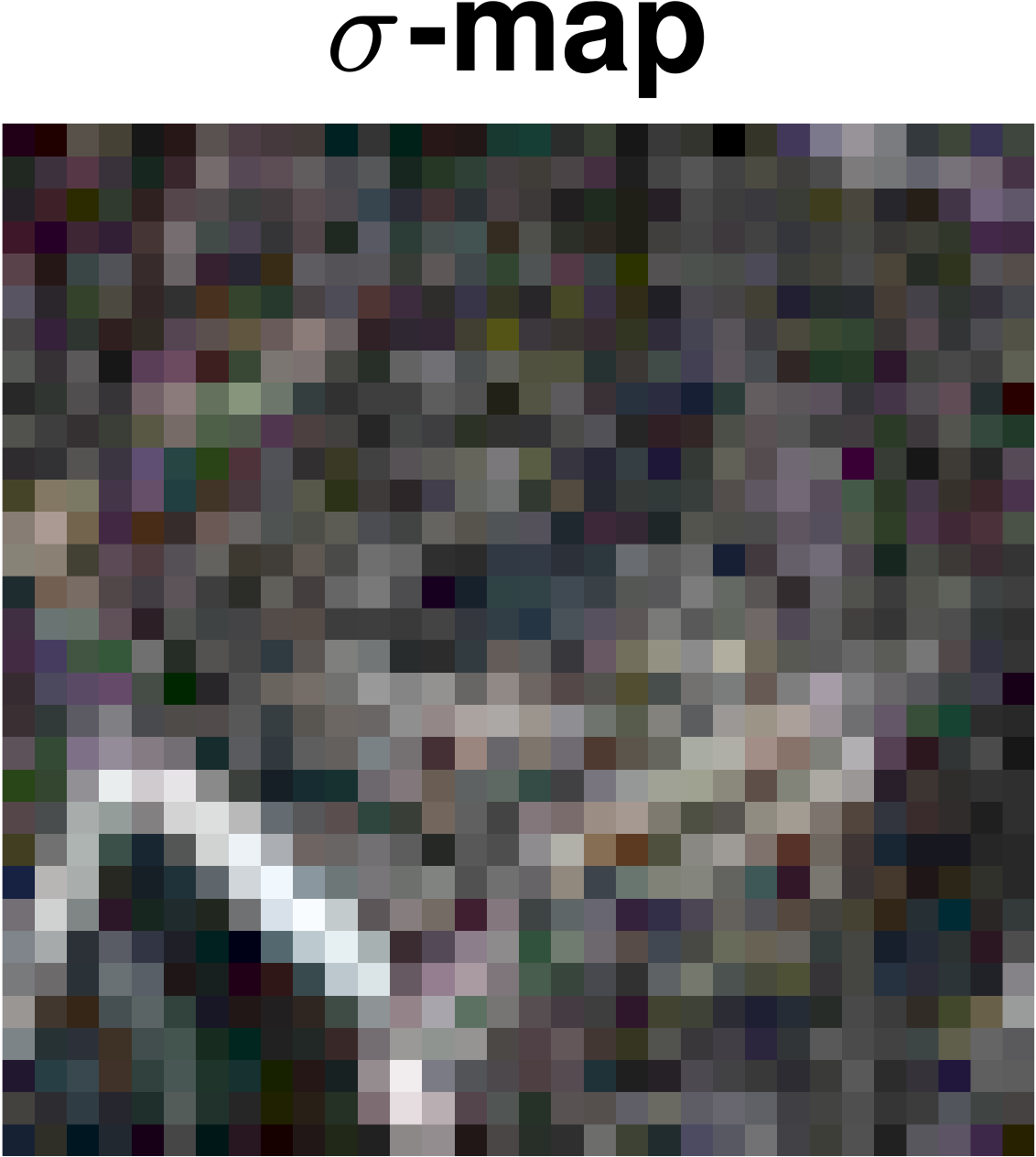}& 
		\includegraphics[width=0.2\columnwidth]{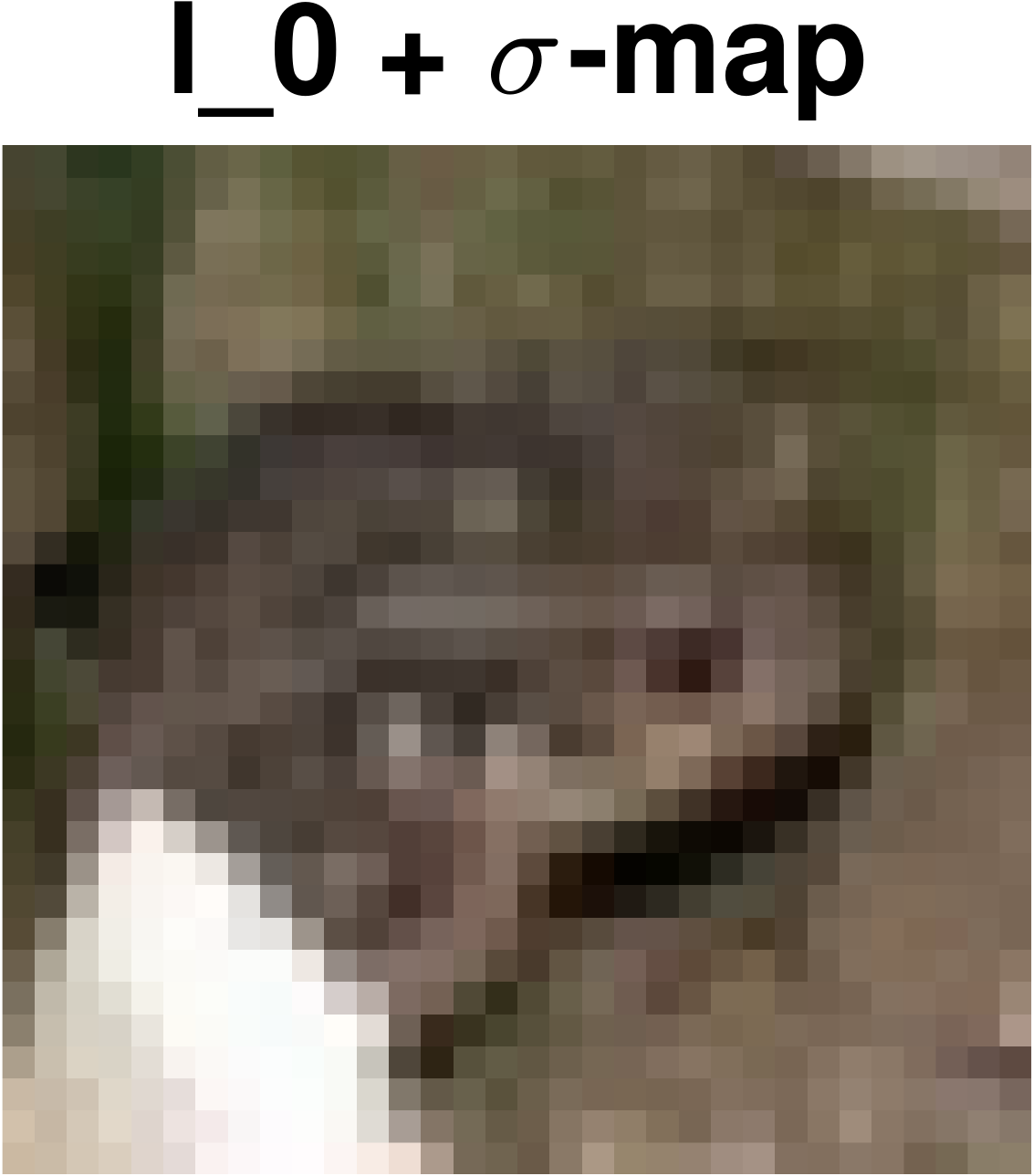}&
		\includegraphics[width=0.2\columnwidth]{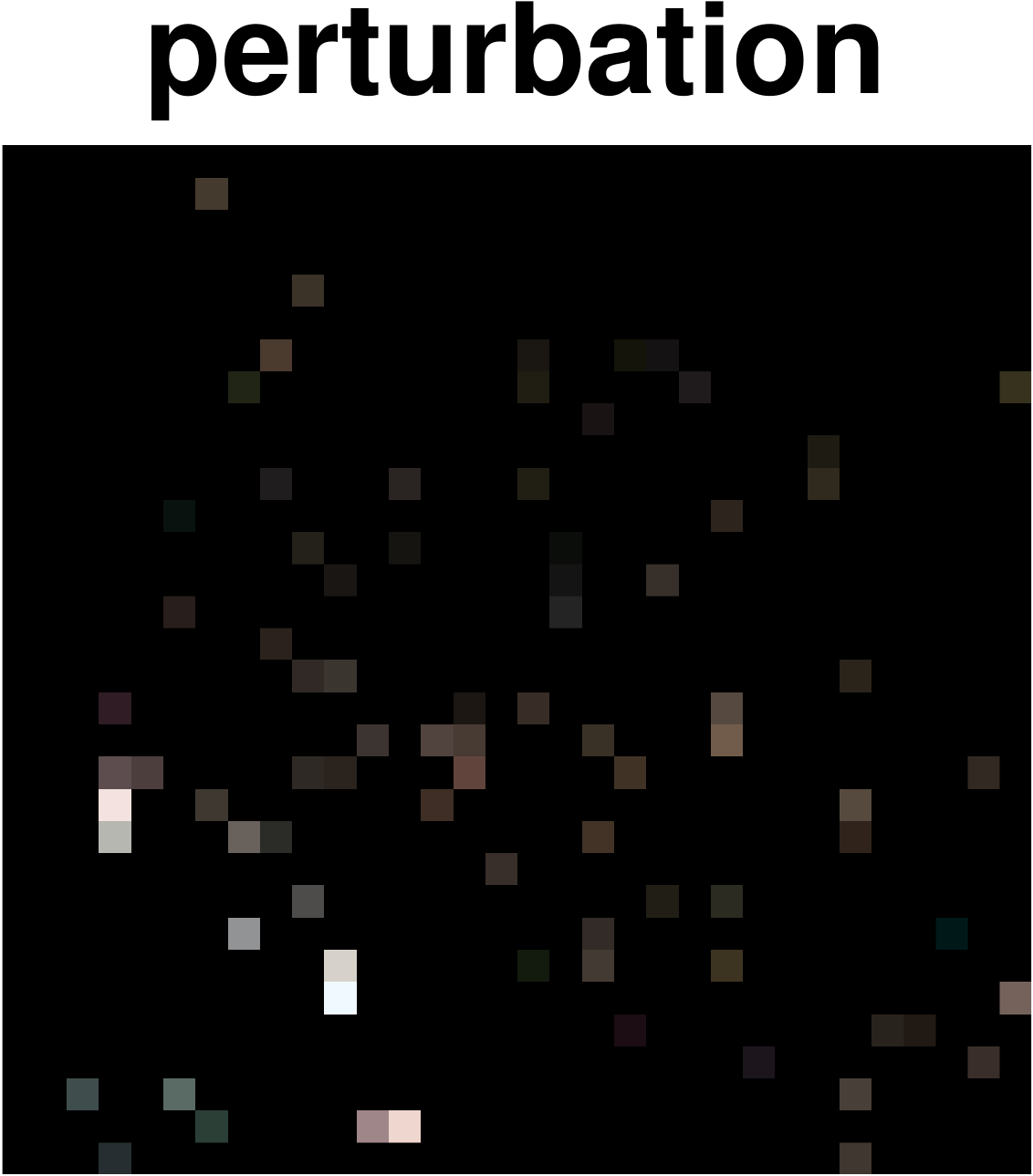}&
		\includegraphics[width=0.2\columnwidth]{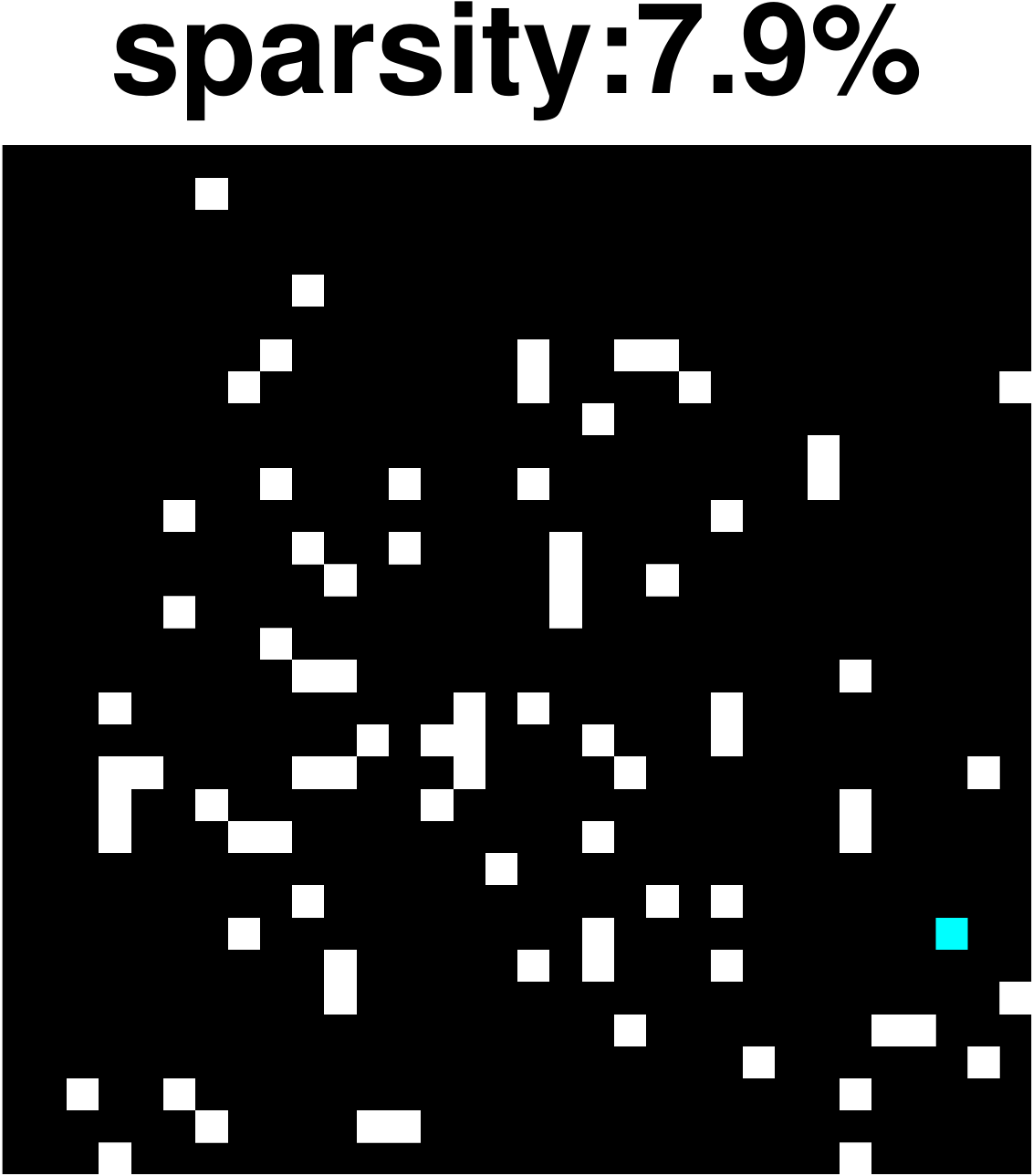}&
		
		\includegraphics[width=0.2\columnwidth]{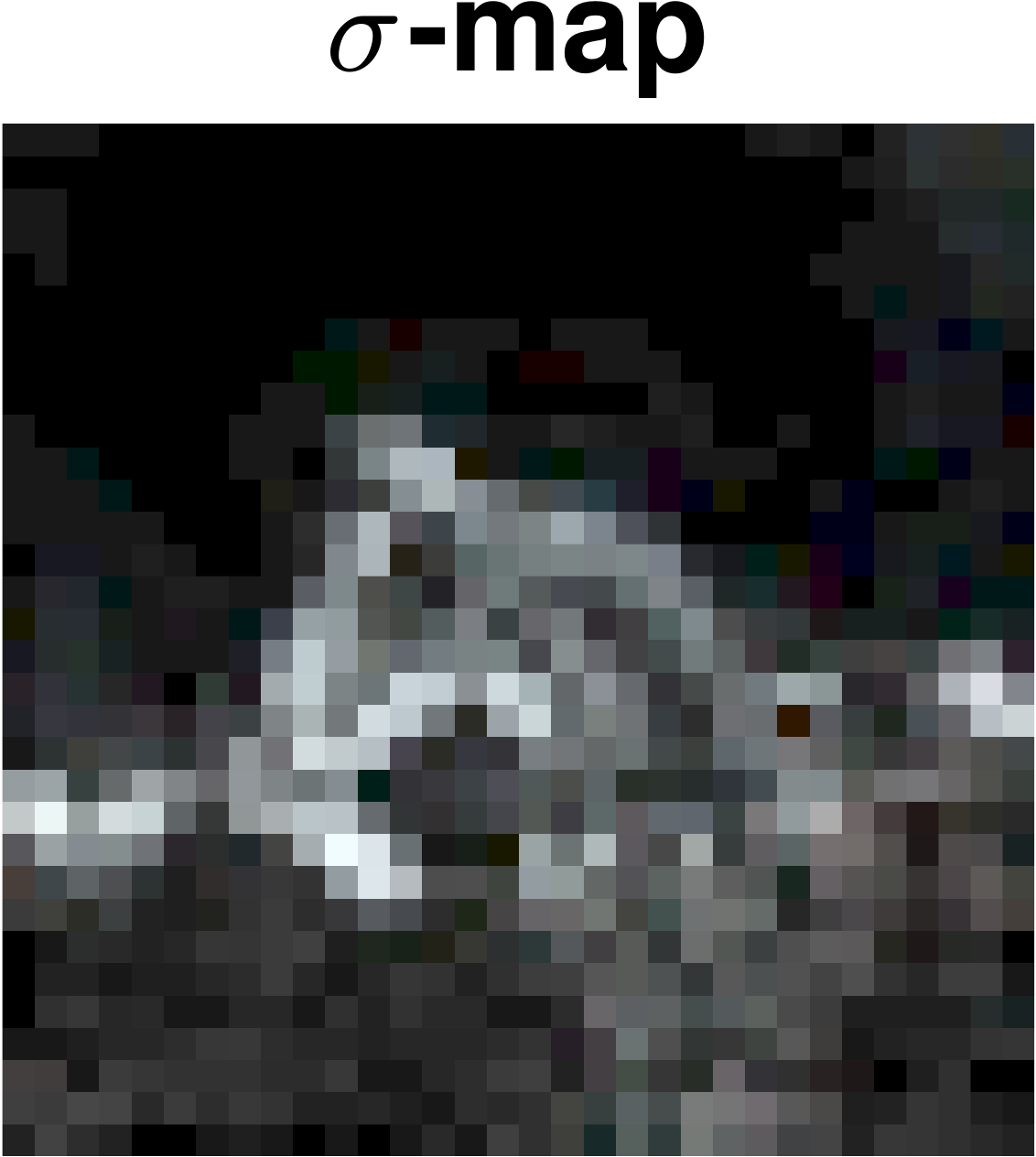}& 
		\includegraphics[width=0.2\columnwidth]{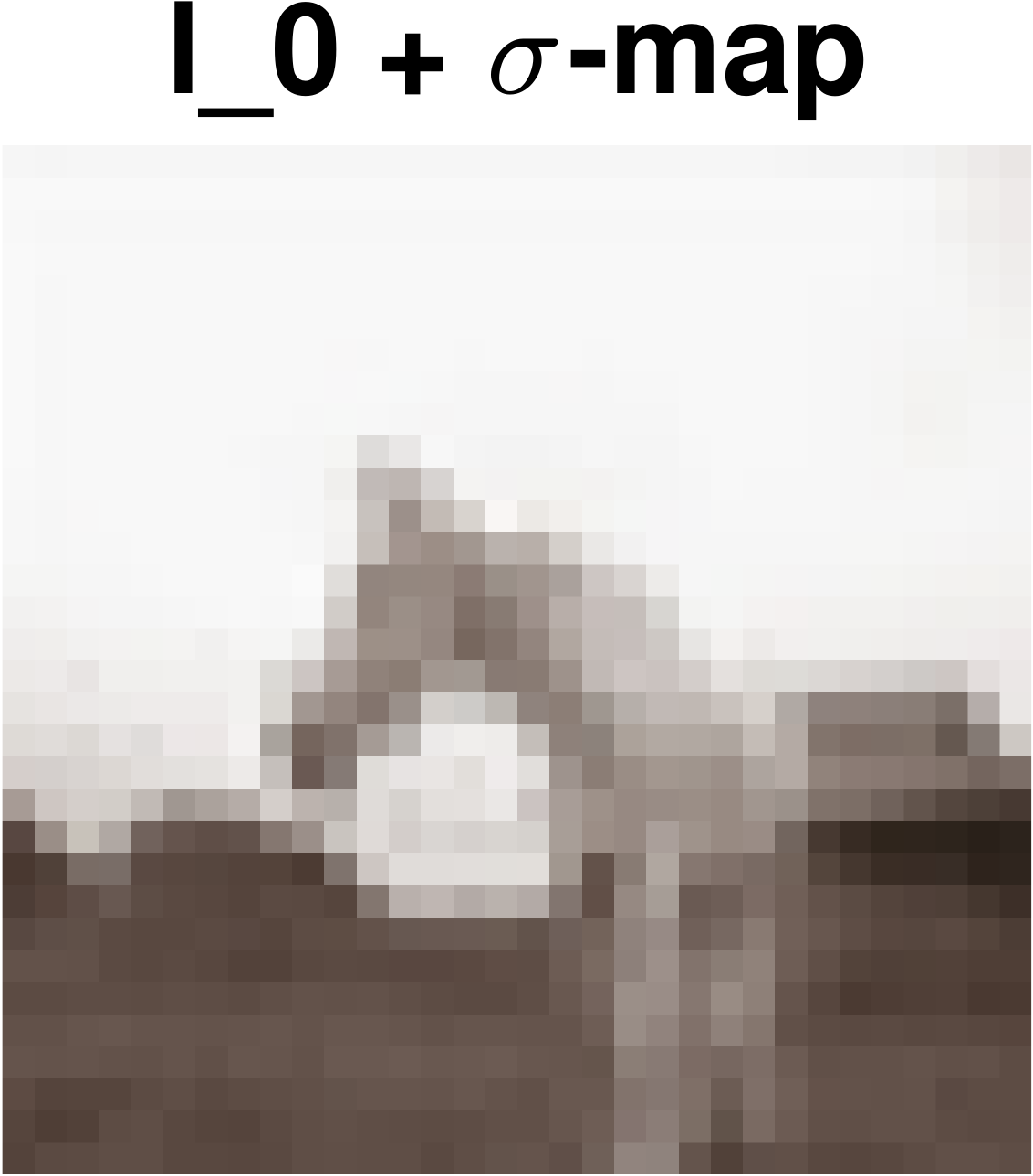}&
		\includegraphics[width=0.2\columnwidth]{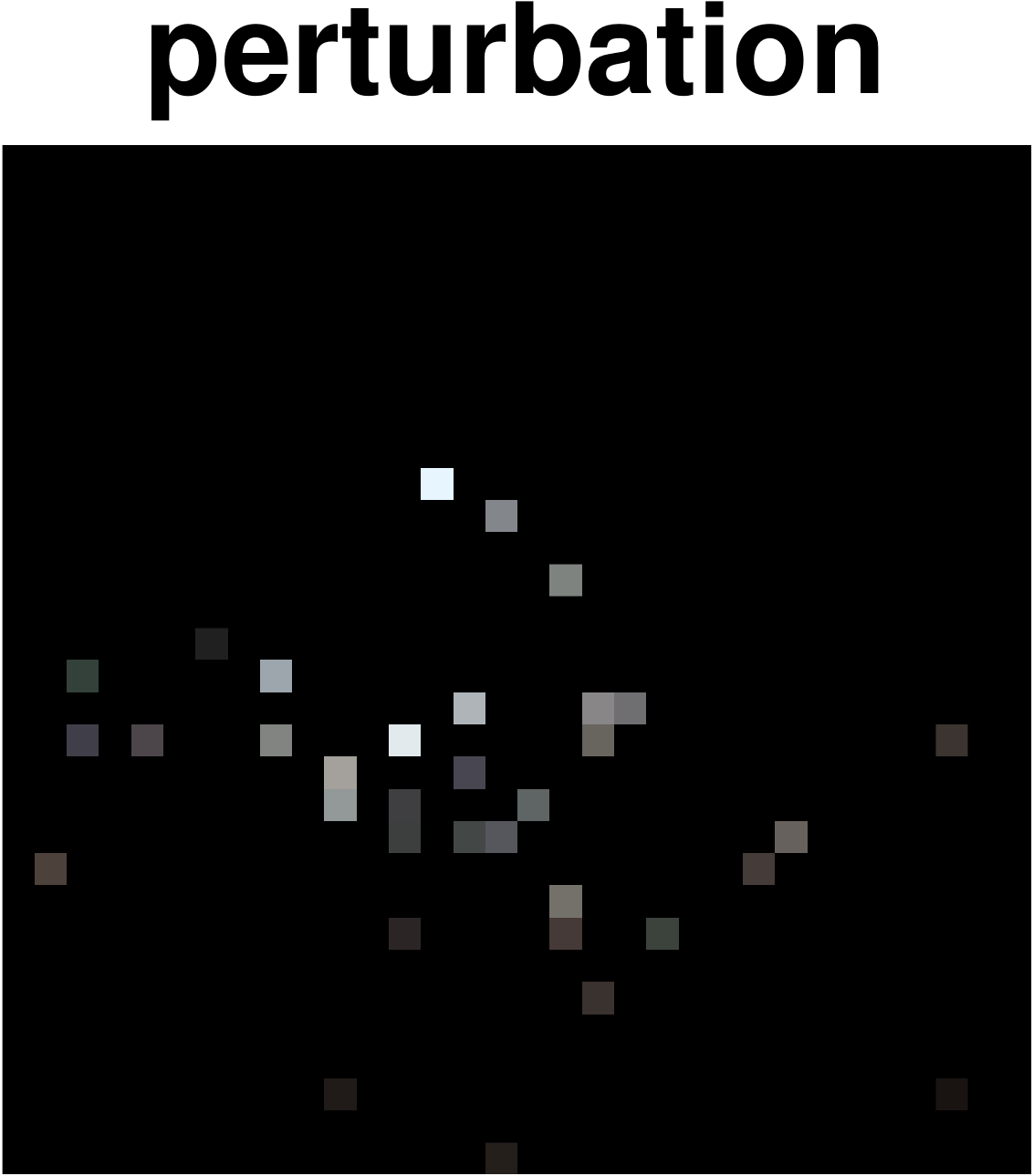}&
		\includegraphics[width=0.2\columnwidth]{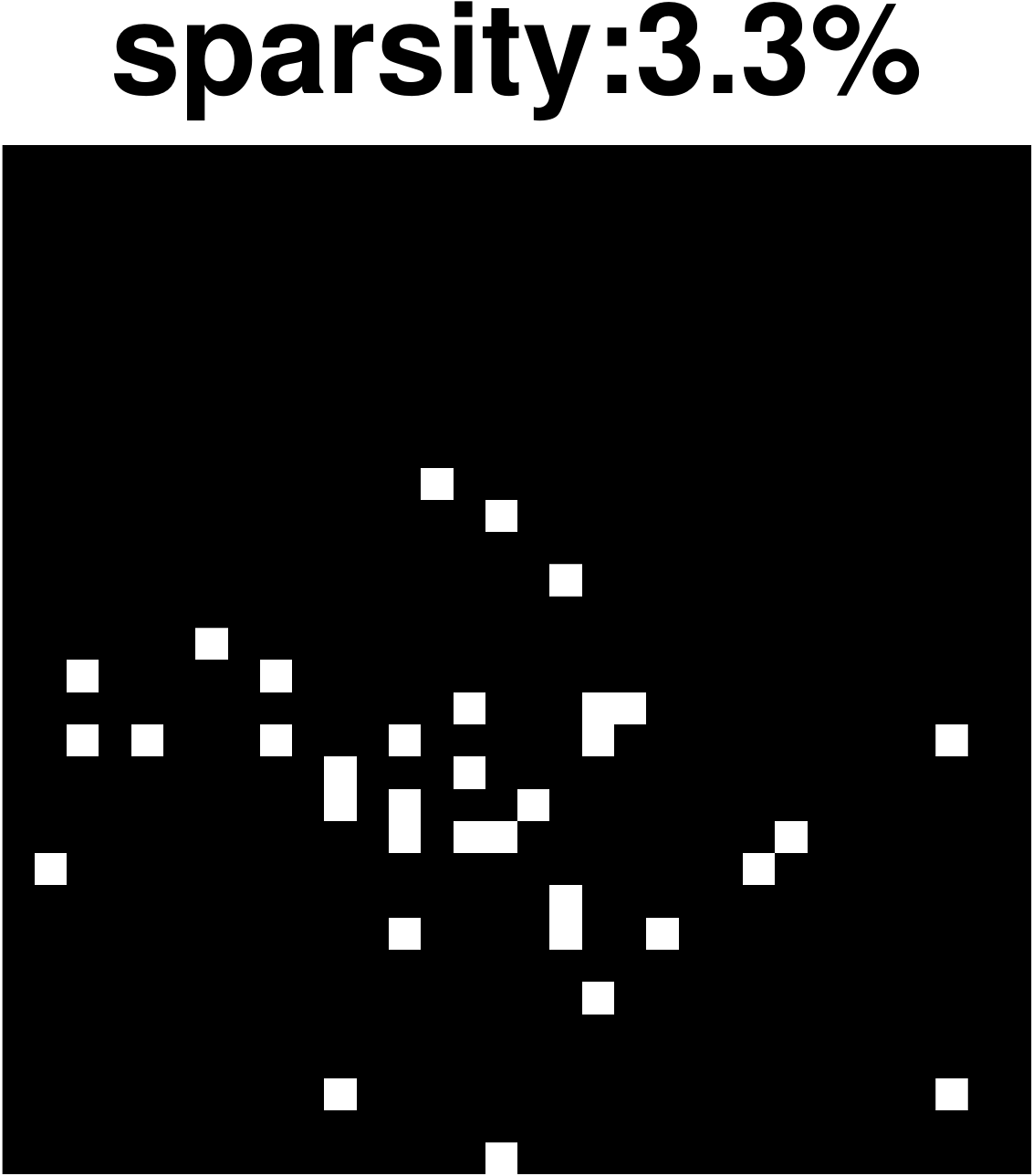} \\
		
		%% im44 frog, dog, dog, dog
		%% im88 horse, cat, deer, cat
		
	\end{tabular}
	\caption{\textbf{Different attacks on CIFAR-10.} We illustrate the differences of the adversarial examples (second column) found by CornerSearch ($l_0$), $l_0+l_\infty$-attack and $\sigma$-CornerSearch respectively first, second and third row. The third column shows the adversarial perturbations rescaled to $[0,1]$, the fourth the map of the modified pixels (\textit{sparsity} column). The original image can be found top left and the RGB representation of the $\sigma$-map bottom left.}\label{fig:imp_CIFAR-10}
\end{figure*}

\begin{figure*}
	\centering
	\begin{tabular}{c c  c c| c c c c}
		\includegraphics[width=0.2\columnwidth]{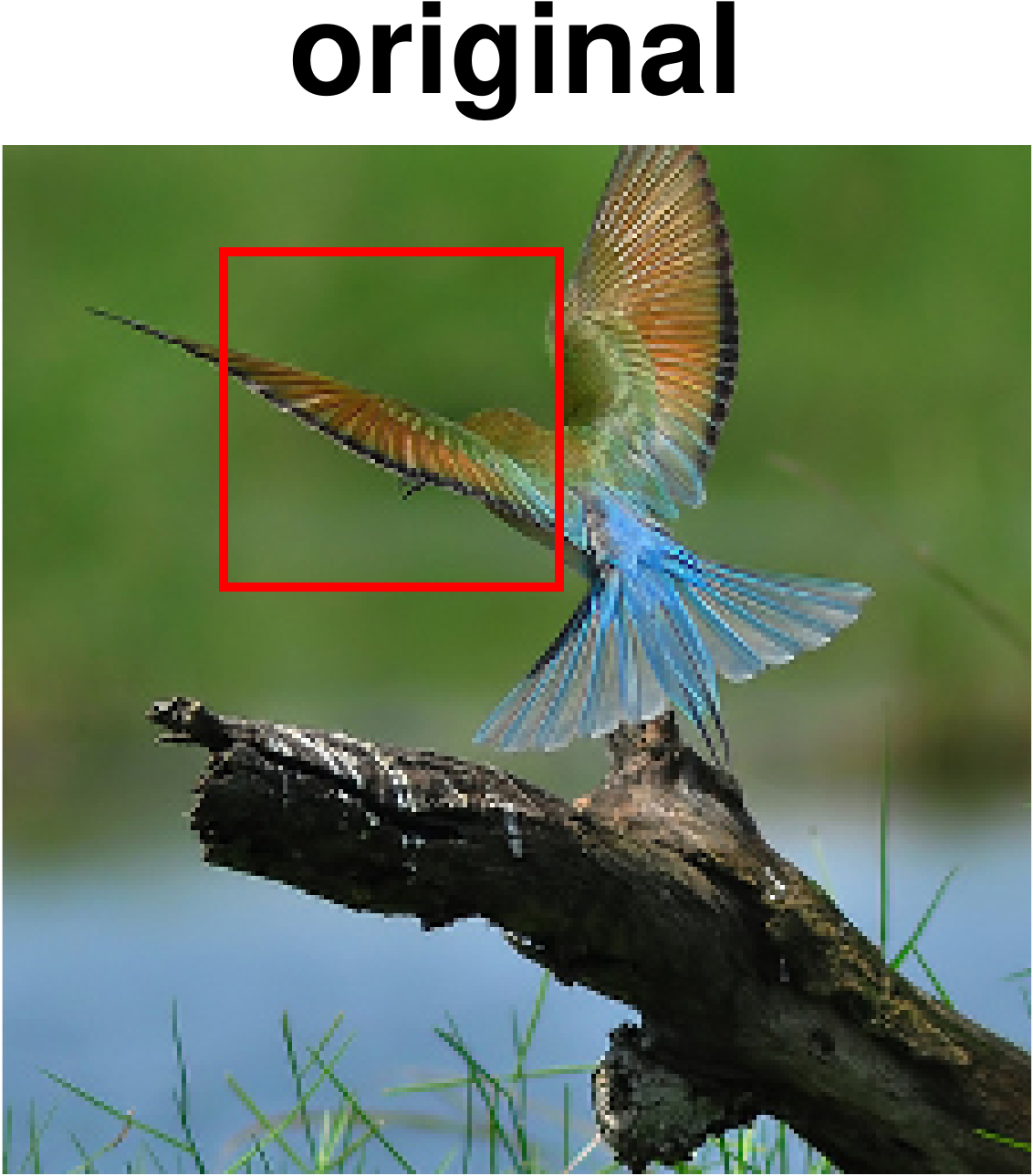}&
		\includegraphics[width=0.2\columnwidth]{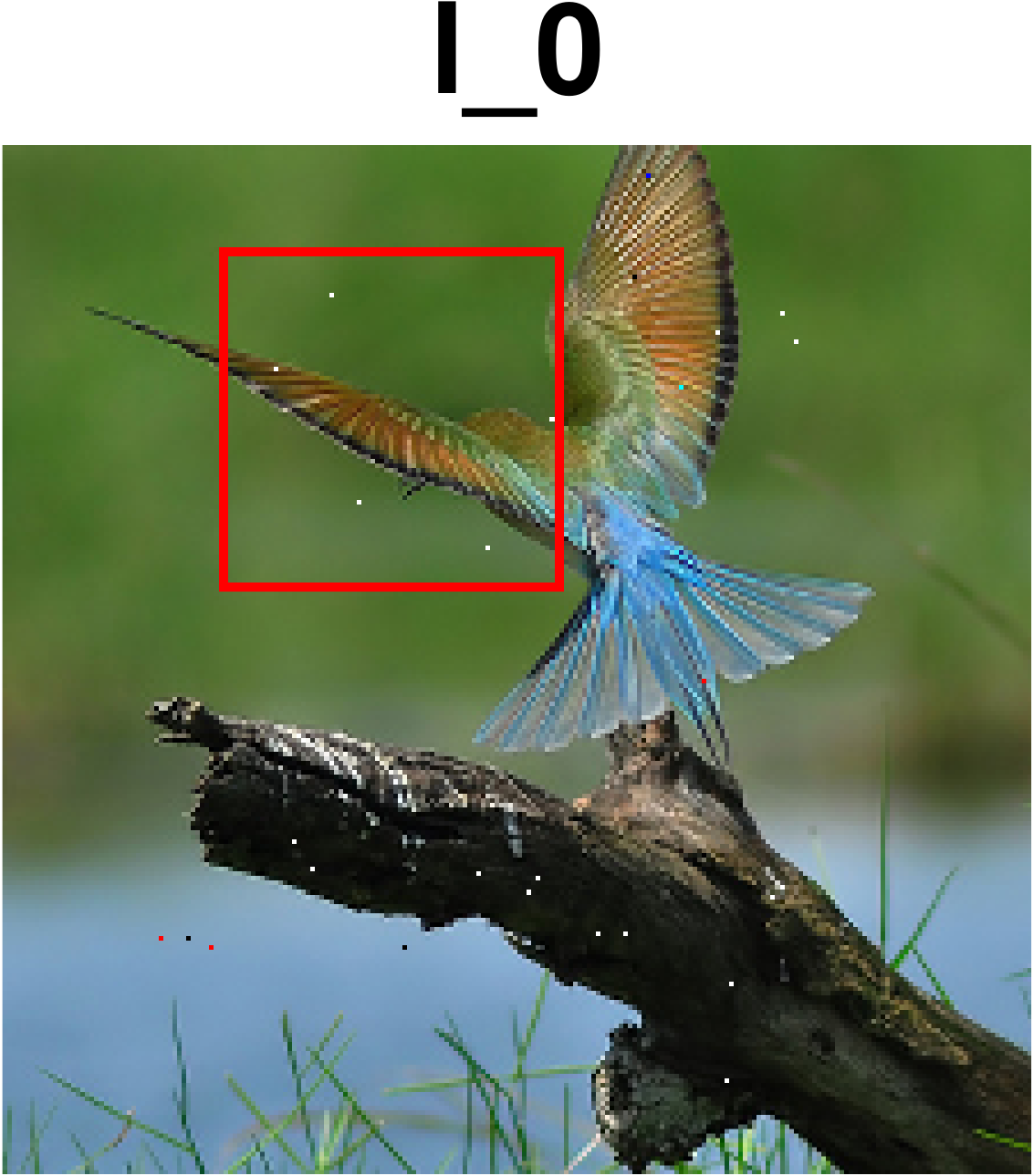}&
		\includegraphics[width=0.2\columnwidth]{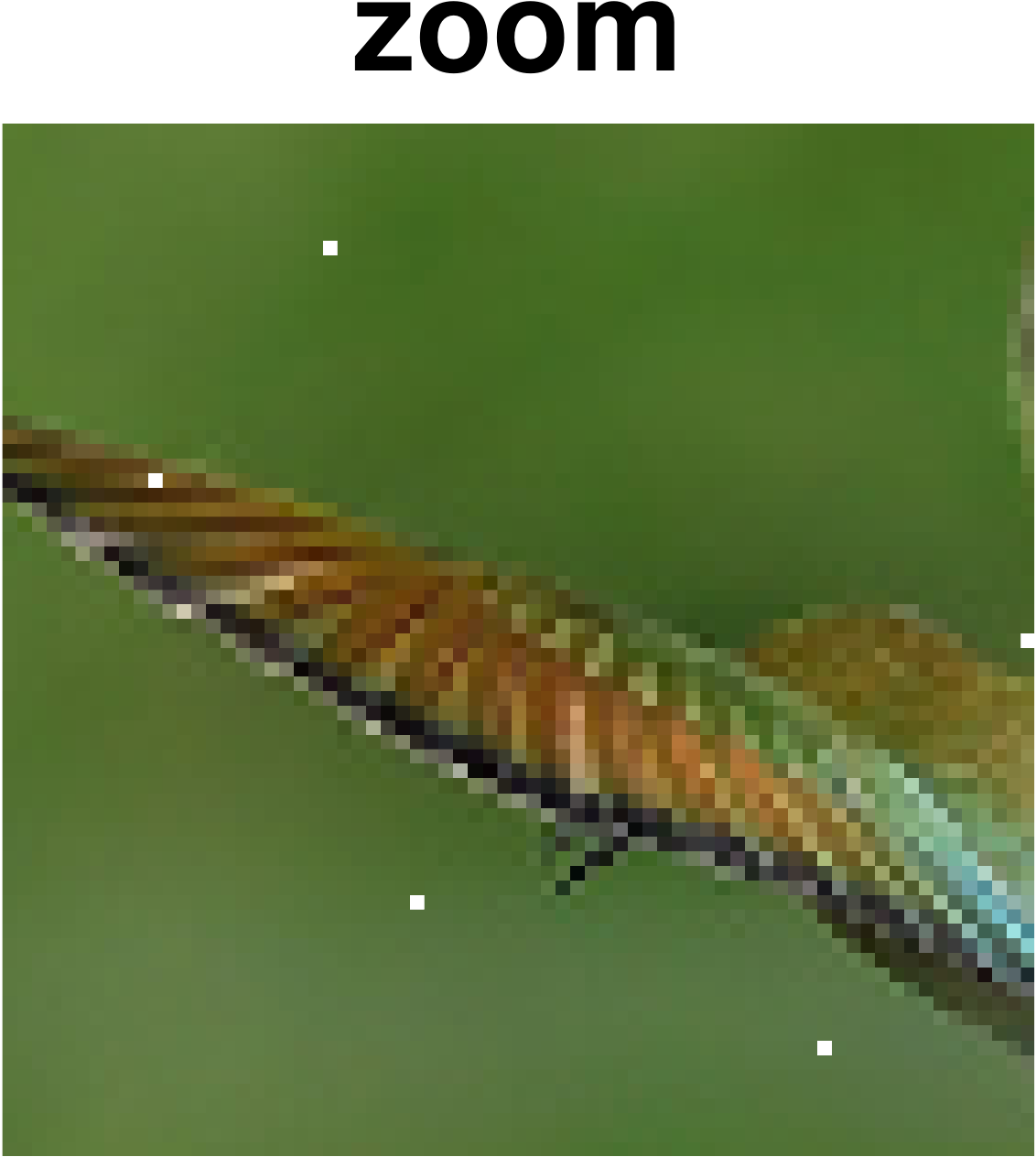}&
		\includegraphics[width=0.2\columnwidth]{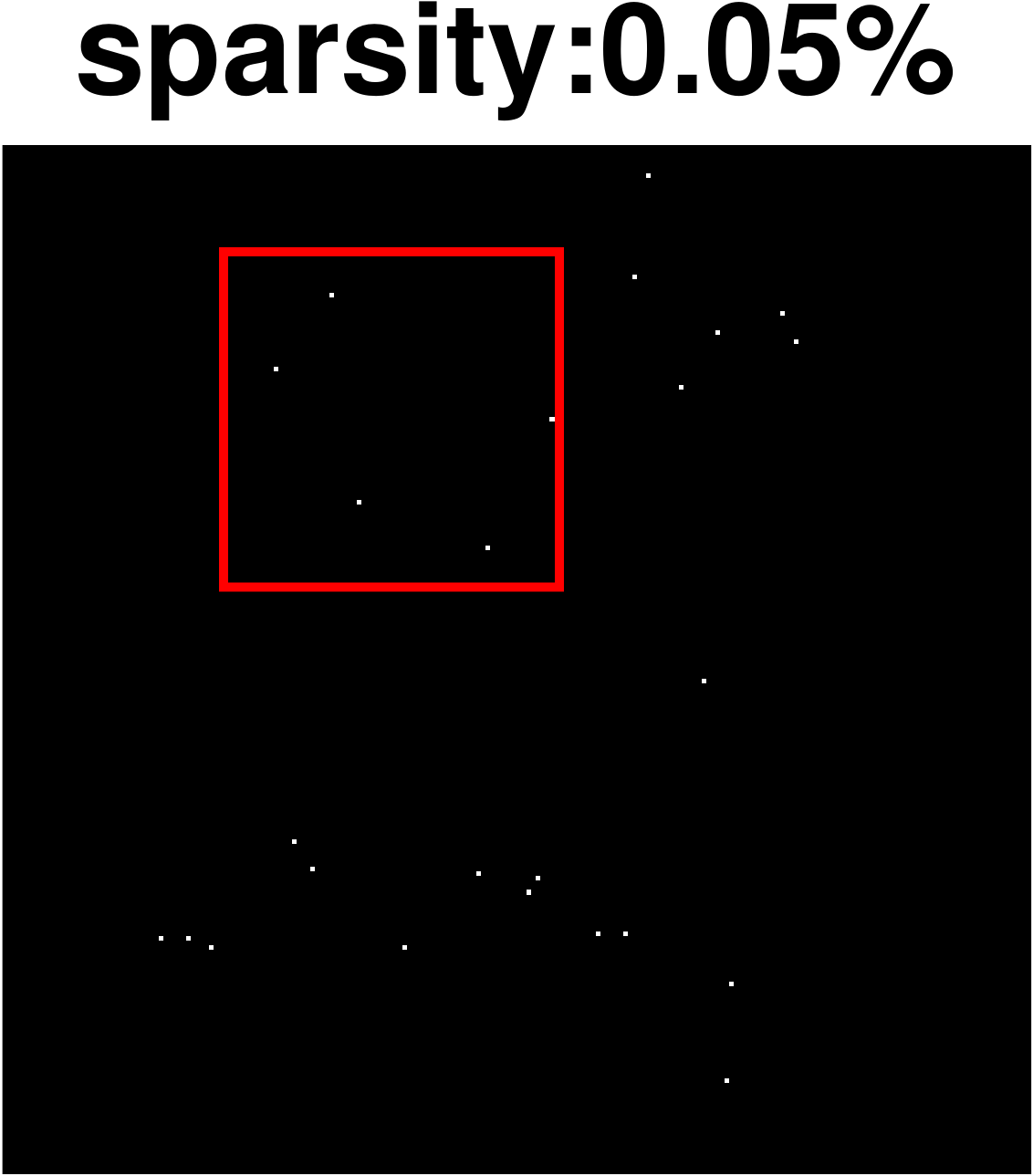}&
		
		\includegraphics[width=0.2\columnwidth]{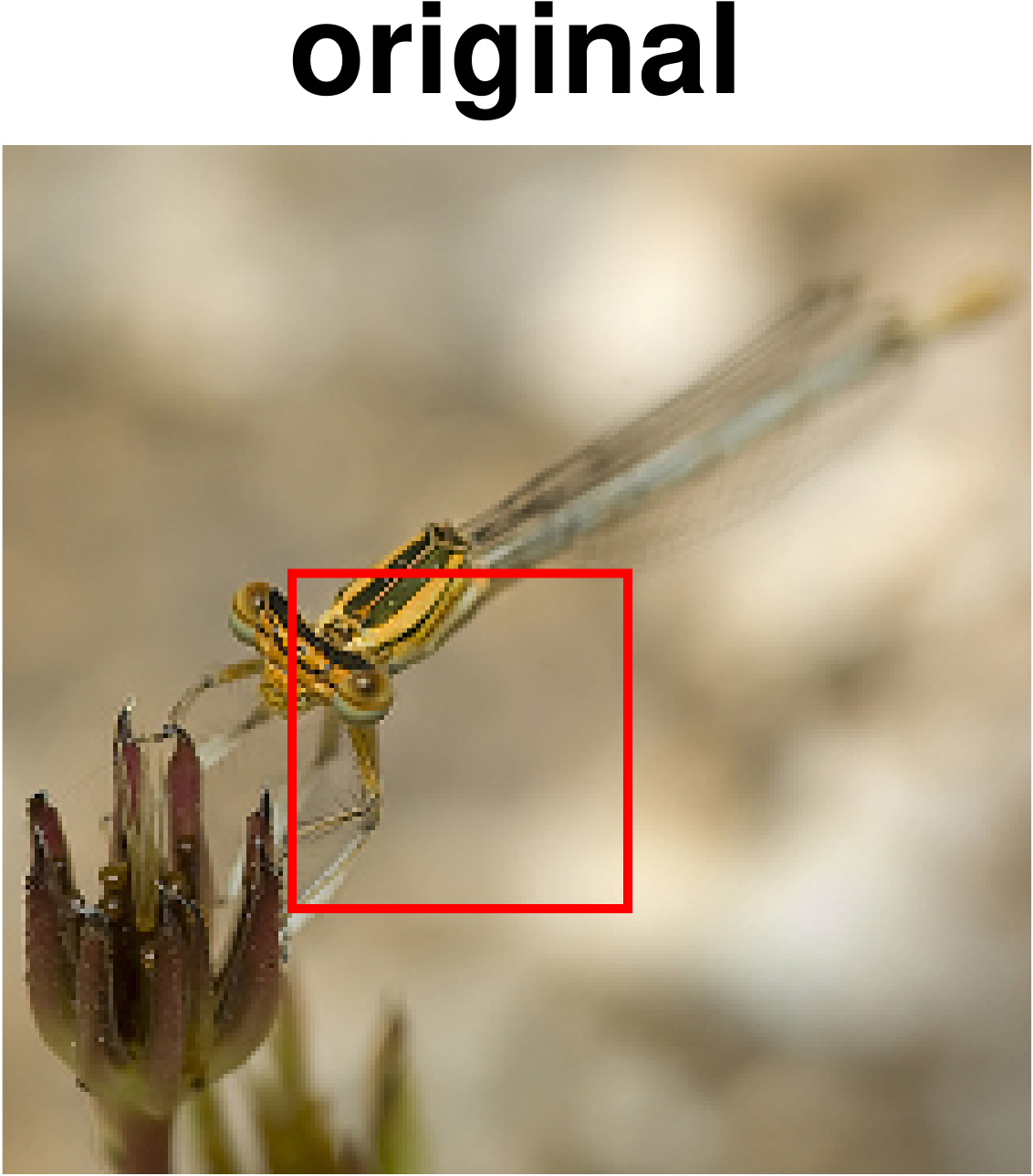}&
		\includegraphics[width=0.2\columnwidth]{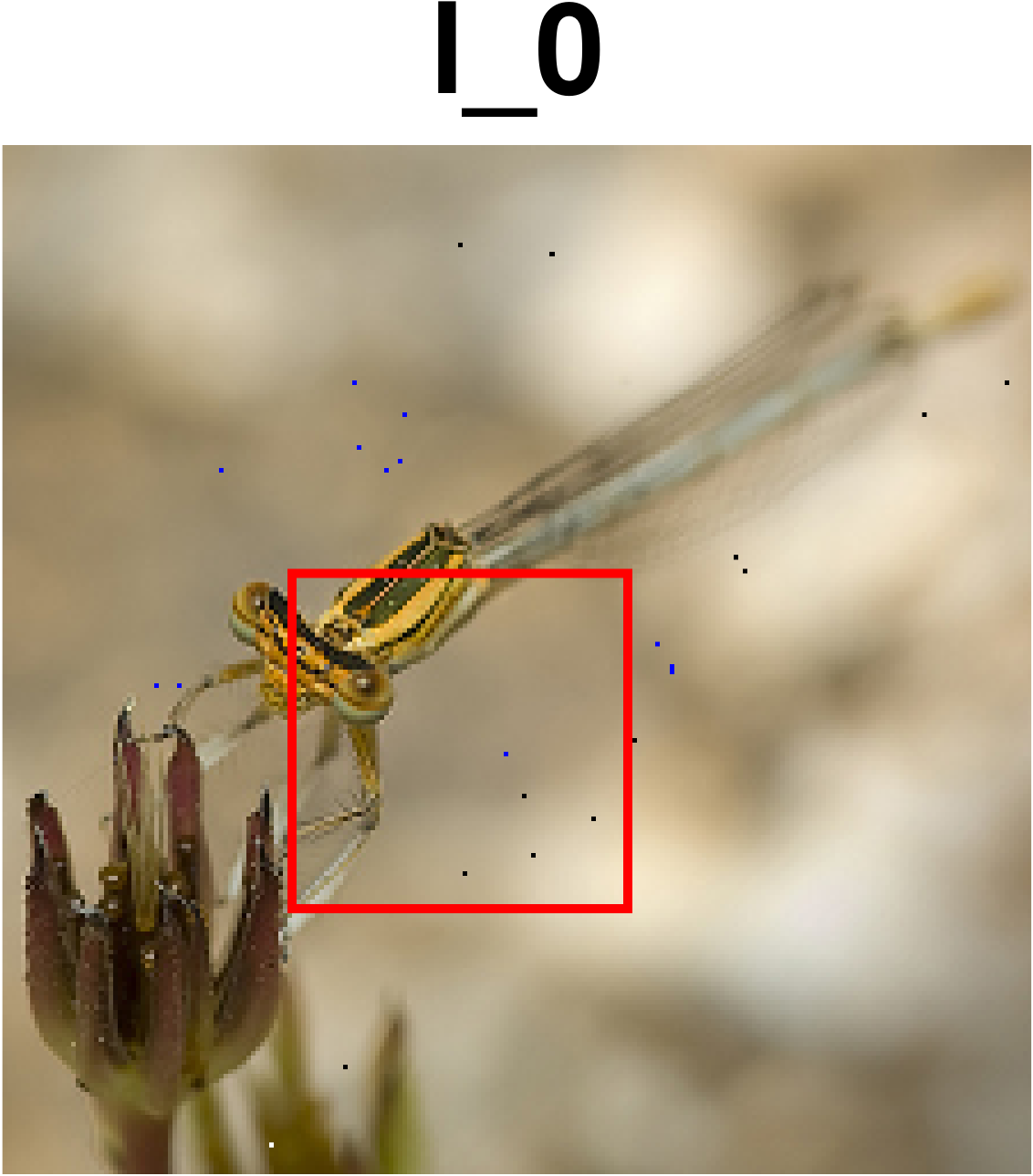}&
		\includegraphics[width=0.2\columnwidth]{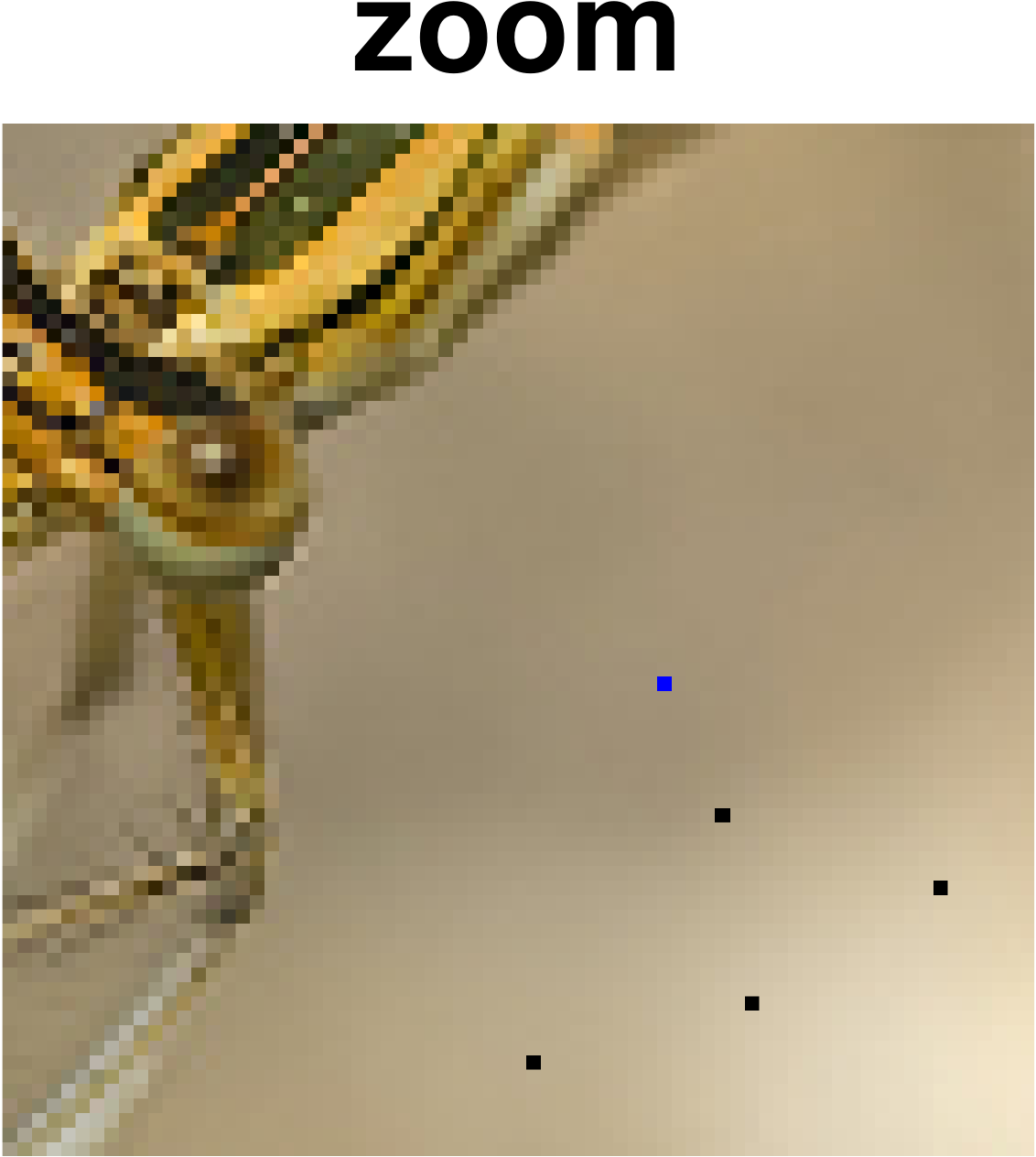}&
		\includegraphics[width=0.2\columnwidth]{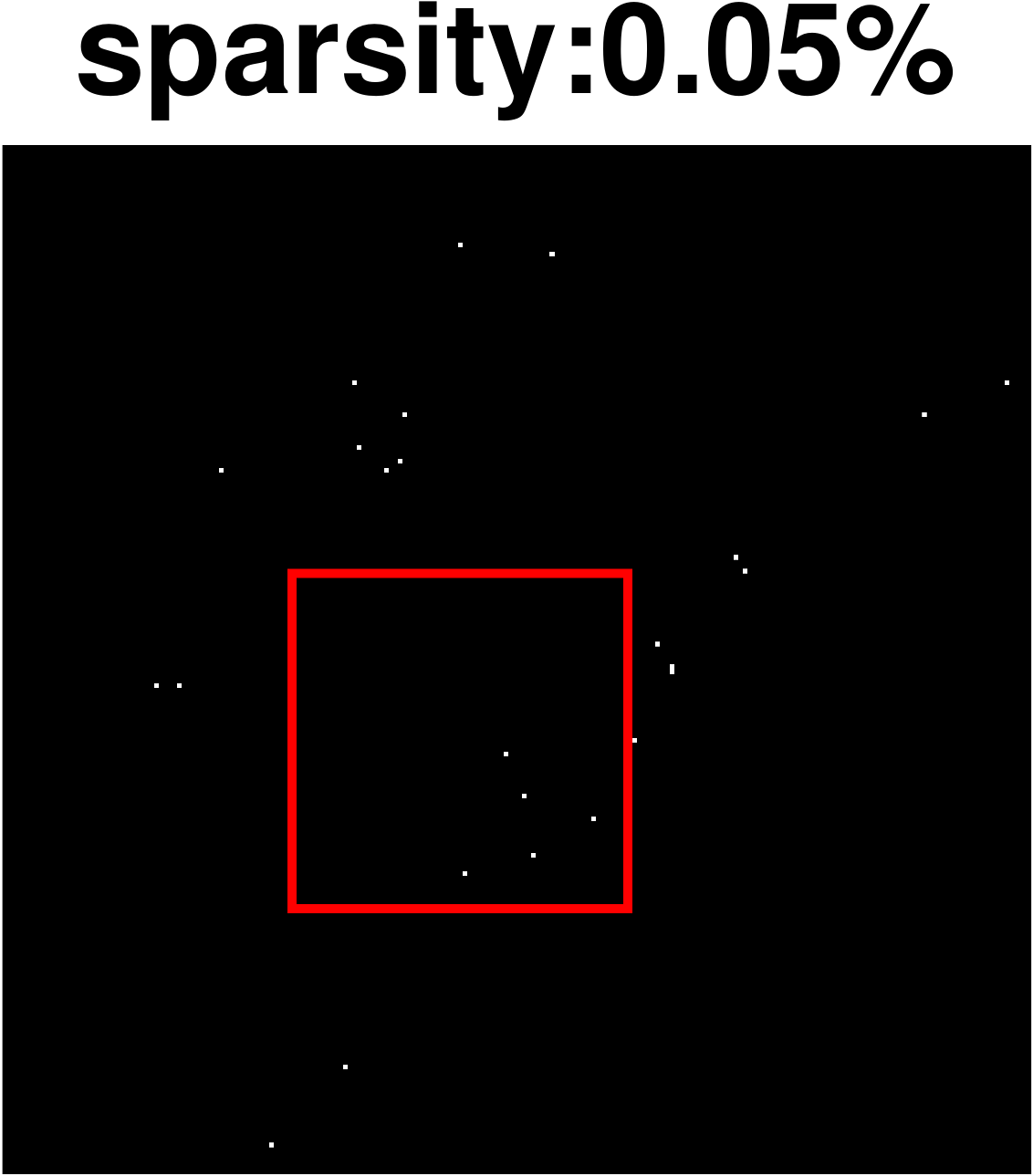}\\
		
		& \includegraphics[width=0.2\columnwidth]{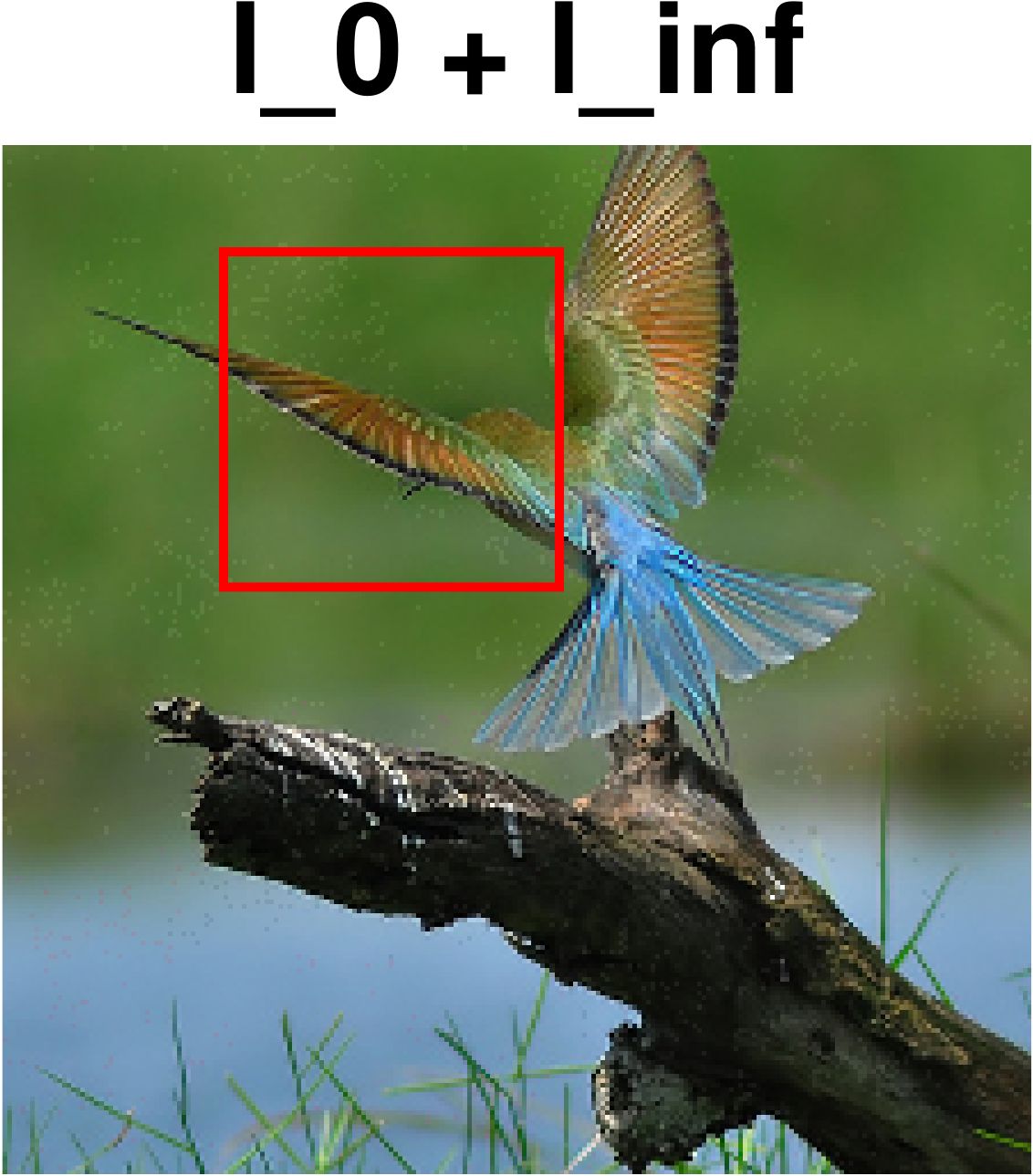}&
		\includegraphics[width=0.2\columnwidth]{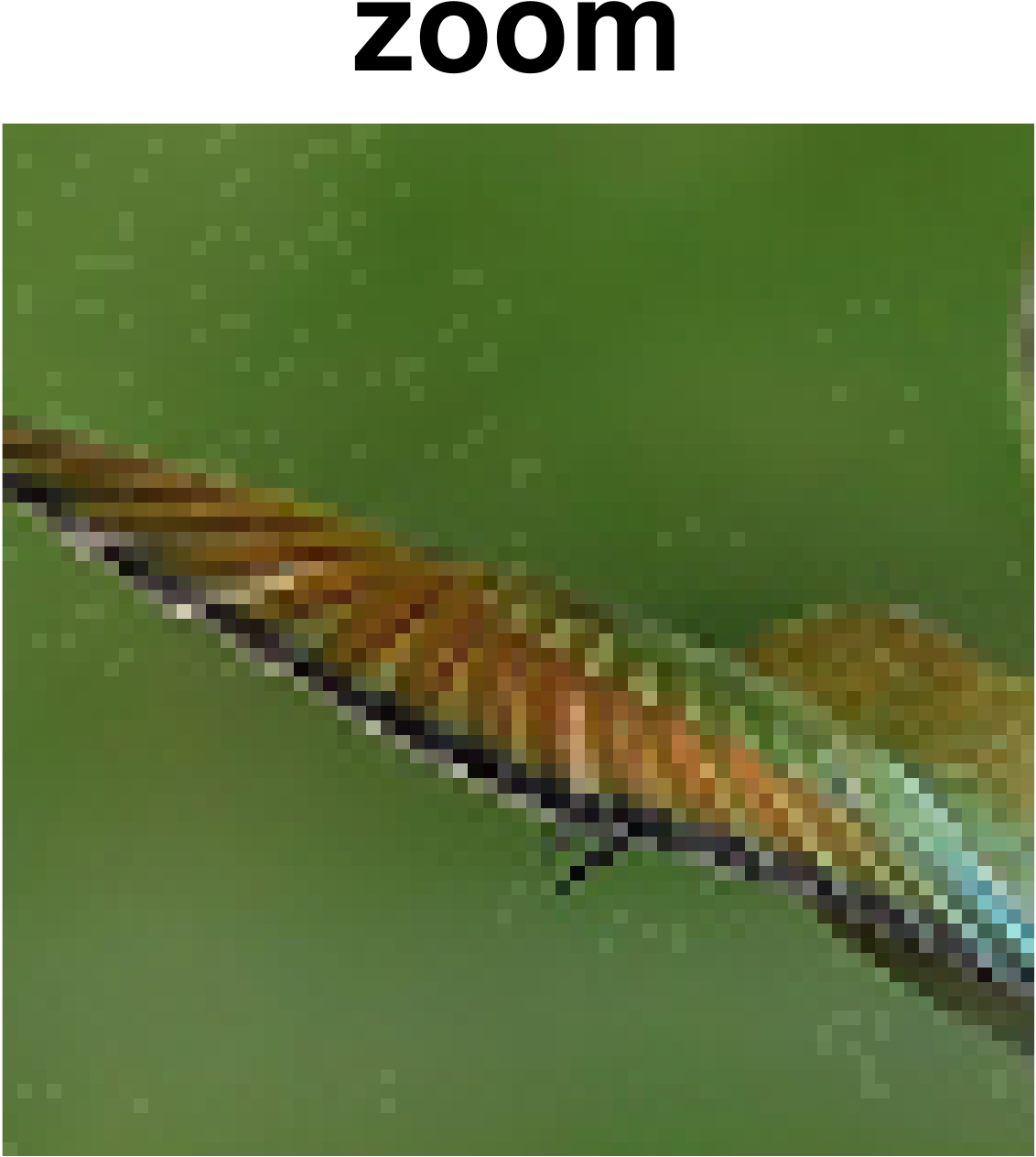}&
		\includegraphics[width=0.2\columnwidth]{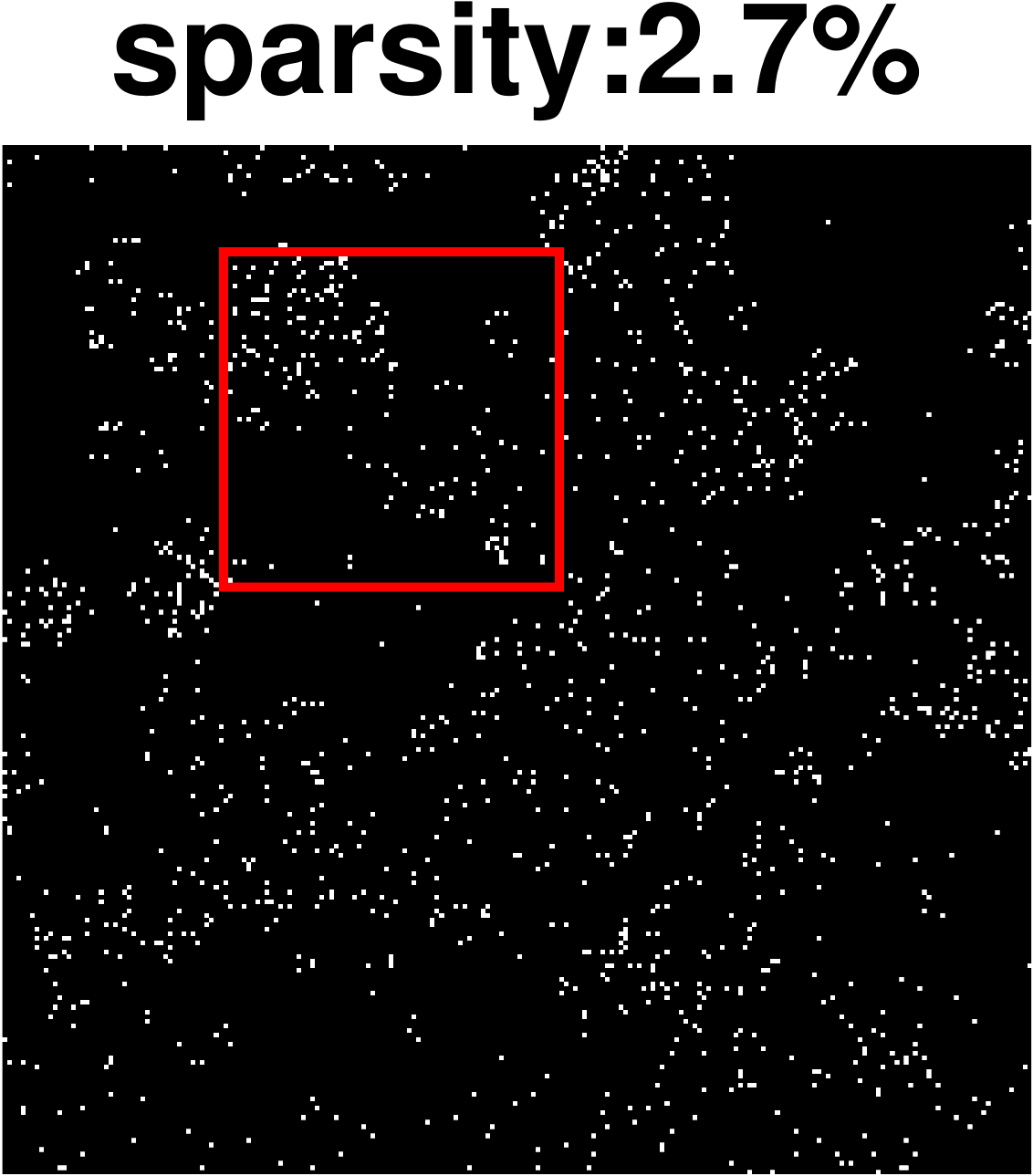}&
		&
		\includegraphics[width=0.2\columnwidth]{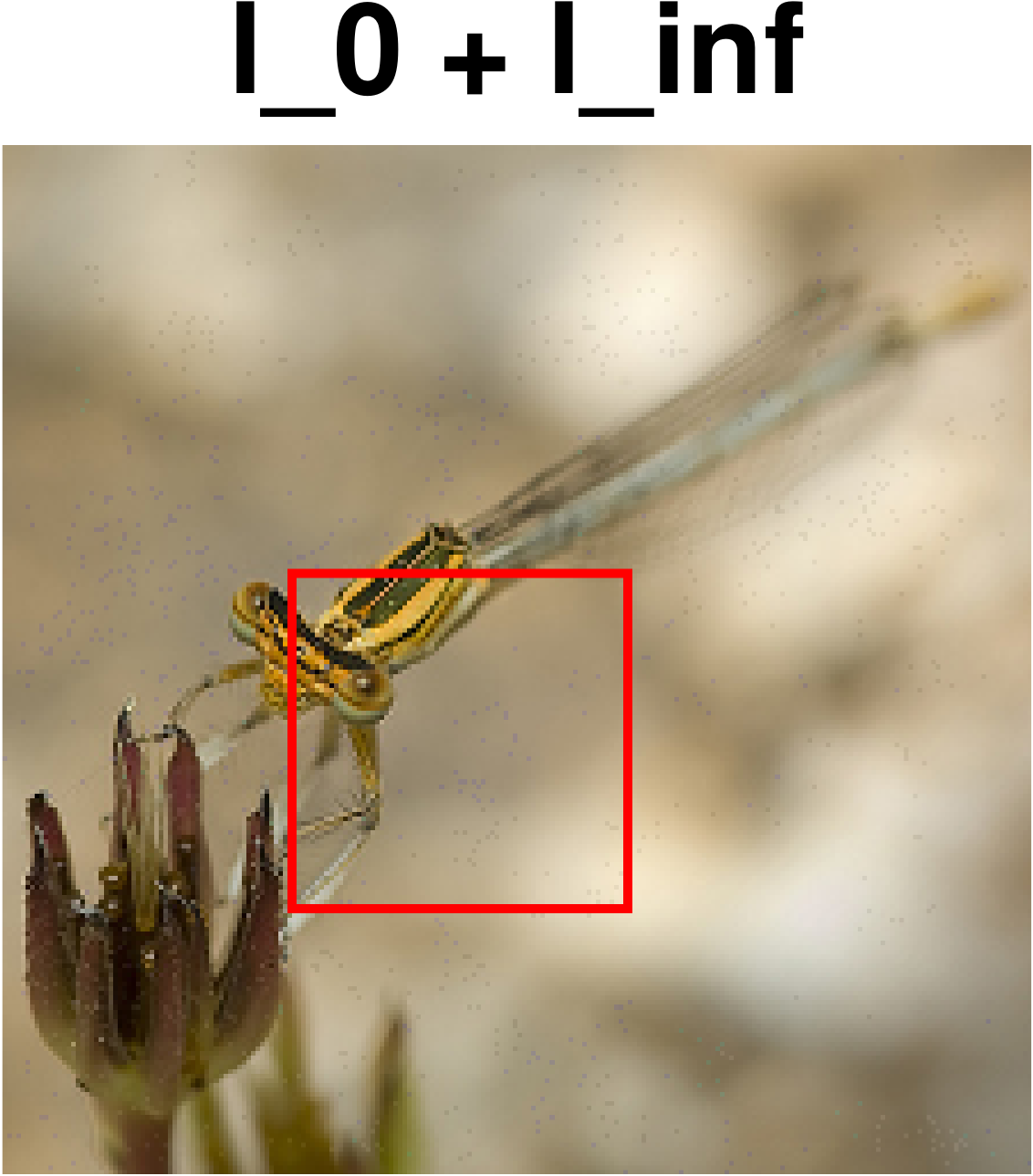}&
		\includegraphics[width=0.2\columnwidth]{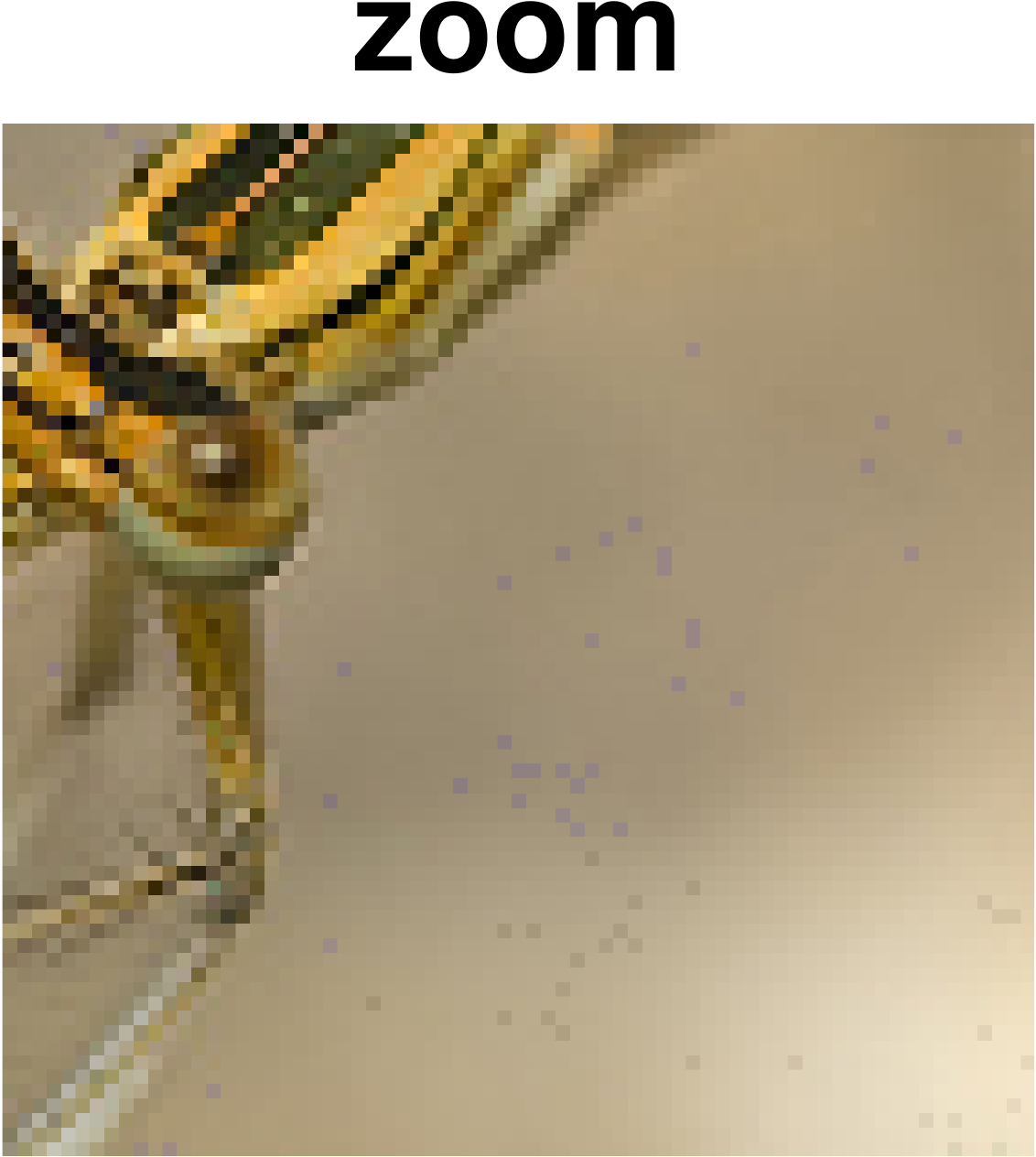}&
		\includegraphics[width=0.2\columnwidth]{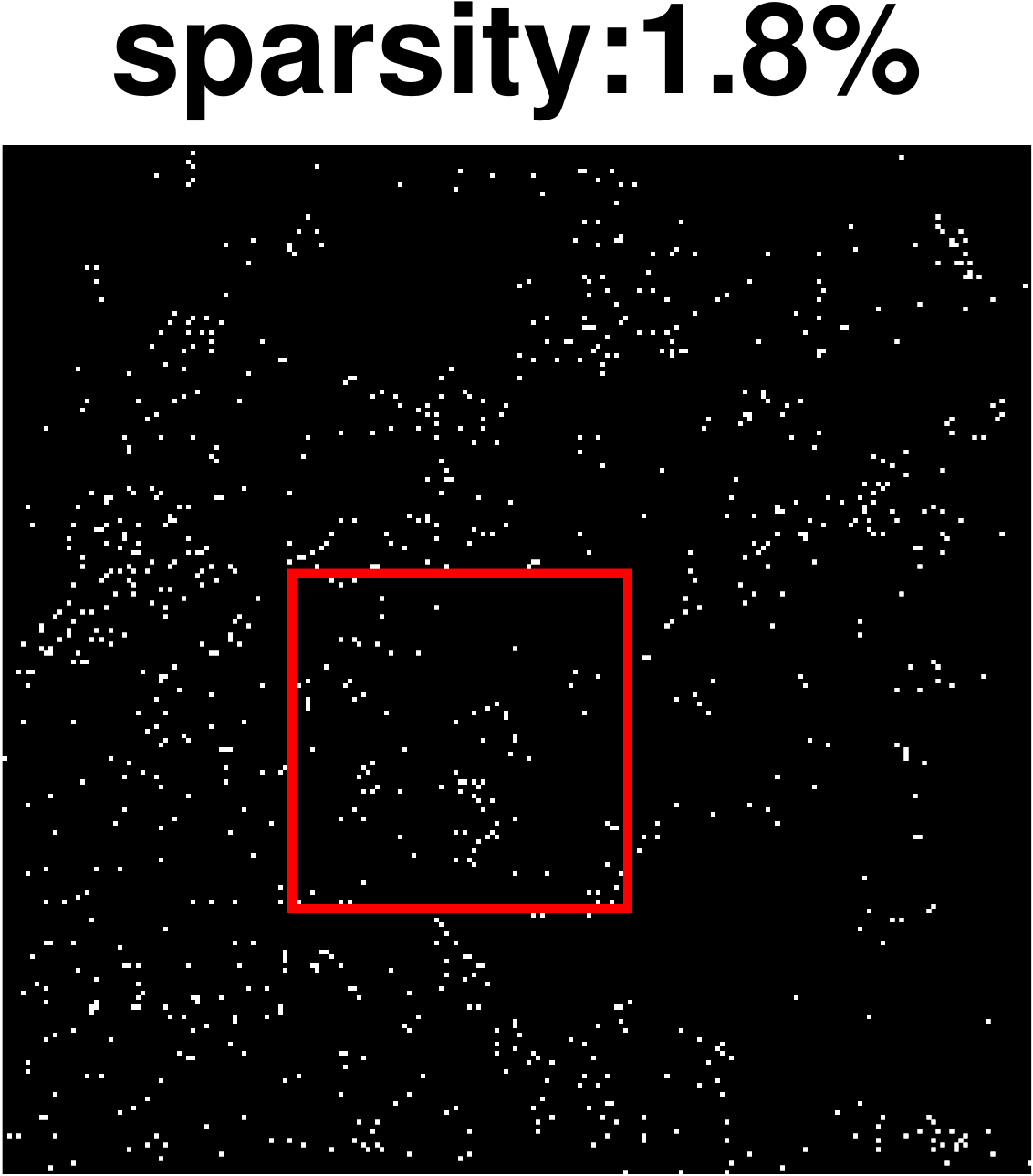}\\
		
		\includegraphics[width=0.2\columnwidth]{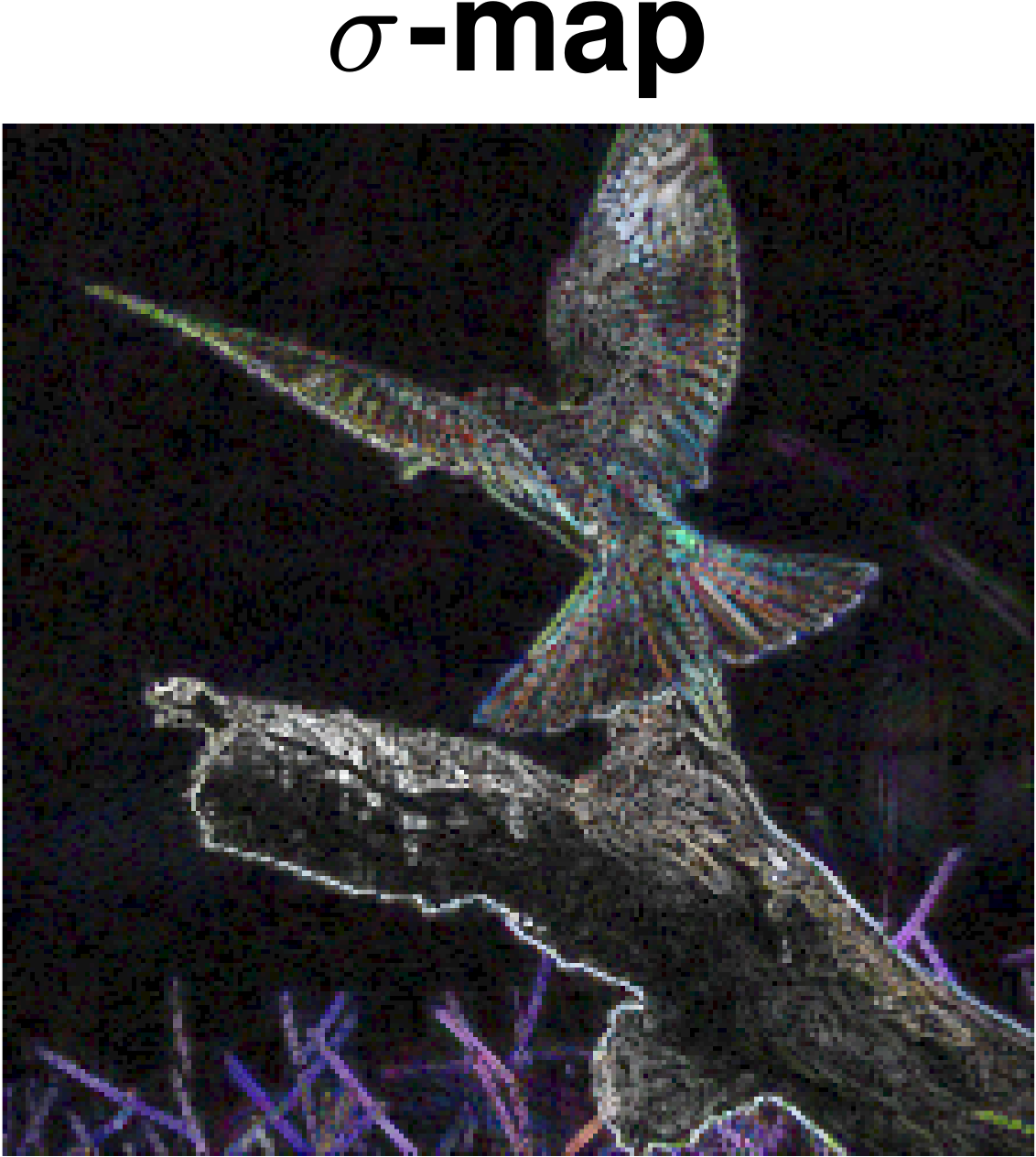}& 
		\includegraphics[width=0.2\columnwidth]{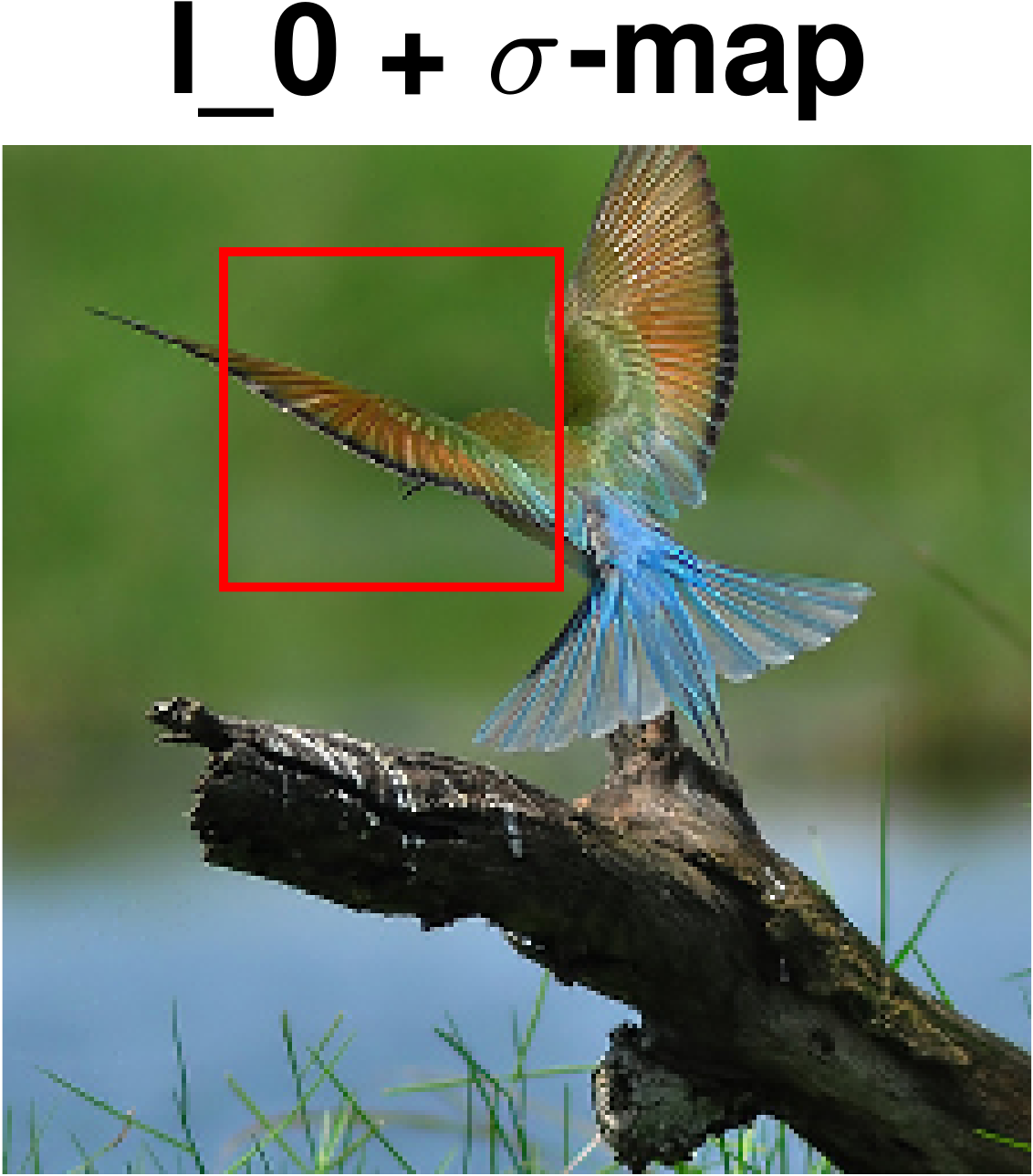}&
		\includegraphics[width=0.2\columnwidth]{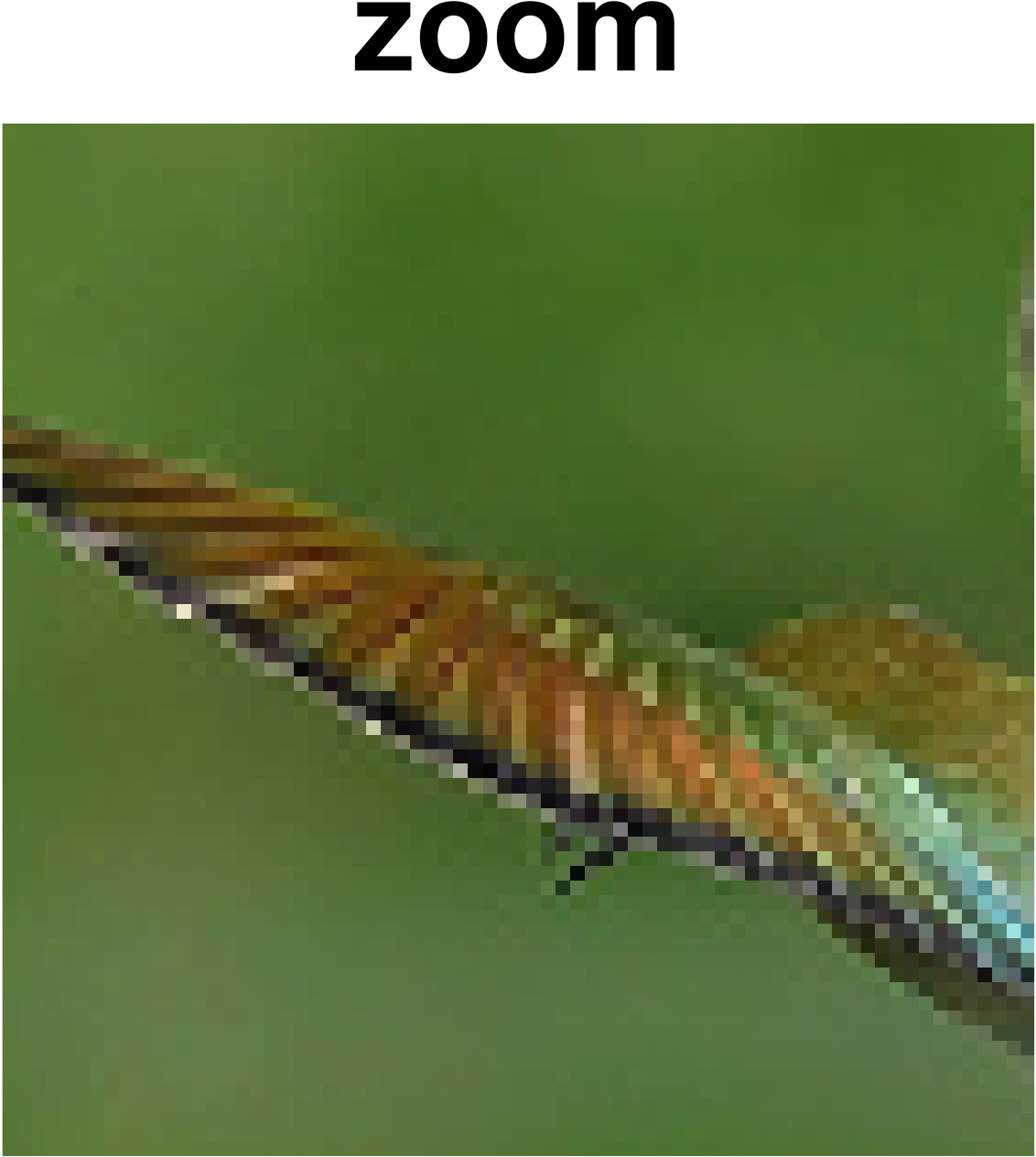}&
		\includegraphics[width=0.2\columnwidth]{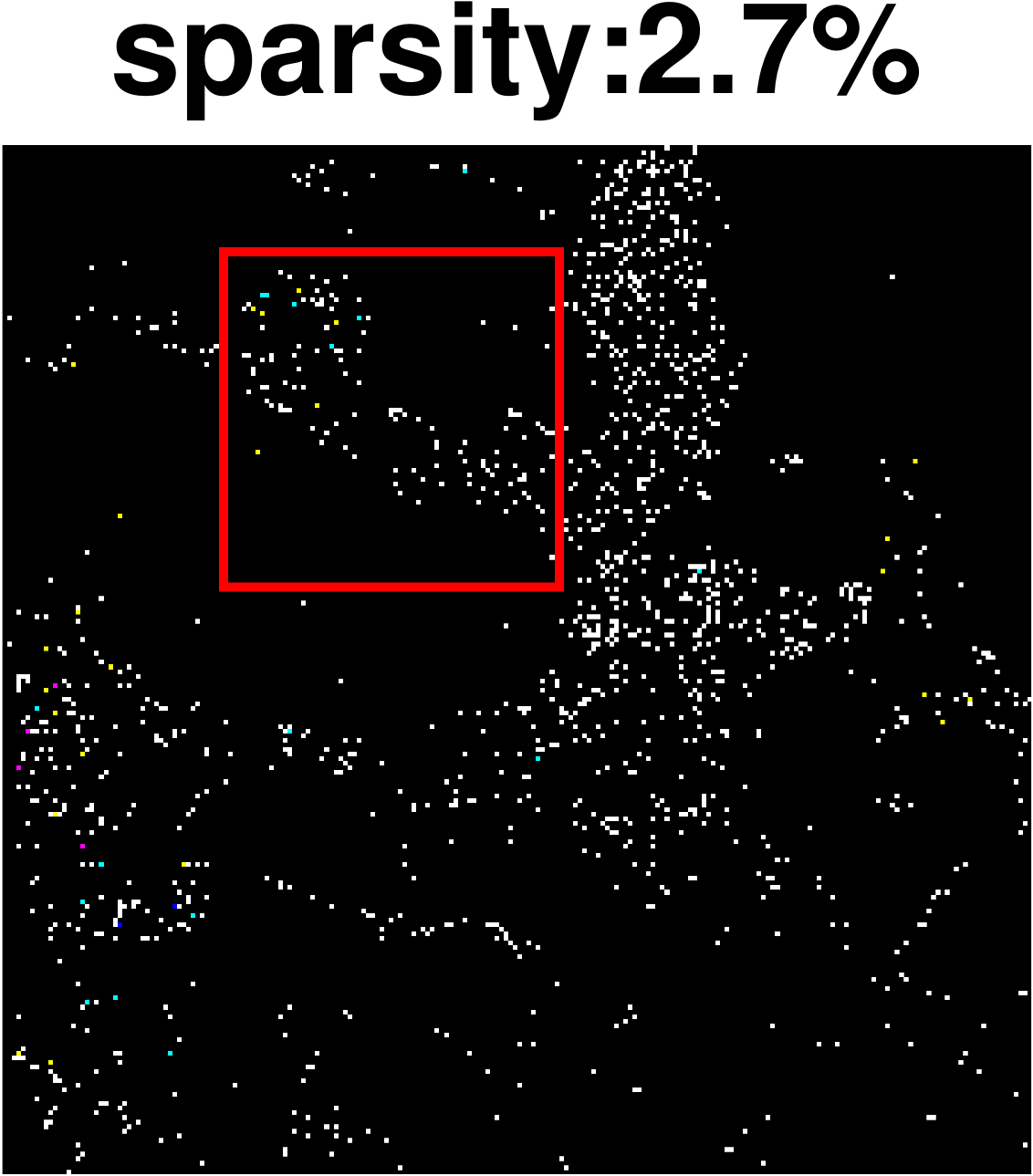}&
		
		\includegraphics[width=0.2\columnwidth]{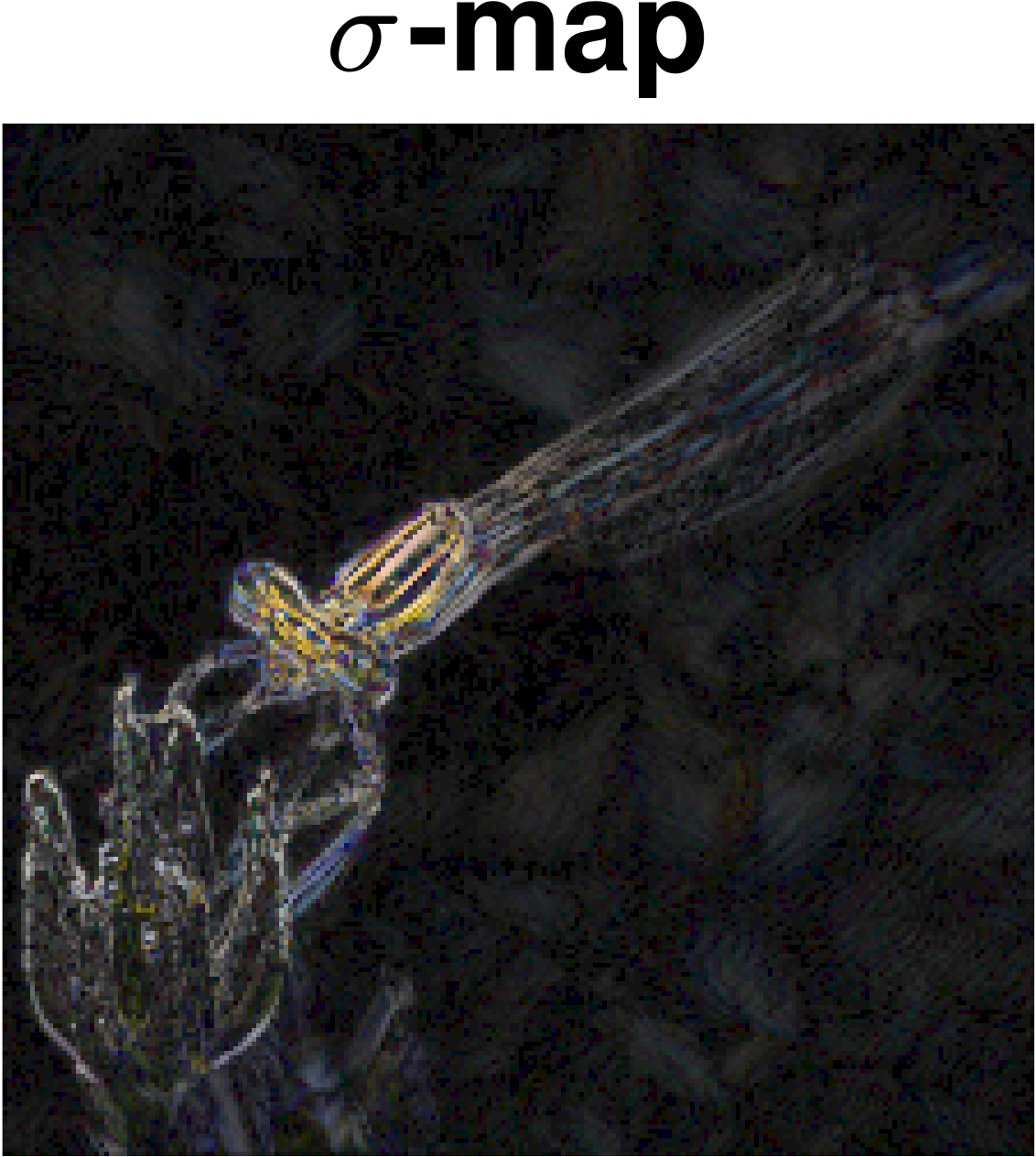}& 
		\includegraphics[width=0.2\columnwidth]{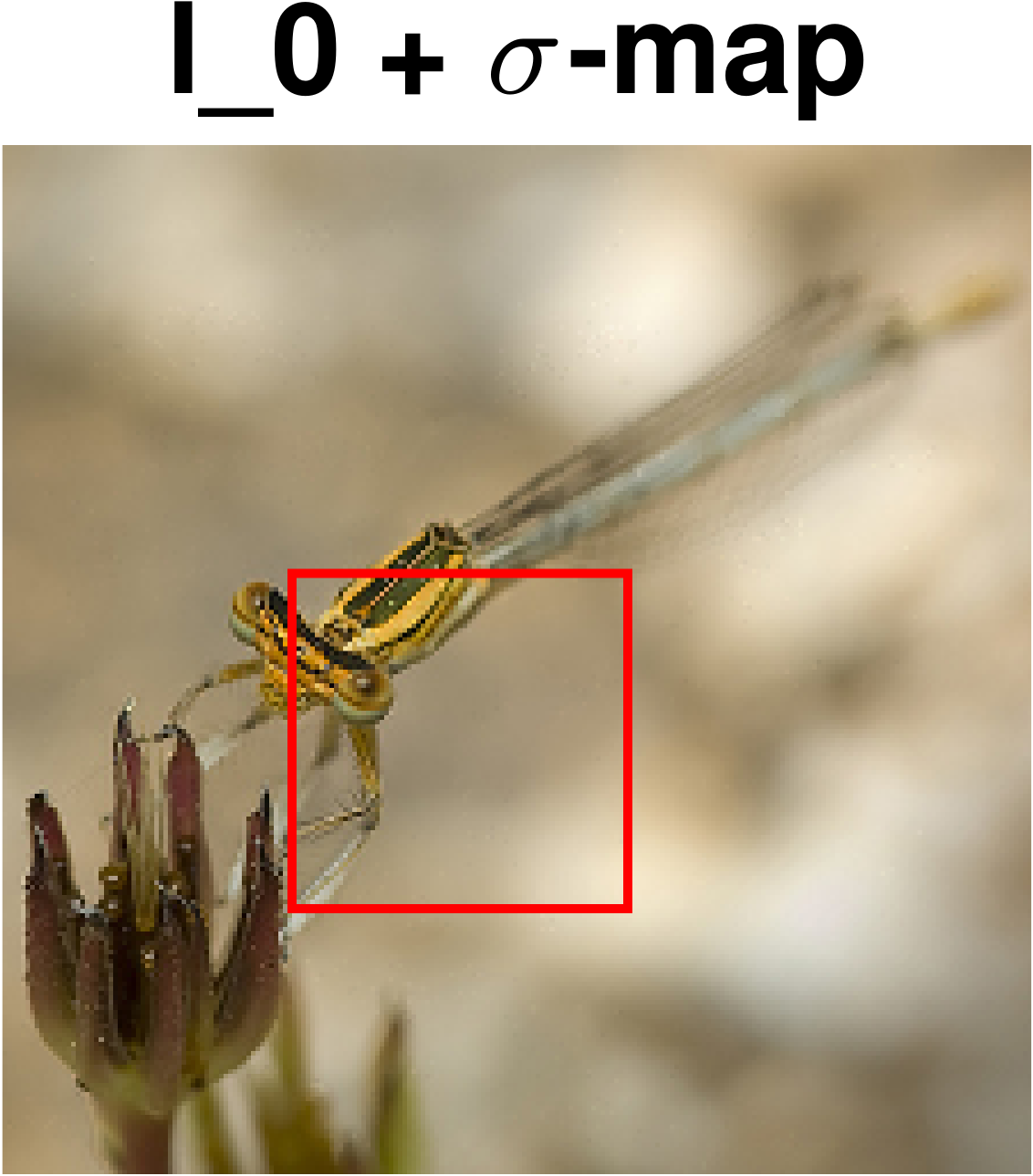}&
		\includegraphics[width=0.2\columnwidth]{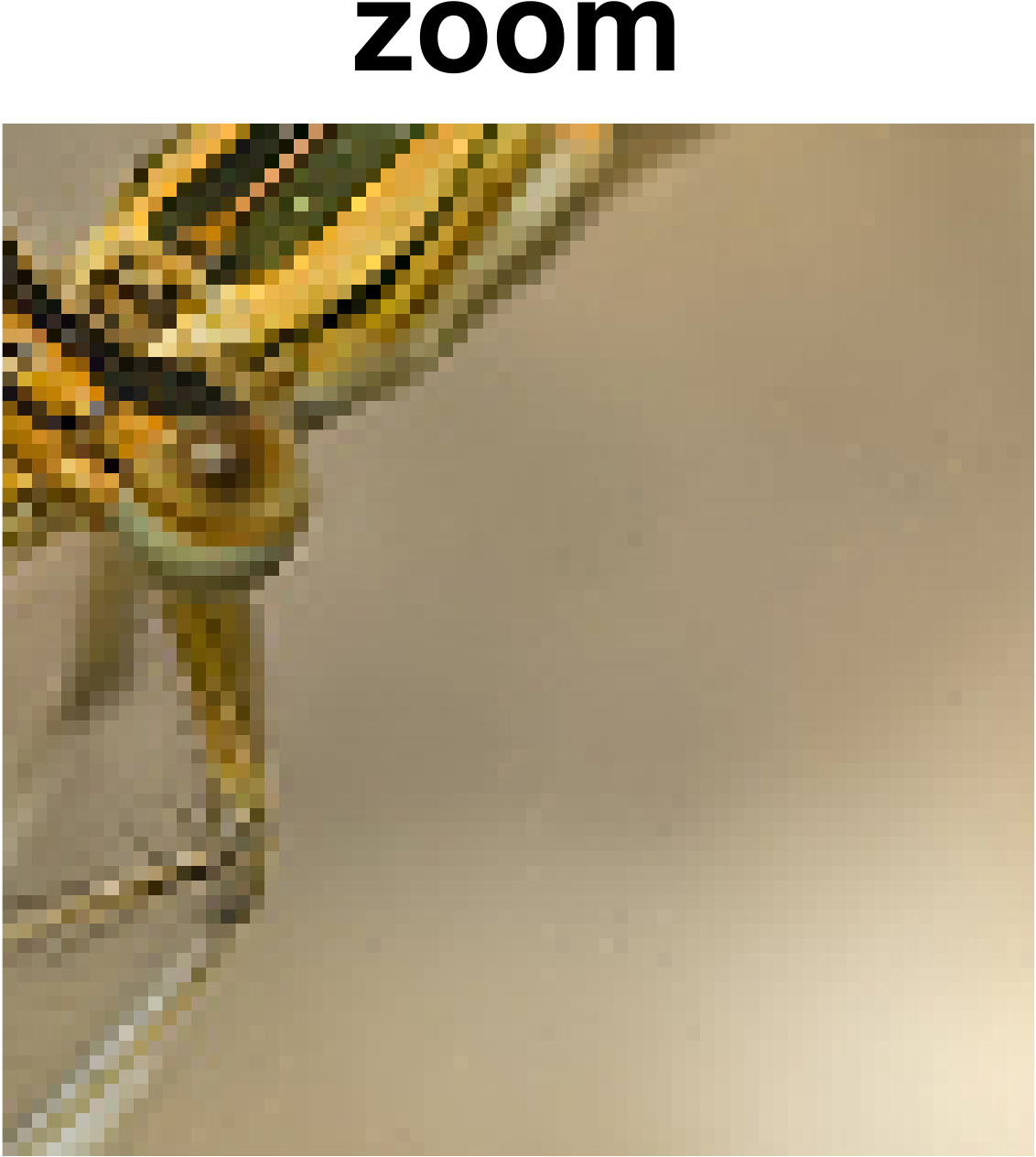}&
		\includegraphics[width=0.2\columnwidth]{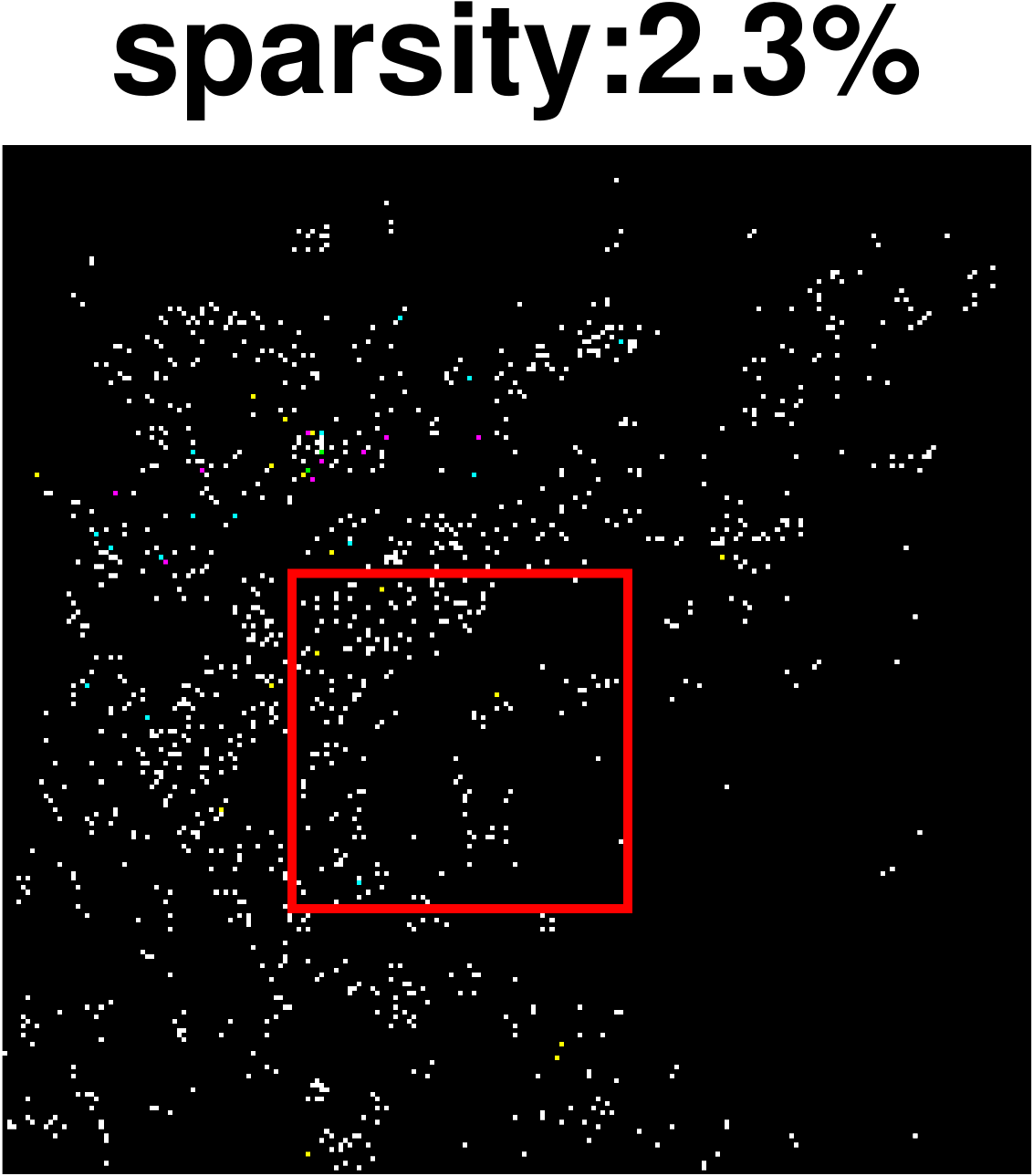} \\
	\end{tabular}
	\caption{\textbf{Different attacks on Restricted ImageNet.} We illustrate the differences of the adversarial examples (second column, zoom in third column) found by CornerSearch ($l_0$), $l_0+l_\infty$-attack and $\sigma$-CornerSearch, respectively first, second and third row. The fourth column shows the map of the modified pixels (\textit{sparsity} column). The original image is in the top left and the RGB representation of the rescaled $\sigma$-map in the bottom left.} \label{fig:imp_ImageNet}
\end{figure*}

Using the derivation in Section \ref{sec:pgd_l0} of the projection onto the $l_0$- resp. $l_0+l_\infty$-ball, we introduce an $l_0$ version of the well-known PGD attack on the cross-entropy function $L$, namely $\textrm{PGD}_0$. The iterative scheme, to be repeated for a fixed number of iterations, is, given an input $x$ assigned to class $c$, \begin{equation}\begin{split}& z^{(i)} = x^{(i-1)} + \eta \cdot\nicefrac{\nabla L(c, f(x^{(i-1)})}{\norm{\nabla L(c, f(x^{(i-1)})}_1}\\ & x^{(i)}= P_k(z^{(i)}), \end{split} \end{equation} where $\eta\in \R_+$, $x^{(0)}=x$, $P_k(z)$ represents the projection onto the $l_0$-ball, with the radius fixed at $k$, and the $l_\infty$-ball defined by the box constraint $x\in[0,1]^d$.
Note that $\textrm{PGD}_0$ needs $k$ to be specified and thus does not aim at the minimal modification to change the decision as in \eqref{eq:adv_opt}. In order to evaluate the robust accuracy, that is the accuracy of the classifier when the goal of the attacker is to change the decision of all correctly classified images using $k$-pixels modifications, one needs to evaluate $\textrm{PGD}_0$ for each value of $k$ separately, whereas all other attacks yield the robust accuracy for all levels of sparsity in one run.\\
%
%onversely, for the other methods a single run is sufficient for multiple $k_0$, since they directly try to solve Problem \eqref{eq:adv_opt}.\\
For comparison we run $\textrm{PGD}_0$, using 20 iterations and 10 random restarts, with 10 sparsity values $k$ on the networks of Table \ref{tab:comp_attacks} (see Appendix for more details). In Figure \ref{fig:pgd} we show the robust accuracy of the different attacks. $\textrm{PGD}_0$ achieves the best results on MNIST for $k\geq 4$, outperforms SparseFool and is even close to CornerSearch on CIFAR-10. As $\textrm{PGD}_0$ is very fast, it is a valuable alternative to our more expensive score-based attack.\\

We further test CornerSearch on Restricted ImageNet,
that is a subset of ImageNet \cite{DenEtAl09} where some of the classes are grouped to form 9 distinct macro-classes. We use the ResNet-50 from \cite{TsiEtAl18} and compare our attack to SparseFool \cite{ModEtAl19} (we do not run the other methods as either no code is available or they do not scale to the size of the images). The images have 50176 pixels.\\
In Table \ref{tab:comp_attacks_IN} we report the statistics on 100 points for SparseFool and our attack with $k_\textrm{max}=1000, N_\textrm{iter}=1000$. As for the other datasets, SparseFool always finds an adversarial example, whereas the smallest \textit{mean} and \textit{median} adversarial modification is achieved by CornerSearch, although with an inferior success rate. The runtime for SparseFool %(recall that we are use our own reimplementation) 
is around 55 times smaller than for CornerSearch. The runtime of our attack directly scales with the number of pixels and the time of a forward pass of the network, both large in this case. However, please note that SparseFool is a white-box attack, whereas ours is a black-box attack. For a comparison to $\textrm{PGD}_0$, given 100 pixels of budget, SF achieves a success rate of 49.4\%, CS 64.0\%, $\textrm{PGD}_0$ 39.3\%.

\subsection{Sparse and Imperceivable manipulations}\label{sec:l0_imperceivable}
We illustrate the differences of the adversarial modifications found by $l_0$-, $l_0+l_\infty$- and $l_0+\sigma$-map attacks. In Figures \ref{fig:imp_CIFAR-10} and \ref{fig:imp_ImageNet} we show some examples. As discussed before the adversarial modifications produced wrt only the $l_0$-norm are the sparsest but also the easiest to recognize. The $l_0+l_\infty$-attack provides images where, although the absolute value of the individual modification is bounded (we use here $\delta=0.1$ for CIFAR-10, $\delta=0.05$ for ImageNet), some perturbations are visible since either colors are not homogeneous with the neighbors (second rows of the left part of Figure \ref{fig:imp_CIFAR-10} and right part of Figure \ref{fig:imp_ImageNet}) or modifications of an uniform background are introduced (second row of the right part of Figure \ref{fig:imp_CIFAR-10} and left part of Figure \ref{fig:imp_ImageNet}). On the other hand, the adversarial modifications of $\sigma$-CornerSearch are imperceivable while still being very sparse (third rows of Figures \ref{fig:imp_CIFAR-10} and \ref{fig:imp_ImageNet}), showing that the $\sigma$-map, also shown in the Figures rescaled so that the largest component is equal to 1, is able to correctly identify the area where a change is difficult to perceive (see in particular the zoomed images).\\
We provide a comparison of the adversarial examples crafted by $\sigma$-CornerSearch and $\sigma$-PGD in the Appendix.

\subsection{Adversarial training}\label{sec:adv_training}
\begin{table}[t]
	{%\footnotesize
		\centering 
		\begin{tabular}{L{15mm} | C{6mm} C{6mm} C{6mm} C{6mm} |C{6mm} C{6mm}}
			%&\multicolumn{4}{c}{MNIST}\\
			& \multicolumn{4}{c|}{$l_0$} &\multicolumn{2}{c}{$l_0+\sigma$}\\
			\cline{2-7}
			\textit{training} & PA & SF & $\textrm{PGD}_0$ & CS & $\sigma$PGD& $\sigma$CS \\
			%\multicolumn{9}{c}{}\\
			\hline
			
			\multicolumn{7}{c}{}\\
			MNIST &\multicolumn{4}{c|}{$k=15$}&\multicolumn{2}{c}{$k=50$} \\
			\hline
			\textit{plain} &25.6&41.2& \textbf{3.6}&9.2&49.2&80.4\\
			\textit{$l_\infty$-at} &1.6&96.0&60.0& \textbf{0.0}&88.2&  90.0\\
			\textit{$l_2$-at} &  73.0&85.0& \textbf{34.0}&55.4&80.2&89.2\\
			\textit{$l_0$-at} &55.8&63.6&74.6& \textbf{39.8}&57.8&73.8\\
			%\textit{$l_0 + l_\infty$-at} &   \\
			\textit{$l_0 + \sigma$-at}&19.4&26.8&\textbf{9.4}& 15.4&93.6&95.4\\
			\hline
			%&\multicolumn{4}{c}{CIFAR-10}\\
			%\multicolumn{7}{c}{}\\
			%CIFAR-10 &\multicolumn{4}{c|}{$k=5$}&\multicolumn{2}{c}{$k=100$} \\
			%\hline
			%\textit{plain}&31.8&75.6&26.4& \textbf{11.8}\\
			%\textit{$l_\infty$-at} &41.0&68.4&36.4& \textbf{24.6}\\
			%\textit{$l_2$-at} &50.0&69.6&41.0& \textbf{32.4}\\
			%\textit{$l_0$-at} &69.2&73.4&70.2& \textbf{61.6}\\
			%\textit{$l_0 + l_\infty$-at} &   \\
			%\hline
			
			\multicolumn{7}{c}{}\\
			CIFAR-10 &\multicolumn{4}{c|}{$k=10$}&\multicolumn{2}{c}{$k=100$}\\
			\hline
			\textit{plain}&15.2&57.0&7.0& \textbf{2.2}& 27.6&52.4\\
			\textit{$l_\infty$-at}&28.6&57.6&22.6& \textbf{10.8}&50.4&63.6\\
			\textit{$l_2$-at} &37.6&60.6&25.4& \textbf{13.8}&53.2&66.0\\
			\textit{$l_0$-at} &64.2&63.8&54.8& \textbf{46.0}&34.6&58.2\\
			%\textit{$l_0 + l_\infty$-at} &   \\              
			\textit{$l_0 + \sigma$-at}&41.6&54.4&6.0&\textbf{5.6} &62.8&67.6\\
			\hline
		\end{tabular}
		\caption{\textbf{Evaluation of adversarial training.} Robust accuracy (\%) given by $l_0$- and $l_0+\sigma$-attacks (changing at most $k$ pixels and for $l_0+\sigma$-attacks fixing $\kappa=0.8$ for MNIST and $\kappa=0.4$ for CIFAR-10) on models adversarially trained wrt different metrics.}
		\label{tab:comp_at_3}}
	
\end{table}

In order to increase robustness of the models to sparse adversarial manipulations, we adapt adversarial training to our cases. We use $\textrm{PGD}_0$ presented above for adversarial training in order to achieve robustness against $l_0$-attacks (\textit{$l_0$-at}), while we use $\sigma$-PGD to enhance robustness against sparse and imperceivable attacks (\textit{$l_0 + \sigma$-at}). With these two techniques we train models on MNIST and CIFAR-10 (more details about the architectures and hyerparameters in the Appendix). We compare them to the models trained on the \textit{plain} training set and with adversarial training wrt the $l_\infty$- and $l_2$-norm (\textit{$l_\infty$-at} and \textit{$l_\infty$-at}). In Table \ref{tab:comp_at_3} we report the robust accuracy
%(that is the percentage of points for which the attack cannot find an adversarial example)
on 500 points (we fix the maximum number of pixels to be modified to $k$, and the parameter of the $l_0+\sigma$ attacks defined in \eqref{eq:sigma_constr} and \eqref{eq:sigma_constr_gray} to $\kappa=0.8$ for MNIST and $\kappa=0.4$ for CIFAR-10).

On MNIST the models trained on $l_2$ and $l_0$ perturbations are the most robust against $l_0$-attacks, while on CIFAR-10 the \textit{$l_0$-at} model is more than 3 times more resistant than all the others. Similarly to \cite{ModEtAl19} we find that \textit{$l_\infty$-at} does not help for $l_0$-robustness. Notably, on both dataset our attacks $\textrm{PGD}_0$ and CornerSearch (CS) achieve the best results and then are the most suitable to evaluate robustness.\\
Regarding the $l_0+\sigma$-map attacks, we see that the \textit{$l_0+\sigma$-at} models are the least vulnerable, but also \textit{$l_\infty$-at} and \textit{$l_2$-at} show some robustness. Note that $\sigma$-PGD is more successful than $\sigma$-CornerSearch but produces less sparse perturbations as it always fully exploits the budget of $k$ pixels to modify while $\sigma$-CS mostly uses just a few of them, making the modifications even more difficult to spot (see Appendix).

{\small
\section*{Acknowledgements}
Supported by the DFG Cluster of Excellence “Machine Learning – New Perspectives for Science”, EXC 2064/1, project number 390727645
and
funded by DFG grant 389792660 as part of \href{https://perspicuous-computing.science}{TRR~248}.
\bibliographystyle{ieee_fullname}

\begin{thebibliography}{10}\itemsep=-1pt

\bibitem{AthEtAl2018}
Anish Athalye, Nicholas Carlini, and David Wagner.
\newblock Obfuscated gradients give a false sense of security: Circumventing
  defenses to adversarial examples.
\newblock {\em arXiv preprint arXiv:1802.00420}, 2018.

\bibitem{BhaEtAl18}
Arjun~Nitin Bhagoji, Warren He, Bo Li, and Dawn~Xiaodong Song.
\newblock Practical black-box attacks on deep neural networks using efficient
  query mechanisms.
\newblock In {\em ECCV}, 2018.

\bibitem{BigEtAl13}
Battista Biggio, Igino Corona, Davide Maiorca, Blaine Nelson, Nedim Srndic,
  Pavel Laskov, Giorgio Giacinto, and Fabio Roli.
\newblock Evasion attacks against machine learning at test time.
\newblock In {\em ECML PKDD}, 2013.

\bibitem{BreRauBet18}
Wieland Brendel, Jonas Rauber, and Matthias Bethge.
\newblock Decision-based adversarial attacks: Reliable attacks against
  black-box machine learning models.
\newblock In {\em ICLR}, 2018.

\bibitem{CarWag2017}
Nicholas Carlini and David~A. Wagner.
\newblock Adversarial examples are not easily detected: Bypassing ten detection
  methods.
\newblock In {\em ACM Workshop on Artificial Intelligence and Security}, 2017.

\bibitem{CarWag2016}
Nicholas Carlini and David~A. Wagner.
\newblock Towards evaluating the robustness of neural networks.
\newblock In {\em IEEE Symposium on Security and Privacy}, 2017.

\bibitem{CheEtAl17}
Pin-Yu {Chen}, Huan {Zhang}, Yash {Sharma}, Jinfeng {Yi}, and Cho-Jui {Hsieh}.
\newblock Zoo: Zeroth order optimization based black-box attacks to deep neural
  networks without training substitute models.
\newblock In {\em 10th ACM Workshop on Artificial Intelligence and Security
  (AISEC)}, 2017.

\bibitem{CroHei18}
Francesco Croce and Matthias Hein.
\newblock A randomized gradient-free attack on {ReLU} networks.
\newblock In {\em GCPR}, 2018.

\bibitem{DalEtAl2004}
Nilesh~N. Dalvi, Pedro~M. Domingos, Mausam, Sumit~K. Sanghai, and Deepak Verma.
\newblock Adversarial classification.
\newblock In {\em KDD}, 2004.

\bibitem{DenEtAl09}
Jia Deng, Wei Dong, Richard Socher, Li-Jia Li, Kehui Li, and Li Fei-Fei.
\newblock Imagenet: A large-scale hierarchical image database.
\newblock In {\em CVPR}, 2009.

\bibitem{EykEtAl2018}
Ivan Evtimov, Kevin Eykholt, Earlence Fernandes, Tadayoshi Kohno, Bo Li, Atul
  Prakash, Amir Rahmati, and Dawn Song.
\newblock Robust physical-world attacks on deep learning visual classification.
\newblock In {\em CVPR}, 2018.

\bibitem{GooShlSze2015}
Ian~J. {Goodfellow}, Jonathon {Shlens}, and Christian {Szegedy}.
\newblock Explaining and harnessing adversarial examples.
\newblock In {\em ICLR}, 2015.

\bibitem{IlyEtAl18}
Andrew Ilyas, Logan Engstrom, Anish Athalye, and Jessy Lin.
\newblock Black-box adversarial attacks with limited queries and information.
\newblock In {\em ICML}, 2018.

\bibitem{KanMooFro18}
Can Kanbak, Seyed{-}Mohsen Moosavi{-}Dezfooli, and Pascal Frossard.
\newblock Geometric robustness of deep networks: analysis and improvement.
\newblock In {\em CVPR}, 2018.

\bibitem{KinEtAl2014}
Diederik~P. Kingma and Jimmy Ba.
\newblock Adam: A method for stochastic optimization.
\newblock preprint, arXiv:1412.6980, 2014.

\bibitem{KurGooBen2016a}
Alexey Kurakin, Ian~J. Goodfellow, and Samy Bengio.
\newblock Adversarial examples in the physical world.
\newblock In {\em ICLR Workshop}, 2017.

\bibitem{LinEtAl14}
Min Lin, Qiang Chen, and Shuicheng Yan.
\newblock Network in network.
\newblock In {\em ICLR}, 2014.

\bibitem{LiuEtAl2016}
Y. Liu, X. Chen, C. Liu, and D. Song.
\newblock Delving into transferable adversarial examples and black-box attacks.
\newblock In {\em ICLR}, 2017.

\bibitem{LowMee2005}
Daniel Lowd and Christopher Meek.
\newblock Adversarial learning.
\newblock In {\em KDD}, 2005.

\bibitem{LuoEtAl18}
Bo Luo, Yannan Liu, Lingxiao Wei, and Qiang Xu.
\newblock Towards imperceptible and robust adversarial example attacks against
  neural networks.
\newblock In {\em AAAI}, 2018.

\bibitem{MadEtAl2018}
Aleksander {Madry}, Aleksandar {Makelov}, Ludwig {Schmidt}, Dimitris {Tsipras},
  and Adrian {Vladu}.
\newblock Towards deep learning models resistant to adversarial attacks.
\newblock In {\em ICLR}, 2018.

\bibitem{ModEtAl19}
Apostolos Modas, Seyed{-}Mohsen Moosavi{-}Dezfooli, and Pascal Frossard.
\newblock Sparsefool: a few pixels make a big difference.
\newblock In {\em CVPR}, 2019.

\bibitem{MooFawFro2016}
Seyed-Mohsen Moosavi-Dezfooli, Alhussein Fawzi, and Pascal Frossard.
\newblock Deepfool: a simple and accurate method to fool deep neural networks.
\newblock In {\em CVPR}, pages 2574--2582, 2016.

\bibitem{NarKas16}
Nina Narodytska and Shiva~Prasad Kasiviswanathan.
\newblock Simple black-box adversarial perturbations for deep networks.
\newblock In {\em CVPR 2017 Workshops}, 2016.

\bibitem{Cleverhans2017}
Nicolas Papernot, Fartash Faghri, Nicholas Carlini, Ian Goodfellow, Reuben
  Feinman, Alexey Kurakin, Cihang Xie, Yash Sharma, Tom Brown, Aurko Roy,
  Alexander Matyasko, Vahid Behzadan, Karen Hambardzumyan, Zhishuai Zhang,
  Yi-Lin Juang, Zhi Li, Ryan Sheatsley, Abhibhav Garg, Jonathan Uesato, Willi
  Gierke, Yinpeng Dong, David Berthelot, Paul Hendricks, Jonas Rauber, and
  Rujun Long.
\newblock cleverhans v2.0.0: an adversarial machine learning library.
\newblock preprint, arXiv:1610.00768, 2017.

\bibitem{PapEtAl15}
N. Papernot, P.~D. McDaniel, S. Jha, M. Fredrikson, Z.~B. Celik, and A. Swami.
\newblock The limitations of deep learning in adversarial settings.
\newblock In {\em 1st IEEE European Symposium on Security \& Privacy}, 2016.

\bibitem{foolbox}
Jonas Rauber, Wieland Brendel, and Matthias Bethge.
\newblock Foolbox: A python toolbox to benchmark the robustness of machine
  learning models.
\newblock In {\em ICML Reliable Machine Learning in the Wild Workshop}, 2017.

\bibitem{SchEtAl19}
Lukas Schott, Jonas Rauber, Wieland Brendel, and Matthias Bethge.
\newblock Towards the first adversarially robust neural network model on
  {MNIST}.
\newblock In {\em ICLR}, 2019.

\bibitem{SuVarKou19}
Jiawei Su, Danilo~Vasconcellos Vargas, and Kouichi Sakurai.
\newblock One pixel attack for fooling deep neural networks.
\newblock {\em arXiv preprint arXiv:1710.08864v5}, 2019.

\bibitem{SzeEtAl2014}
C. Szegedy, W. Zaremba, I. Sutskever, J. Bruna, D. Erhan, I. Goodfellow, and R.
  Fergus.
\newblock Intriguing properties of neural networks.
\newblock In {\em ICLR}, pages 2503--2511, 2014.

\bibitem{TsiEtAl18}
Dimitris {Tsipras}, Shibani {Santurkar}, Logan {Engstrom}, Alexander {Turner},
  and Aleksander {Madry}.
\newblock Robustness may be at odds with accuracy.
\newblock In {\em ICLR}, 2019.

\bibitem{XiaEtAl18}
Chaowei Xiao, Jun{-}Yan Zhu, Bo Li, Warren He, Mingyan Liu, and Dawn Song.
\newblock Spatially transformed adversarial examples.
\newblock In {\em ICLR}, 2018.

\end{thebibliography}

}

\ifapp
%\clearpage
\appendix

\section{Projection onto the intersection of the $l_0$-ball and the $\sigma$-map constraints}
We here present the projection step which is used for the $\sigma$-PGD attack, where we allow perturbations on at most $k$ pixels and respecting the $\sigma$-map constraints which are defined in \eqref{eq:sigma_constr} and \eqref{eq:sigma_constr_gray}.

\subsection{Color images}
Given a color image $x \in [0,1]^{d \times 3}$, we want to project a given point $y \in \R^{d \times 3}$ onto the set
\begin{align*} C(x)=\big\{ z \in &\R^{d\times 3}\,\big|\, \sum_{i=1}^d \maxop_{j=1,\dots,3} \Id_{|z_{ij}-x_{ij}|>0} \leq k,\\
& (1 - \kappa \sigma_{ij})x_{ij} \leq z_{ij} \leq (1 +\kappa \sigma_{ij})x_{ij},\\ &0\leq z_{ij}\leq 1
\big\},
\end{align*}
where $d$ is the number of pixels, $\sigma_{ij}$ are the pixelwise, channel-specific bounds defined in Section \ref{sec:attack-models} and $\kappa>0$ a given parameter. We can write
the projection problem as
\begin{align*}
\minop_{\lambda \in \R^d}\quad &  \sum_{i=1}^d \sum_{j=1}^3 (y_{ij} - (1+\lambda_i \sigma_{ij})x_{ij})^2\\
\textrm{s. th.} \quad & -\kappa \leq \lambda_i \leq \kappa, \quad i=1,\ldots,d \\
\quad & 0 \leq (1+\lambda_i \sigma_{ij})x_{ij} \leq 1, \; i=1,\ldots,d, \, j=1,\ldots,3\\
%\quad & |\{i\,|\, |\lambda_i|>0\}|\leq k
\quad & \sum_{i=1}^d \Id_{|\lambda_i|>0} \leq k
\end{align*}
Ignoring the combinatorial constraint, we first solve for each pixel the problem
\begin{align*}
\minop_{\lambda_i \in \R}\quad &  \sum_{j=1}^3 (y_{ij} - (1+\lambda_i \sigma_{ij})x_{ij})^2\\
\textrm{s. th.} \quad & -\kappa \leq \lambda_i \leq \kappa \\
\quad & 0 \leq (1+\lambda_i \sigma_{ij})x_{ij} \leq 1, \quad  j=1,\ldots,3\\
\end{align*}
We first note that the last constraint is always fulfilled if $x_{ij}=0$ or $\sigma_{ij}=0$. In the other case we can rewrite the constraint as
\[ -\frac{1}{\sigma_{ij}} \,\leq\, \lambda_i \,\leq \,\frac{1}{\sigma_{ij}}\Big(\frac{1}{x_{ij}}-1\Big), \quad j=1,\ldots,3.\]
Combining all constraints yields
\begin{align*}\lambda_i^{(l)}:=&\max\Big\{-\kappa,\maxop_{\stackrel{j}{x_{ij}\neq 0, \sigma_{ij}\neq 0}} -\frac{1}{\sigma_{ij}} \Big\} \, \leq \,\lambda_i\\ &\leq \,\min\Big\{\kappa ,\minop_{\stackrel{j}{x_{ij}\neq 0, \sigma_{ij}\neq 0}} \frac{1}{\sigma_{ij}}\Big(\frac{1}{x_{ij}}-1\Big)\Big\}:=\lambda_i^{(u)}.\end{align*}
The unconstrained solution is given by
\[ \lambda_i'= \frac{\sum_{j=1}^3 \sigma_{ij}x_{ij}(y_{ij}-x_{ij})}{\sum_{j=1}^3 \sigma_{ij}^2 x_{ij}^2}.\]
Thus the optimal solution for each pixel $i$ is given by
\[ \lambda_i^* = \max\{ \lambda_i^{(l)}, \min\{\lambda_i',\lambda_i^{(u)}\}\}.\]
The final solution of the original problem allows only to choose $k$ pixels to be changed.
%All pixels can be optimized independently of each other as the objective is separable. By optimizing a pixel $i$ we the objective decreases by
For each pixel $i$ the quantity
\[ \phi_i:=\sum_{j=1}^3 (y_{ij}-x_{ij})^2 - \sum_{j=1}^3 (y_{ij}-(1+\lambda_i^* \sigma_{ij})x_{ij})^2\]
represents the difference in how much the objective increases between the cases $\lambda_i=0$ (that is $y_i$ is projected to $x_i$) and $\lambda_i= \lambda_i^*$. Since we want to minimize the objective function, the optimal solution is obtained by sorting $(\phi_i)_{i=1}^d$ in decreasing order $\pi$ and setting
\[ \lambda^{(final)}_{\pi_i} = \begin{cases} \lambda_{\pi_i}^* & \textrm{ if } i=1,\ldots,k,\\ 0 & \textrm{ else }.\end{cases}.\]

Finally, the point belonging to $C(x)$ onto which $y$ is projected is $z \in \R^{d\times 3}$ defined componentwise by
\[z_{ij}= (1 + \lambda_{i}^{(final)}\sigma_{ij})x_{ij}, \; i=1,\ldots,d, \, j=1,\ldots,3. \]

\subsection{Gray-scale images}
Since gray-scale images have only one color channel and, to get imperceivable manipulations, we use additive modifications as defined in \eqref{eq:sigma_constr_gray}, we project onto the set, given the original image $x$, \begin{align*} C(x)=\big\{ z \in \R^{d}\,\big|\, &\sum_{i=1}^d \Id_{|z_{i}-x_{i}|>0} \leq k,\\
& x_{i} - \kappa \sigma_{i} \leq z_{i} \leq x_{i} +\kappa \sigma_{i},\\ &0\leq z_{i}\leq 1
\big\}.
\end{align*}
Defining \begin{align*} &l_i := \max\{x_i - \kappa \sigma_{i}, 0\}, \quad u_i := \min\{x_{i} + \kappa \sigma_{i}, 1\},
\end{align*}
we can see that in this case the problem is equivalent to the projection onto the intersection of an $l_0$-ball and box constraints and then solved as illustrated in Section \ref{sec:proj_1}.

\section{Experiments}
\begin{table}[t]
	\centering
	\begin{tabular}{L{17mm}| L{18mm}| L{20mm}| R{11mm}}
		\multicolumn{4}{c}{\textbf{test accuracy of the attacked models}}\\[2mm]
		
		section & dataset & model & accuracy\\
		\hline
		\multirow{3}{*}{Section 5.1} & MNIST & NiN \cite{LinEtAl14} & 99.66\%\\
		\cline{2-4}
		& CIFAR-10 & NiN \cite{LinEtAl14} & 90.62\%\\
		\cline{2-4}
		& R-ImageNet & ResNet-50 \cite{TsiEtAl18} & 94.46\%\\
		\hline
		\multirow{11}{*}{\shortstack[l]{Sections\\5.2, 5.3, \ref{sec:app_imp}}} & \multirow{5}{*}{MNIST} & \textit{plain} \cite{MadEtAl2018}& 99.17\%\\
		& & \textit{$l_\infty$-at} \cite{MadEtAl2018} & 98.53\%\\
		& & \textit{$l_2$-at} & 98.95\%\\
		& & \textit{$l_0$-at} & 96.38\%\\
		& & \textit{$l_0+\sigma$-at} & 99.29\%\\
		
		\cline{2-4}
		&\multirow{5}{*}{CIFAR-10} & \textit{plain} & 88.38\% \\
		& & \textit{$l_\infty$-at} &79.90\% \\
		& & \textit{$l_2$-at} &80.44\% \\
		& & \textit{$l_0$-at} &82.31\%\\
		& & \textit{$l_0+\sigma$-at} &76.24\% \\
		\cline{2-4}
		
		& R-ImageNet & ResNet-50 \cite{TsiEtAl18} &94.46\% \\
		\hline
	\end{tabular}
	\caption{\textbf{Accuracy of the attacked models.} We here report the accuracy on the test (validation set for Restricted ImageNet) set of the models introduced in Section 5.}
	\label{tab:acc}
\end{table}
We here report the details about the attacks, the attacked models and the parameters used in Section 5. The test accuracy (validation accuracy for Restricted ImageNet) of every model introduced in the paper is reported in Table \ref{tab:acc}.

\subsection{Evaluation of $l_0$-attacks}
The architecture used for this experiment is the Network in Network from \cite{LinEtAl14}, which we trained according to the code available at \url{https://github.com/BIGBALLON/cifar-10-cnn}, adapting it also to the case of MNIST (which has different input dimension).\\
We run $\textrm{PGD}_0$ with ten thresholds $k$ (that is the maximum number of pixels that can be modified), which are $k \in \{2, 3, 4, 5, 6, 8, 10, 12, 15, 20\}$ for MNIST and $k \in \{1, 2, 3, 4, 6, 8, 10, 12, 15, 20\}$ for CIFAR-10.
%We use the following hyperparameters: 20 random restarts, 10 iterations and step size $\eta=\nicefrac{140}{255}$ (see Equation (5)).

\paragraph{Running times} We report the average running times for one image for the experiments in Section \ref{sec:experiments_l0_attacks} (the times for SparseFool are those given by our reimplementation, which uses DeepFool as implemented in \cite{foolbox}).
MNIST: LocSearchAdv 0.6s (from [25]), PA 21s, CW 300s, SparseFool 2.5s, CornerSearch 9.8s, $\textrm{PGD}_0$ 0.06s (one threshold). CIFAR-10: LocSearchAdv 0.7s [25], PA 22s, CW 283s, SparseFool 1.0s, CornerSearch 3.6s, $\textrm{PGD}_0$ 0.19s (one threshold). ImageNet: SparseFool 17s, CornerSearch 953s, PGD$_0$ 13s (one threshold).

\paragraph{Stability of CornerSearch} Since Algorithm \ref{alg:algorithm-label} involves a component of random sampling, we want to analyse here how the performance of CornerSearch depends on it. Then, we run CornerSearch for 10 times on the models used in Table \ref{tab:comp_attacks}
and get the following statistics: MNIST, success rate (\%) $97.37 \pm 0.13$, \textit{mean} $9.12 \pm 0.05$, \textit{median} $7 \pm 0$. CIFAR-10, success rate (\%) $99.33 \pm 0.12$, \textit{mean} $2.71 \pm 0.02$, \textit{median} $2\pm 0$. This means that our attack is stable across different runs.

\subsection{Sparse and Imperceivable manipulations}\label{sec:app_imp}
\begin{figure}[t]\centering \includegraphics[scale=0.15]{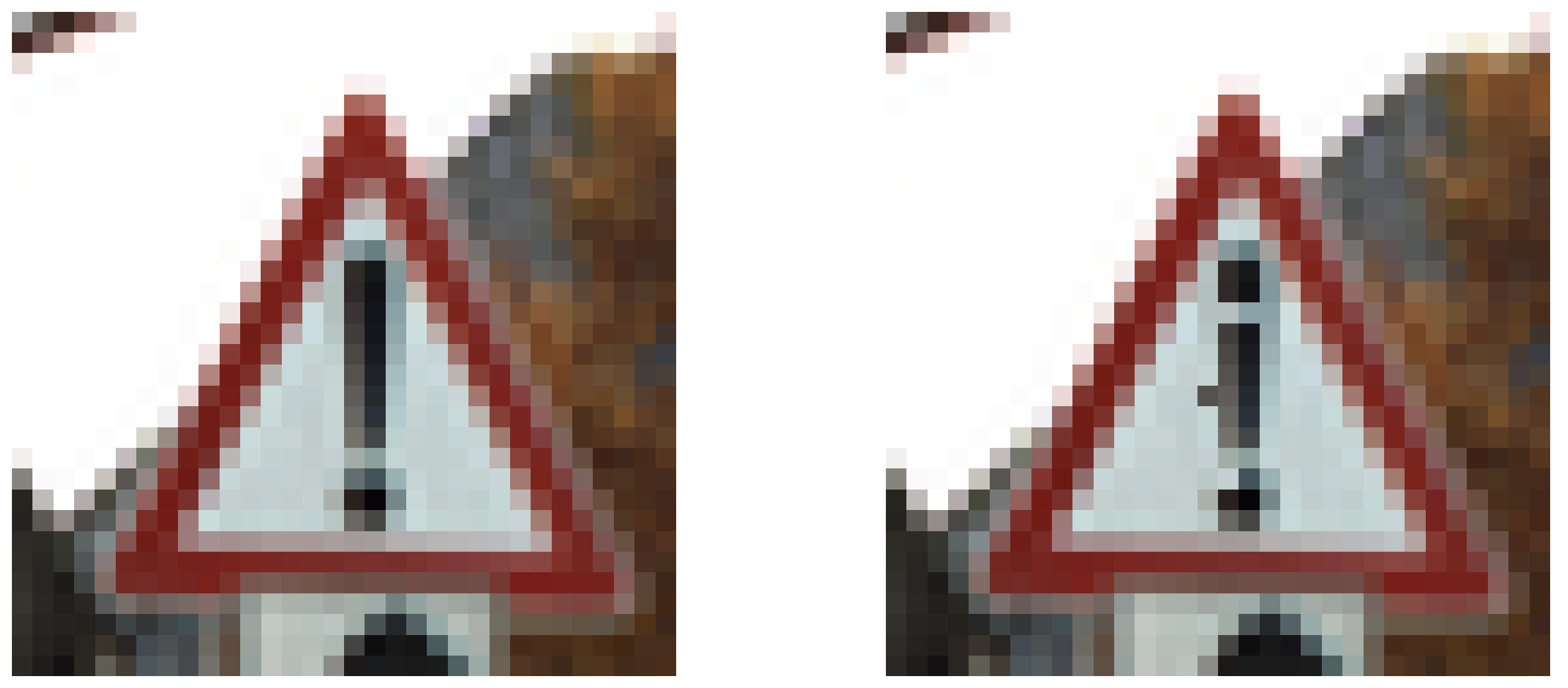} \caption{Left: original. Right: $l_0$-adversarial example with clearly visible changes along axis-aligned edges.} \label{fig:vis_1}\end{figure}
In Figure \ref{fig:vis_1} we show an example of how changes along axis-aligned edges are evident and easy to detect, even if the color is similar to that of some of the neighboring pixels. This provides a further justification of the heuristic we use to decide where the images can be perturbed in an invisible way.

In Figures \ref{fig:imp_MNIST}, \ref{fig:imp_CIFAR-10_app} and \ref{fig:imp_ImageNet_app} we illustrate how the $l_0+\sigma$-map attack produces sparse and imperceivable adversarial perturbations, while both the $l_0$- and the $l_0+l_\infty$-attack either introduce colors which are non-homogeneous with those of the neighbors or modify pixels in an uniform background, which makes them easily visible. Further examples can be found at \url{https://github.com/fra31/sparse-imperceivable-attacks}.

\paragraph{MNIST} In Figure \ref{fig:imp_MNIST} we show the differences among the adversarial examples found by our attacks CornerSearch ($l_0$-attack), $l_0+l_\infty$-attack and $\sigma$-CornerSearch. We see that our $l_0+\sigma$-map attack does not modify pixels in the background, far from the digit or in areas of uniform color in the interior of the digit itself.\\
%Moreover, while, in the example shown in the lower left quadrant of Figure \ref{fig:imp_MNIST}, the original image is correctly classified as a "4", we notice that the adversarial example found by $l_0+l_\infty$-attack shows features of a new class. The modified pixels bridge the gap between the vertical segments in the upper part of the digit, making it similar to a "9". This means that this image does not clearly belong to any class and thus cannot be considered as an obvious adversarial sample.
%Conversely, the $l_0+\sigma$-attack does not change the shape of the digit, which can still be clearly recognized as belonging to the original class "4".\\
The attacked model is the \textit{plain} model from Section 5.3 (more details on the architecture below). For the $l_0+l_\infty$-attack we use a bound on the $l_\infty$-norm of the perturbation of $\delta=0.2$.

\paragraph{CIFAR-10} We show in Figure \ref{fig:imp_CIFAR-10_app} more examples built as those in Figure 3 of Section 5. The attacked model is the \textit{plain} classifier from Section 5.3 (more details on the architecture below). For the $l_0+l_\infty$-attack we use a bound on the $l_\infty$-norm of the perturbation of $\delta=0.1$.

\paragraph{Retricted ImageNet} We show in Figure \ref{fig:imp_ImageNet_app} more examples created as those in Figure 4 of Section 5. The attacked model is the ResNet-50 from \cite{TsiEtAl18} (both weights and code are available at \url{https://github.com/MadryLab/robust-features-code}) already introduced in Section 5.1. For the $l_0+l_\infty$-attack we use $\delta=0.05$ as bound on the $l_\infty$-norm of the perturbation.

\subsection{Adversarial training}
\paragraph{MNIST} The architecture used is the same as in \cite{MadEtAl2018} (available at \url{https://github.com/MadryLab/mnist_challenge}), consisting of 2 convolutional layers, each followed by a max-pooling operation, and 2 dense layers. We trained our classifiers for 100 epochs with Adam \cite{KinEtAl2014}.\\
The \textit{plain} and \textit{$l_\infty$-at} models are those provided by \cite{MadEtAl2018} at \url{https://github.com/MadryLab/mnist_challenge}, while we trained \textit{$l_2$-at} using the plain gradient as direction of the update for PGD, in contrast to the sign of the gradient which is used for adversarial training wrt the $l_\infty$-norm.\\
For adversarial training wrt $l_0$-norm we use $k=20$ (maximal number of pixels to be changed), 40 iterations and step size $\eta=\nicefrac{30000}{255}$.
For \textit{$l_0+\sigma$-at} we set $k=100$, $\kappa=0.9$ (for the bounds given by the $\sigma$-map), 40 iterations of gradient descent with step size $\eta=\nicefrac{30000}{255}$.

\paragraph{CIFAR-10} We use a CNN with 8 convolutional layers, consisting of 96, 96, 192, 192, 192, 192, 192 and 384 feature maps respectively, and 2 dense layers of 1200 and 10 units. ReLU activation function is applied on the output of each layer, apart from the last one. We perform the training with data augmentation (in particular, random crops and random mirroring are applied) for 100 epochs and with Adam optimizer \cite{KinEtAl2014}.\\
For adversarial training wrt $l_0$-norm we use $k=20$ (number of pixels to be changed), 10 iterations of PGD with step size $\eta=\nicefrac{30000}{255}$. For \textit{$l_0+\sigma$-at}, we use $k=120$, $\kappa=0.6$, 10 iterations of PGD with step size $\eta=\nicefrac{30000}{255}$. %The models achieve the following accuracy on the full test set: \textit{plain} 88.38\%, \textit{$l_\infty$-at} 79.90\%, \textit{$l_2$-at} 80.44\%, \textit{$l_0$-at} 83.21\%, \textit{$l_0 + l_\infty$-at} 81.28\%.

\begin{figure*}
	\centering
	\begin{tabular}{c c  c c| c c c c}
		\includegraphics[width=0.2\columnwidth]{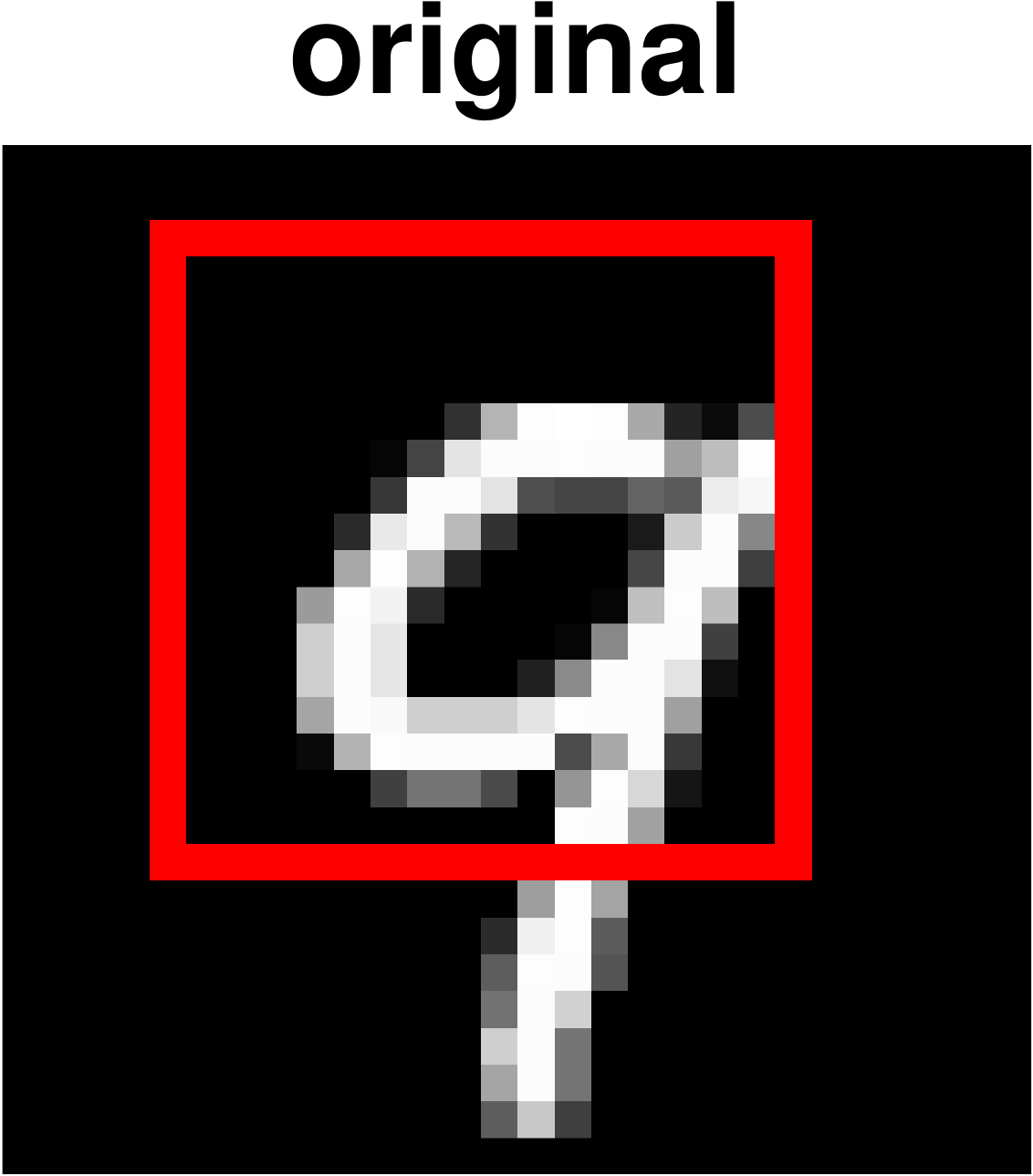}&
		\includegraphics[width=0.2\columnwidth]{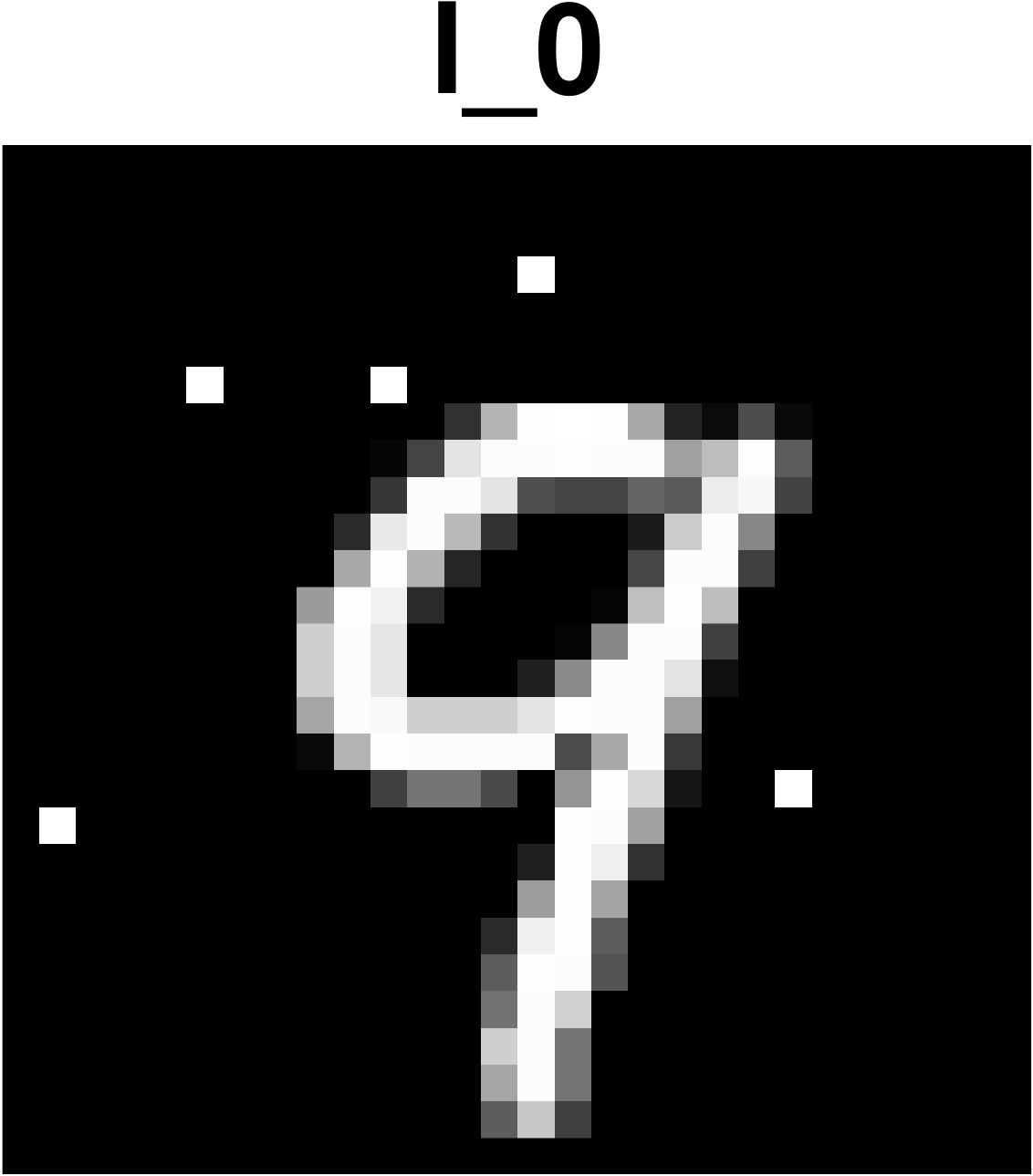}&
		\includegraphics[width=0.2\columnwidth]{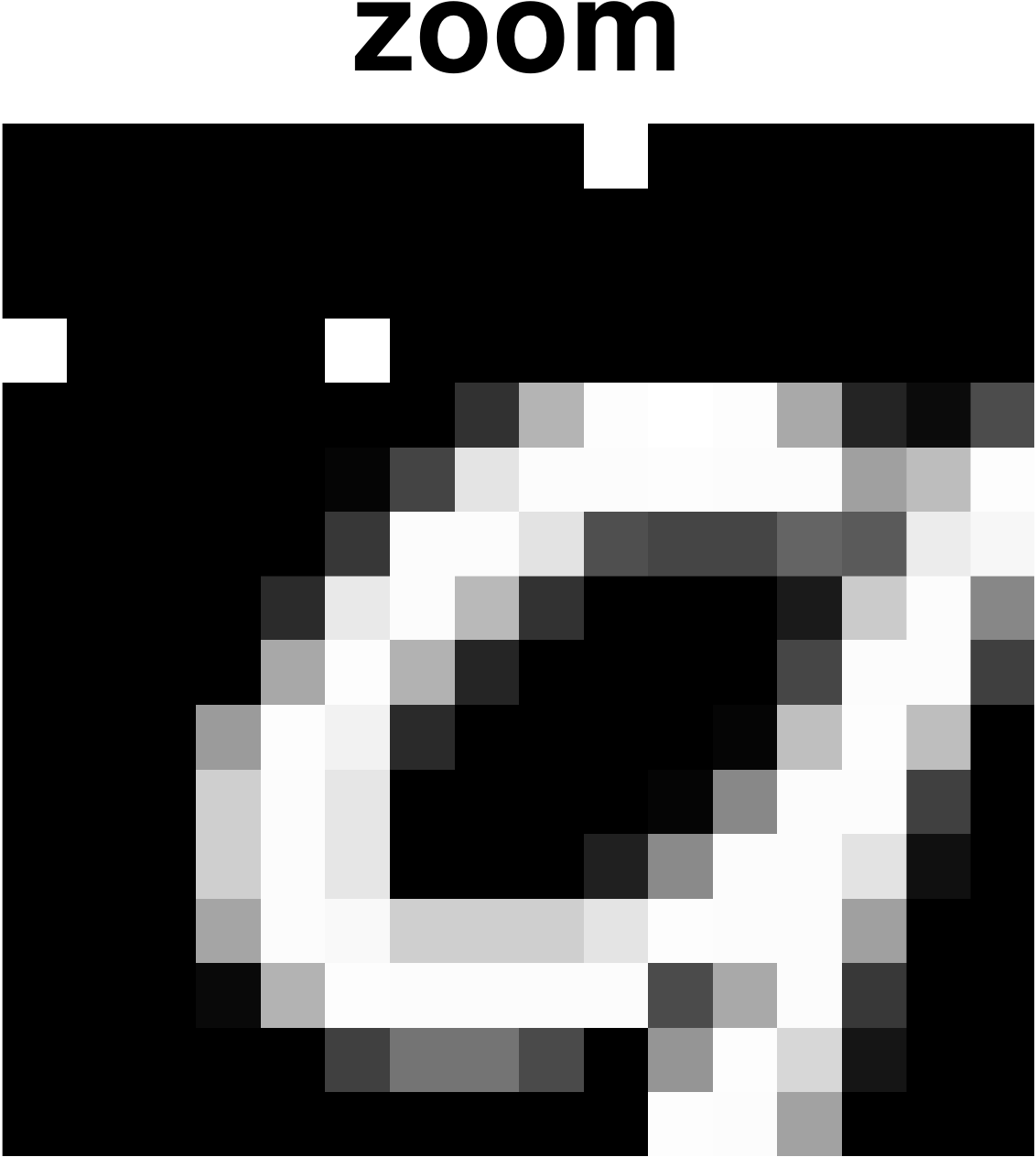}&
		\includegraphics[width=0.2\columnwidth]{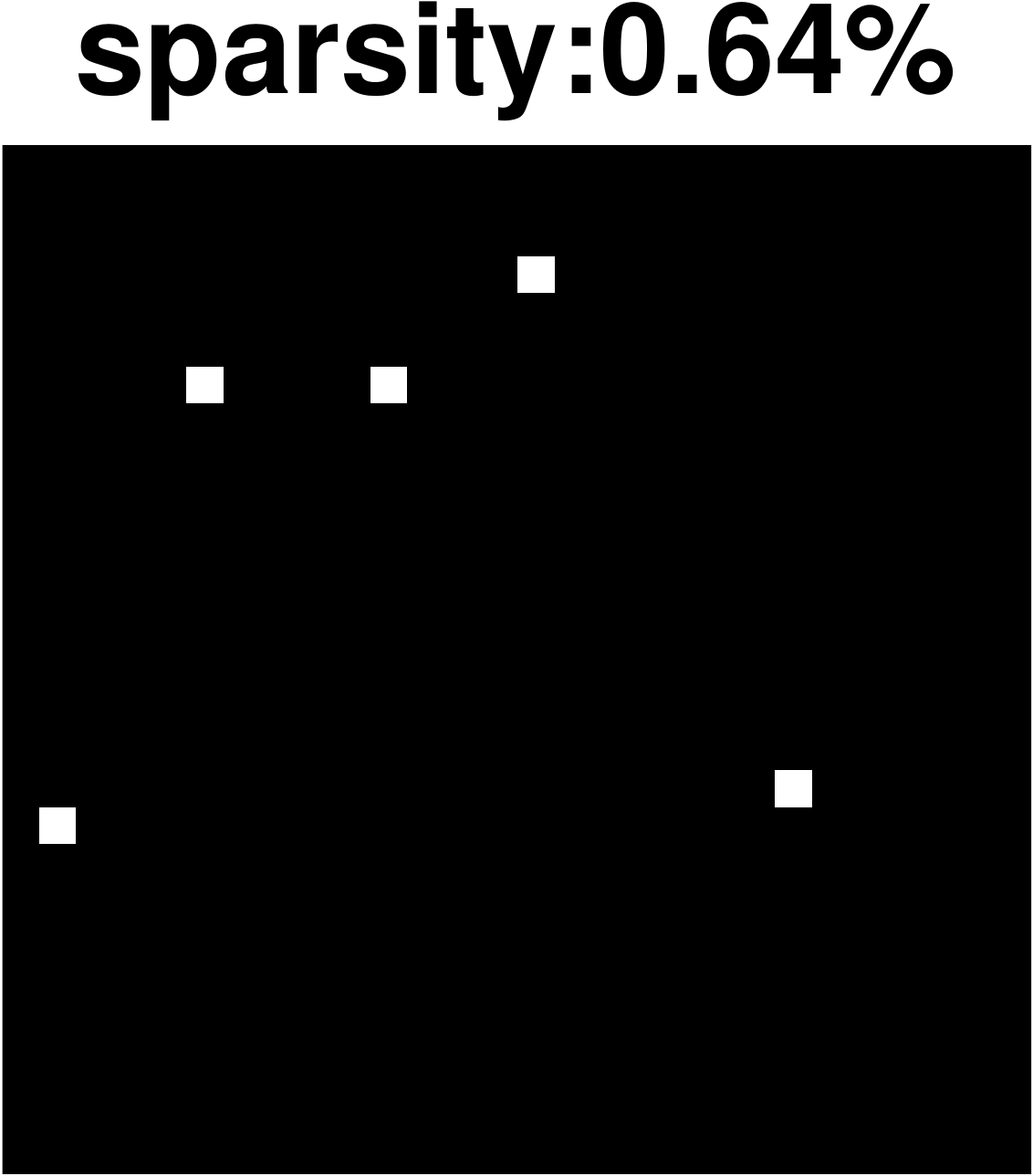}&
		
		\includegraphics[width=0.2\columnwidth]{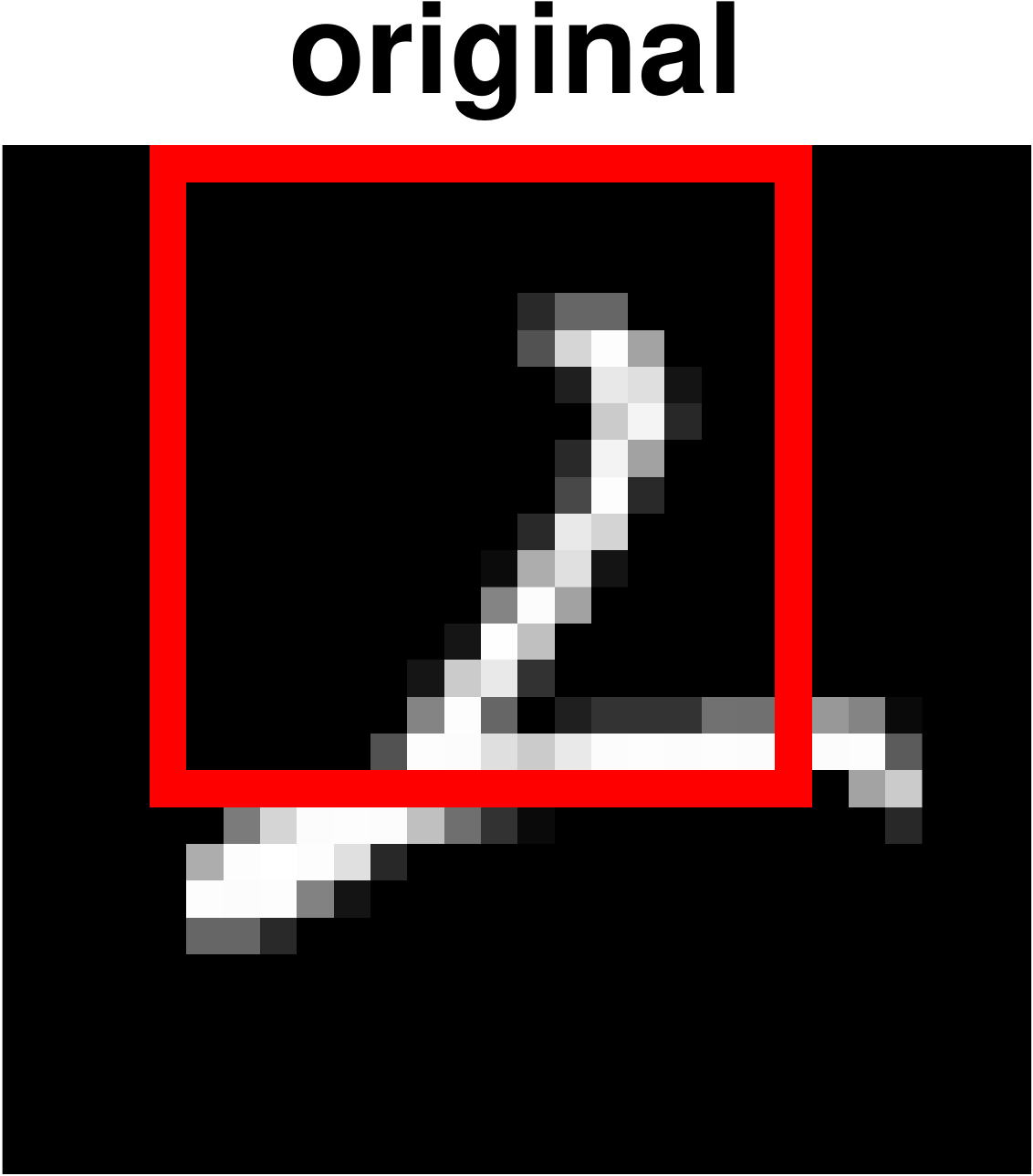}&
		\includegraphics[width=0.2\columnwidth]{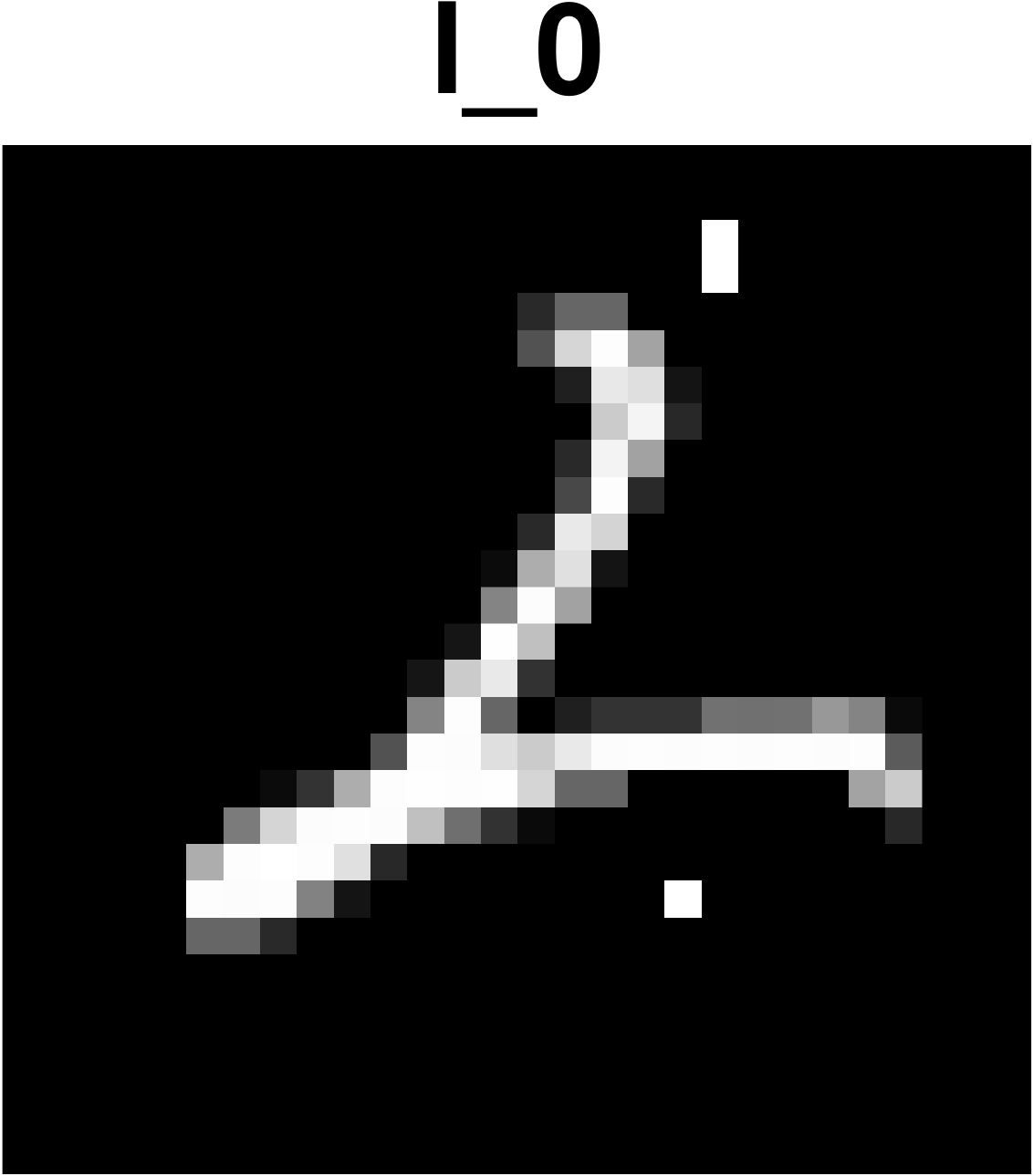}&
		\includegraphics[width=0.2\columnwidth]{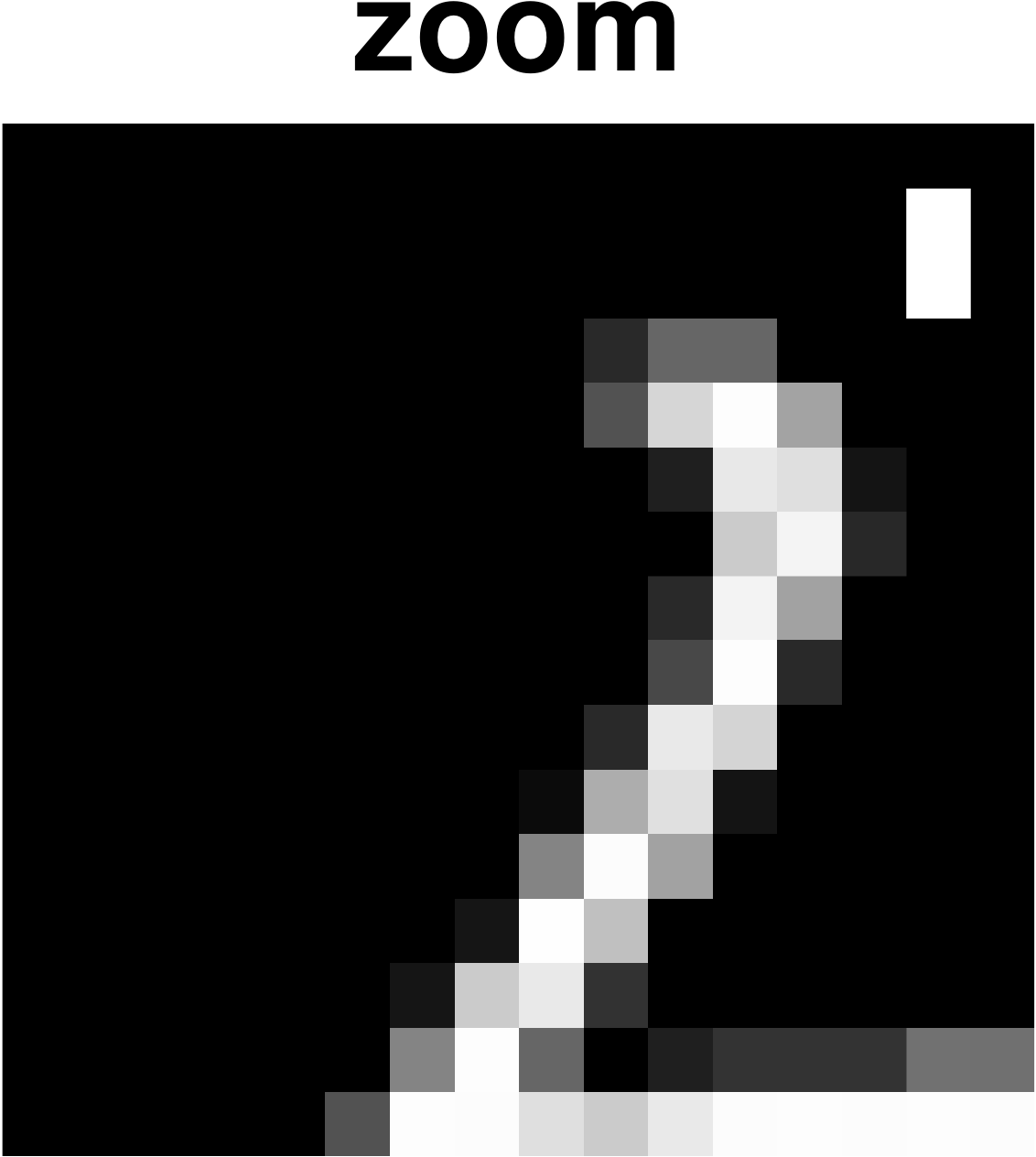}&
		\includegraphics[width=0.2\columnwidth]{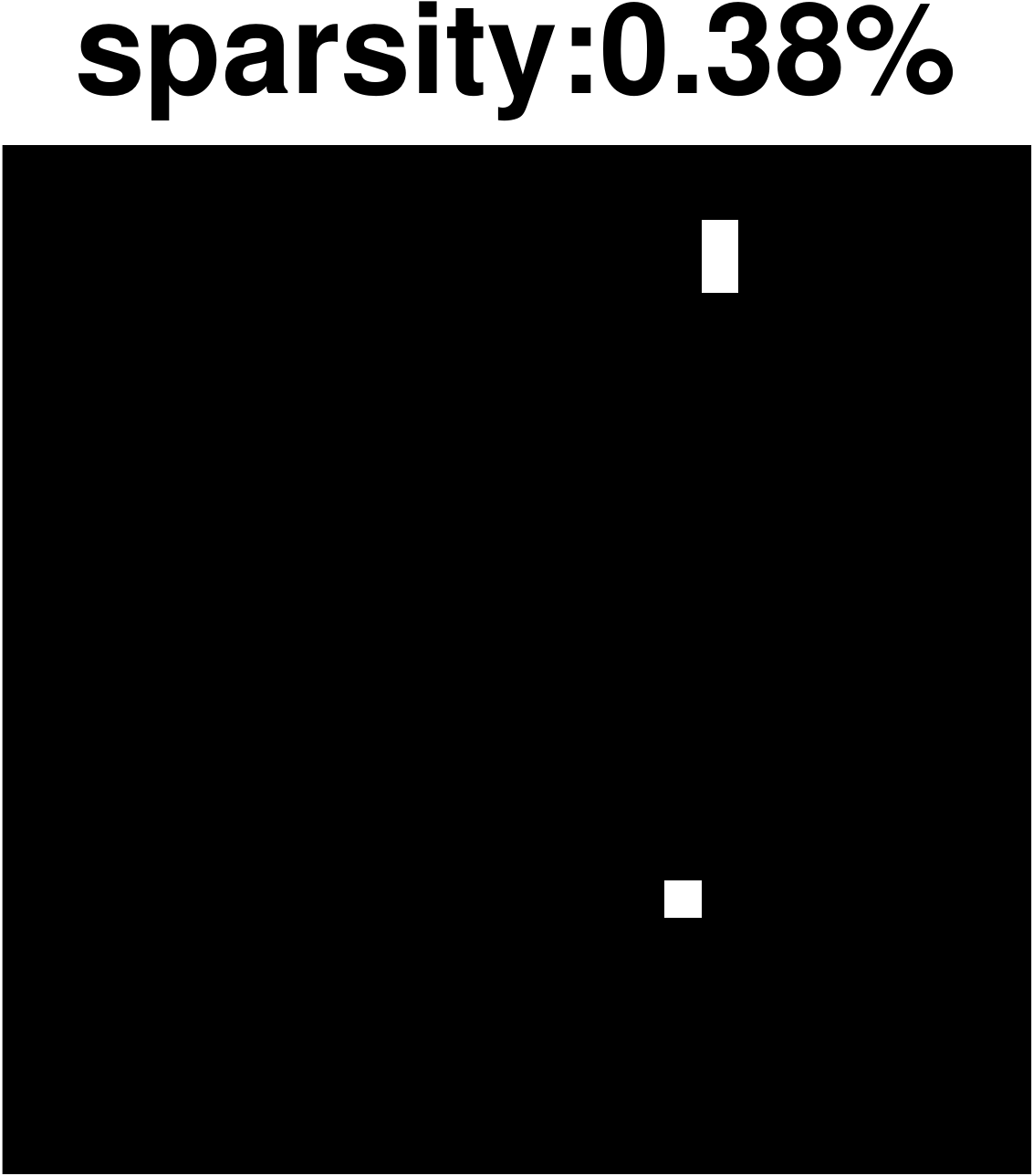}\\
		
		& \includegraphics[width=0.2\columnwidth]{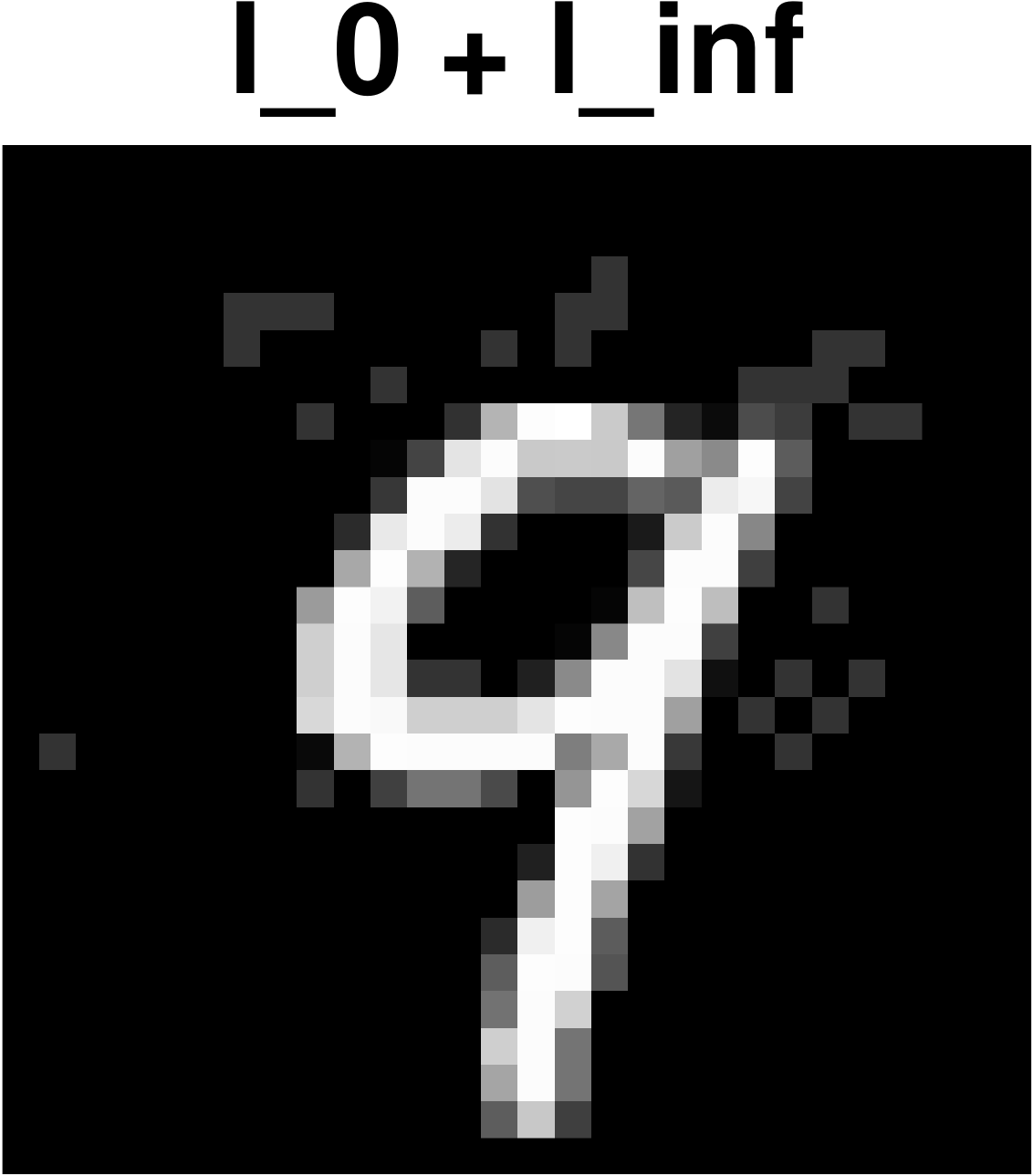}&
		\includegraphics[width=0.2\columnwidth]{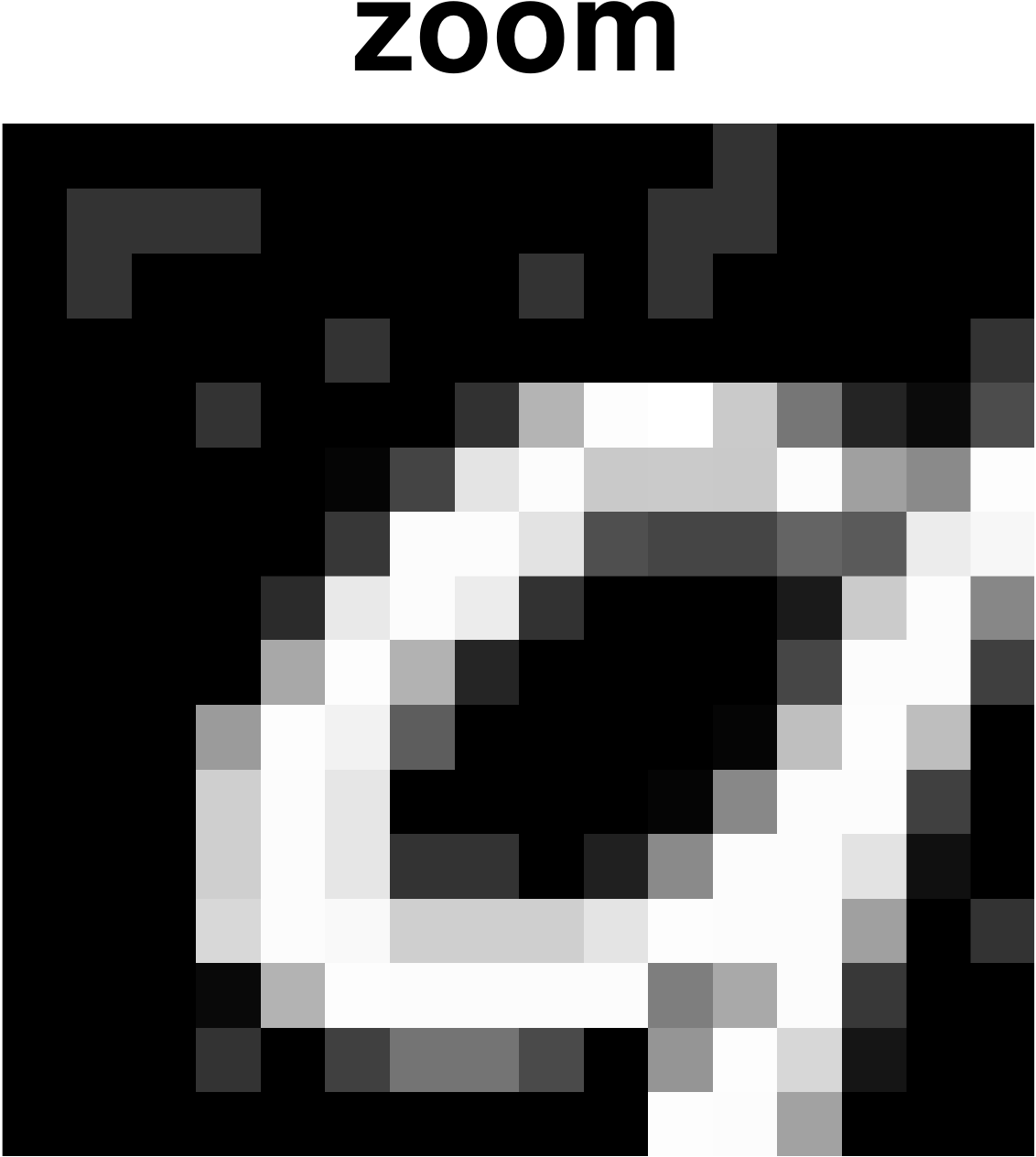}&
		\includegraphics[width=0.2\columnwidth]{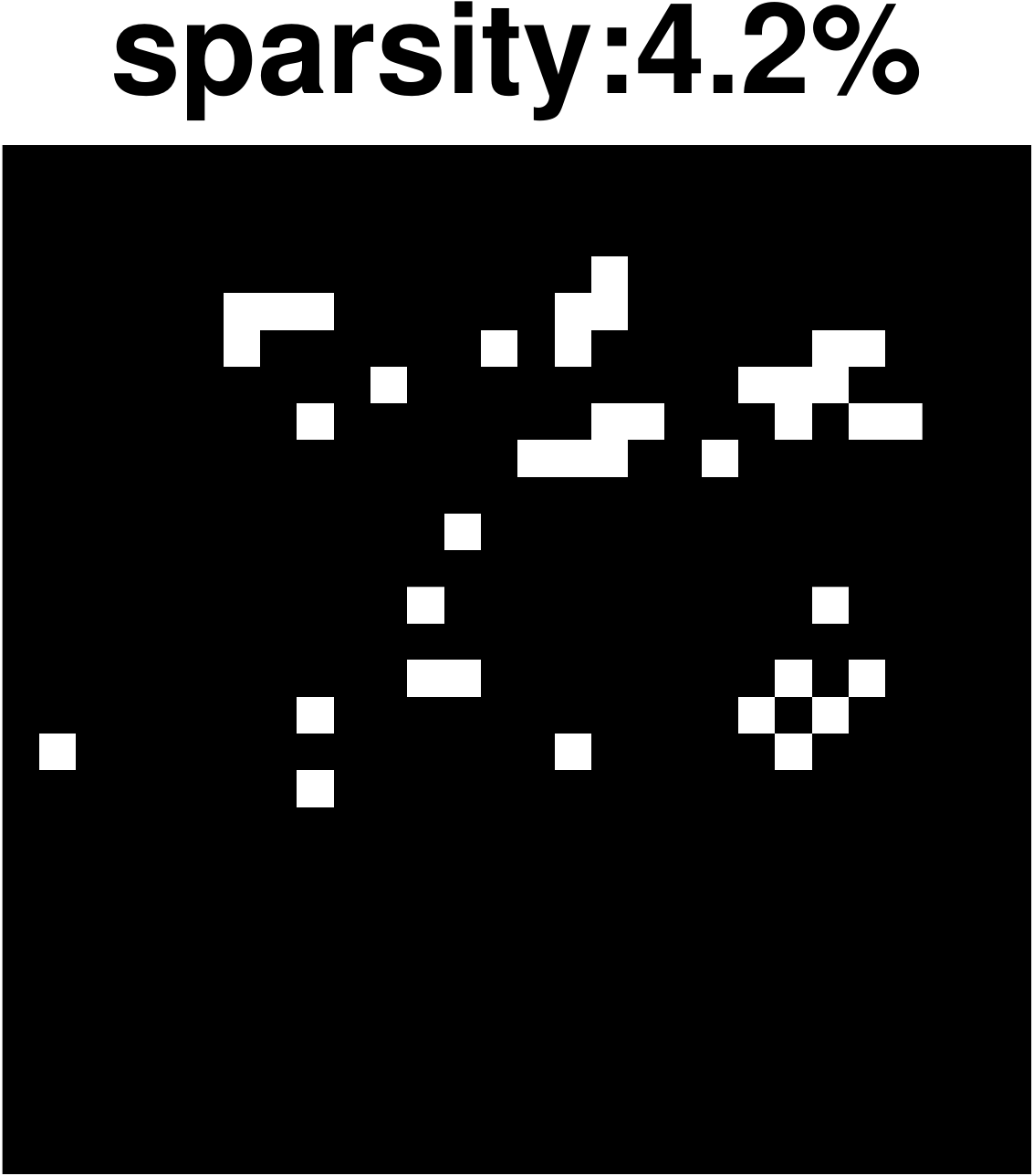}&
		&
		\includegraphics[width=0.2\columnwidth]{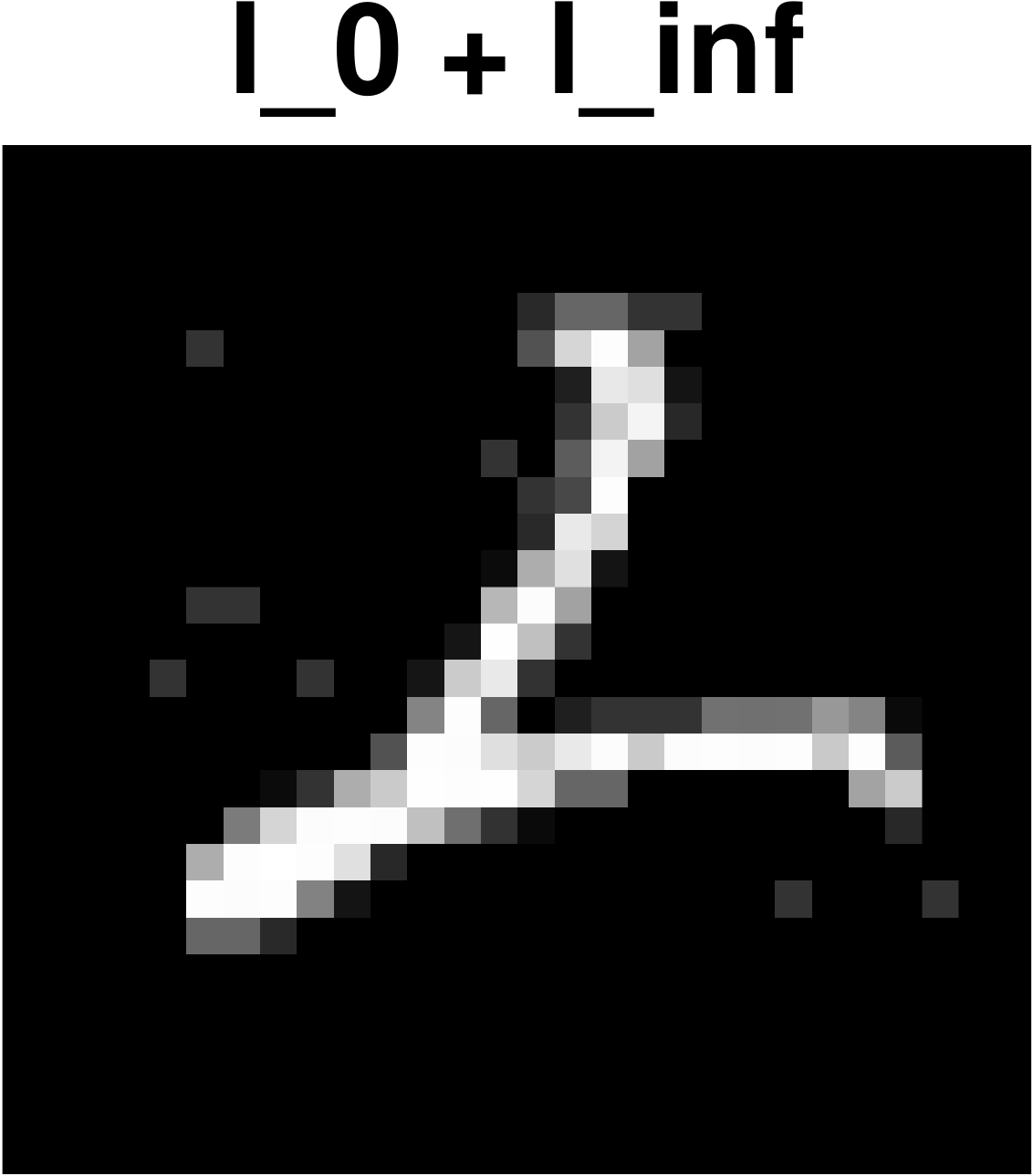}&
		\includegraphics[width=0.2\columnwidth]{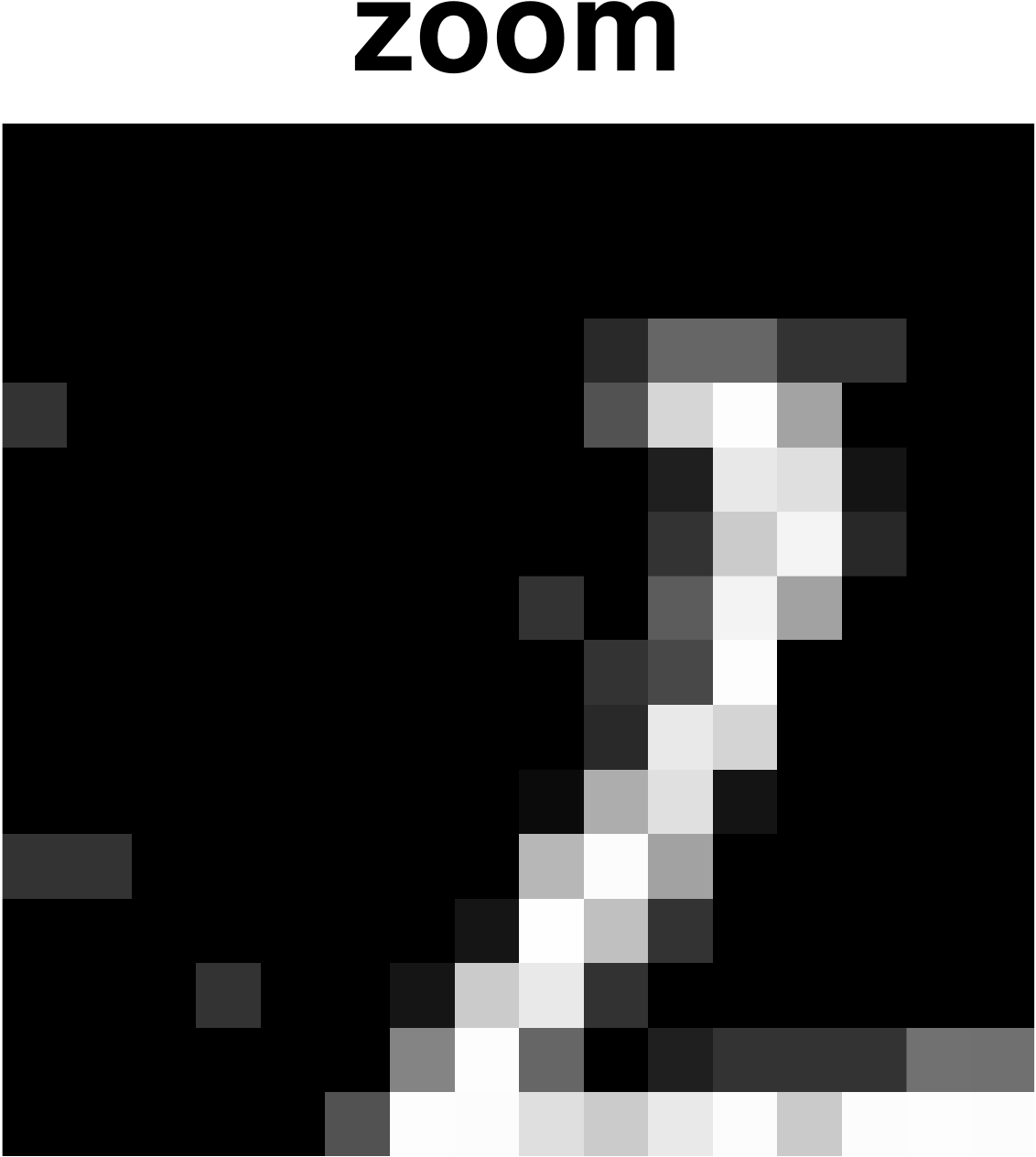}&
		\includegraphics[width=0.2\columnwidth]{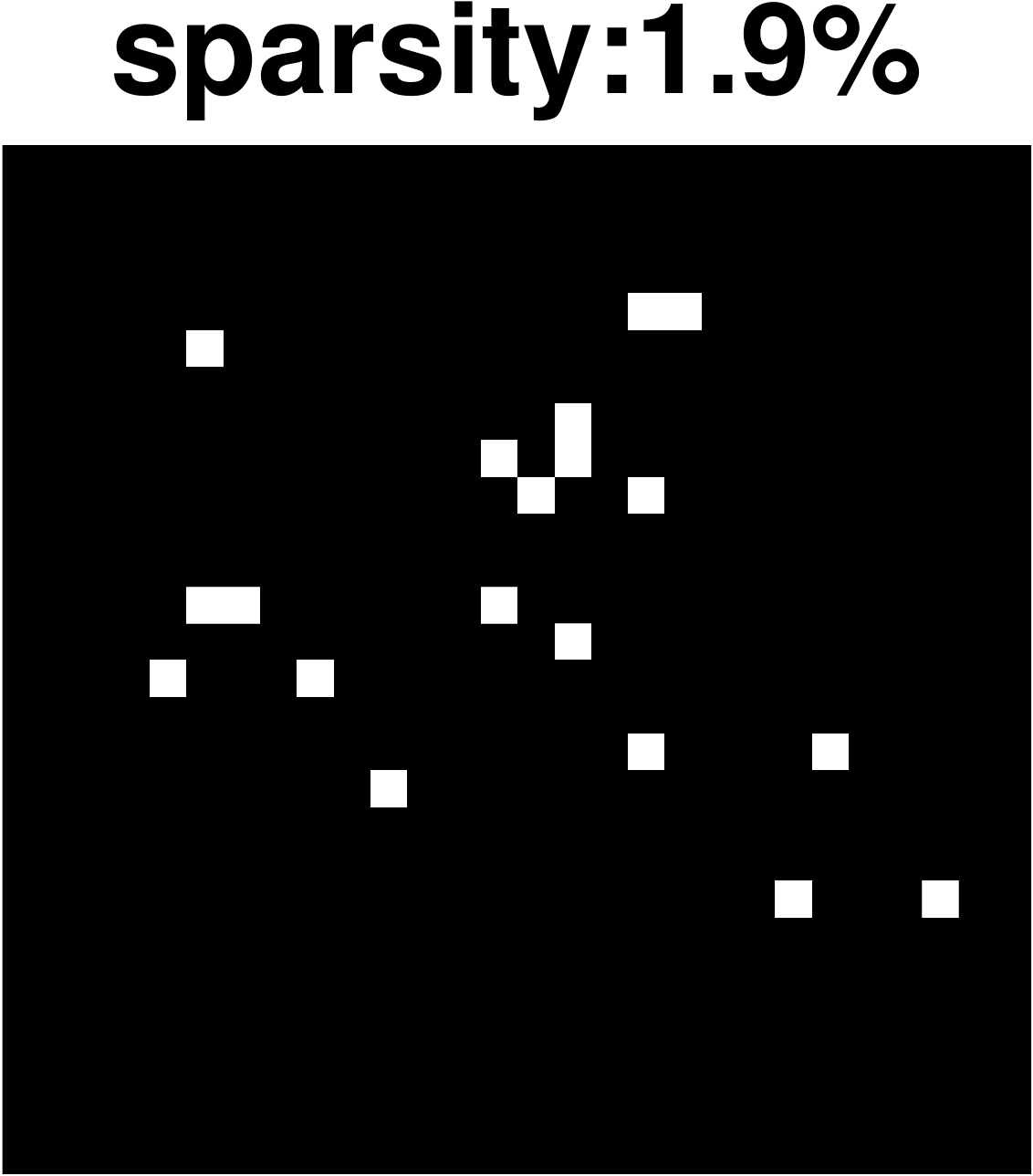}\\
		
		\includegraphics[width=0.2\columnwidth]{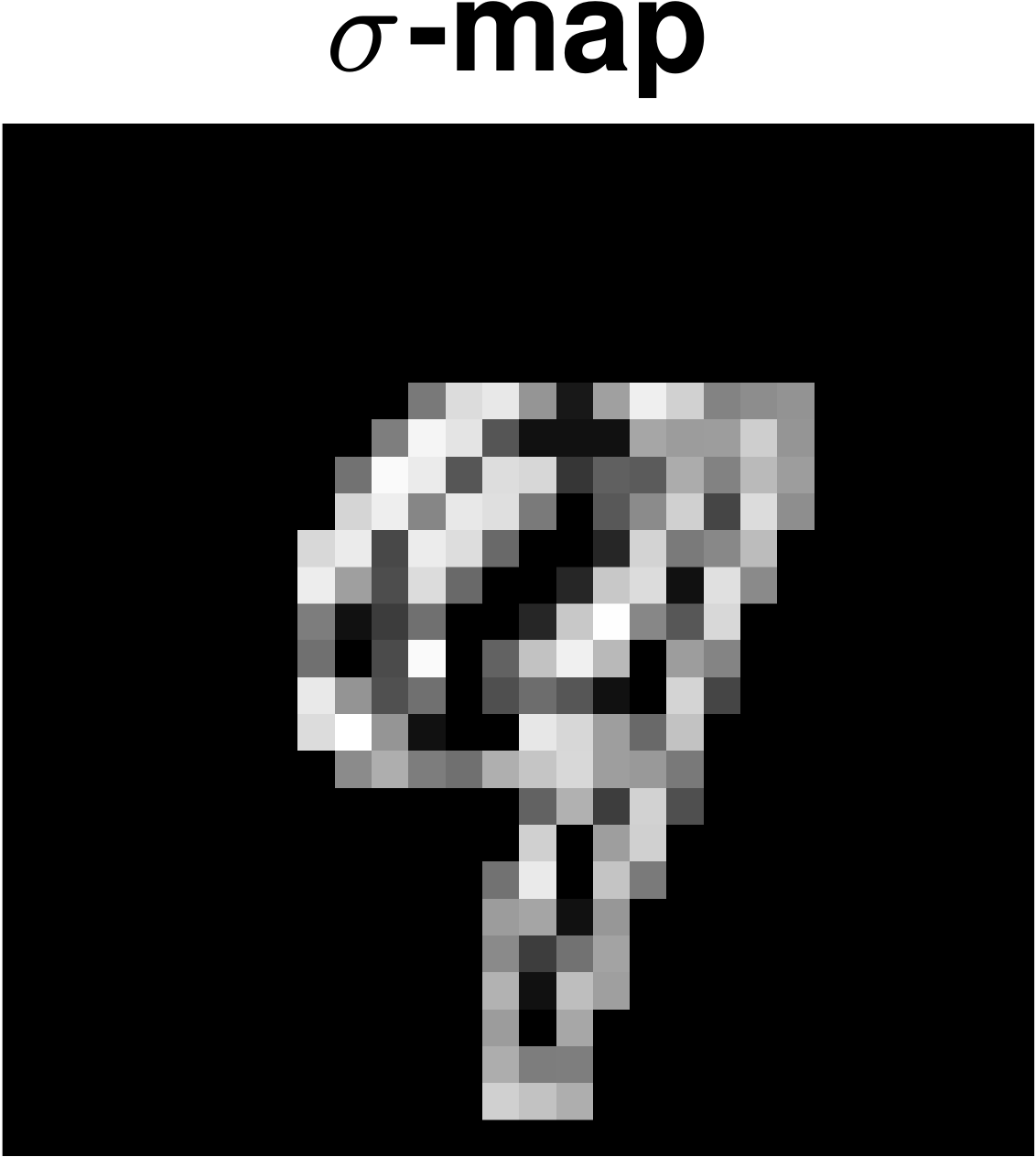}& 
		\includegraphics[width=0.2\columnwidth]{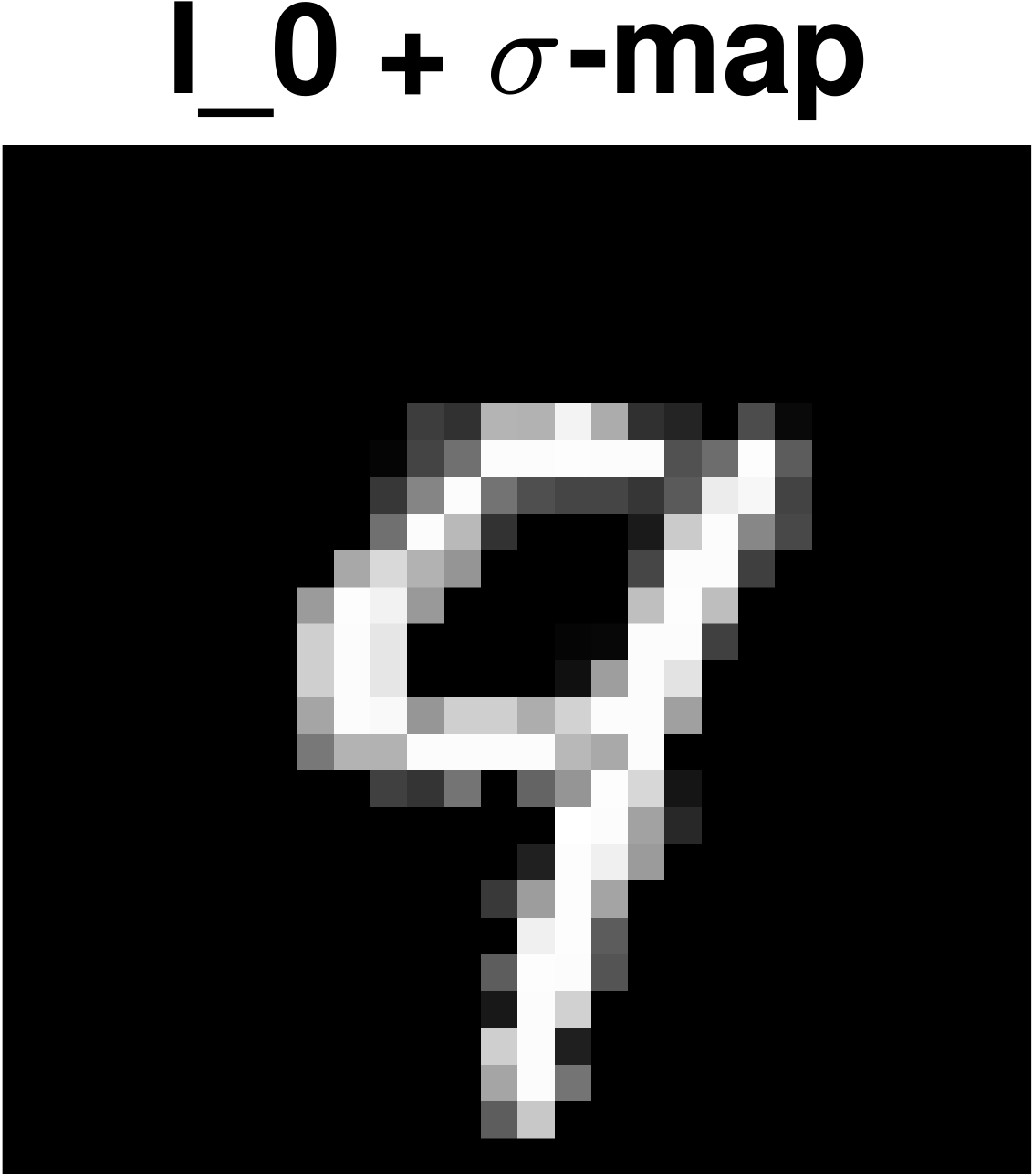}&
		\includegraphics[width=0.2\columnwidth]{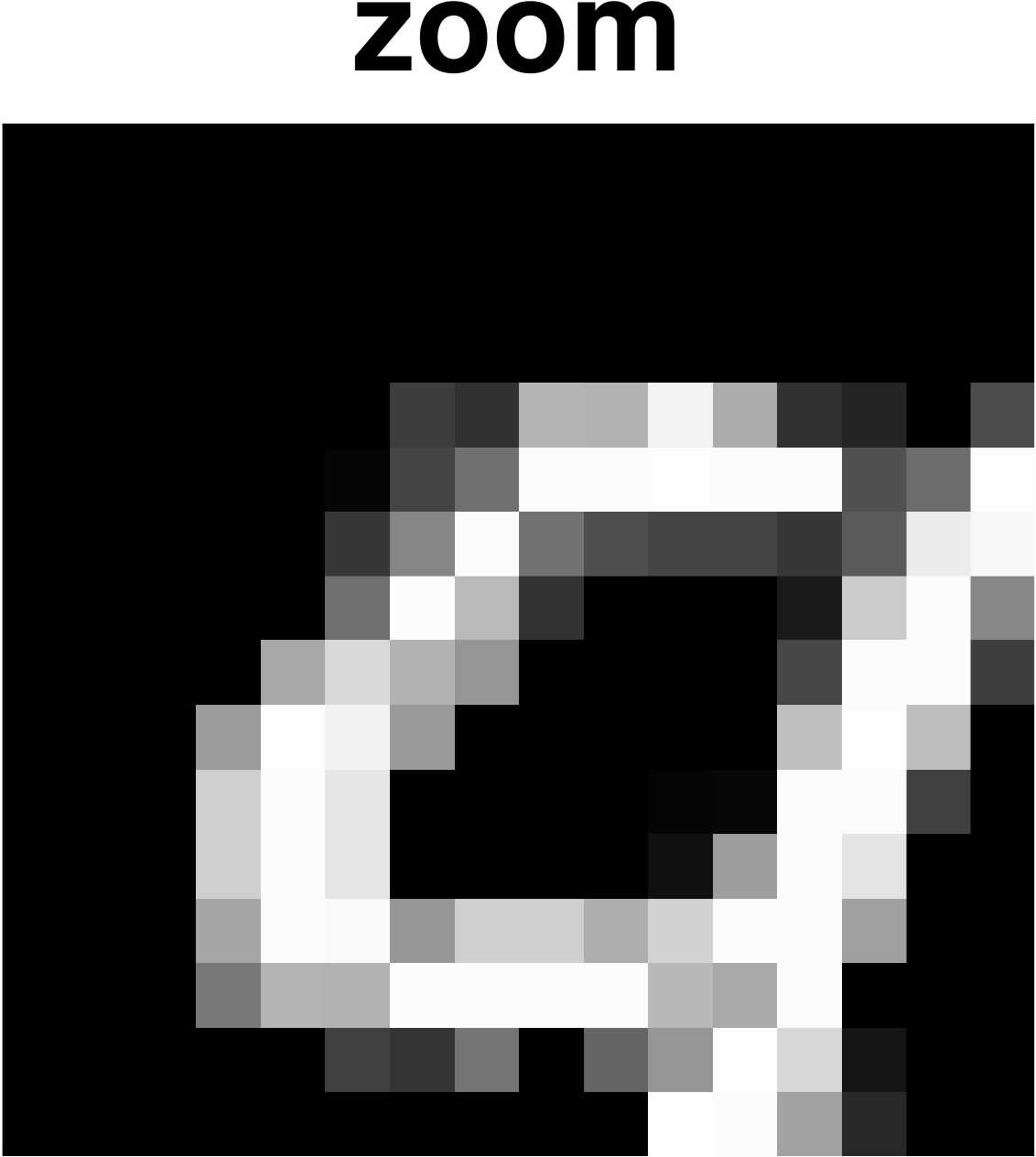}&
		\includegraphics[width=0.2\columnwidth]{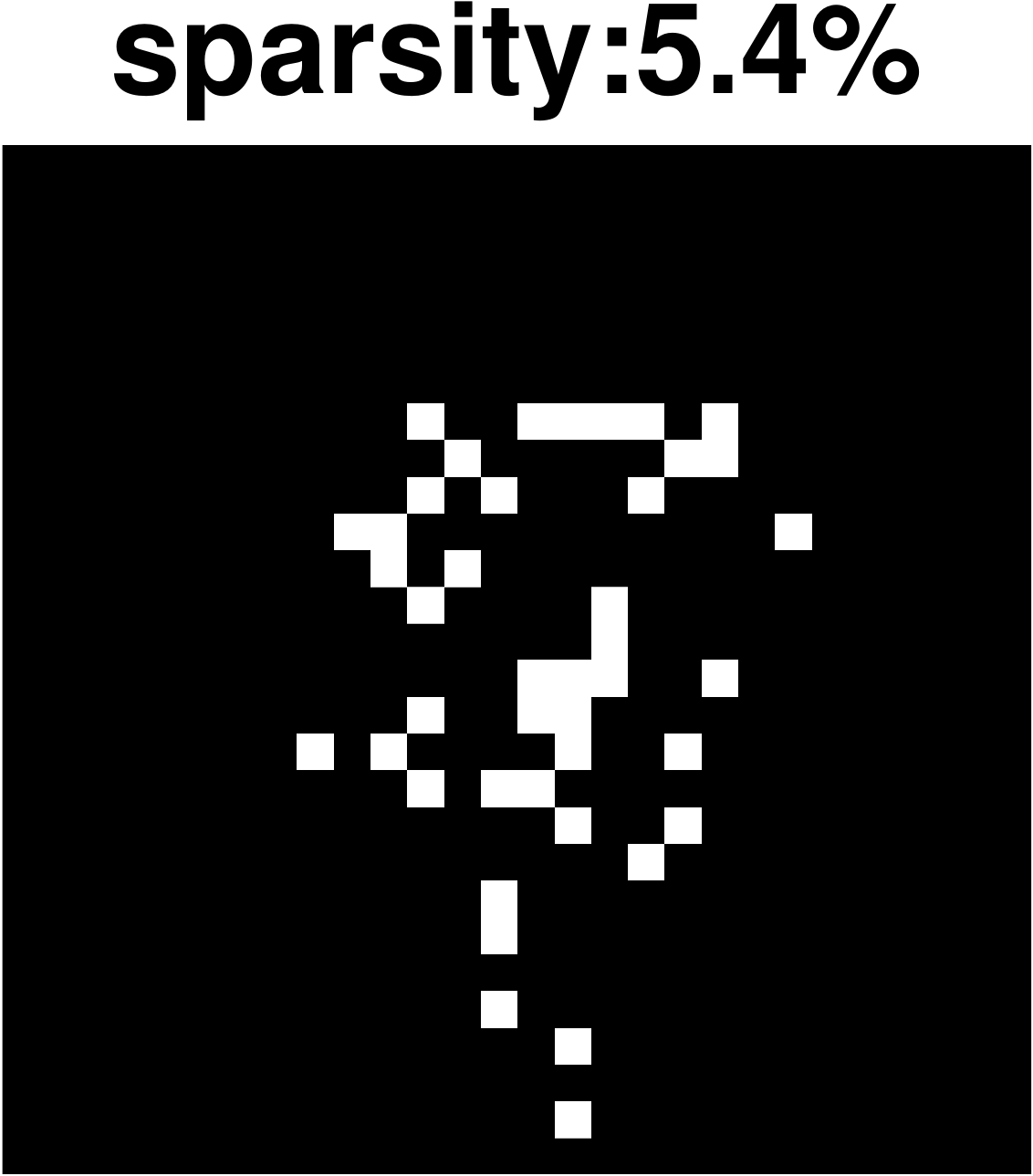}&
		
		\includegraphics[width=0.2\columnwidth]{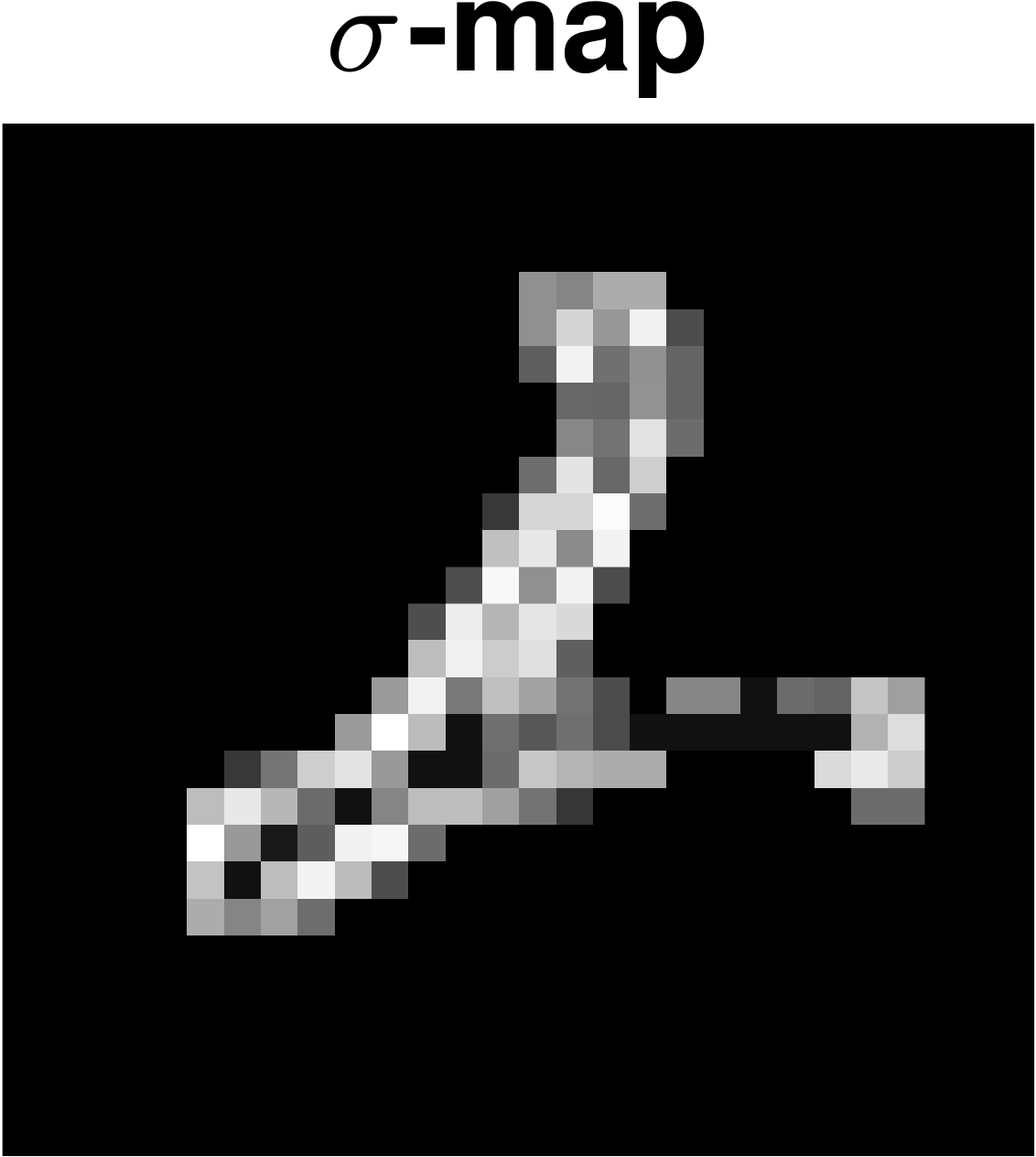}& 
		\includegraphics[width=0.2\columnwidth]{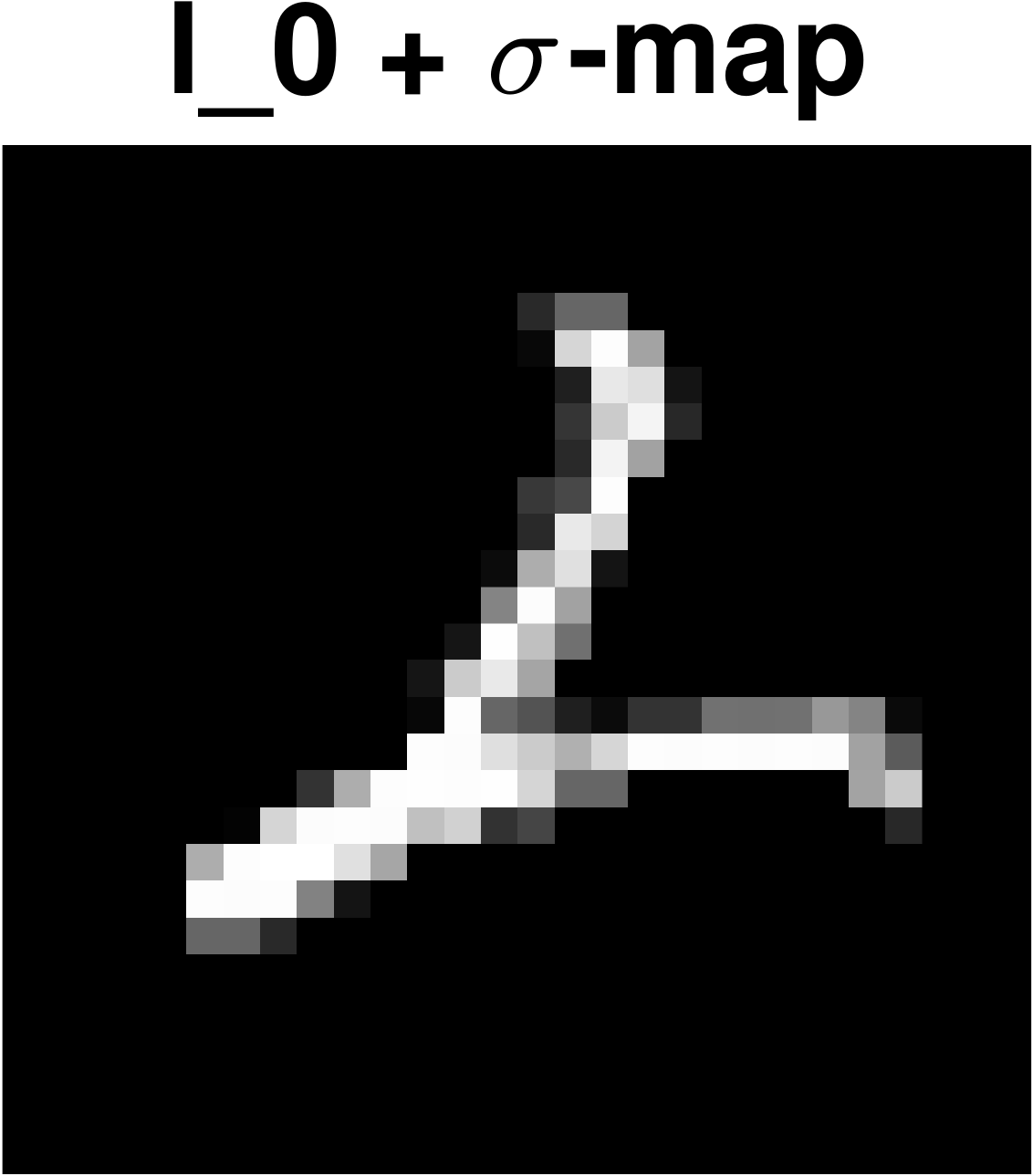}&
		\includegraphics[width=0.2\columnwidth]{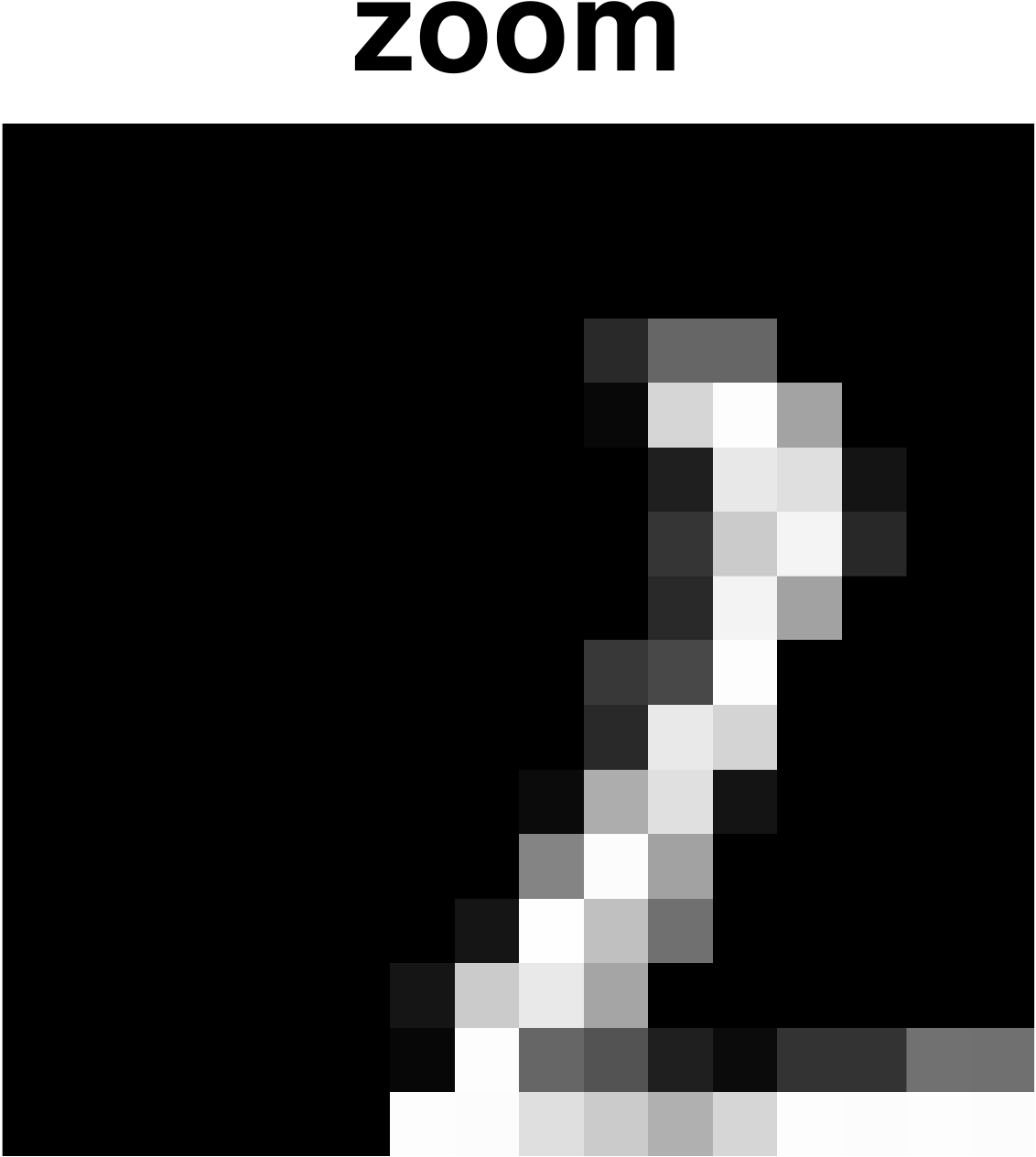}&
		\includegraphics[width=0.2\columnwidth]{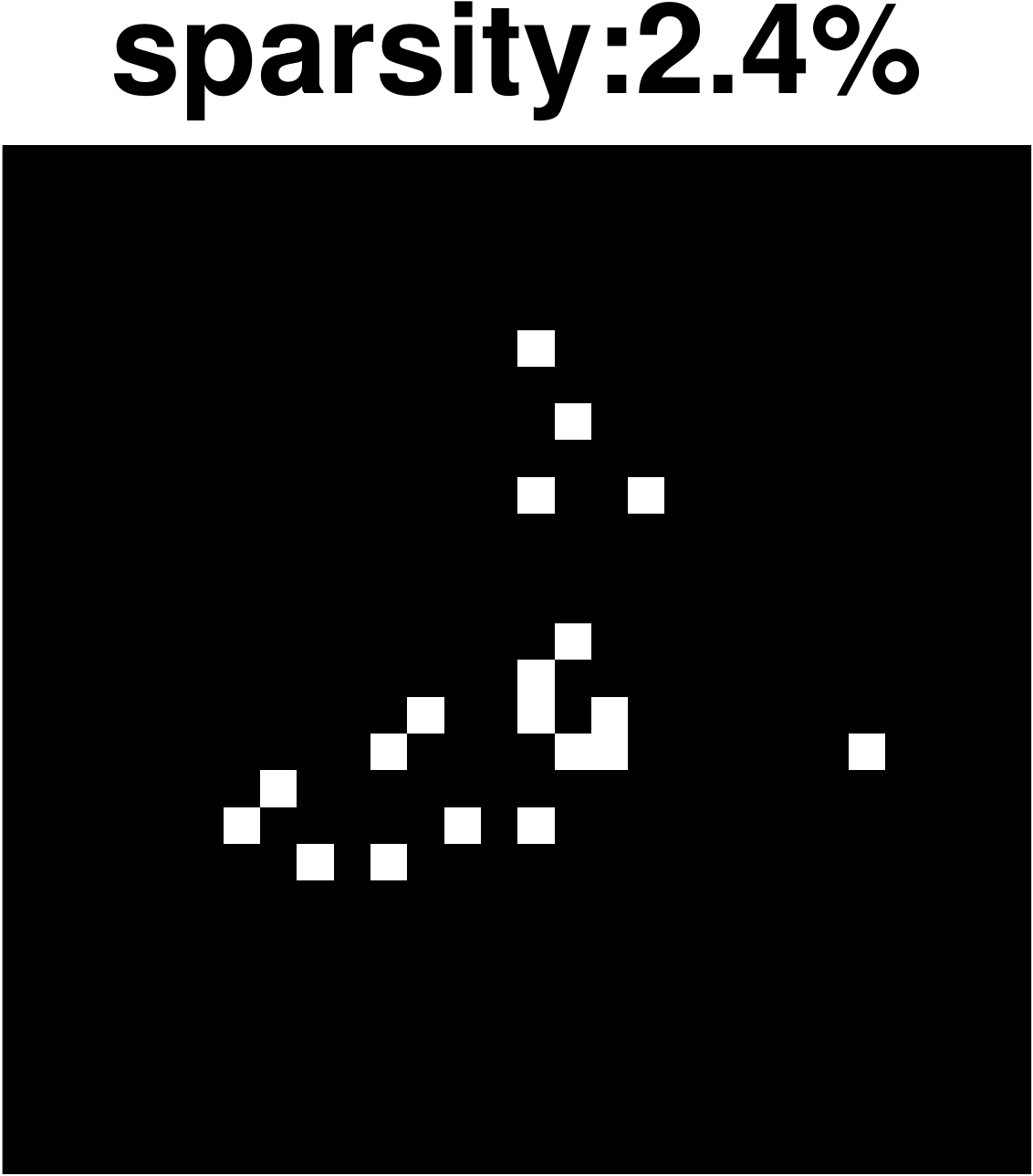} \\

	\end{tabular}
	\caption{\textbf{Different attacks on MNIST.} We illustrate the differences of the adversarial examples (second column) found by CornerSearch ($l_0$), $l_0+l_\infty$- and $\sigma$-CornerSearch, respectively first, second and third row. The third column shows the zoom of the area highlighted by the red box while the fourth column contains the map of the modified pixels (\textit{sparsity} column). The original image can be found top left and the visualization of the $\sigma$-map (rescaled so that $\max_{i}\sigma_{i}=1$) bottom left.}\label{fig:imp_MNIST}
\end{figure*}

\begin{figure*}
	\centering

	\begin{tabular}{c c  c c| c c c c}
		\includegraphics[width=0.2\columnwidth]{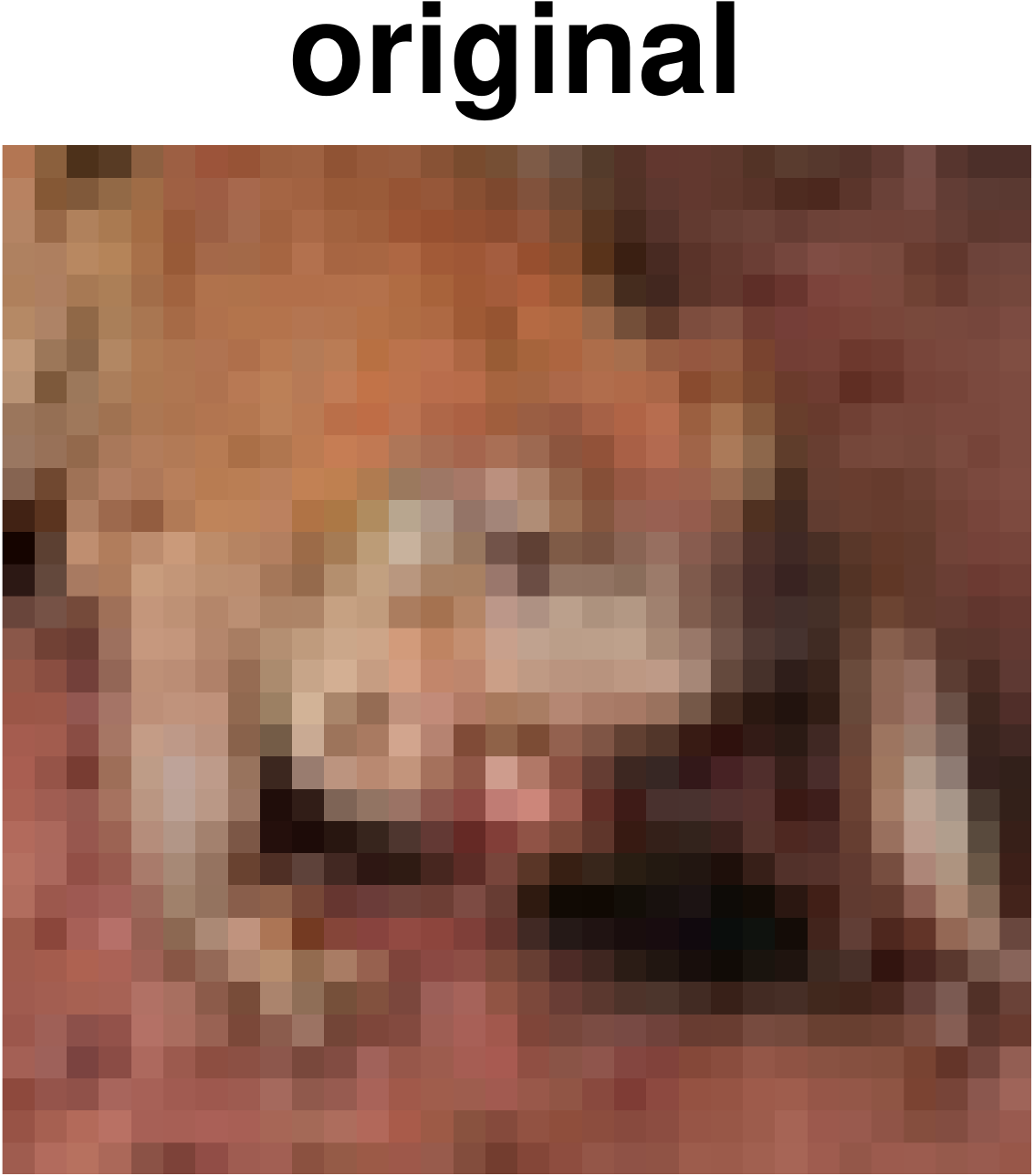}&
		\includegraphics[width=0.2\columnwidth]{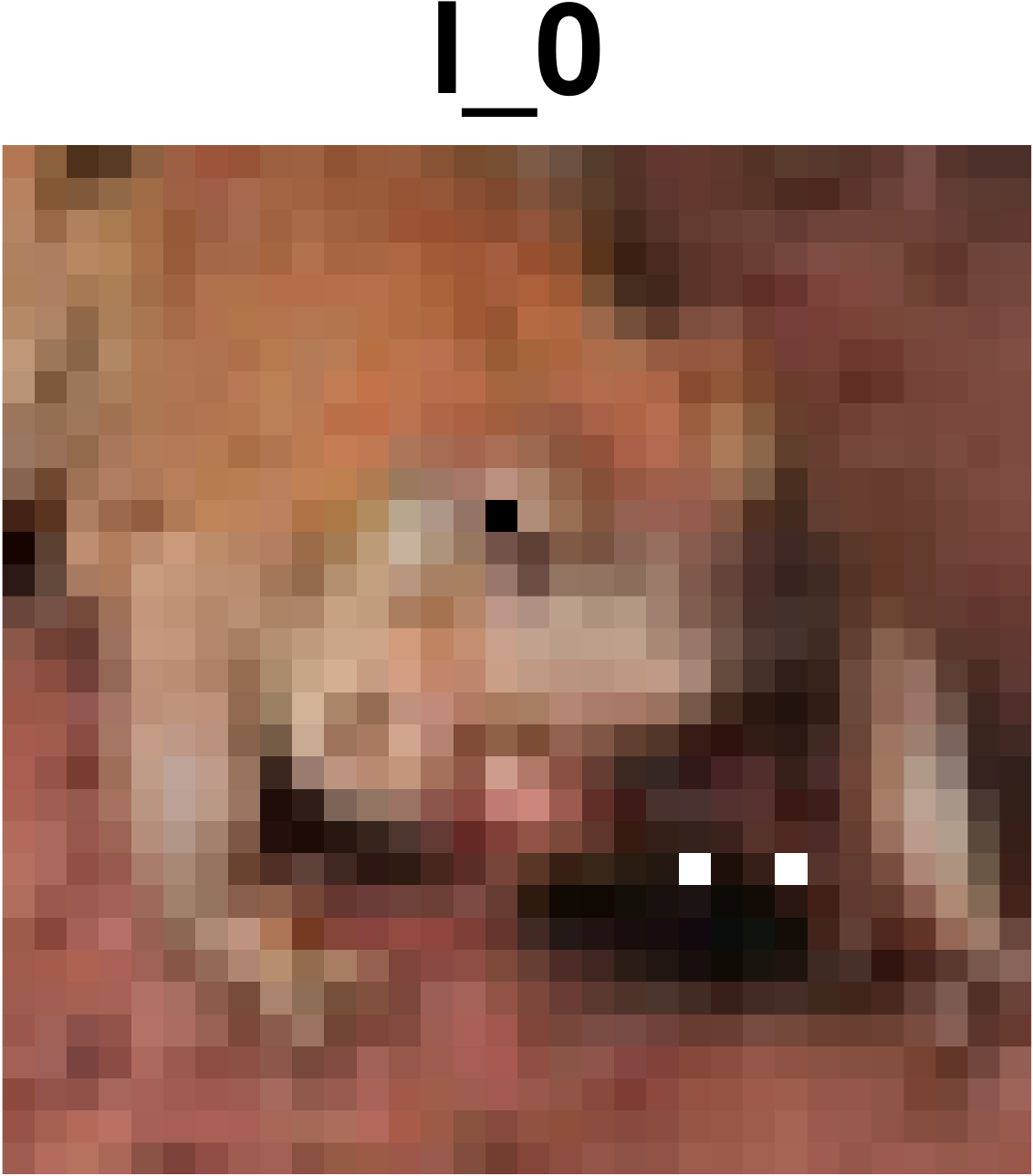}&
		\includegraphics[width=0.2\columnwidth]{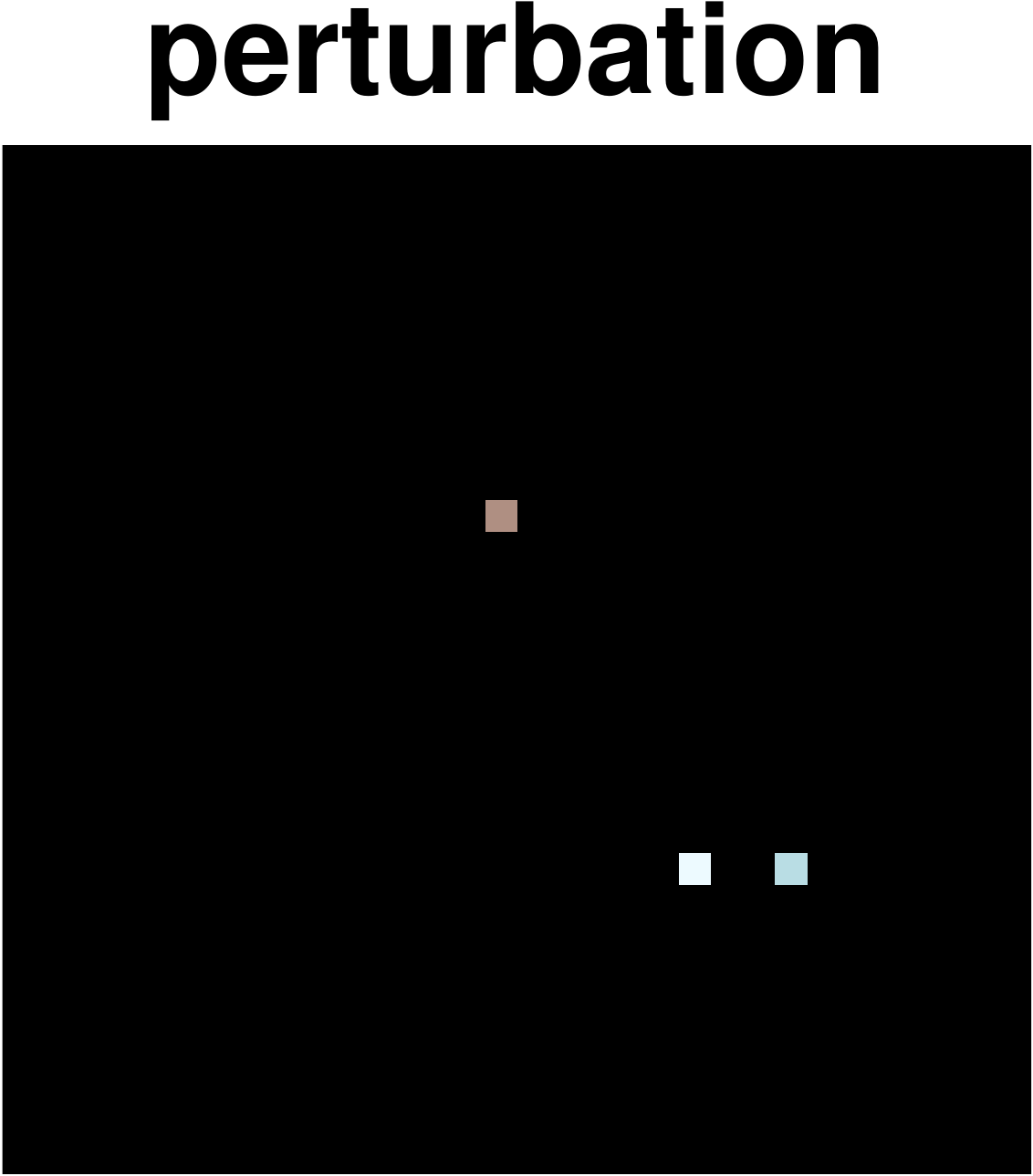}&
		\includegraphics[width=0.2\columnwidth]{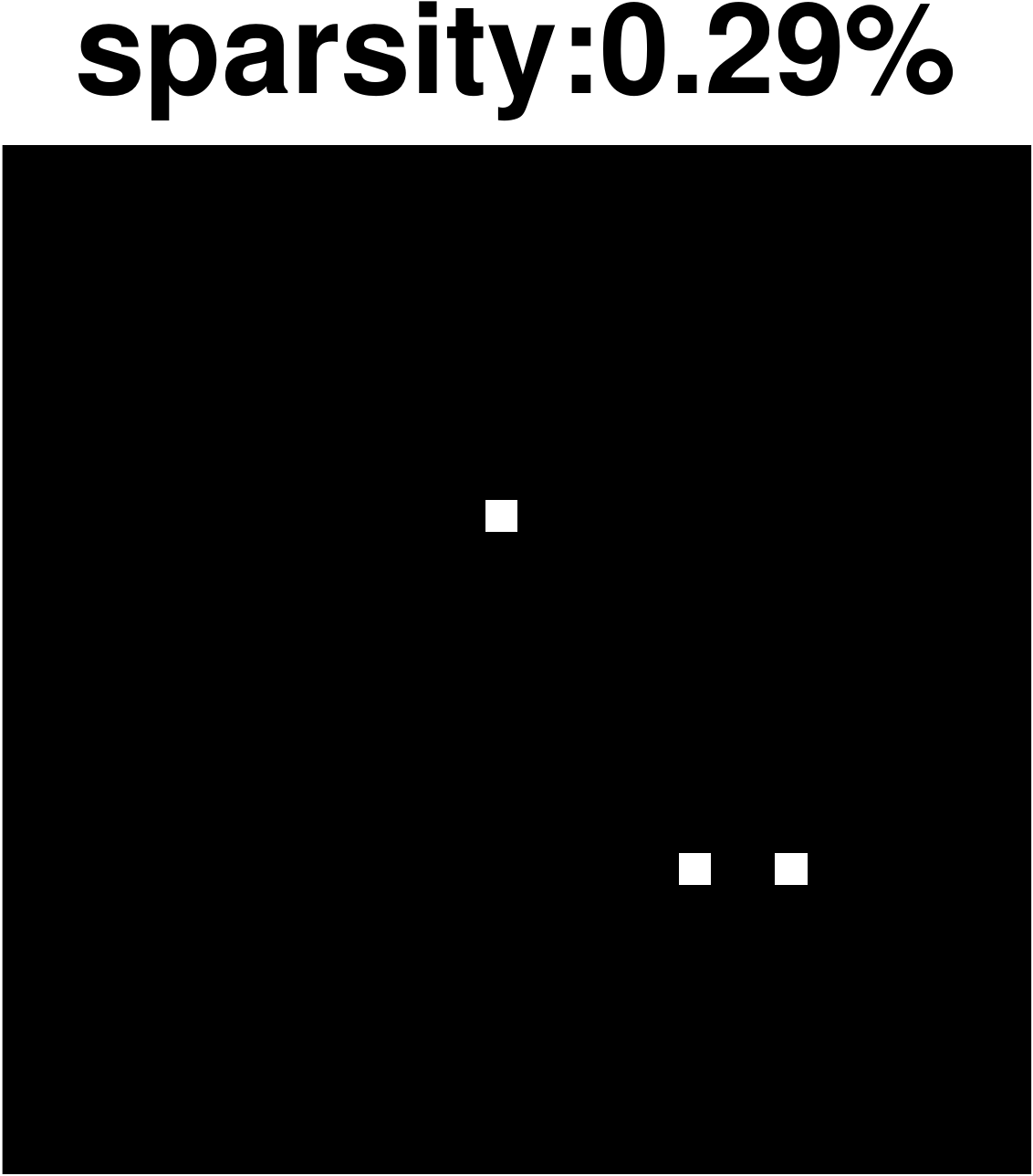}&
		
		\includegraphics[width=0.2\columnwidth]{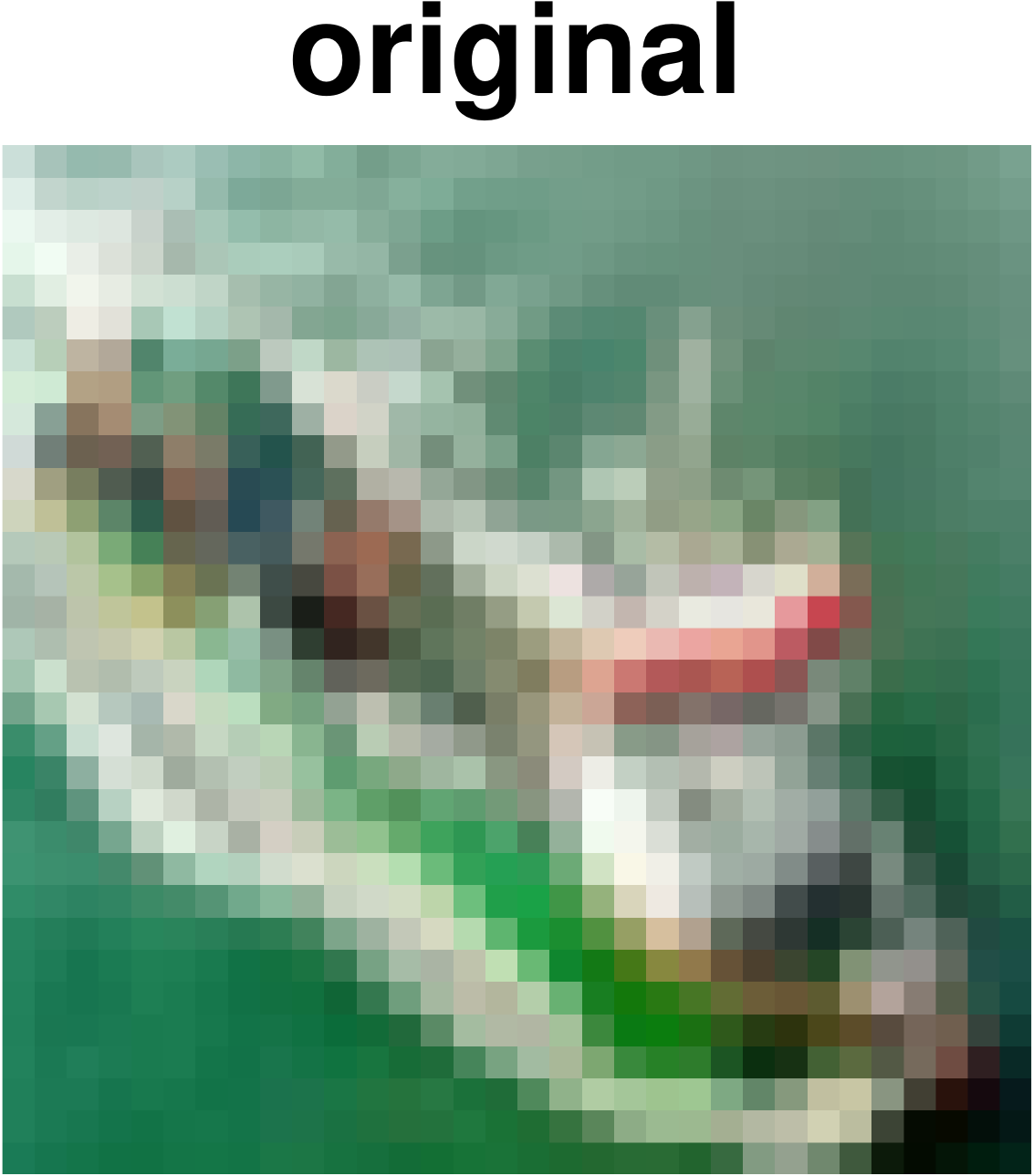}&
		\includegraphics[width=0.2\columnwidth]{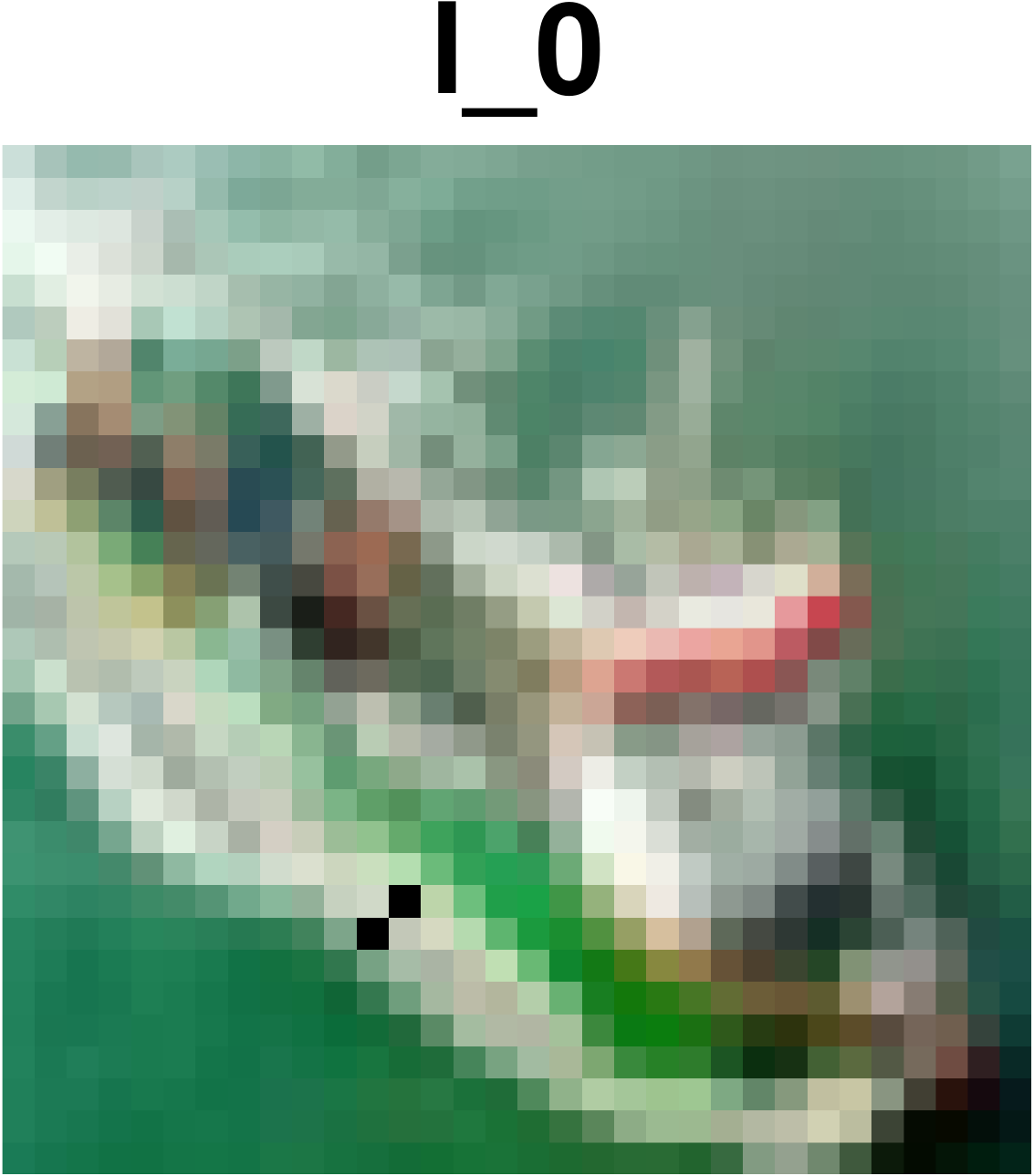}&
		\includegraphics[width=0.2\columnwidth]{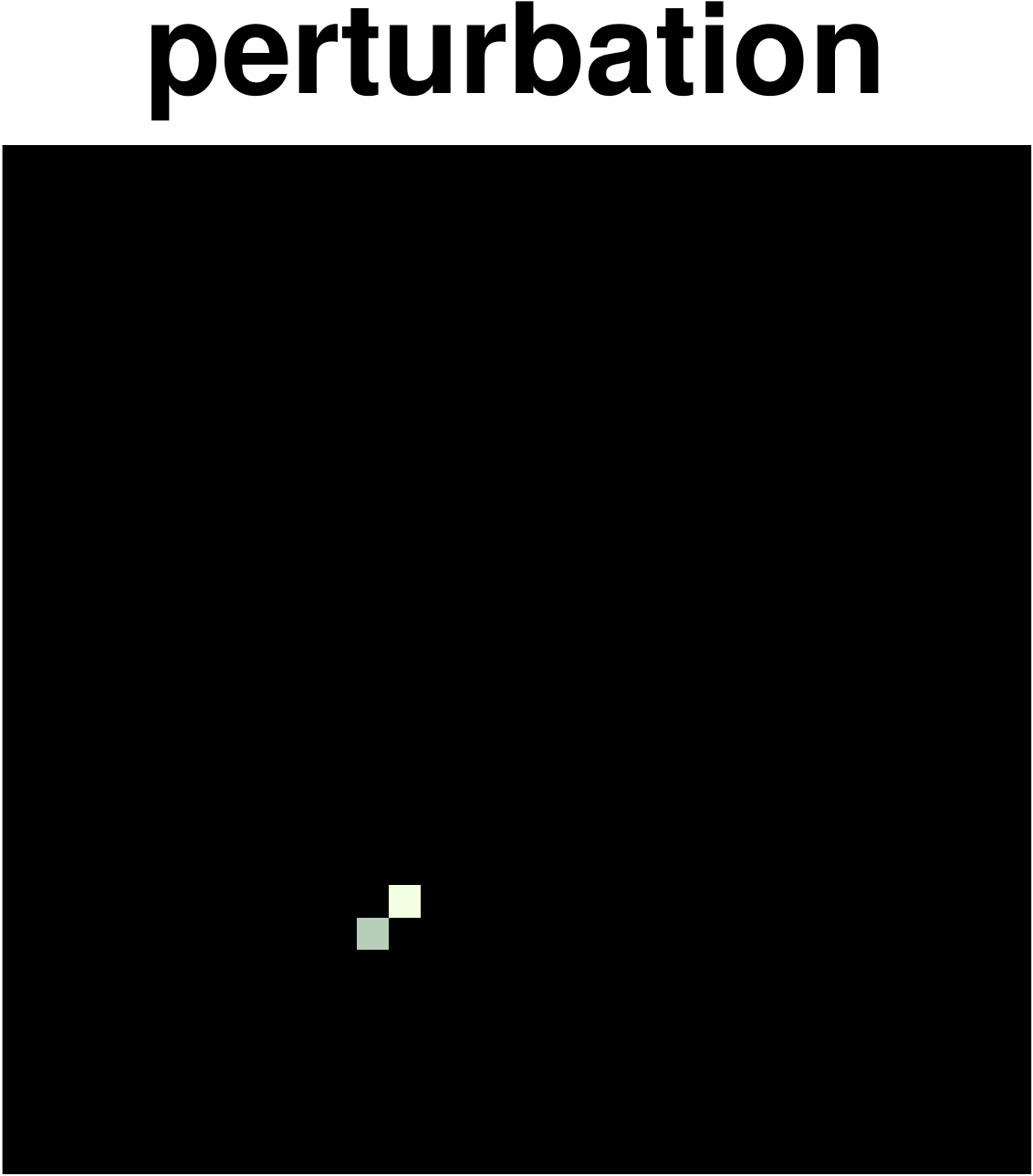}&
		\includegraphics[width=0.2\columnwidth]{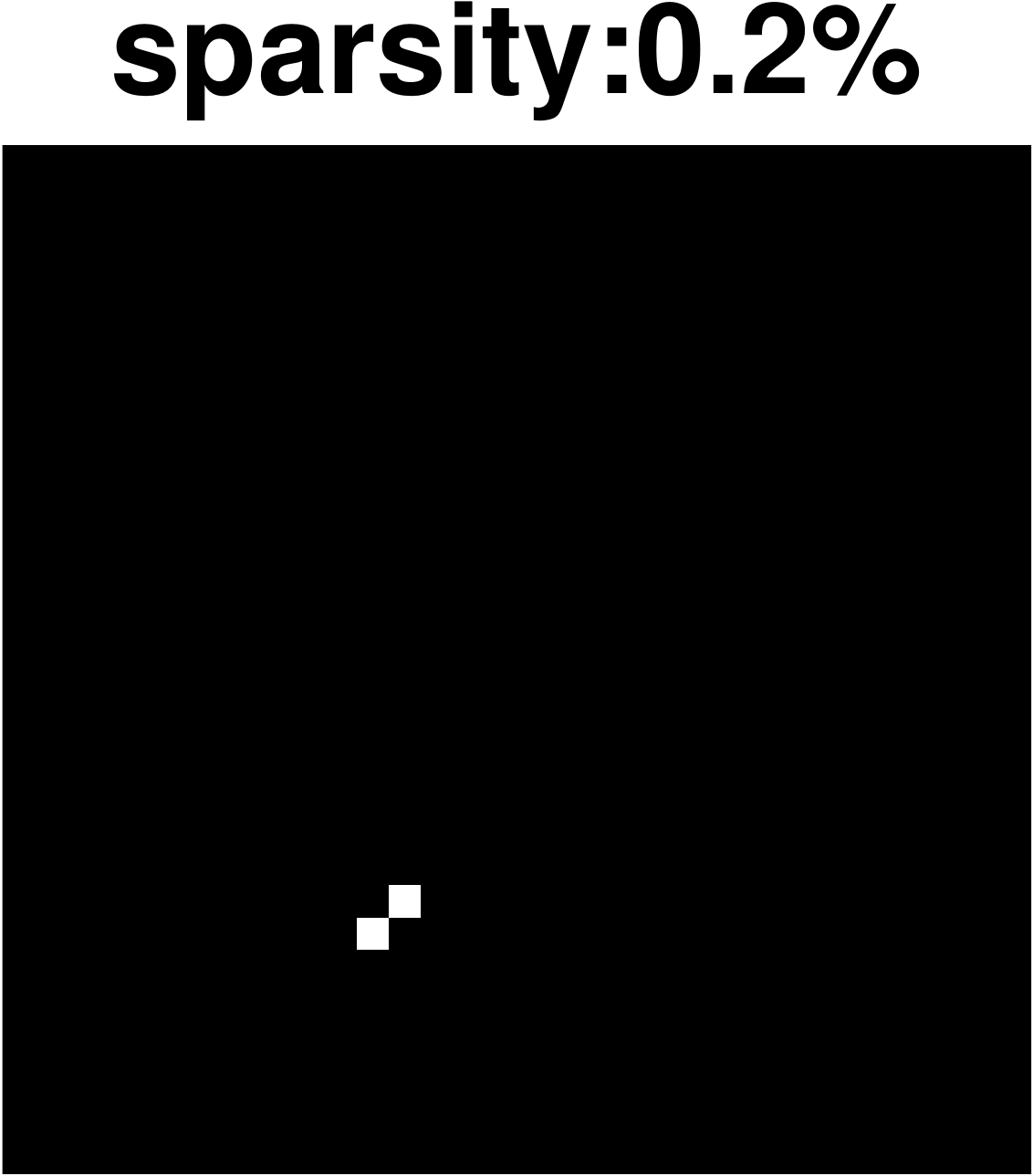}\\
		
		& \includegraphics[width=0.2\columnwidth]{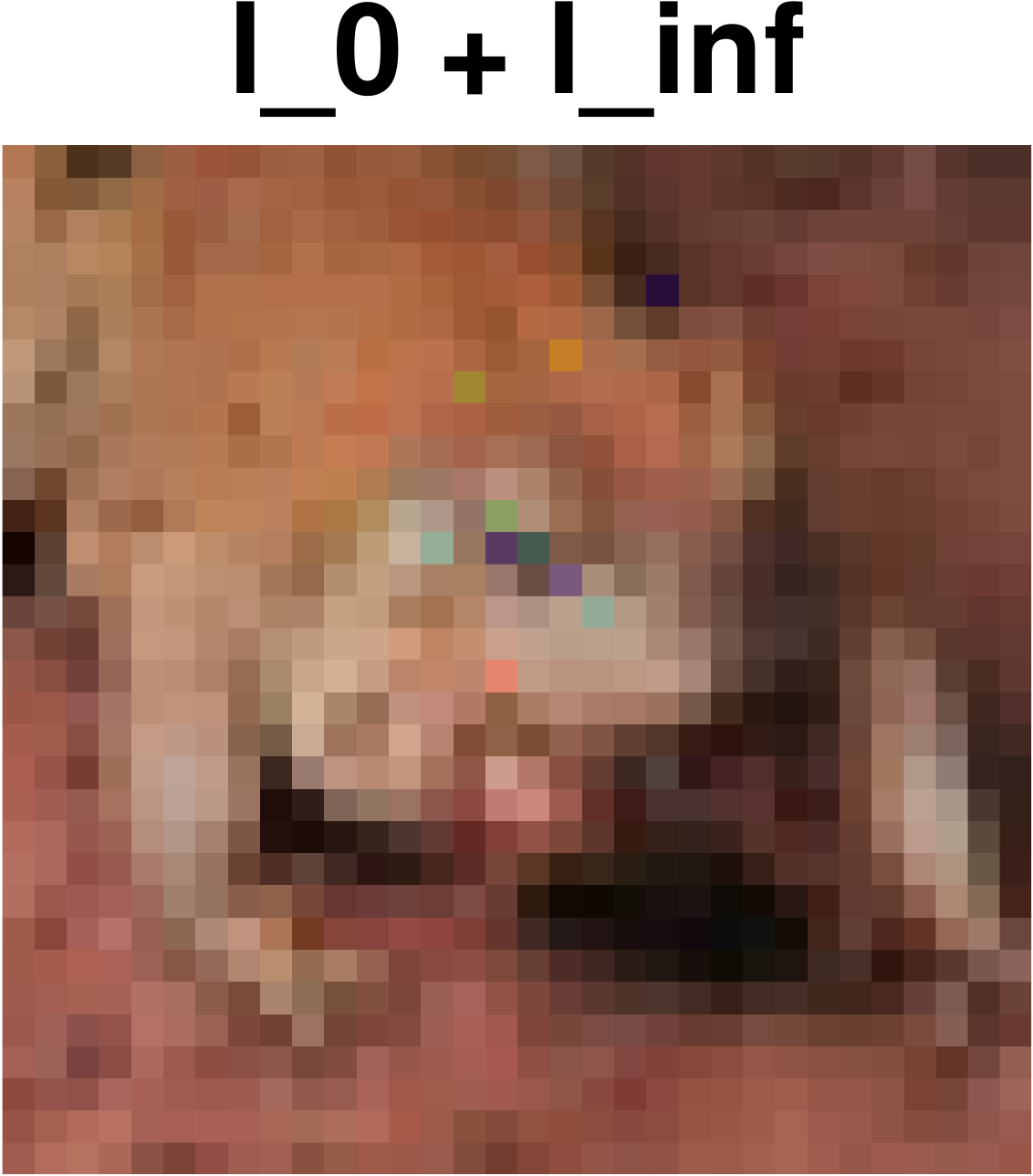}&
		\includegraphics[width=0.2\columnwidth]{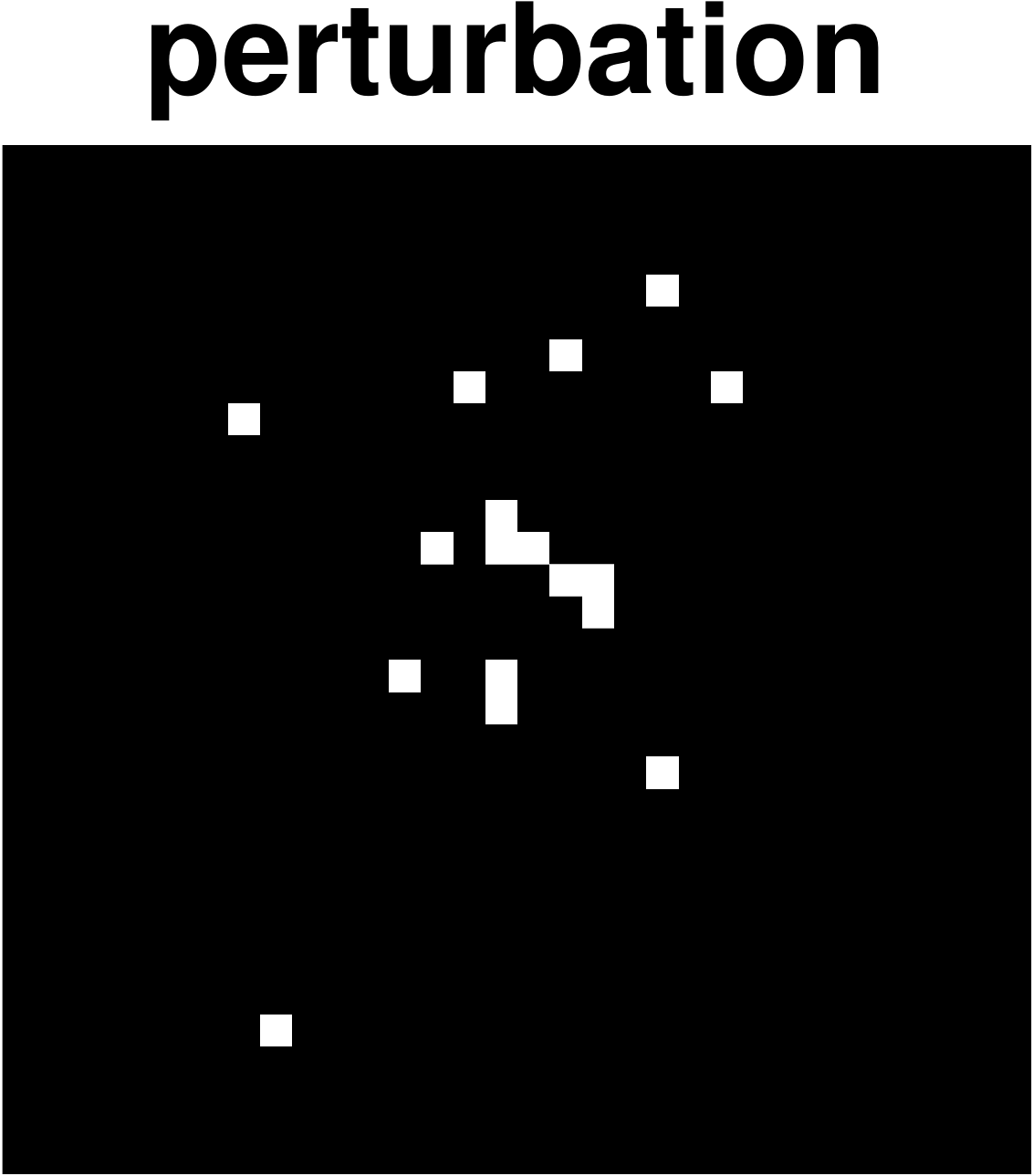}&
		\includegraphics[width=0.2\columnwidth]{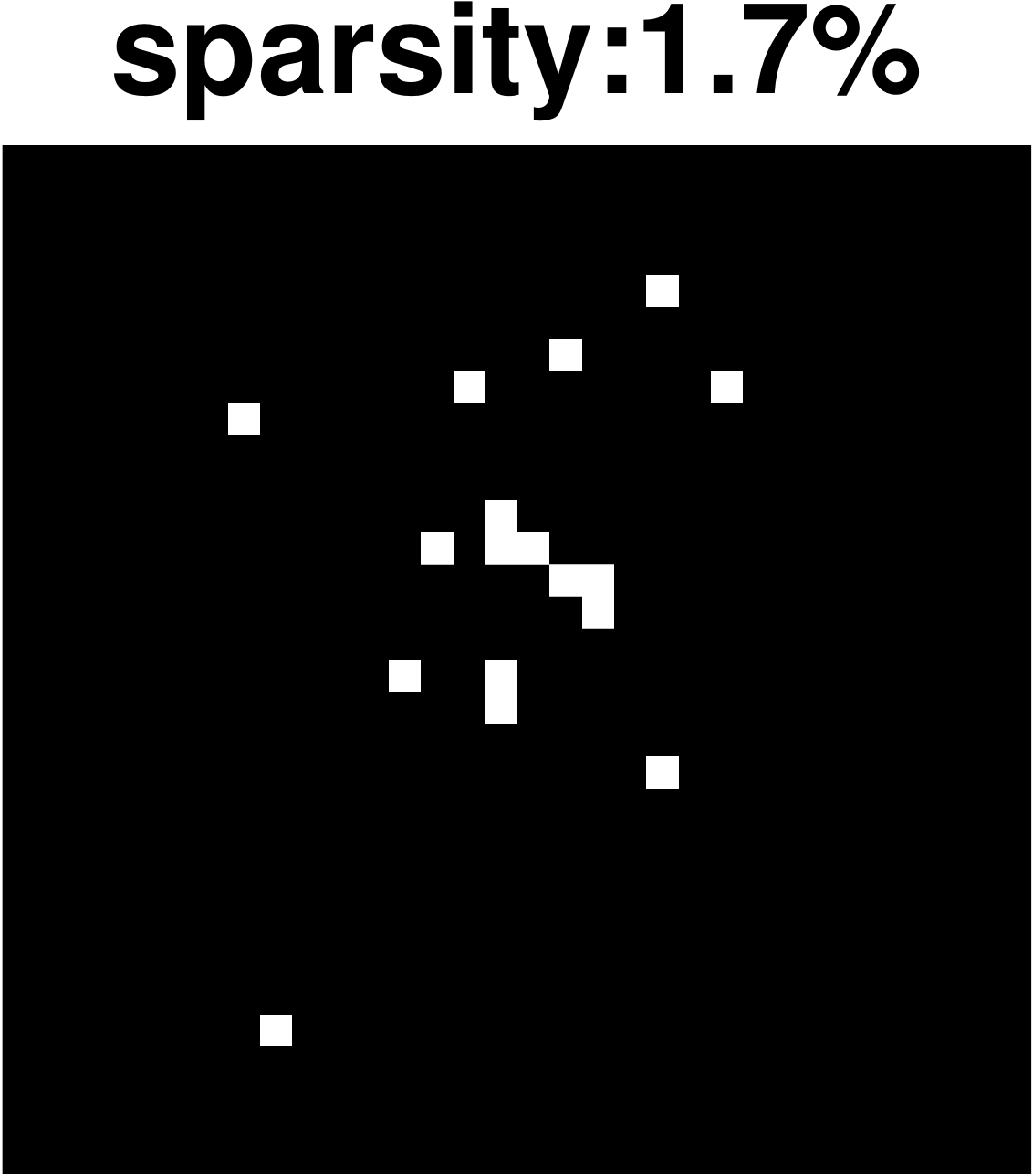}&
		&
		\includegraphics[width=0.2\columnwidth]{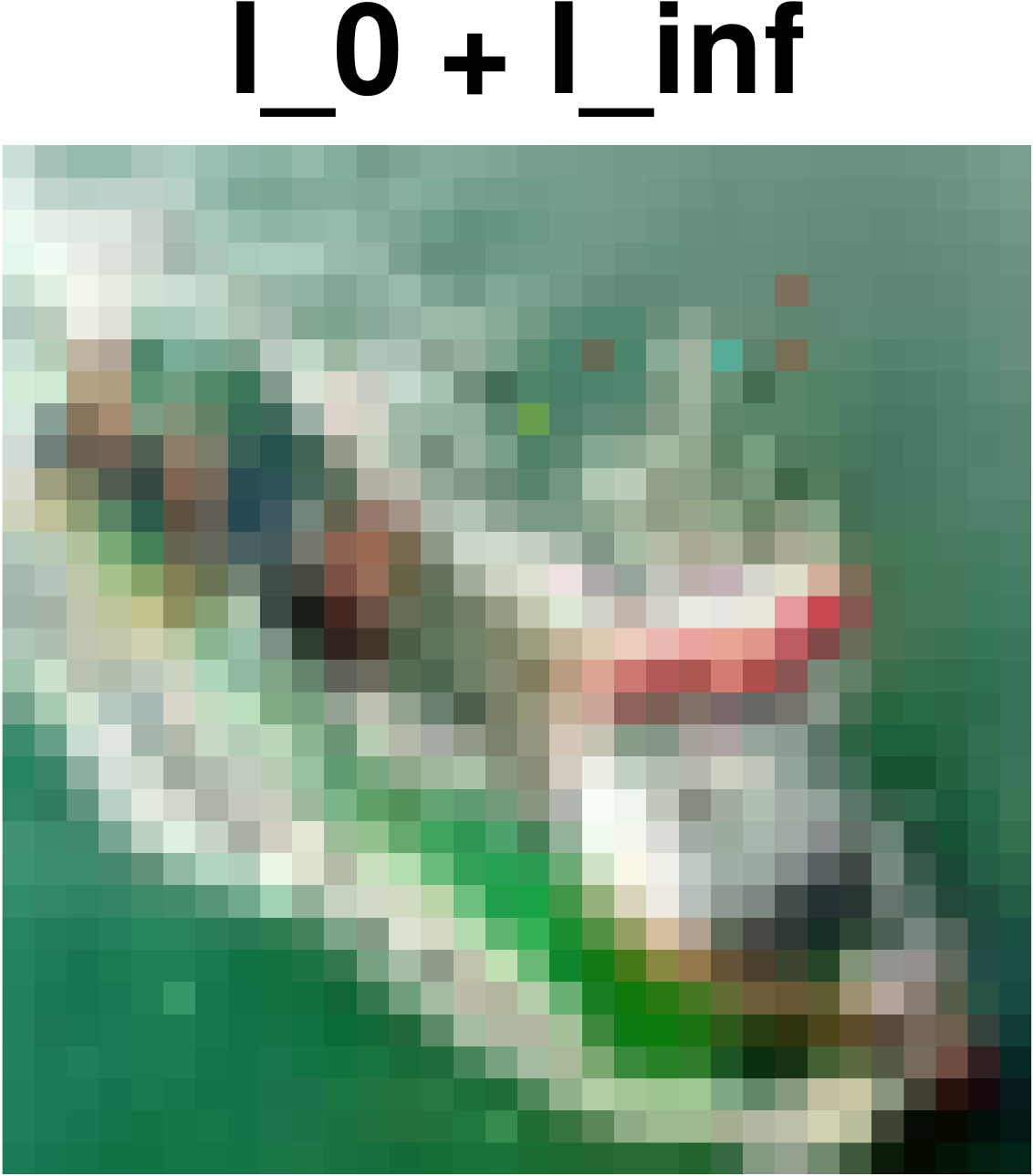}&
		\includegraphics[width=0.2\columnwidth]{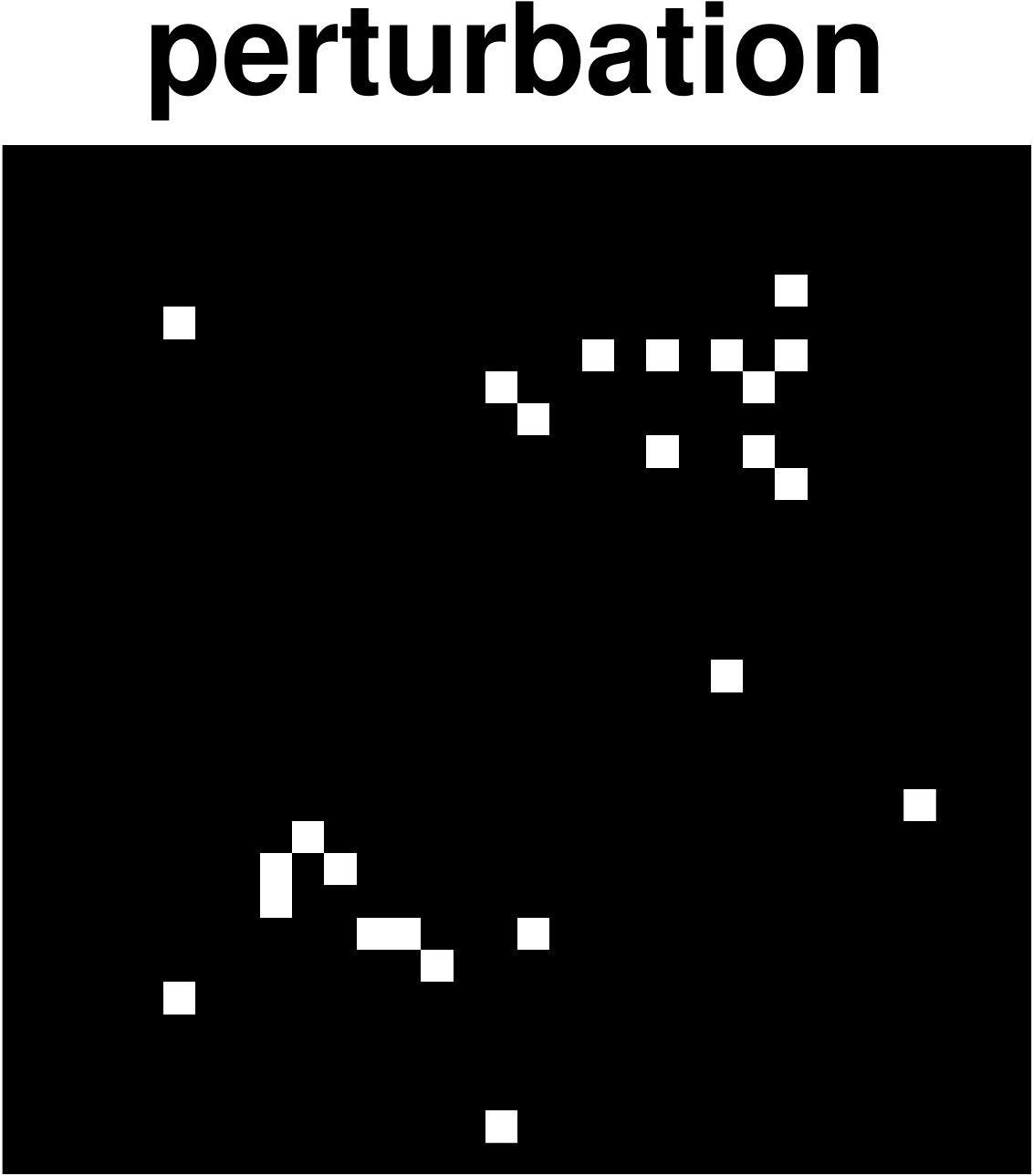}&
		\includegraphics[width=0.2\columnwidth]{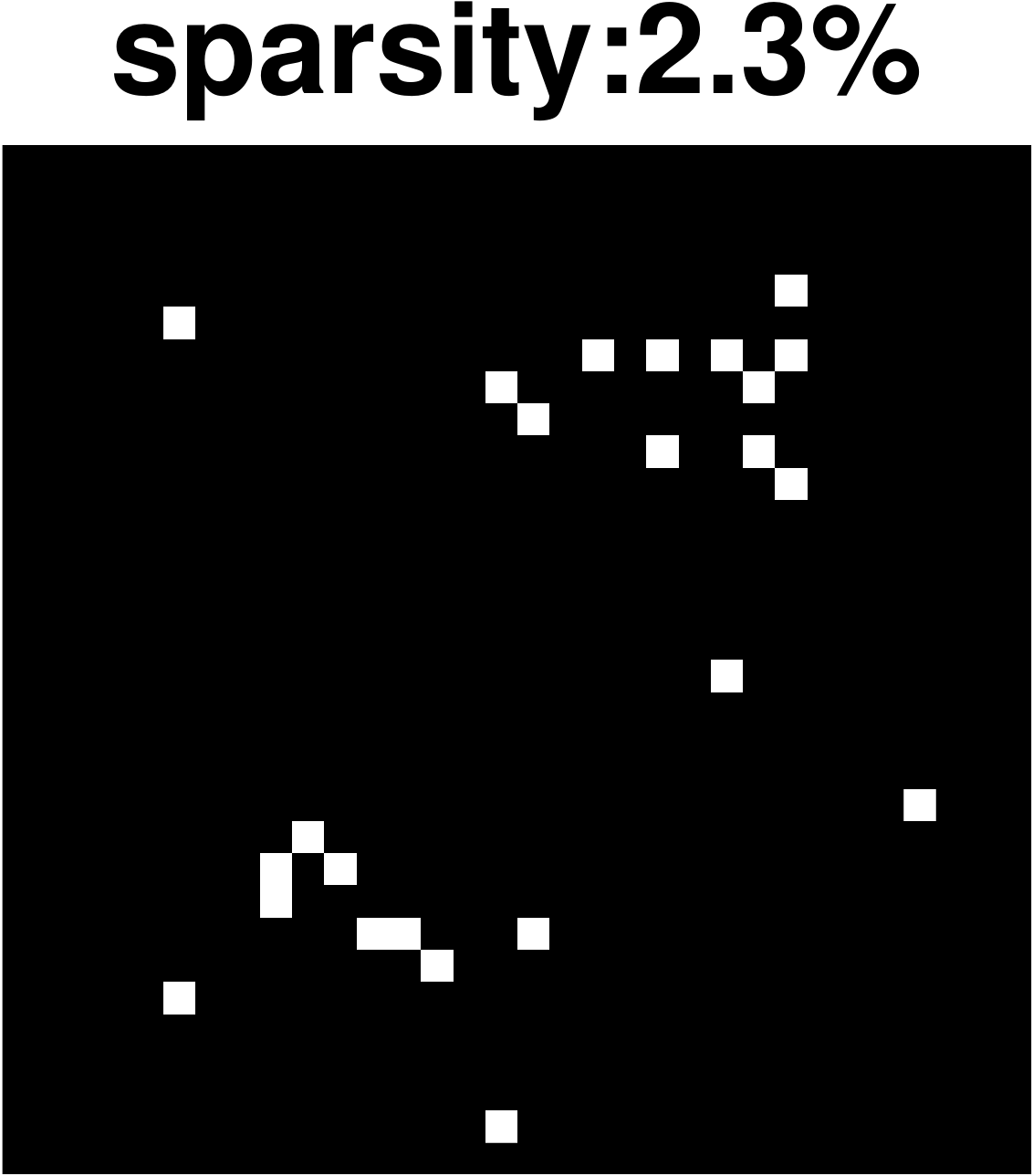}\\
		
		\includegraphics[width=0.2\columnwidth]{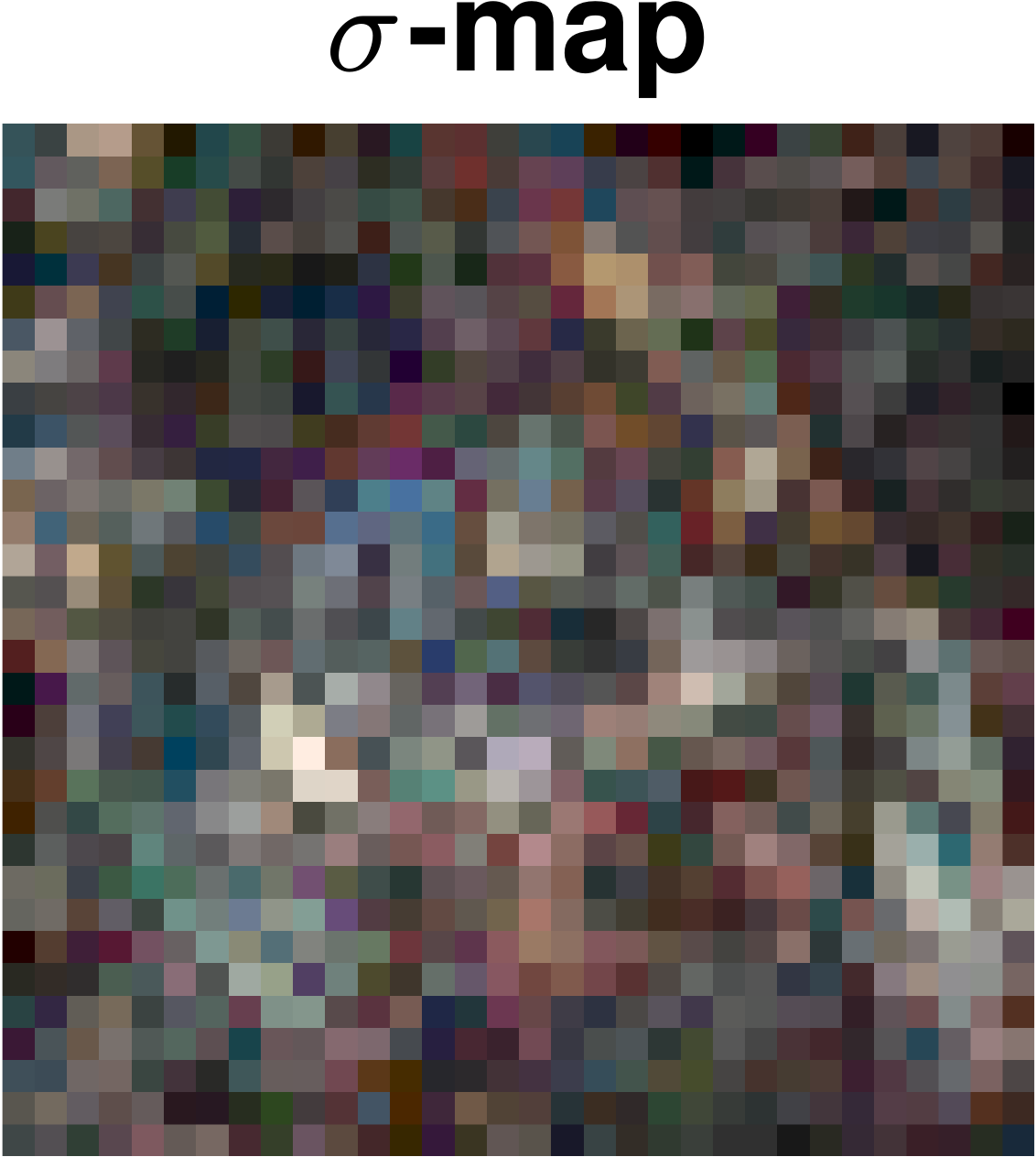}& 
		\includegraphics[width=0.2\columnwidth]{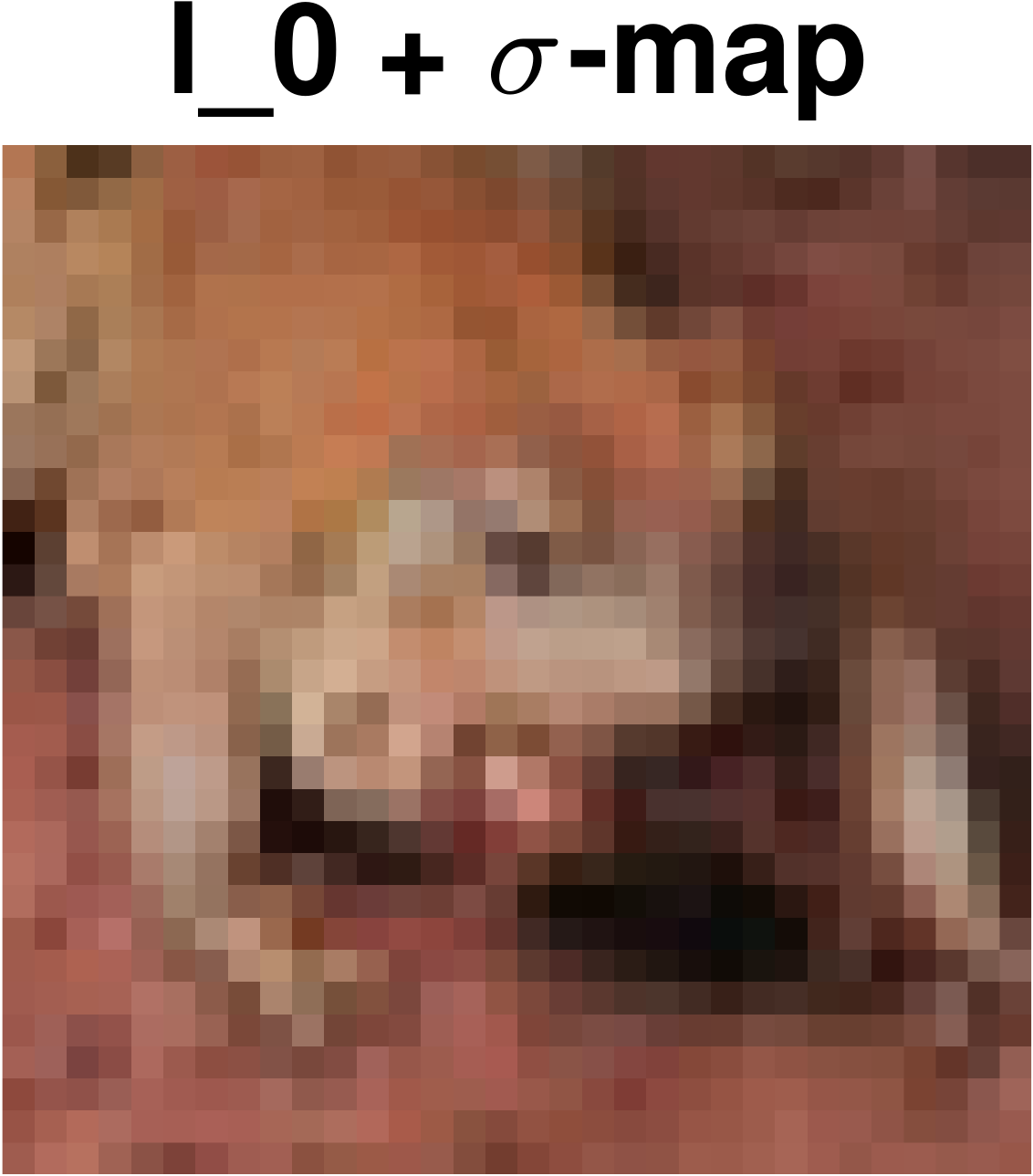}&
		\includegraphics[width=0.2\columnwidth]{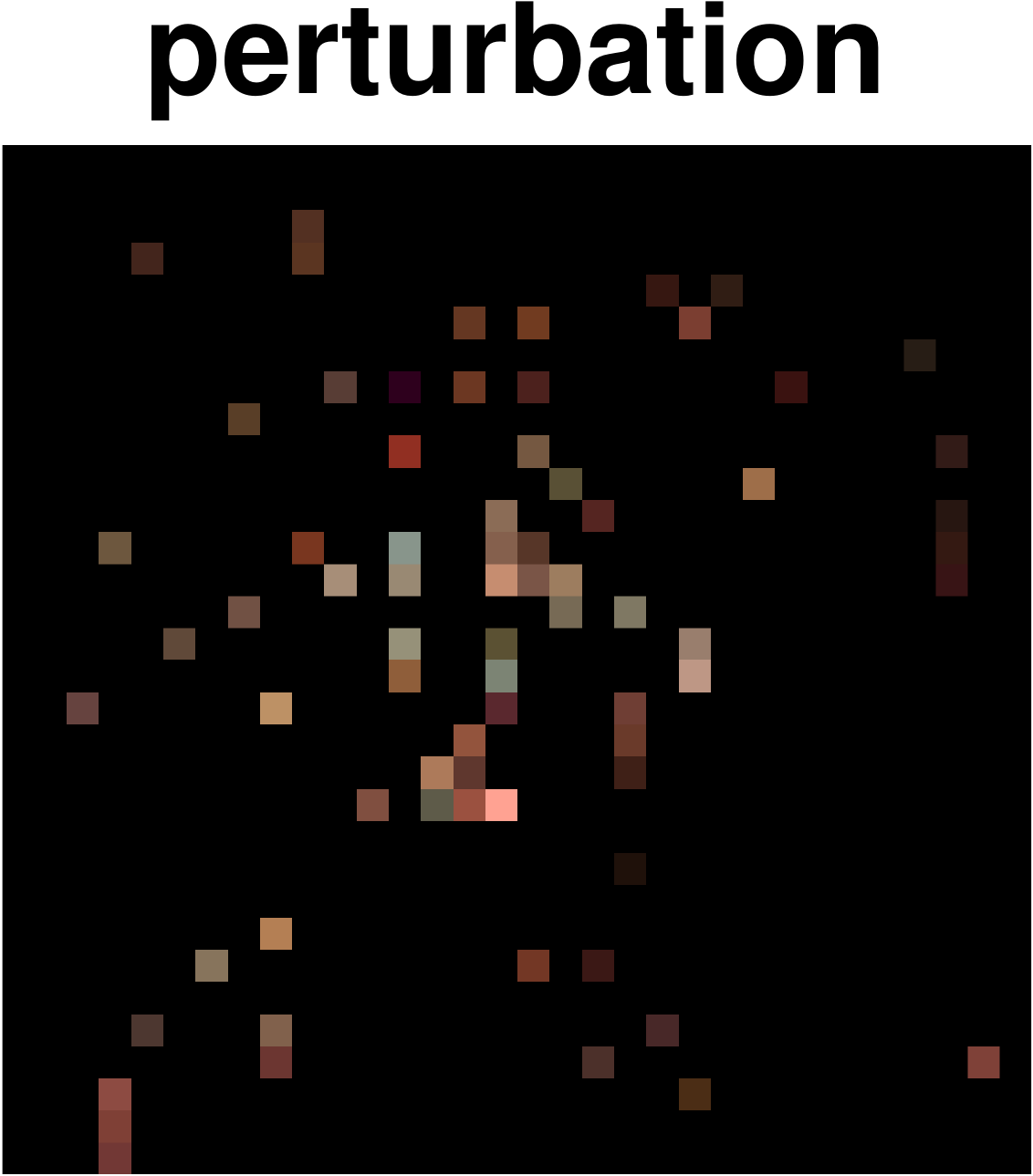}&
		\includegraphics[width=0.2\columnwidth]{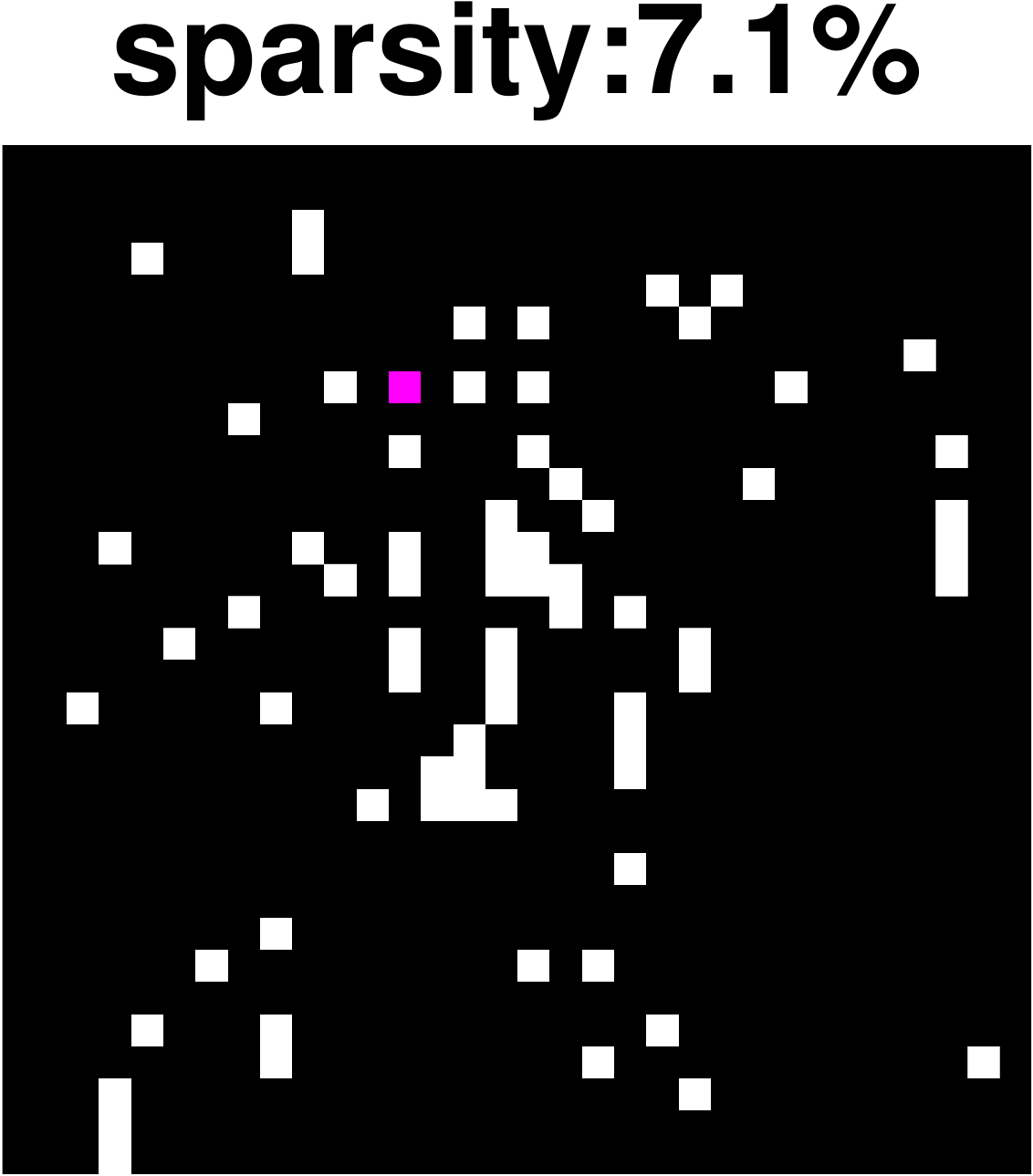}&
		
		\includegraphics[width=0.2\columnwidth]{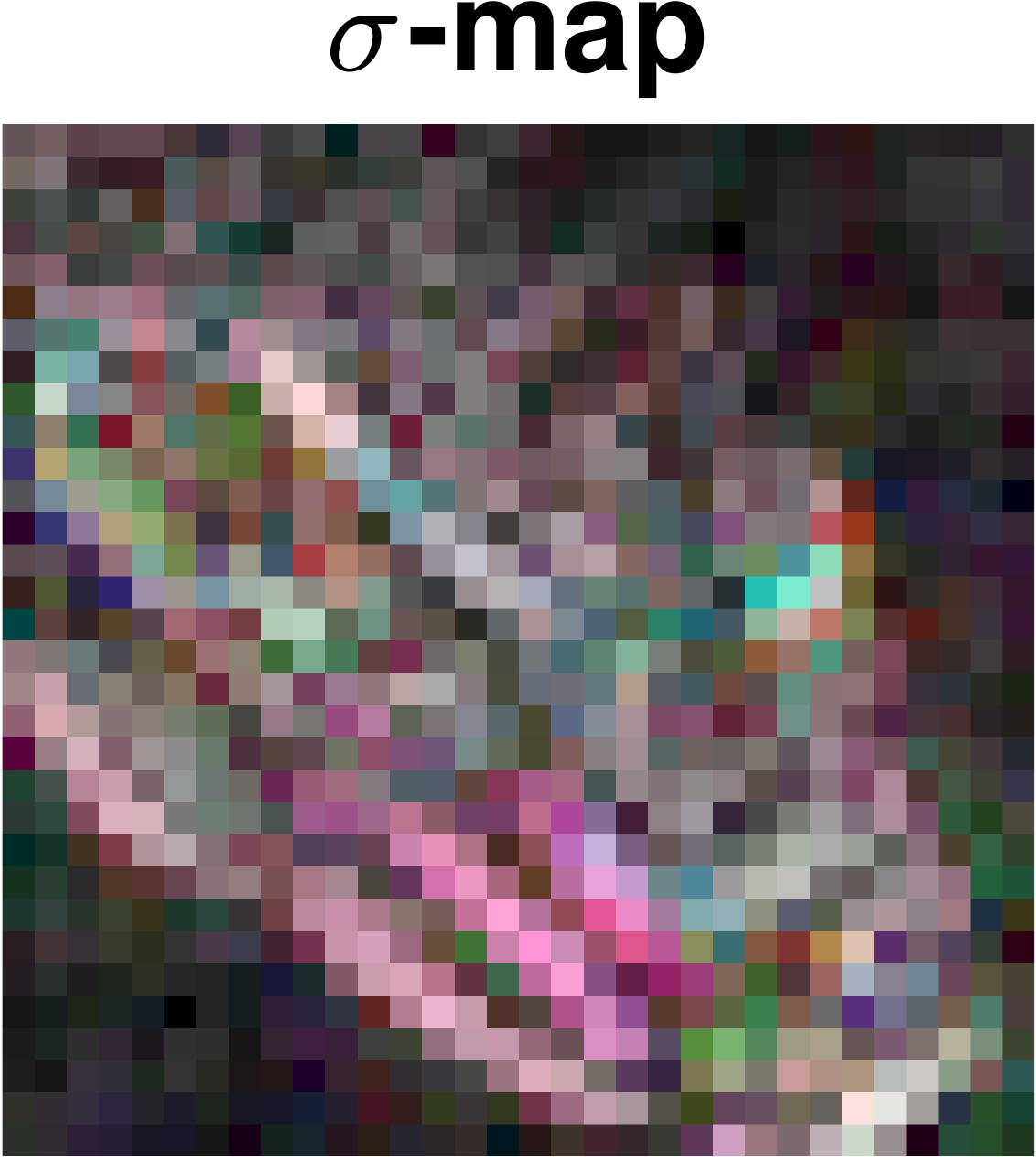}& 
		\includegraphics[width=0.2\columnwidth]{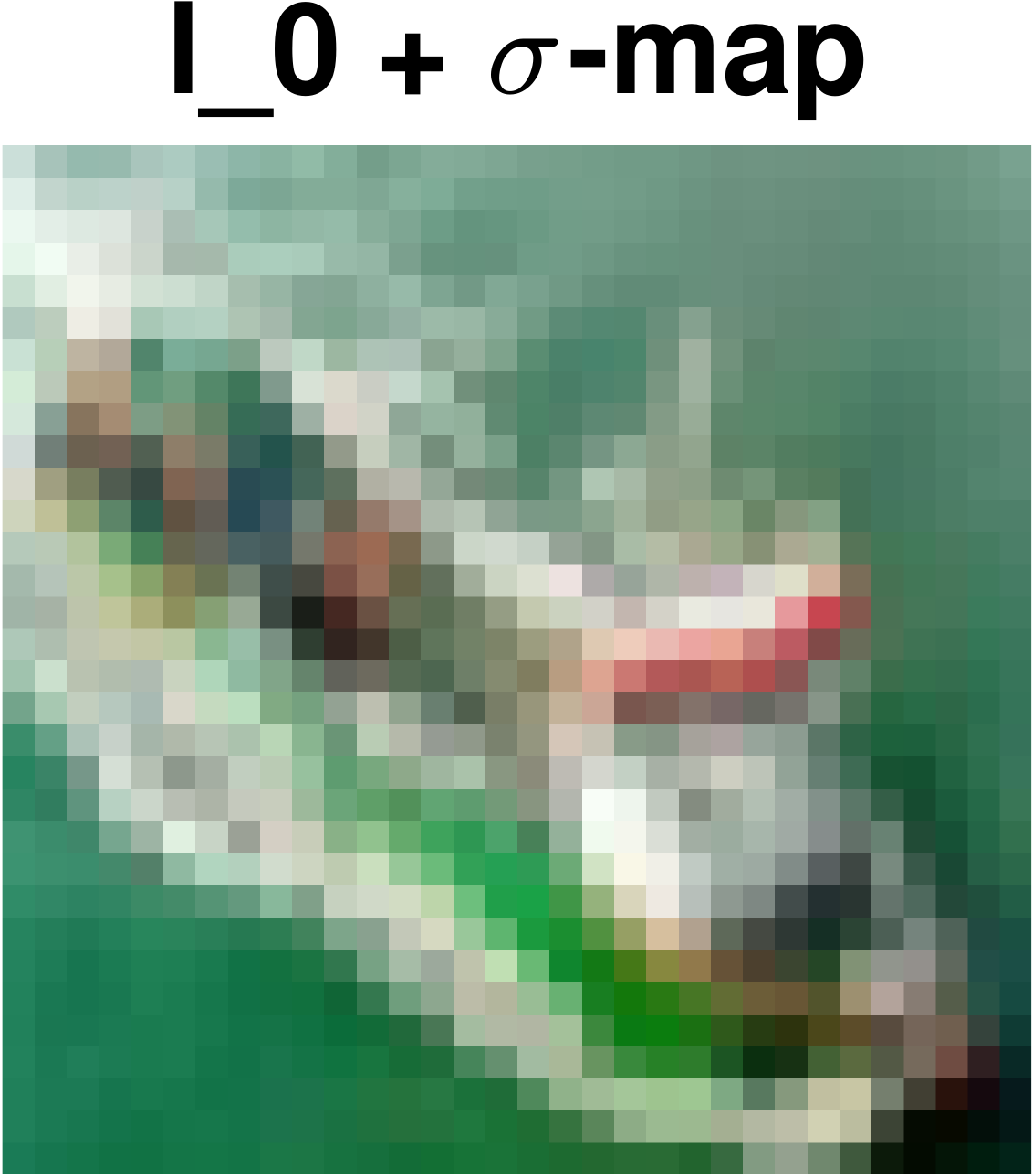}&
		\includegraphics[width=0.2\columnwidth]{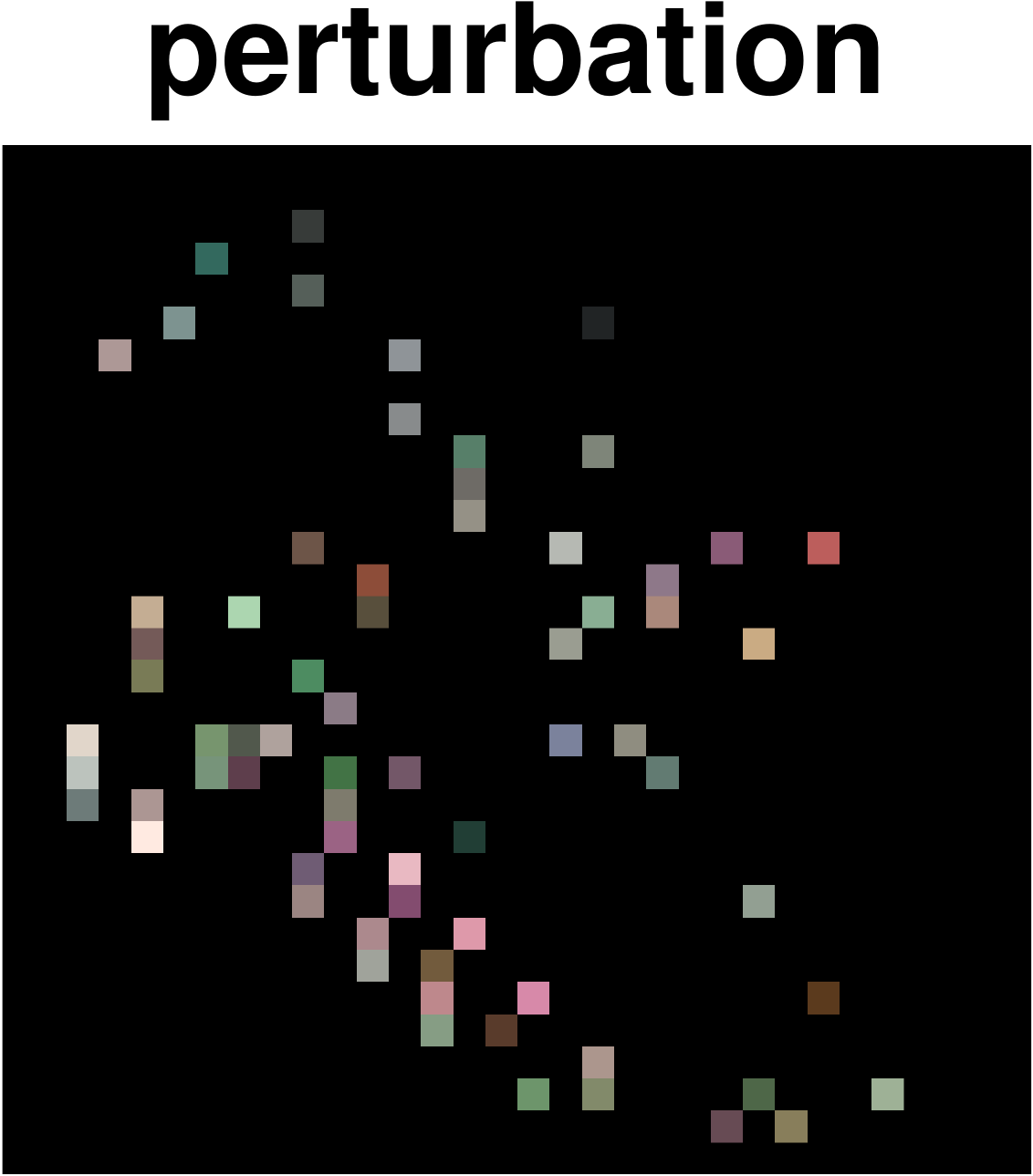}&
		\includegraphics[width=0.2\columnwidth]{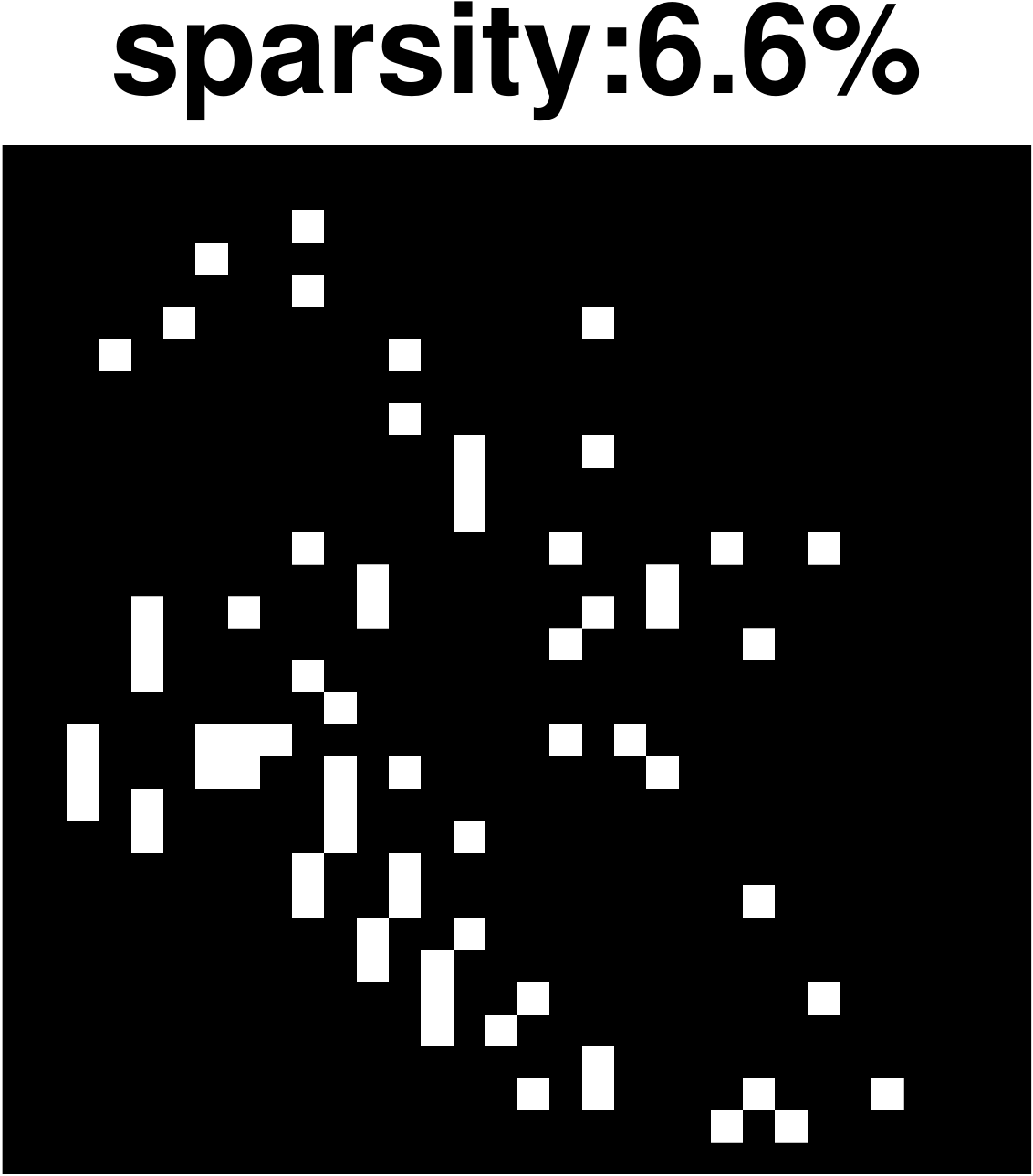} \\
		
		%% im44 frog, dog, dog, dog
		%% im88 horse, cat, deer, cat

	\end{tabular}
	\caption{\textbf{Different attacks on CIFAR-10.} We illustrate the differences of the adversarial examples (second column) found by CornerSearch ($l_0$), $l_0+l_\infty$-attack and $\sigma$-CornerSearch, respectively first, second and third row. The third column shows the adversarial perturbations rescaled to $[0,1]$, the fourth the map of the modified pixels (\textit{sparsity} column). The original image can be found top left and the RGB representation of the $\sigma$-map, rescaled so that $\max_{i,j}\sigma_{ij}=1$, bottom left.}\label{fig:imp_CIFAR-10_app}
\end{figure*}

\begin{figure*}[p]
	\centering
	\begin{tabular}{c c  c c| c c c c}
		\includegraphics[width=0.2\columnwidth]{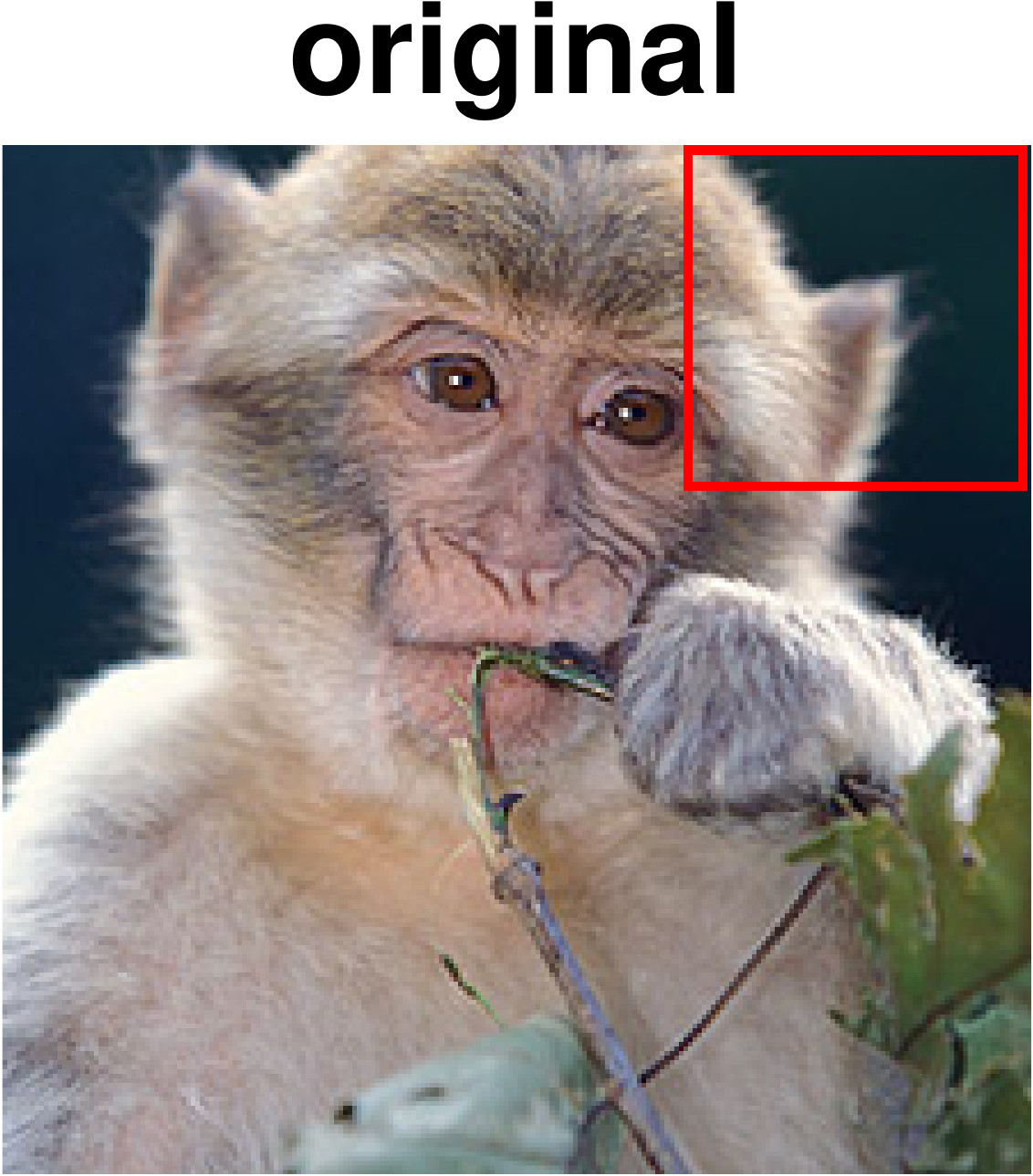}&
		\includegraphics[width=0.2\columnwidth]{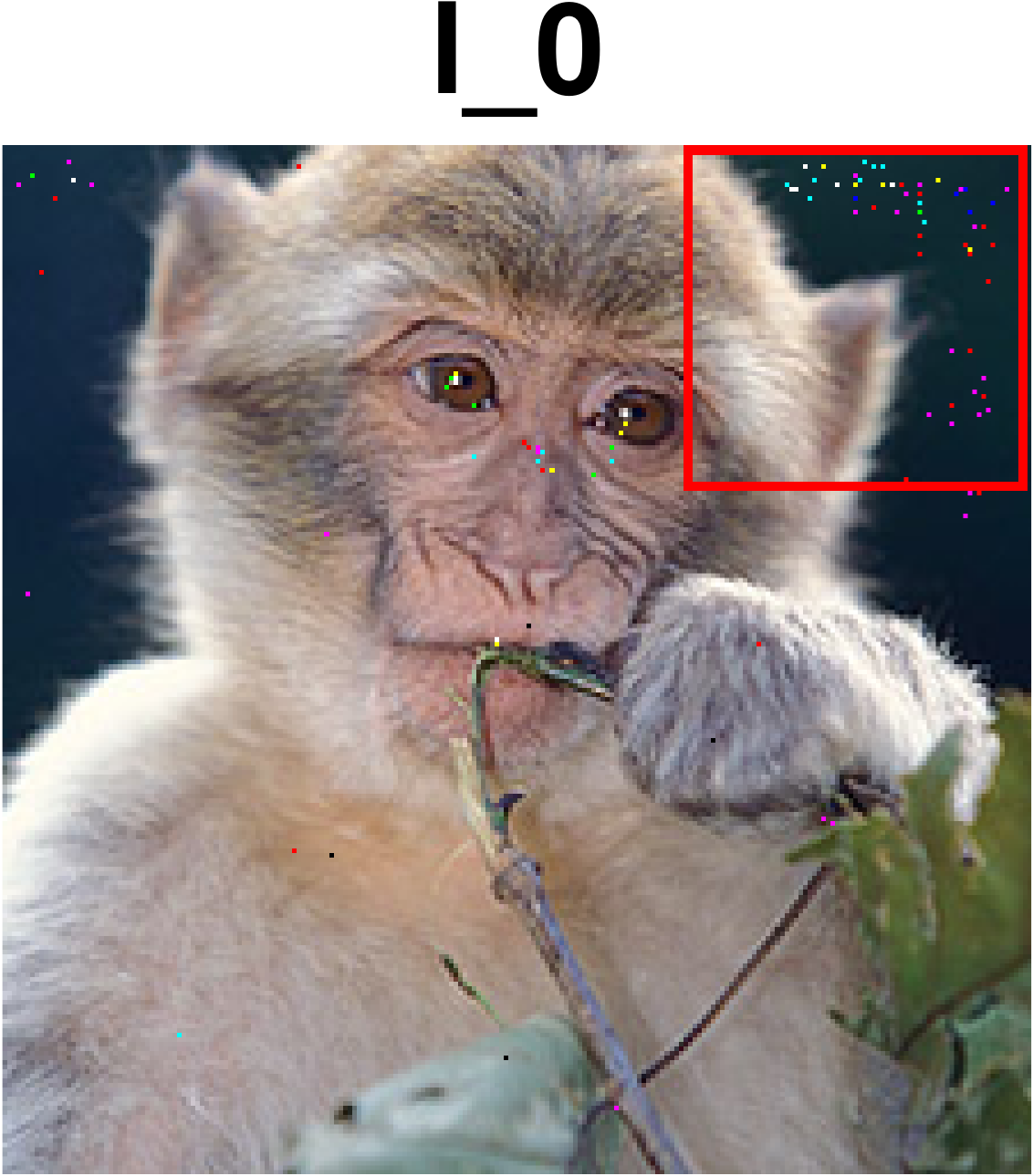}&
		\includegraphics[width=0.2\columnwidth]{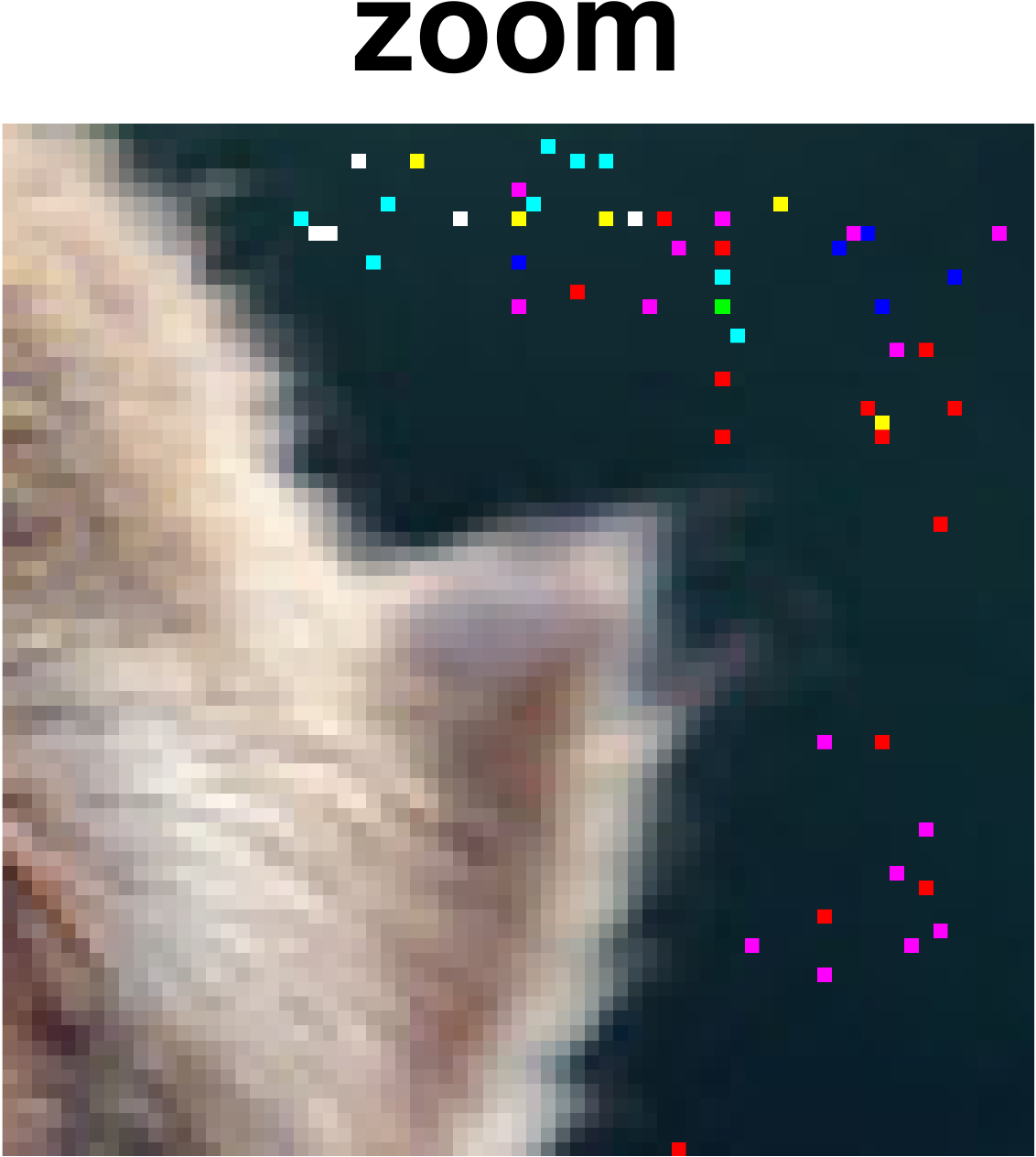}&
		\includegraphics[width=0.2\columnwidth]{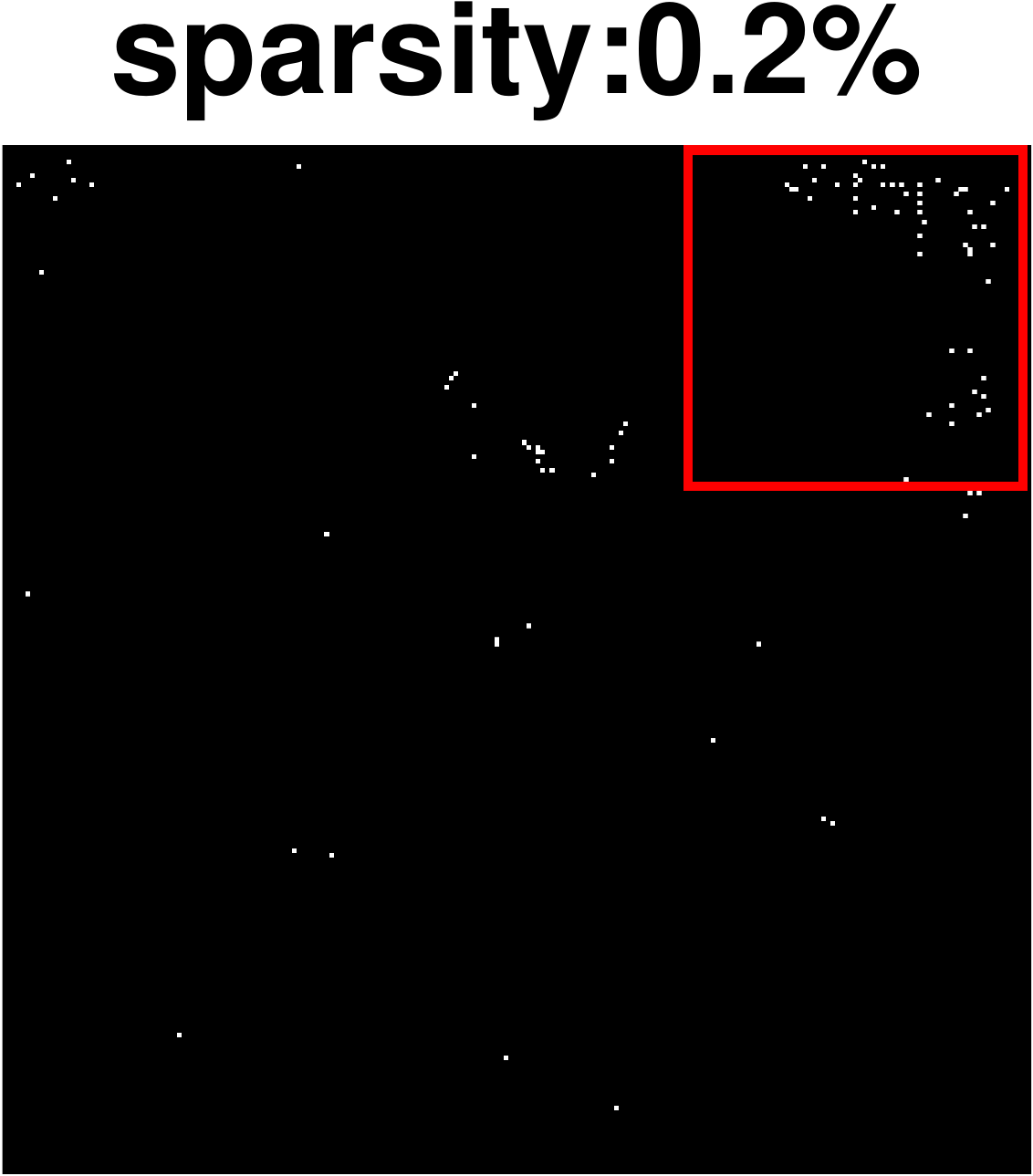}&
		
		\includegraphics[width=0.2\columnwidth]{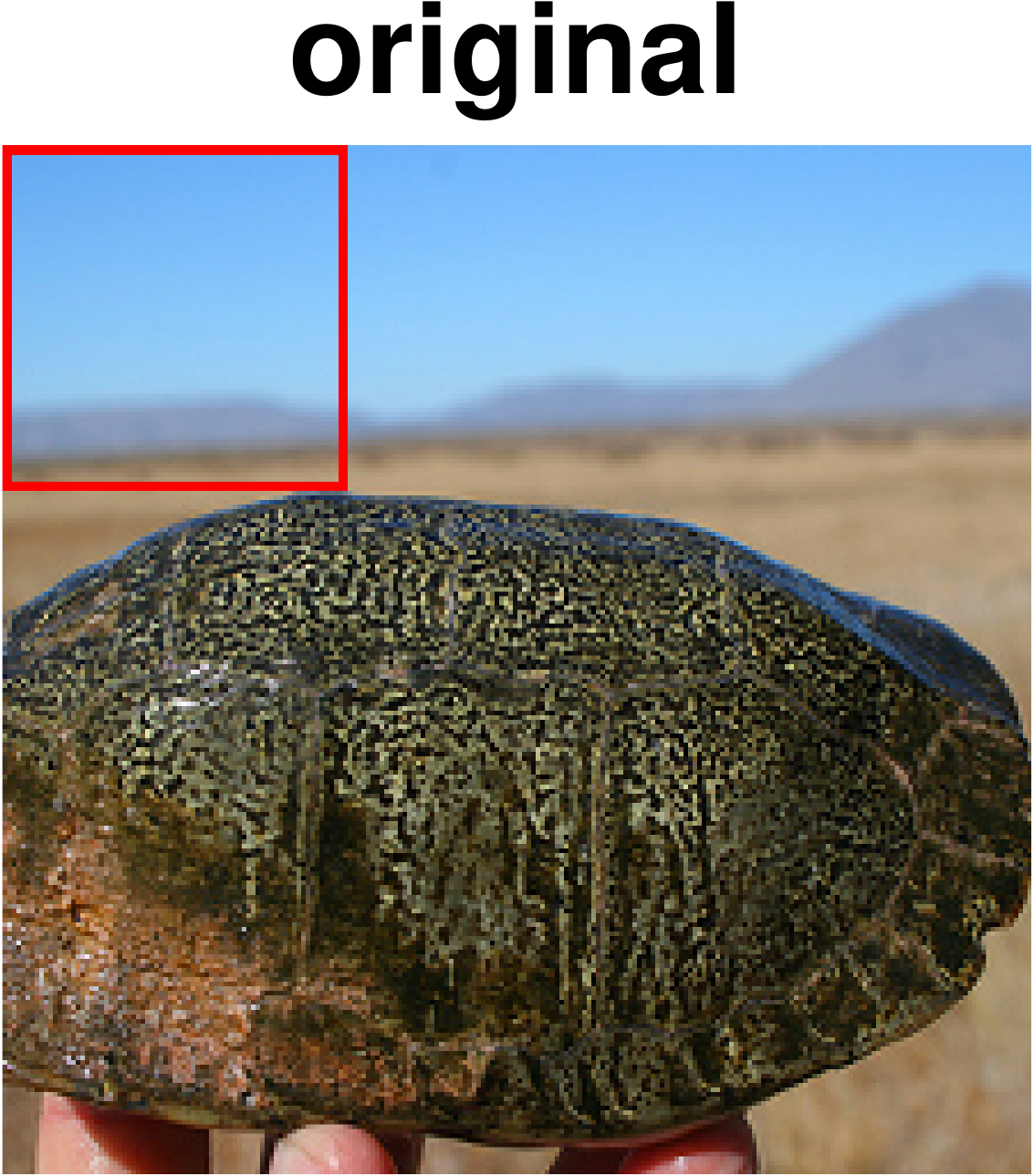}&
		\includegraphics[width=0.2\columnwidth]{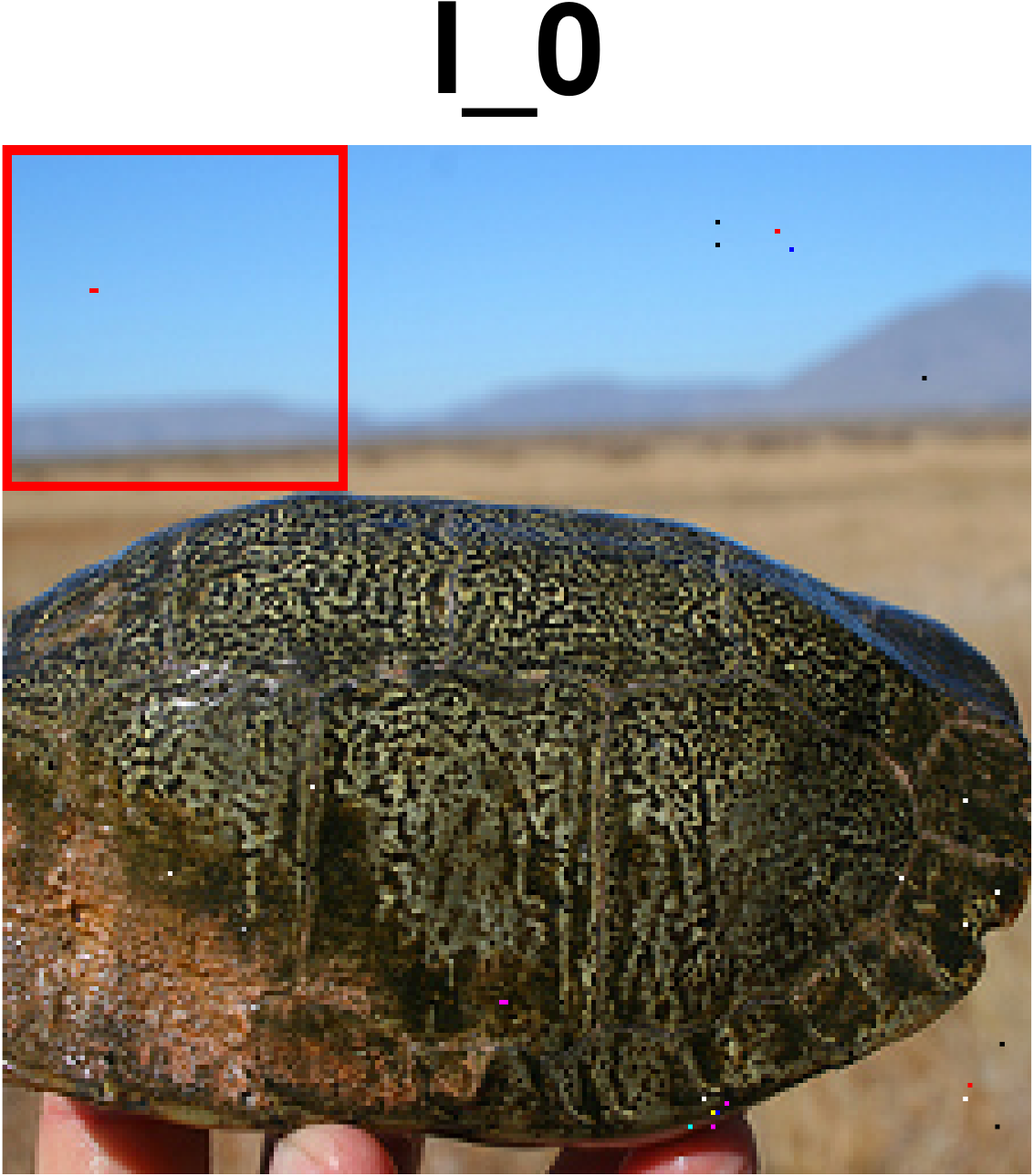}&
		\includegraphics[width=0.2\columnwidth]{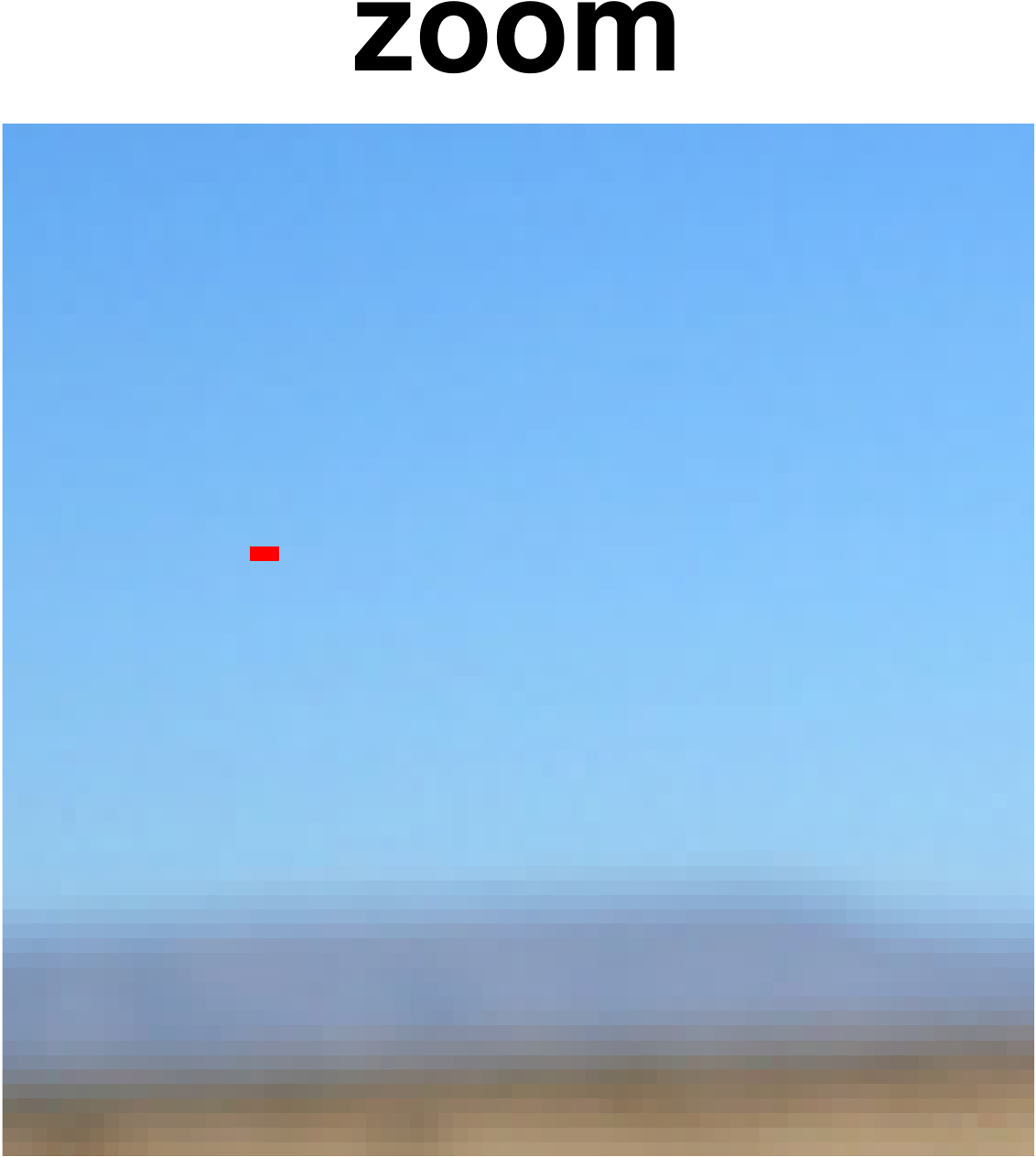}&
		\includegraphics[width=0.2\columnwidth]{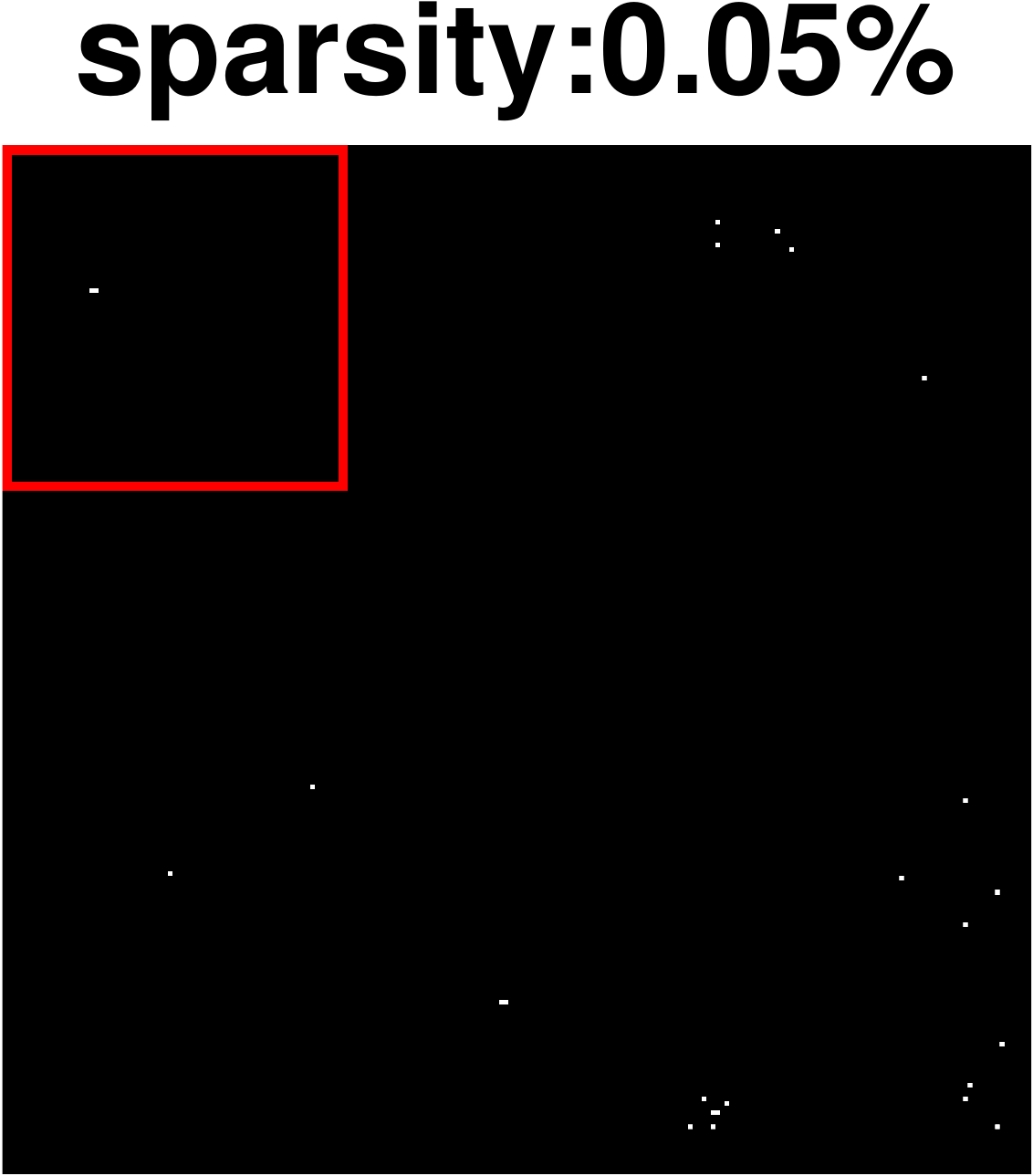}\\
		
		& \includegraphics[width=0.2\columnwidth]{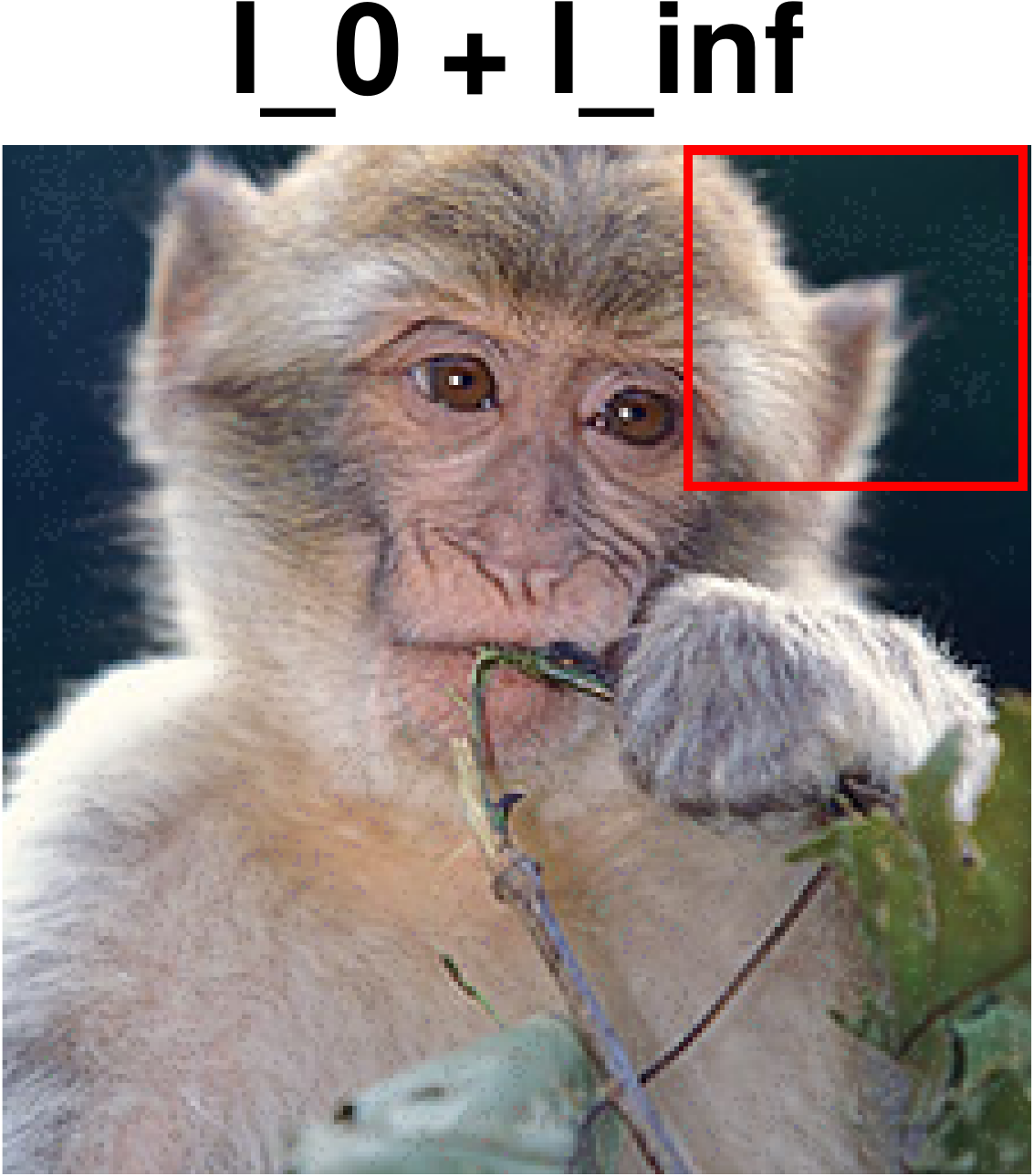}&
		\includegraphics[width=0.2\columnwidth]{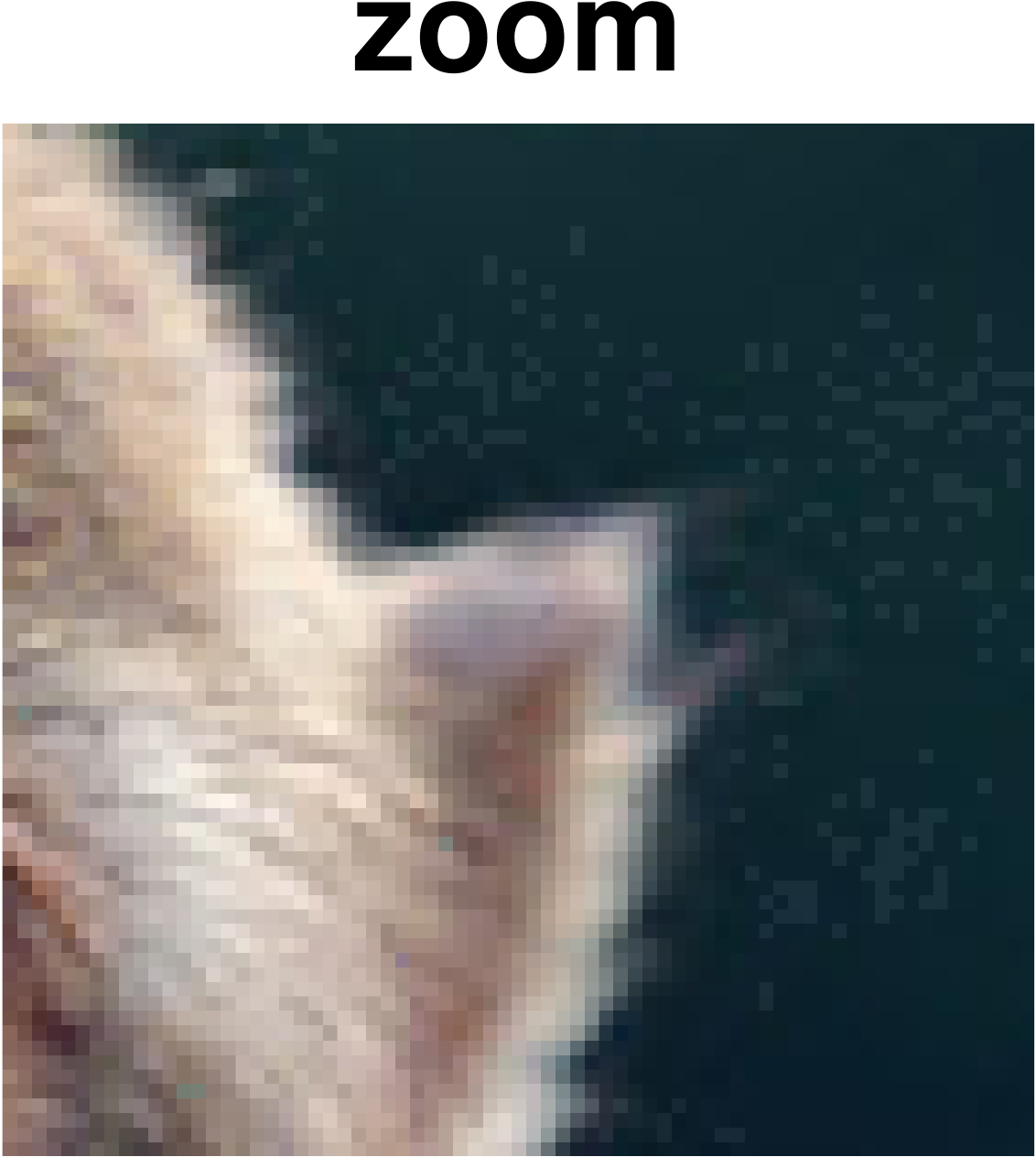}&
		\includegraphics[width=0.2\columnwidth]{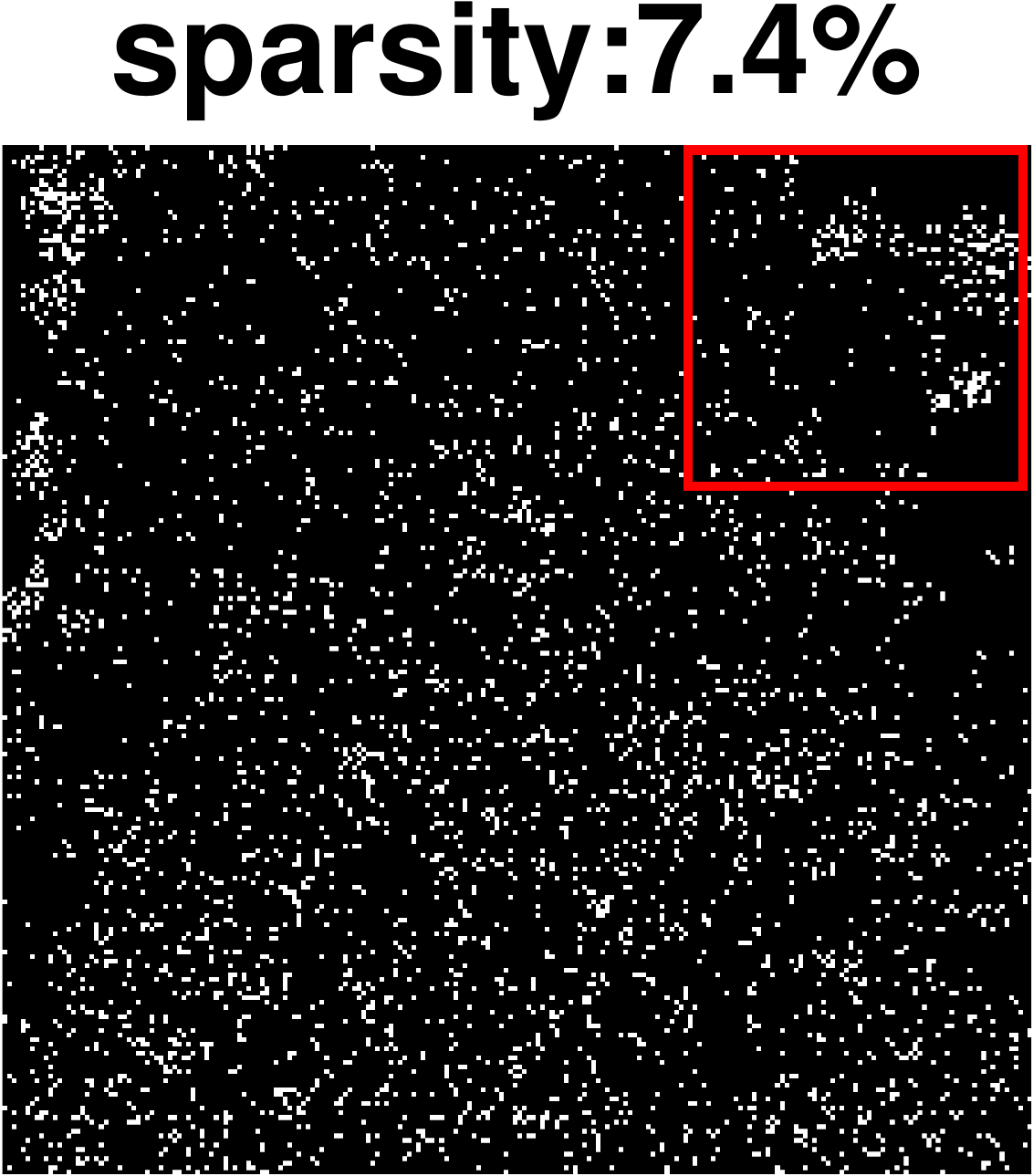}&
		&
		\includegraphics[width=0.2\columnwidth]{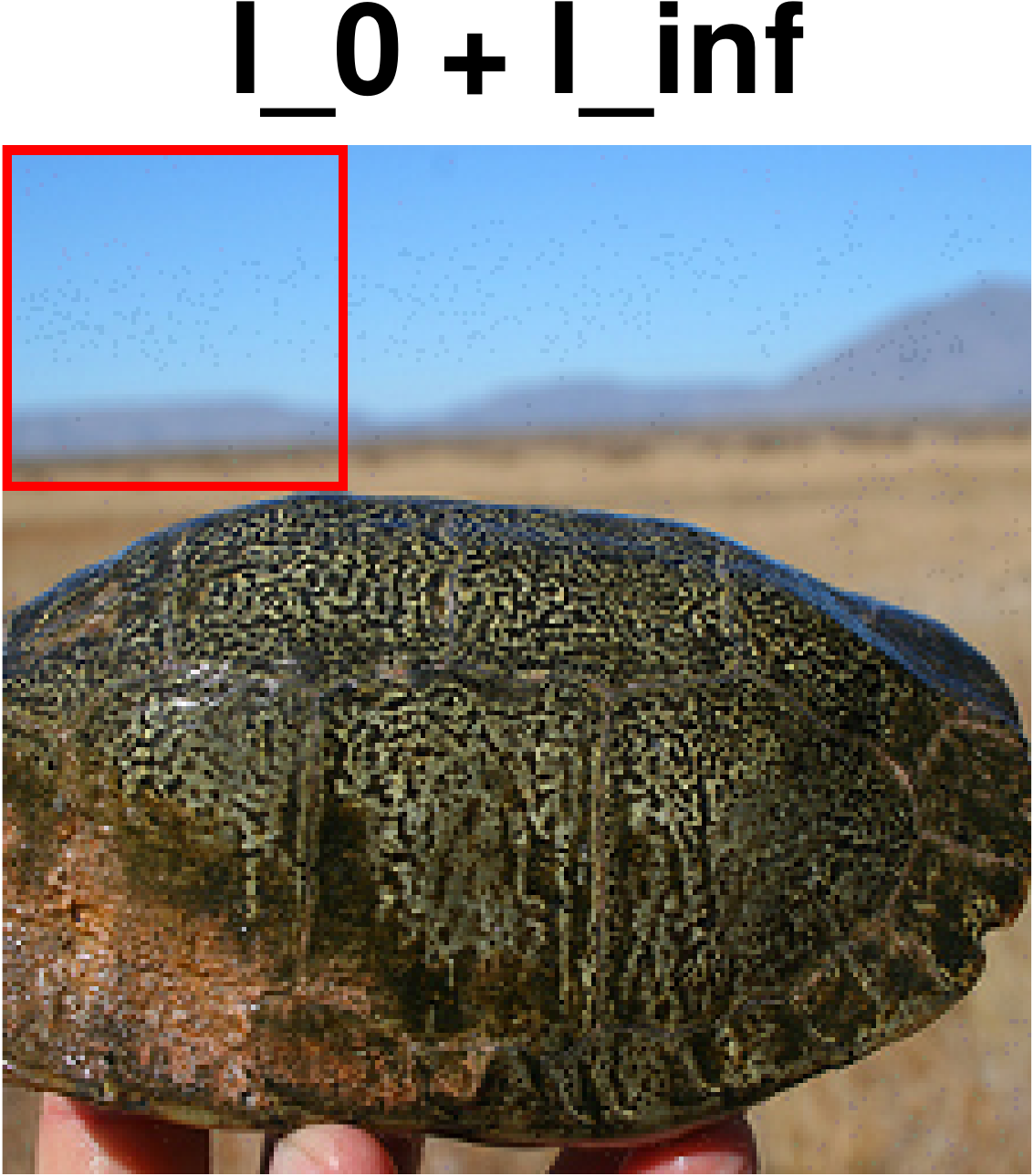}&
		\includegraphics[width=0.2\columnwidth]{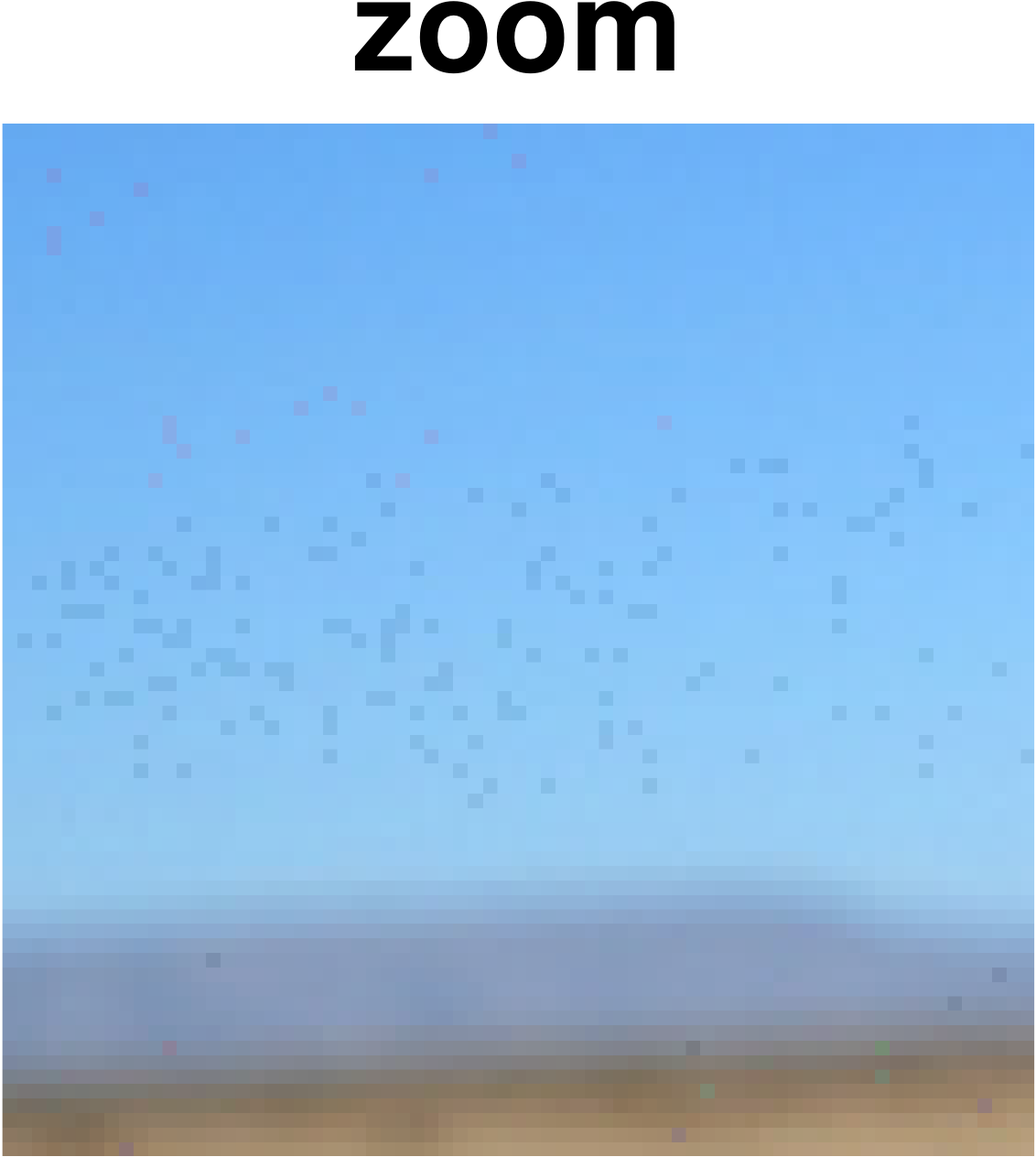}&
		\includegraphics[width=0.2\columnwidth]{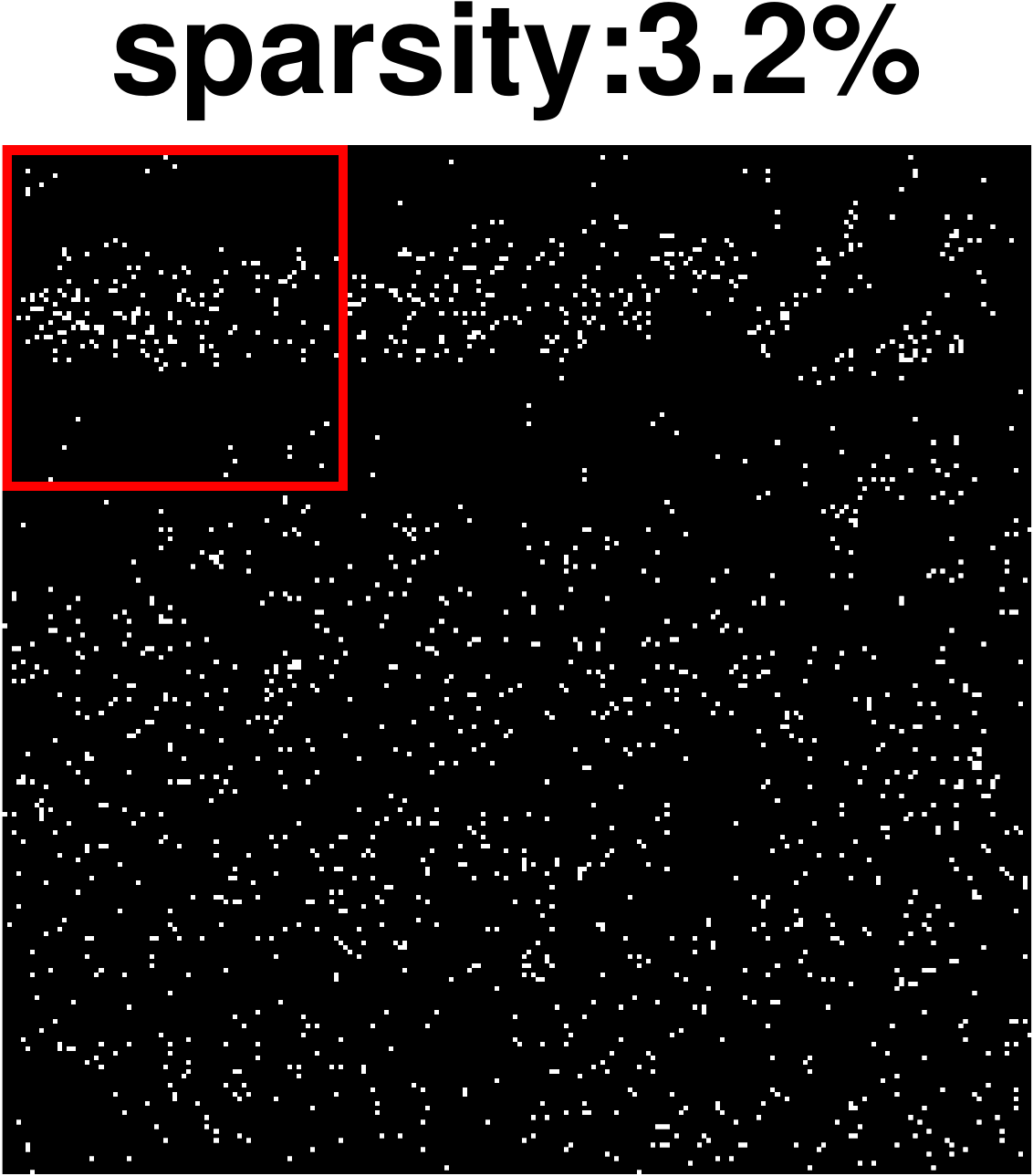}\\
		
		\includegraphics[width=0.2\columnwidth]{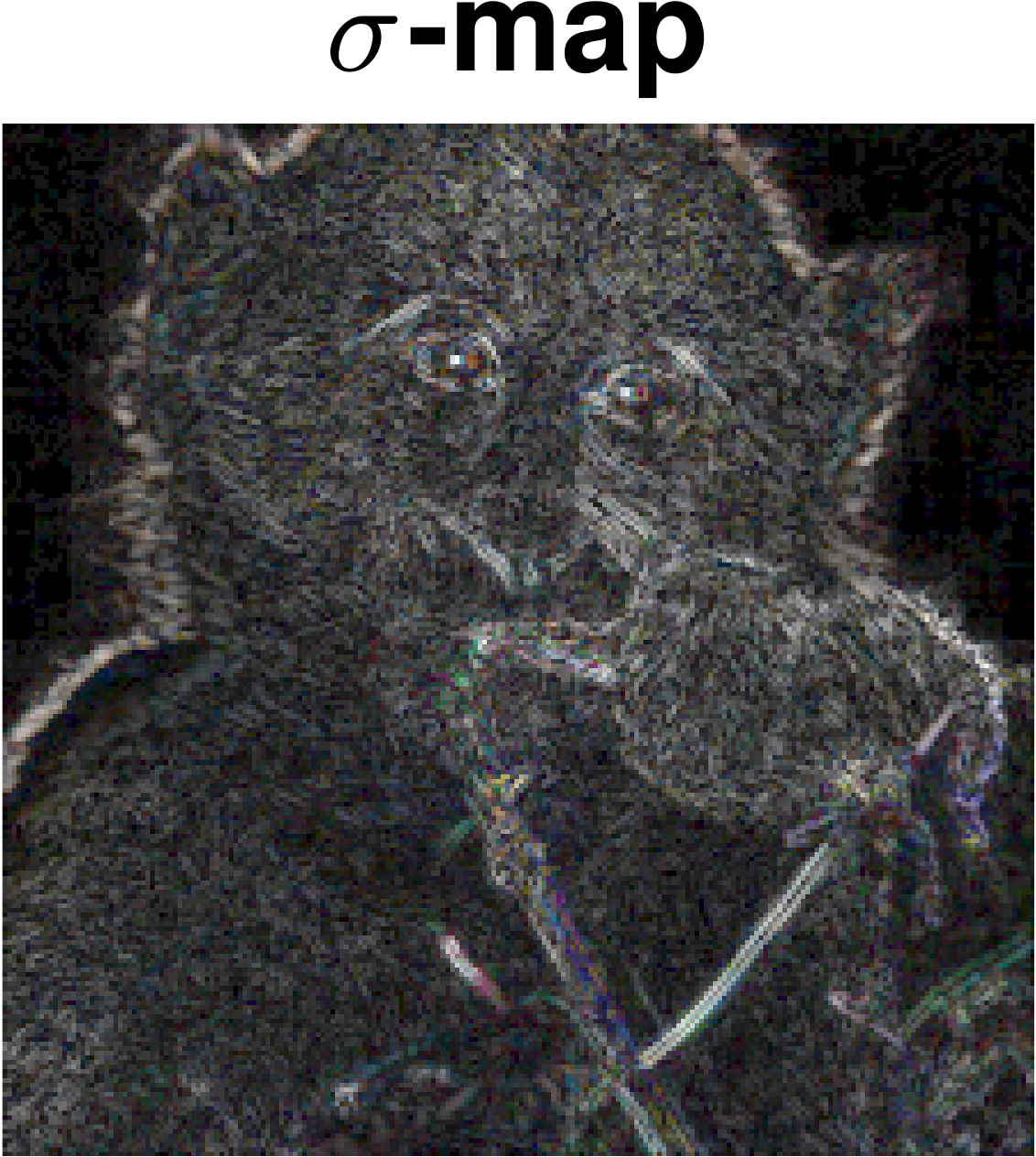}& 
		\includegraphics[width=0.2\columnwidth]{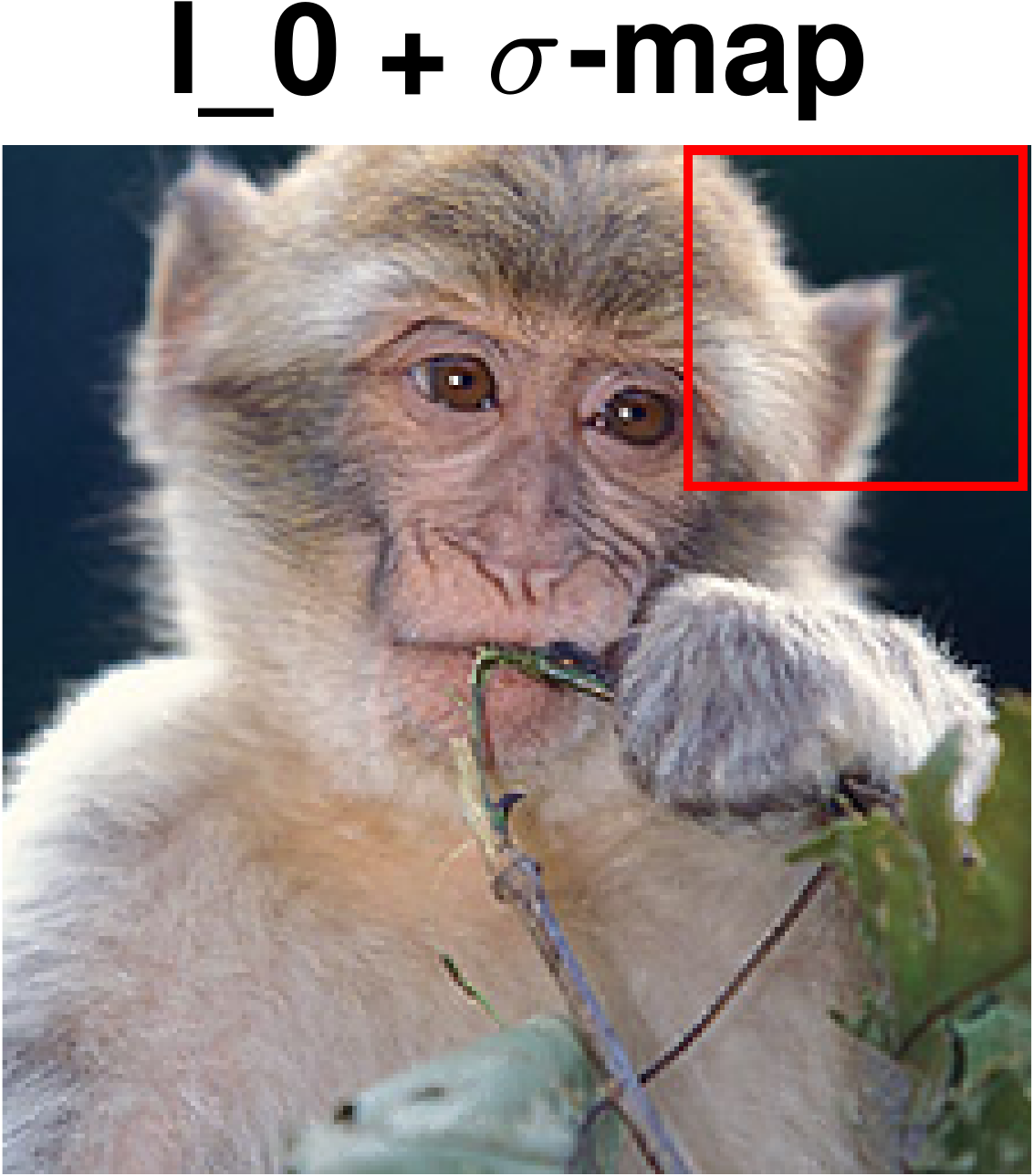}&
		\includegraphics[width=0.2\columnwidth]{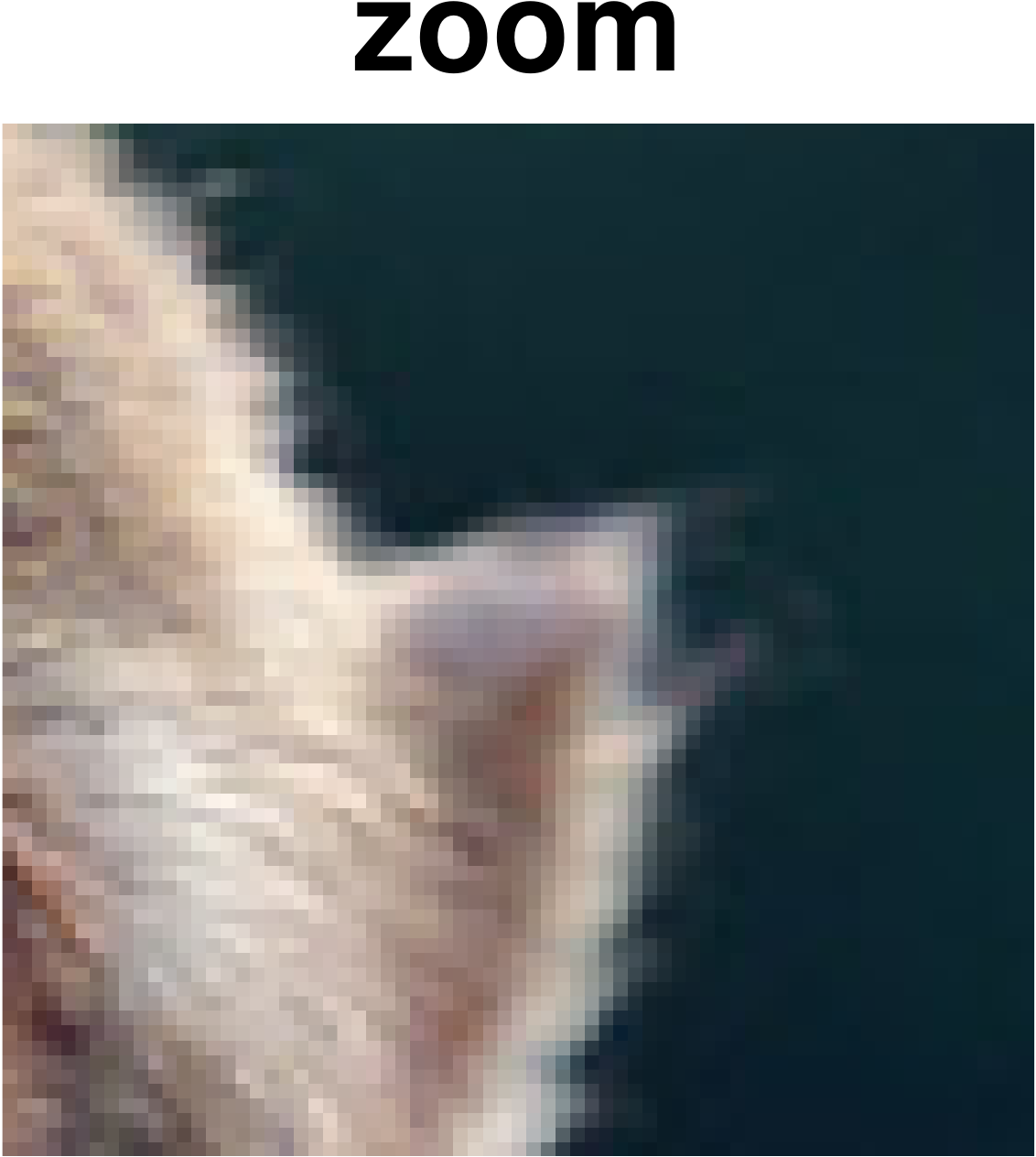}&
		\includegraphics[width=0.2\columnwidth]{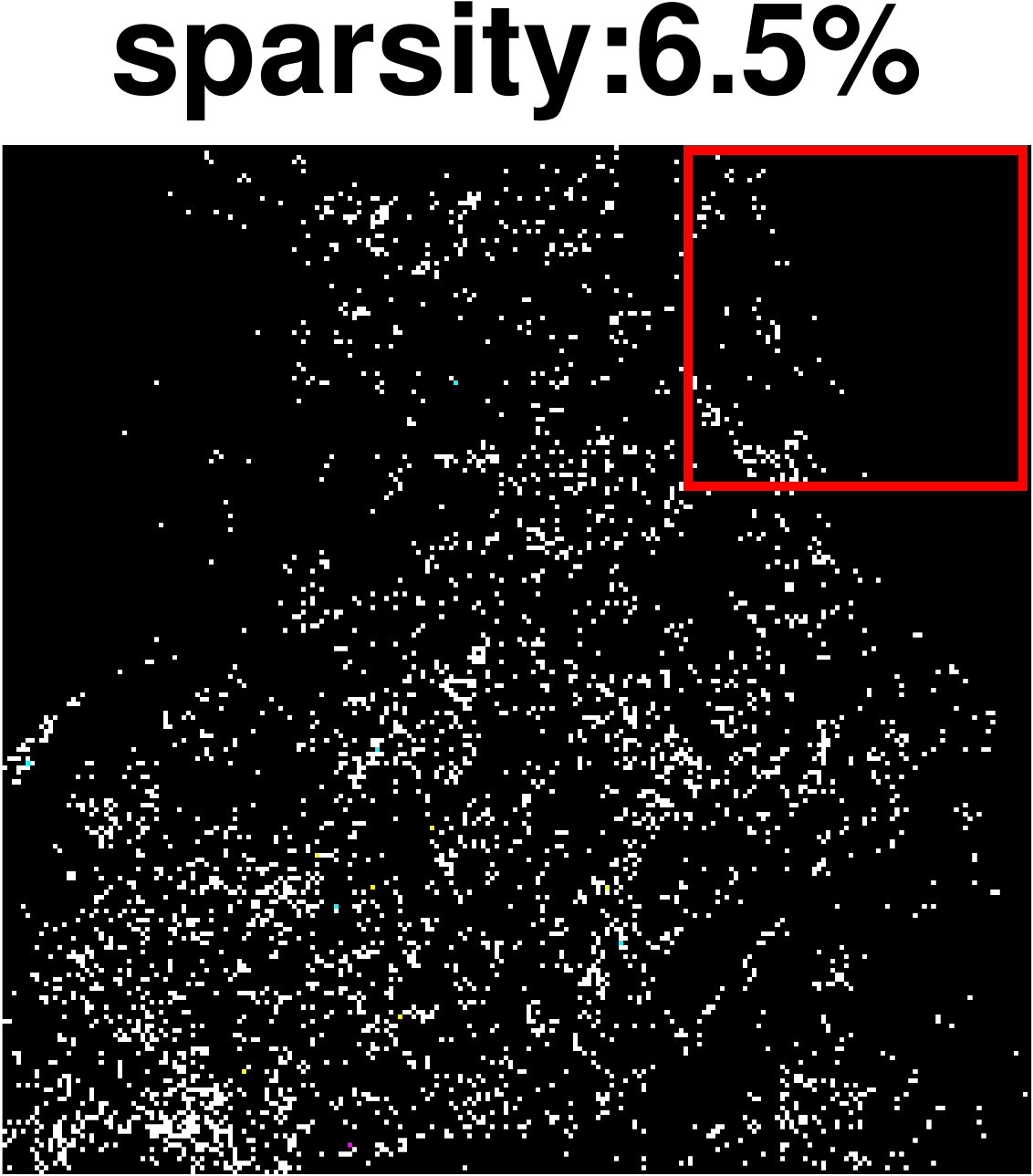}&
		
		\includegraphics[width=0.2\columnwidth]{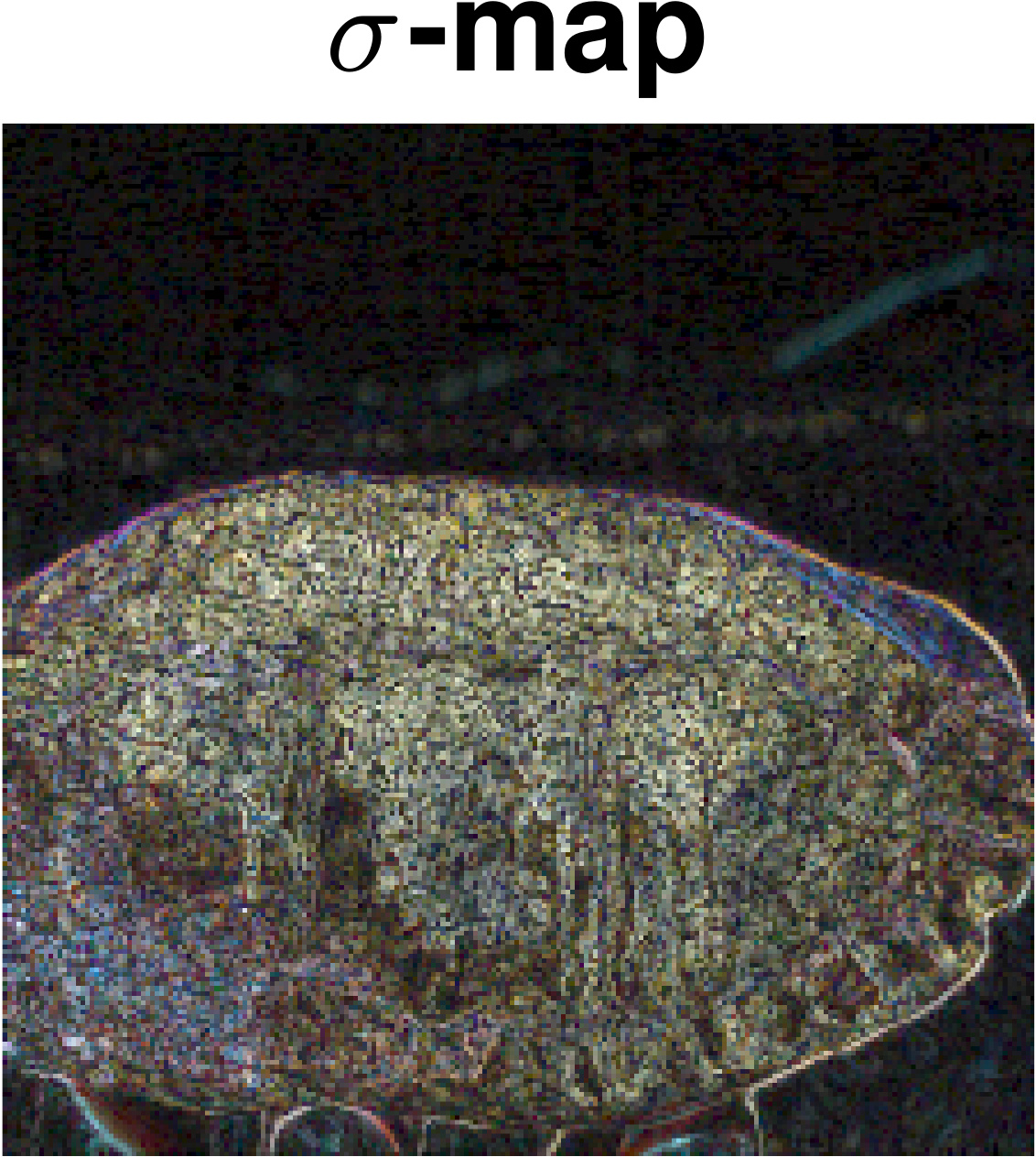}& 
		\includegraphics[width=0.2\columnwidth]{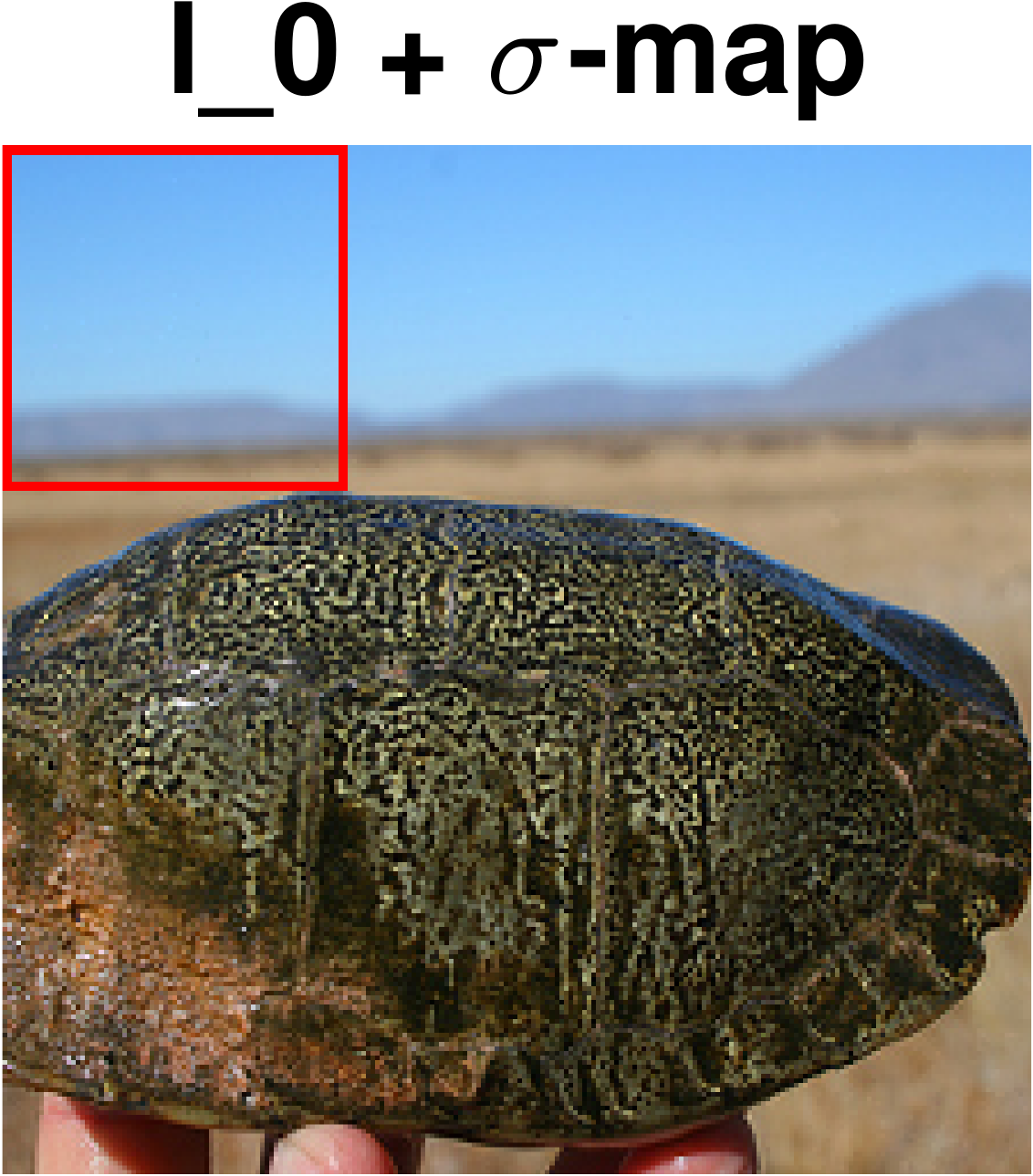}&
		\includegraphics[width=0.2\columnwidth]{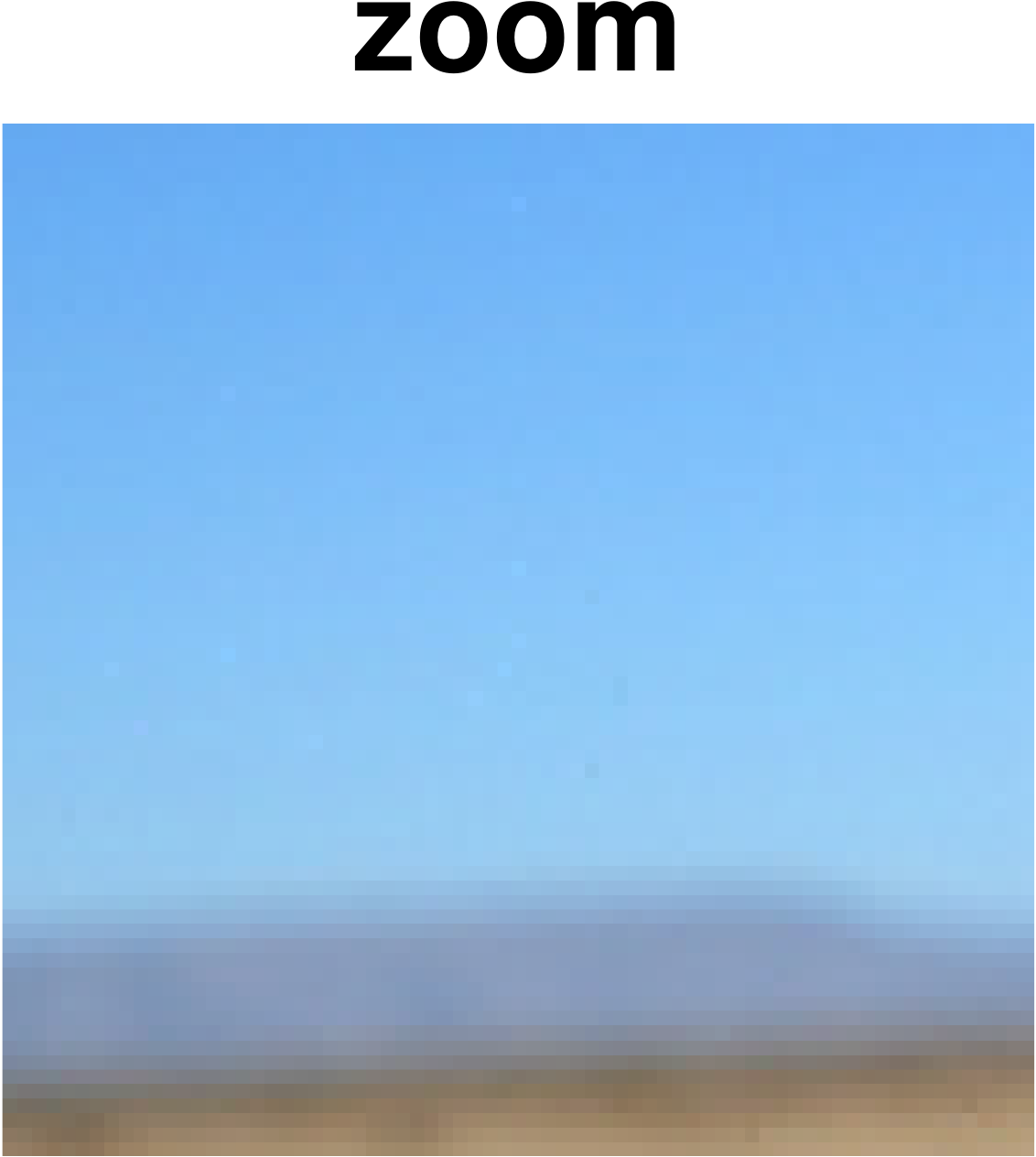}&
		\includegraphics[width=0.2\columnwidth]{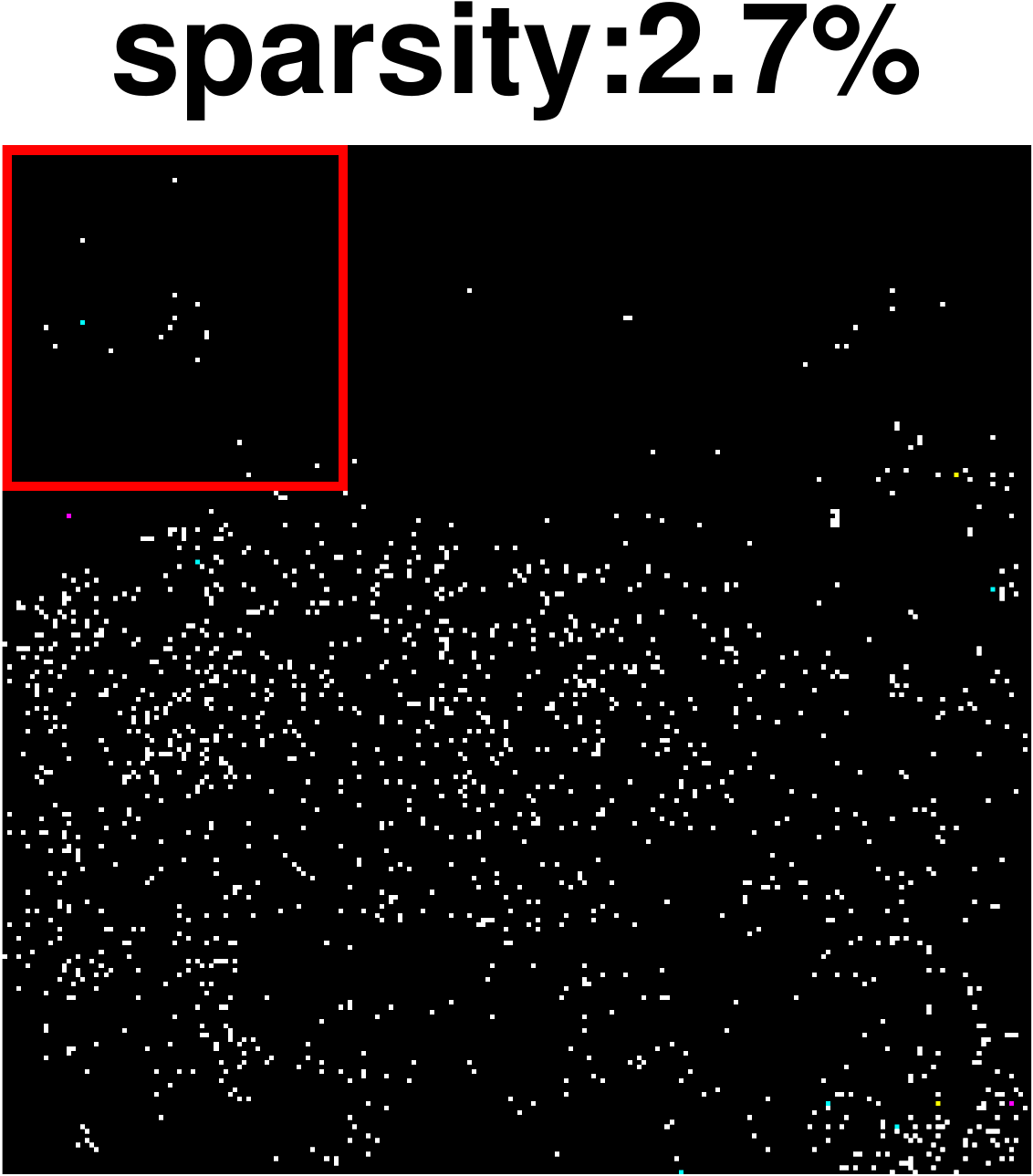} \\
		\multicolumn{8}{c}{}\\
		\hline
		\multicolumn{8}{c}{}\\
		\includegraphics[width=0.2\columnwidth]{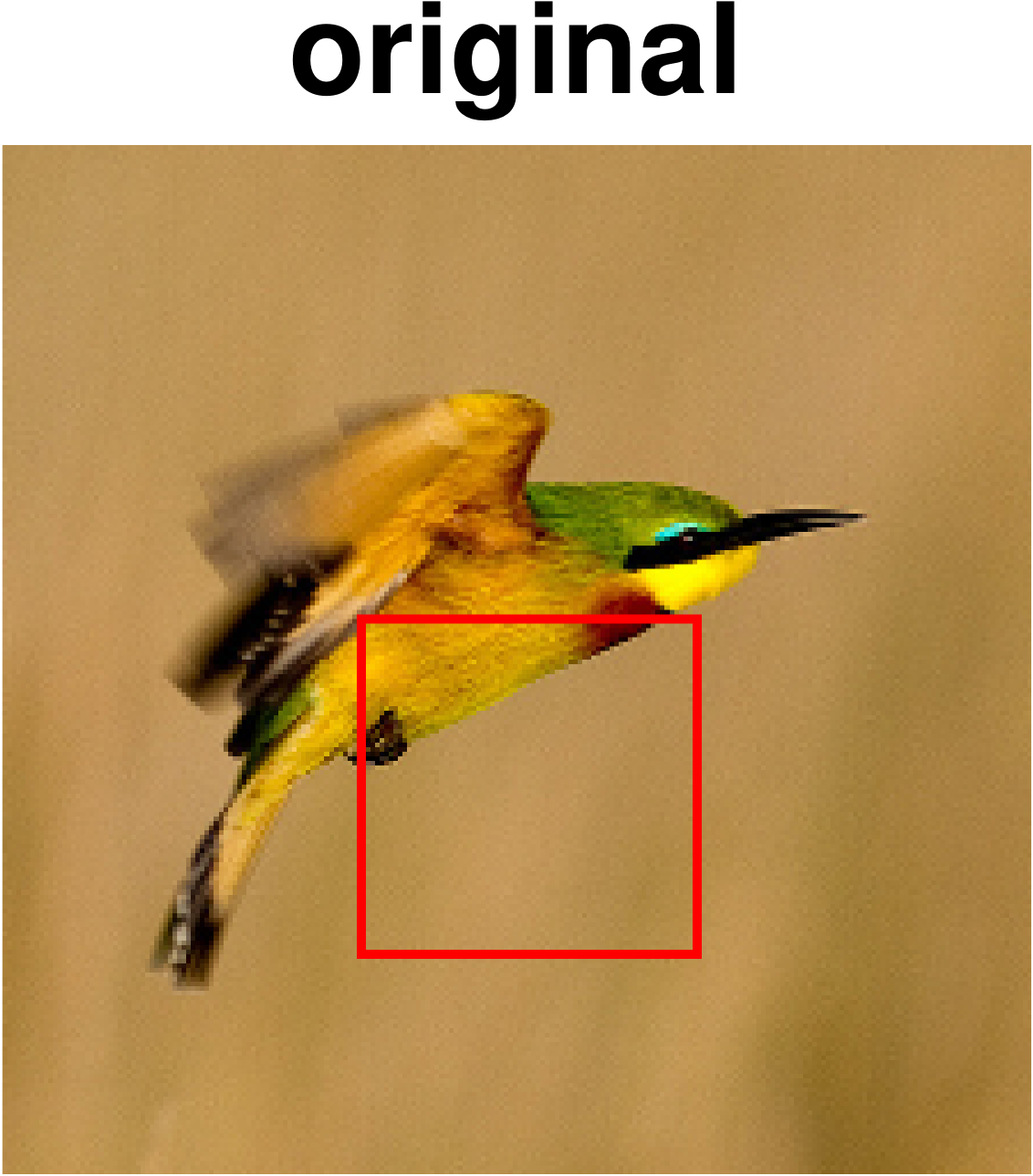}&
		\includegraphics[width=0.2\columnwidth]{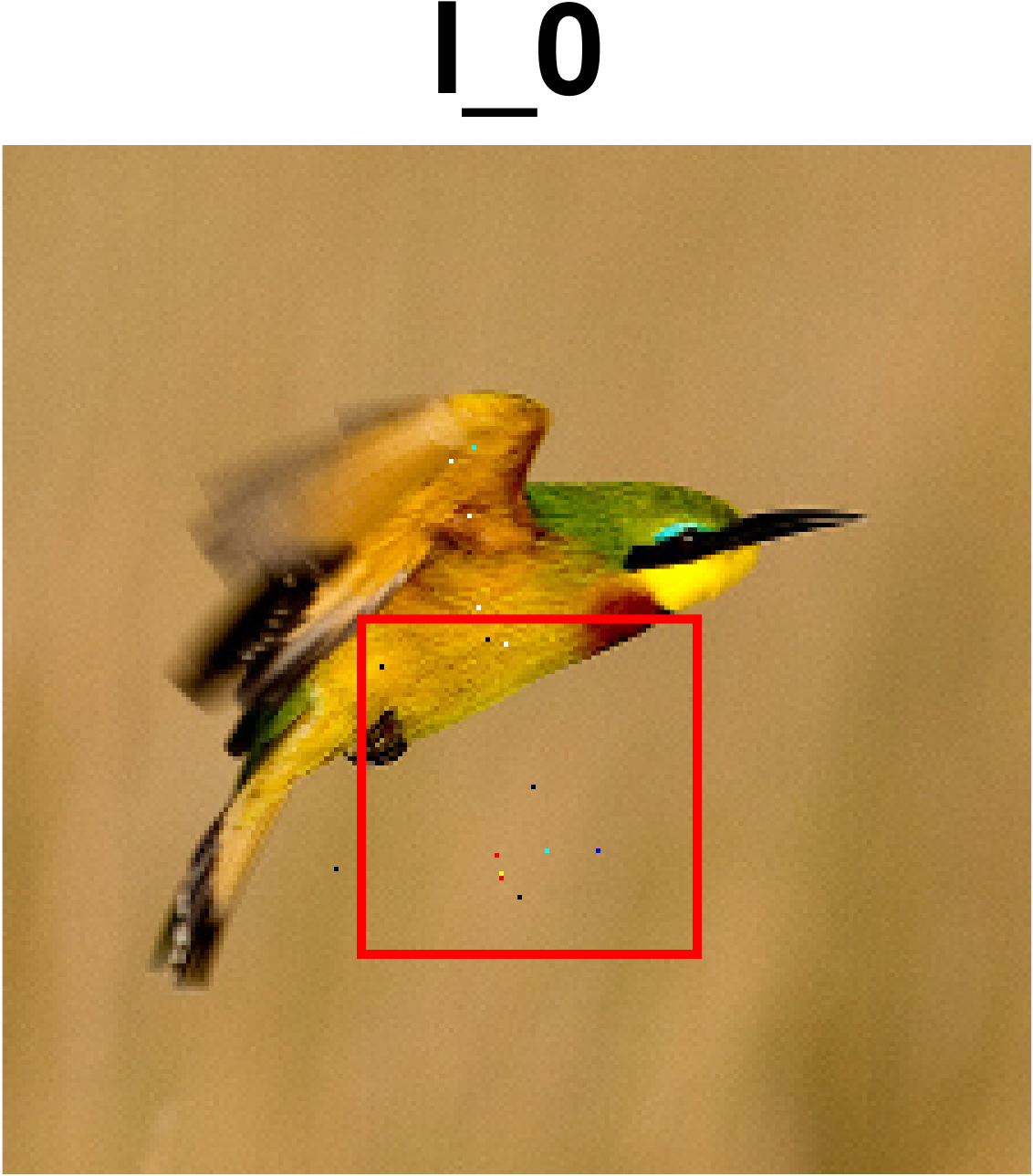}&
		\includegraphics[width=0.2\columnwidth]{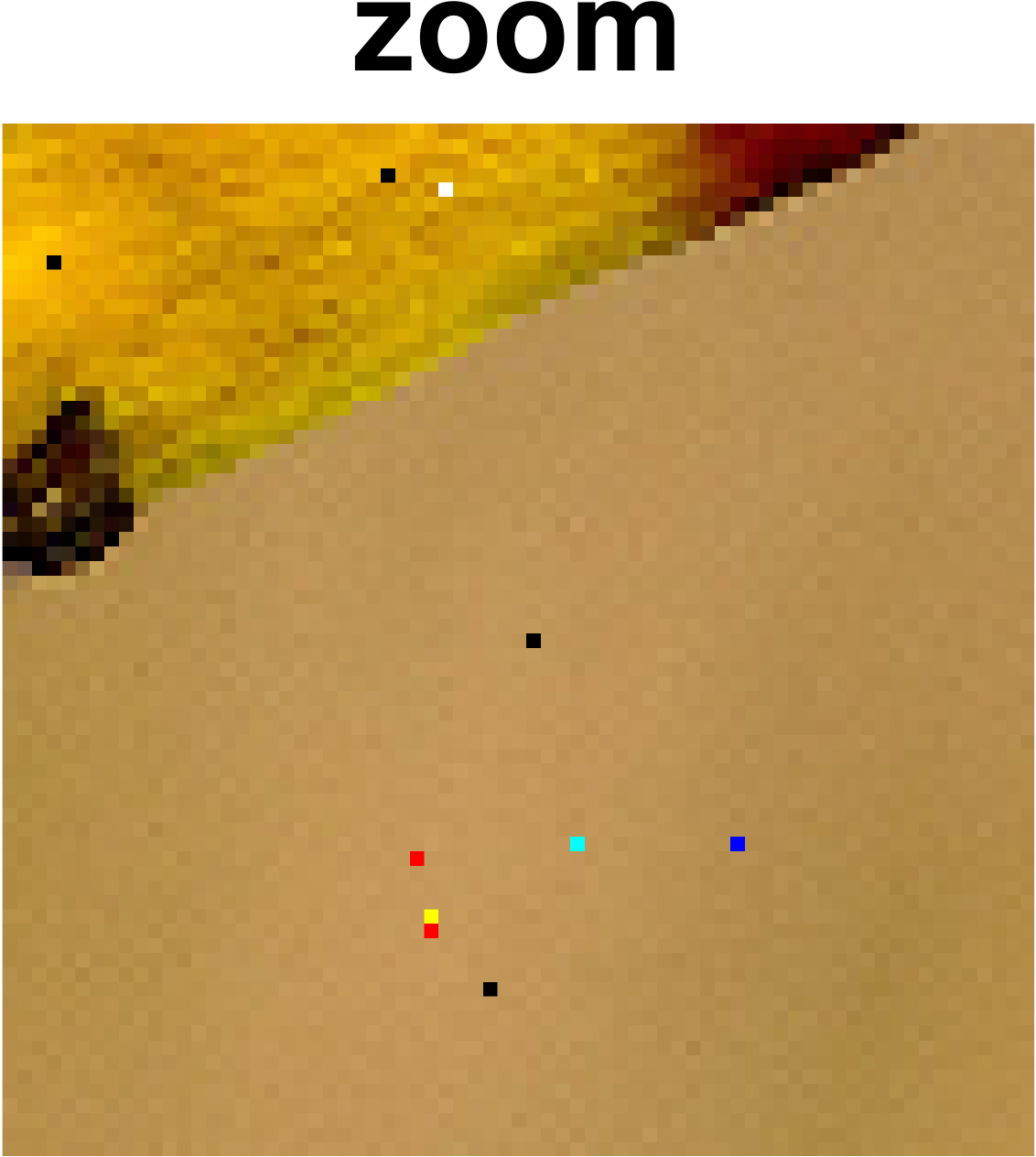}&
		\includegraphics[width=0.2\columnwidth]{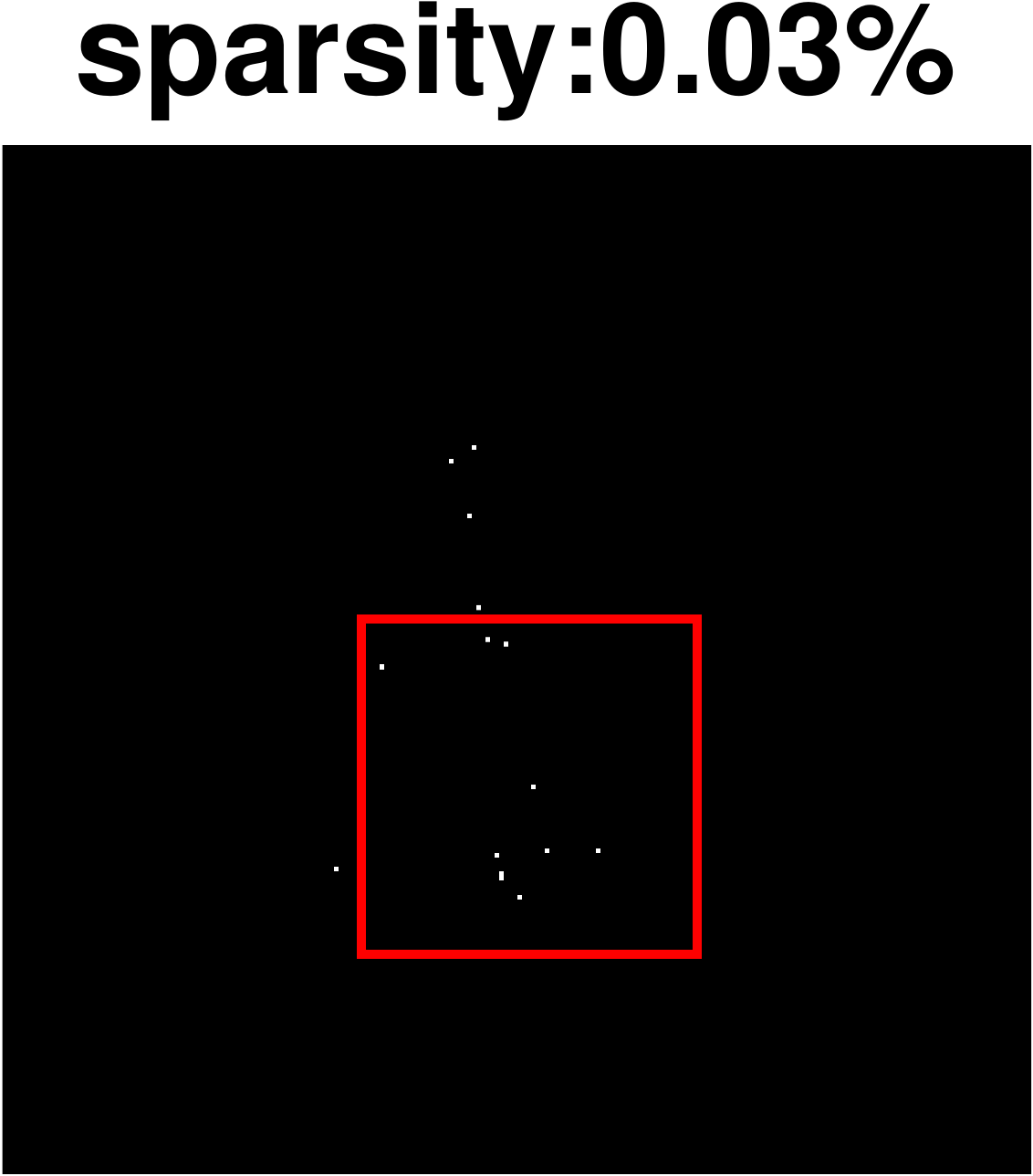}&

		\includegraphics[width=0.2\columnwidth]{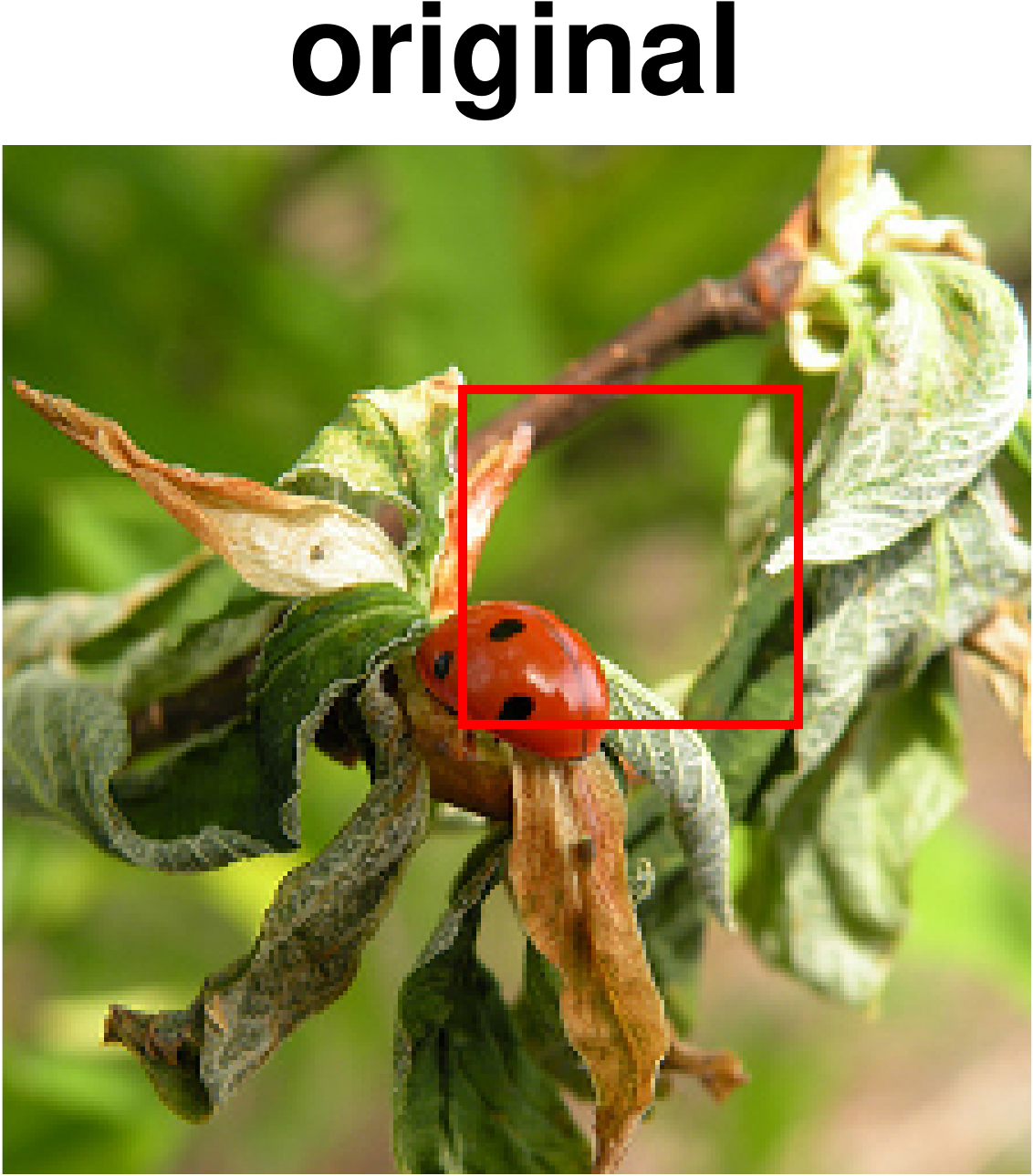}&
		\includegraphics[width=0.2\columnwidth]{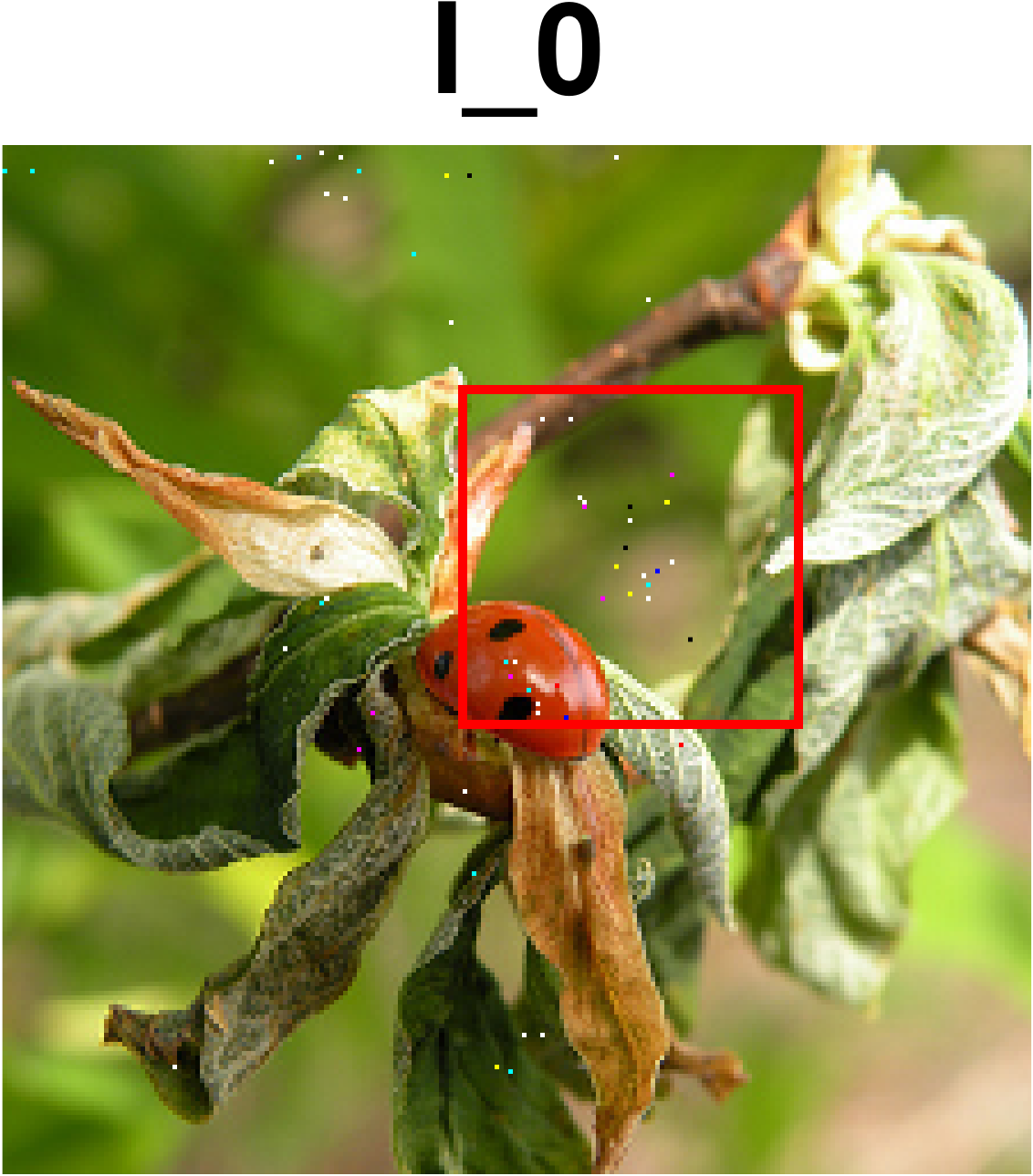}&
		\includegraphics[width=0.2\columnwidth]{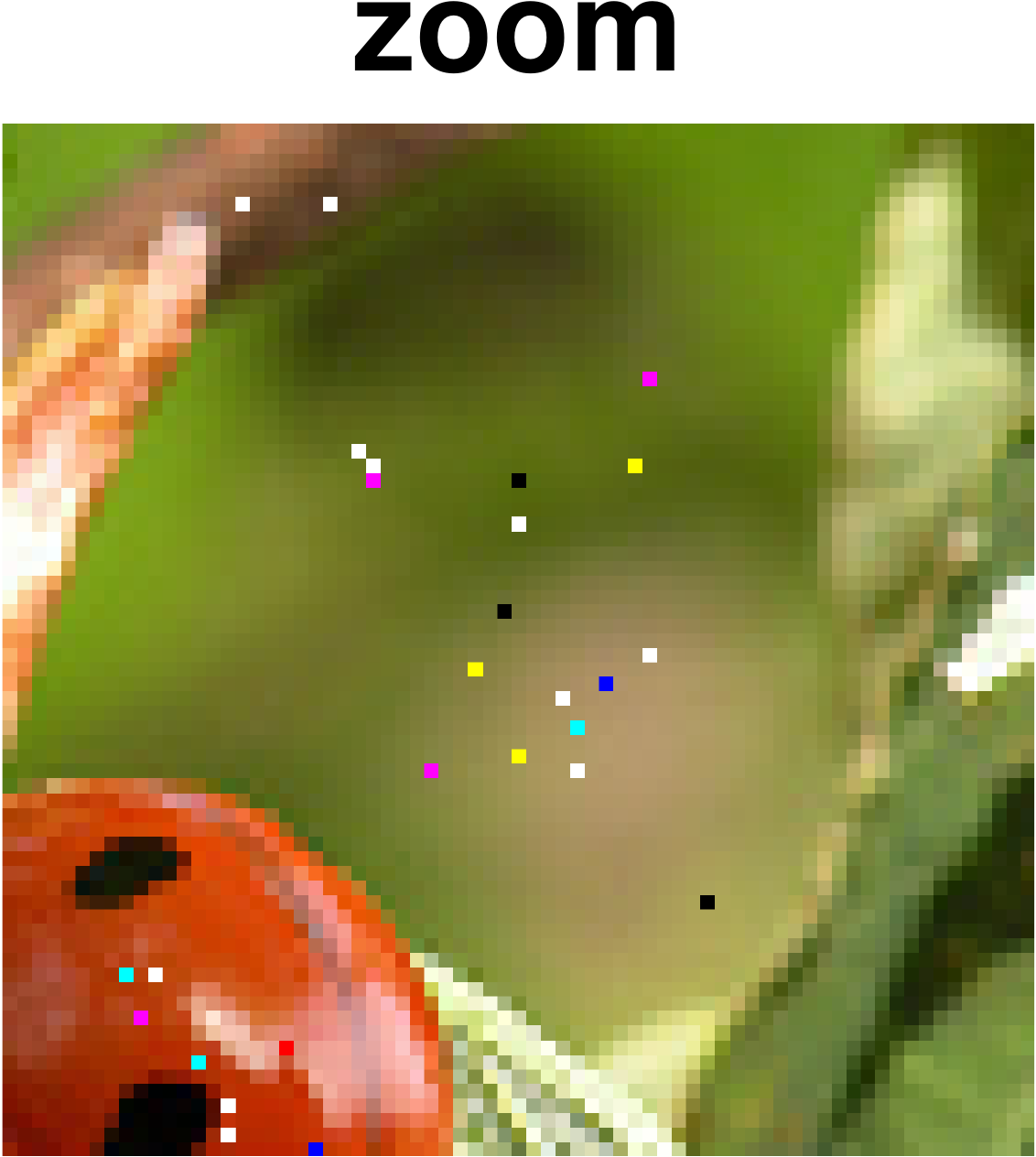}&
		\includegraphics[width=0.2\columnwidth]{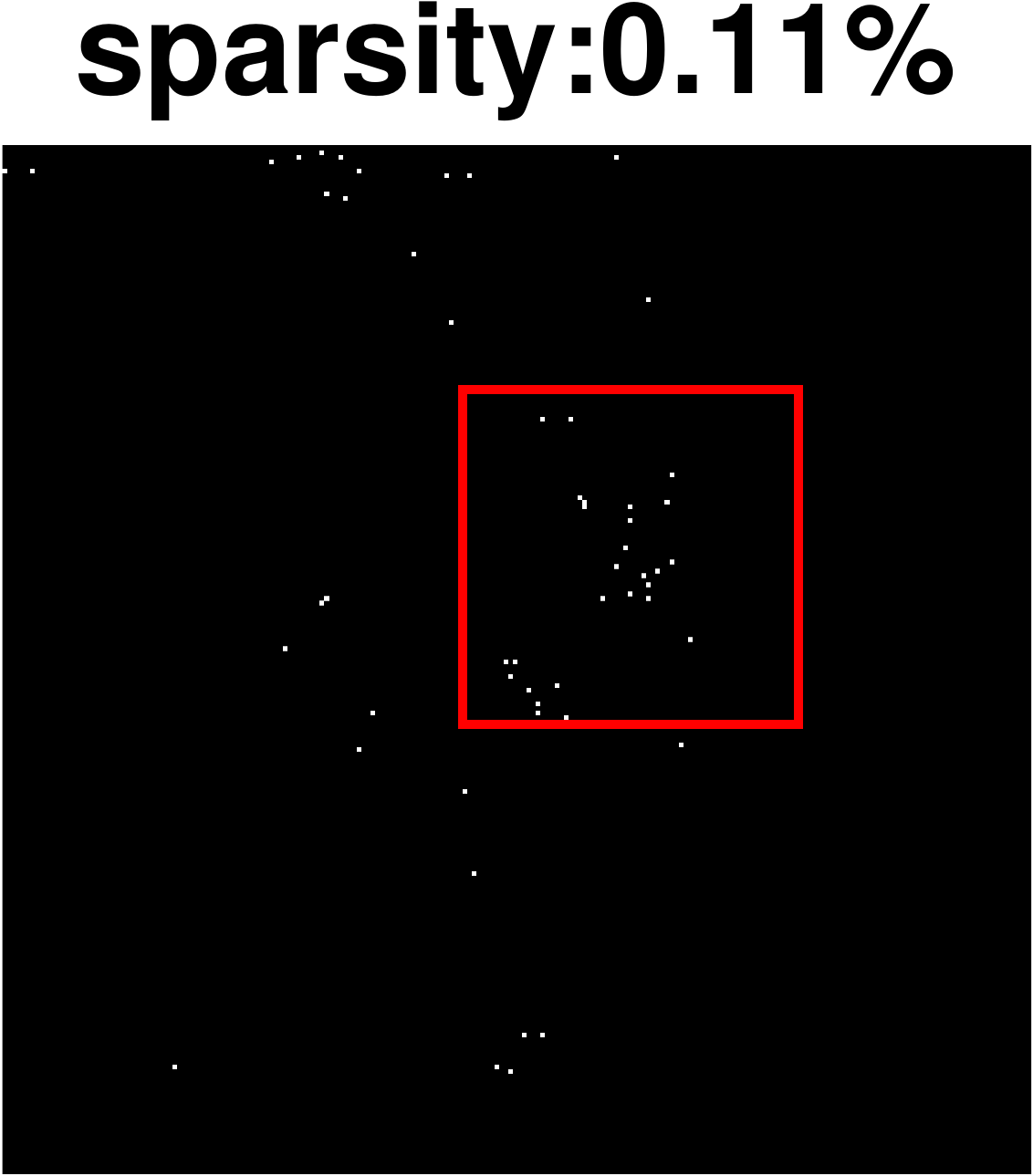}\\
		
		& \includegraphics[width=0.2\columnwidth]{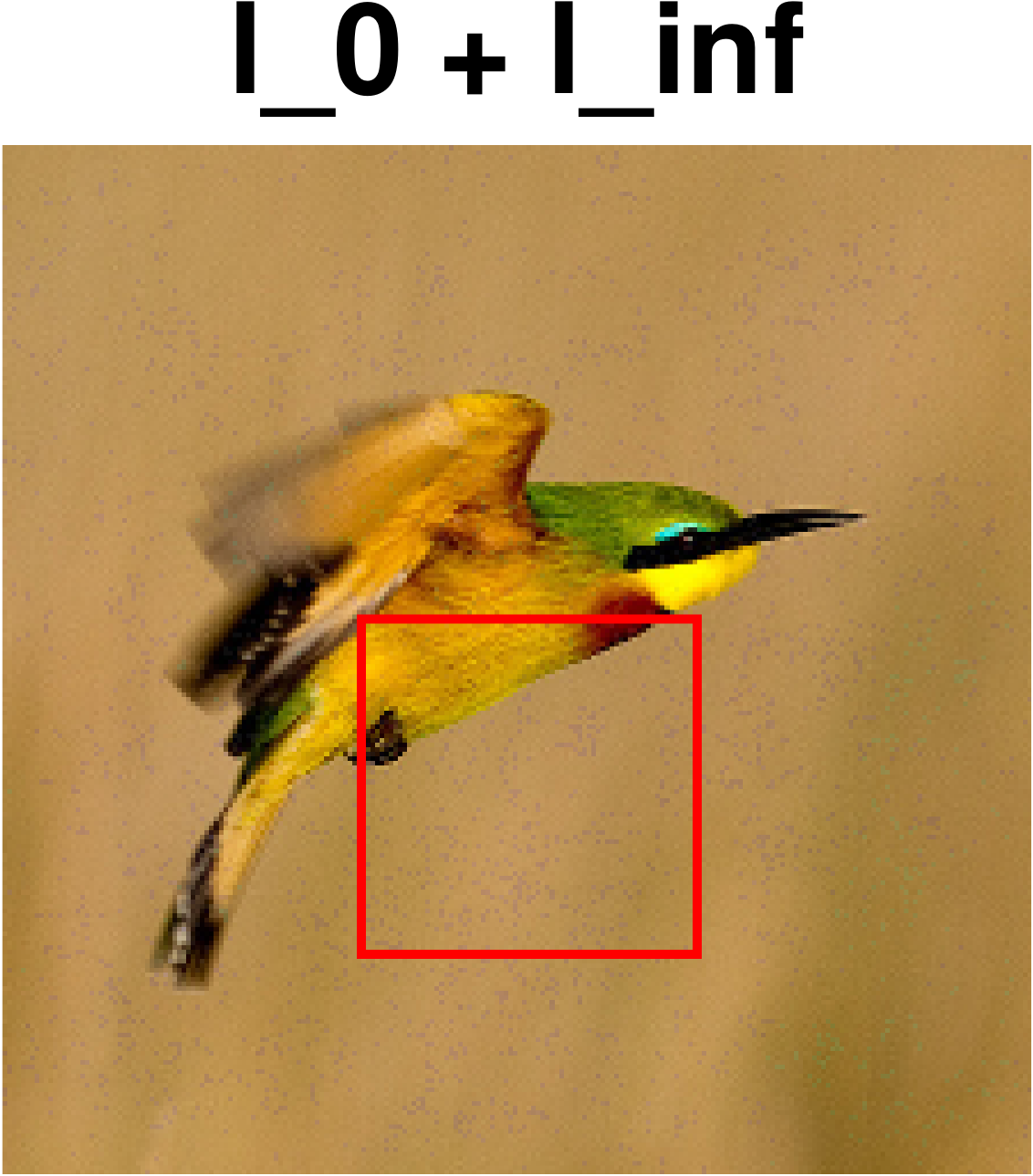}&
		\includegraphics[width=0.2\columnwidth]{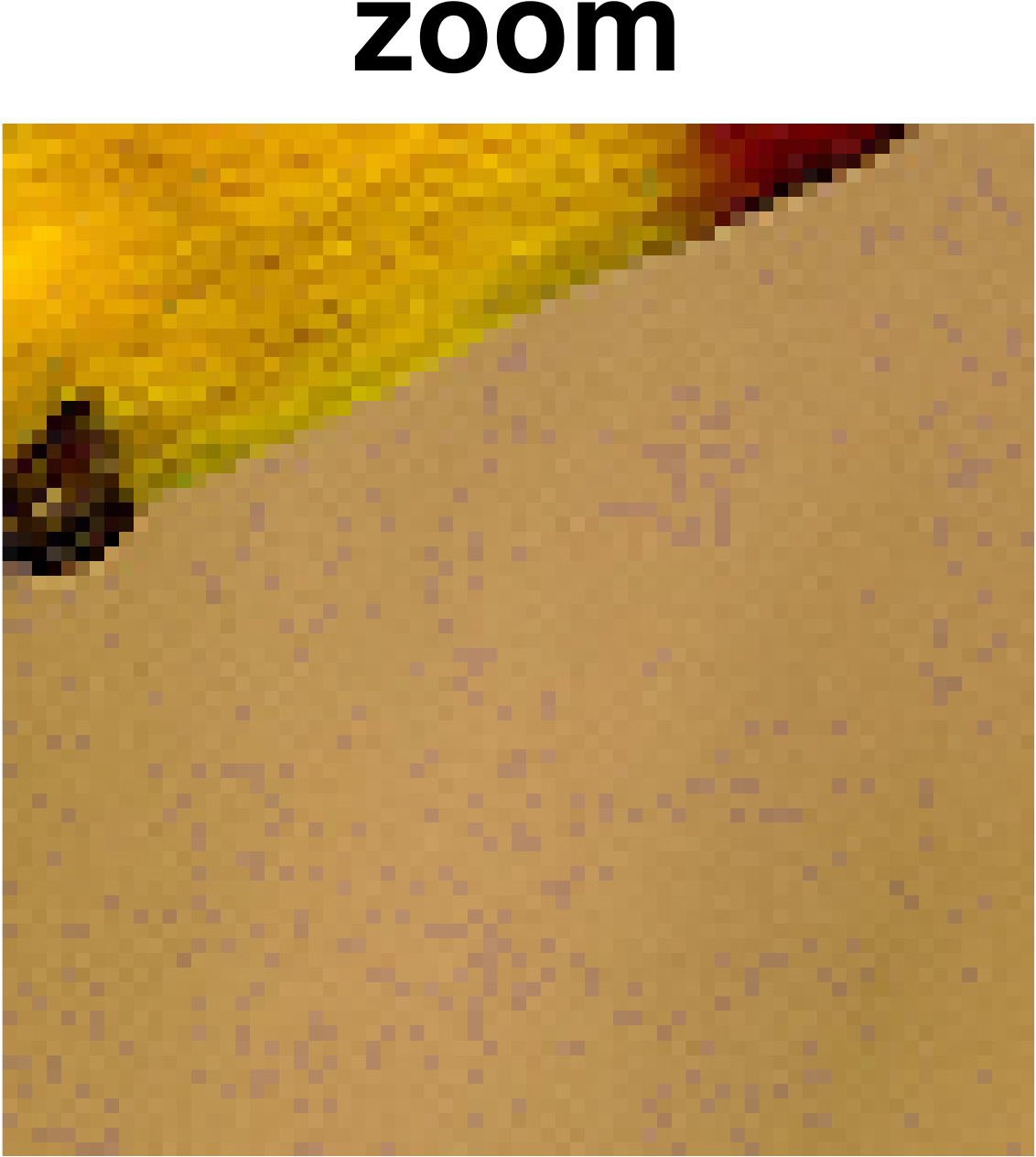}&
		\includegraphics[width=0.2\columnwidth]{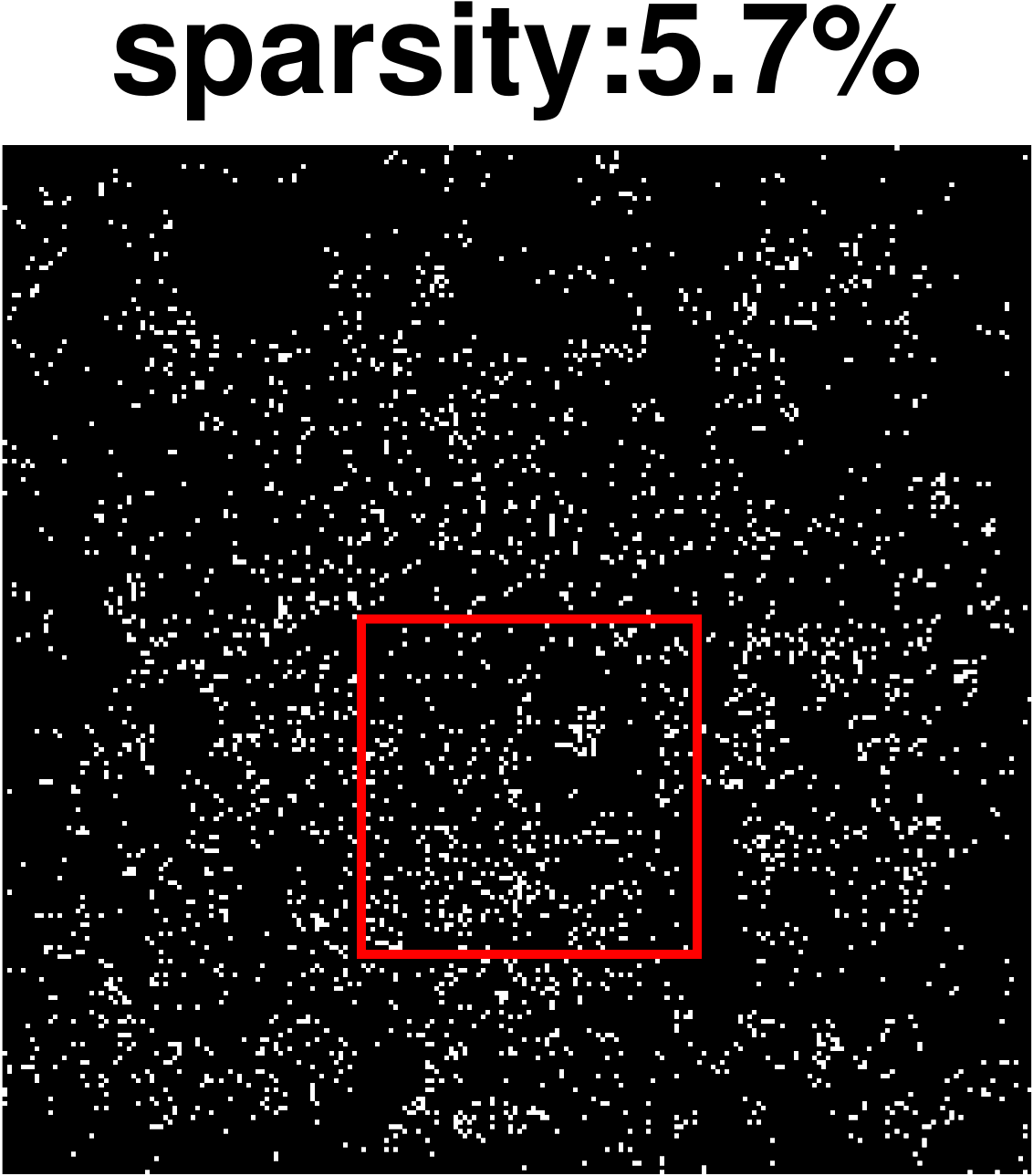}&
		&
		\includegraphics[width=0.2\columnwidth]{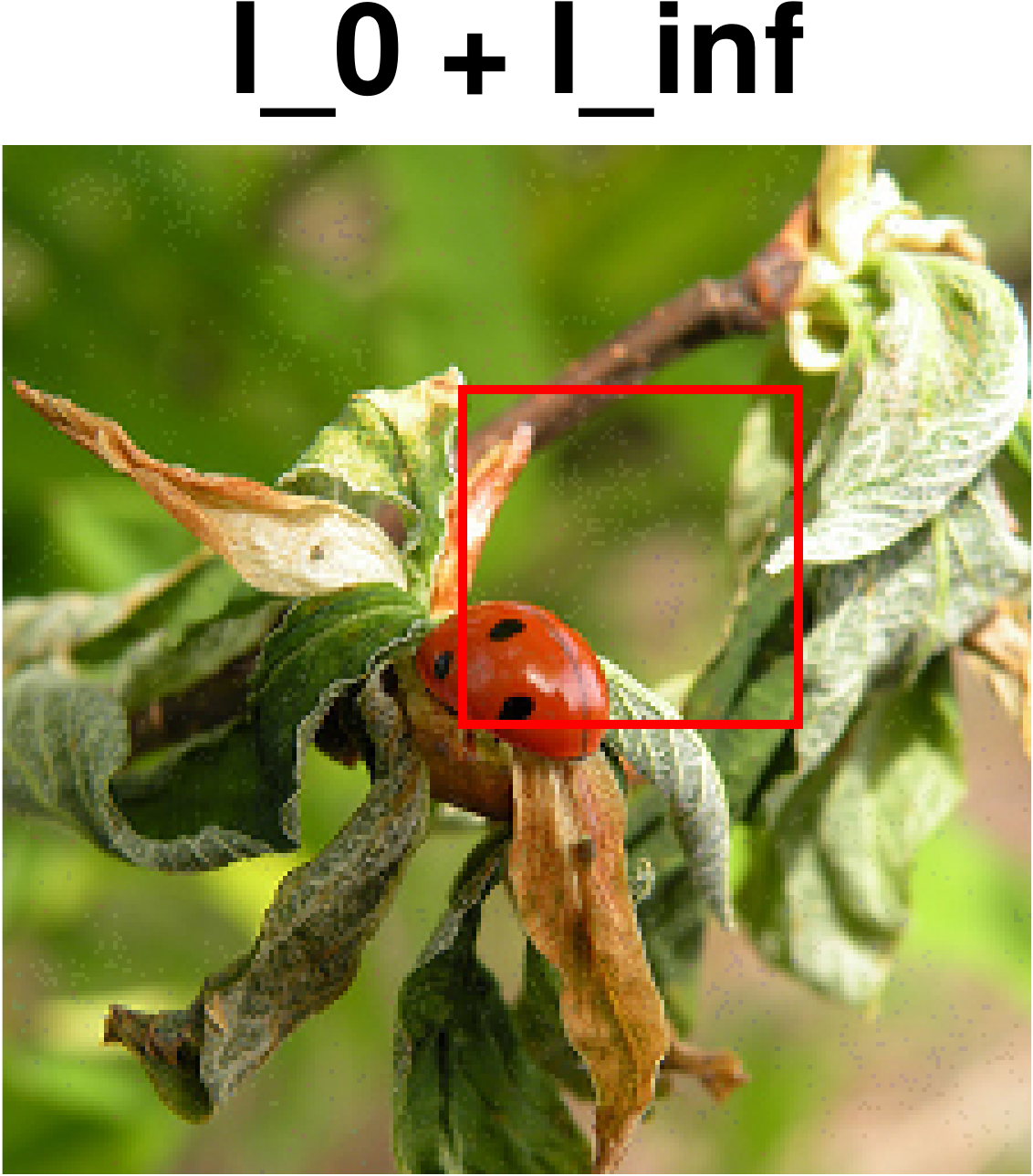}&
		\includegraphics[width=0.2\columnwidth]{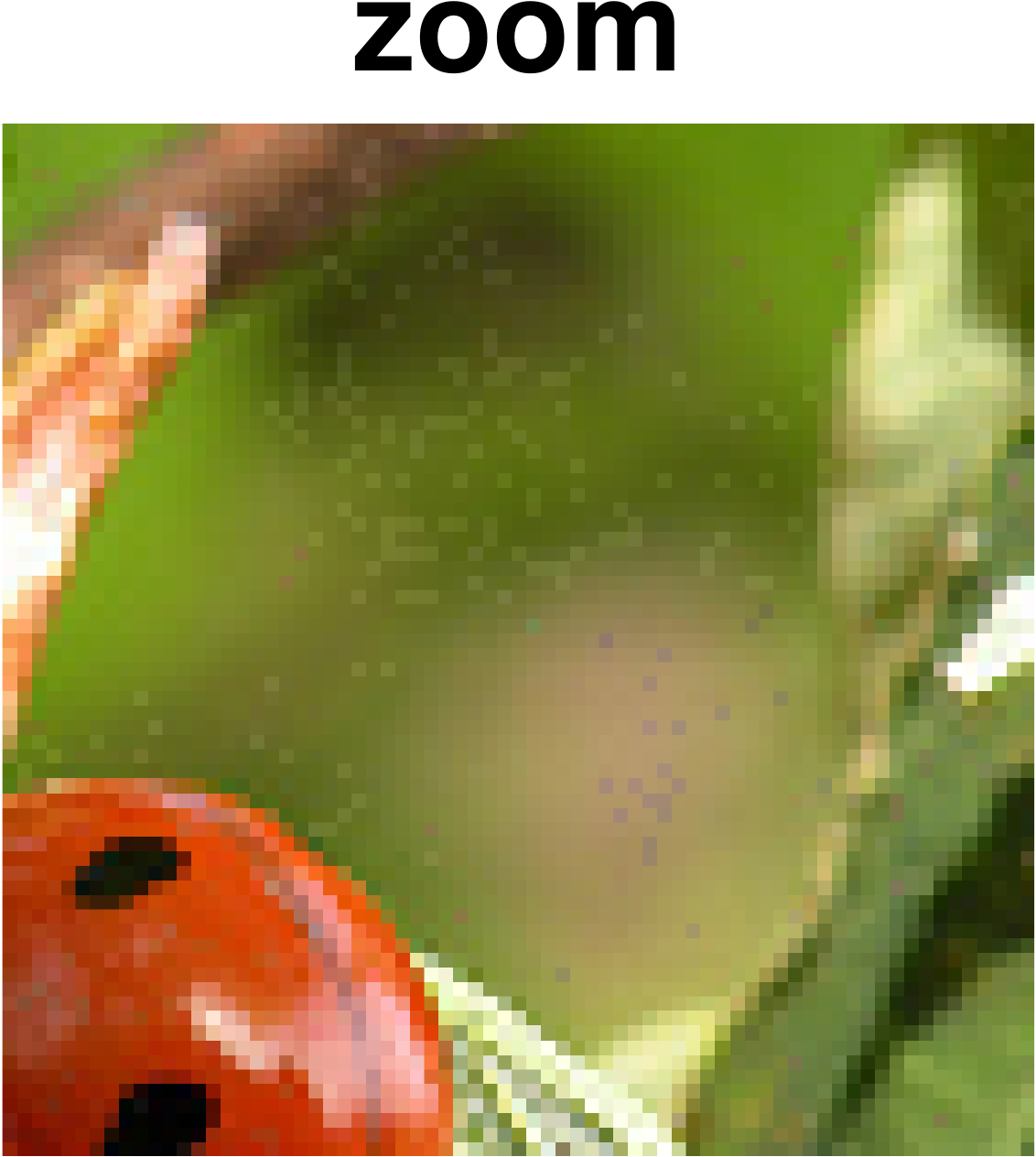}&
		\includegraphics[width=0.2\columnwidth]{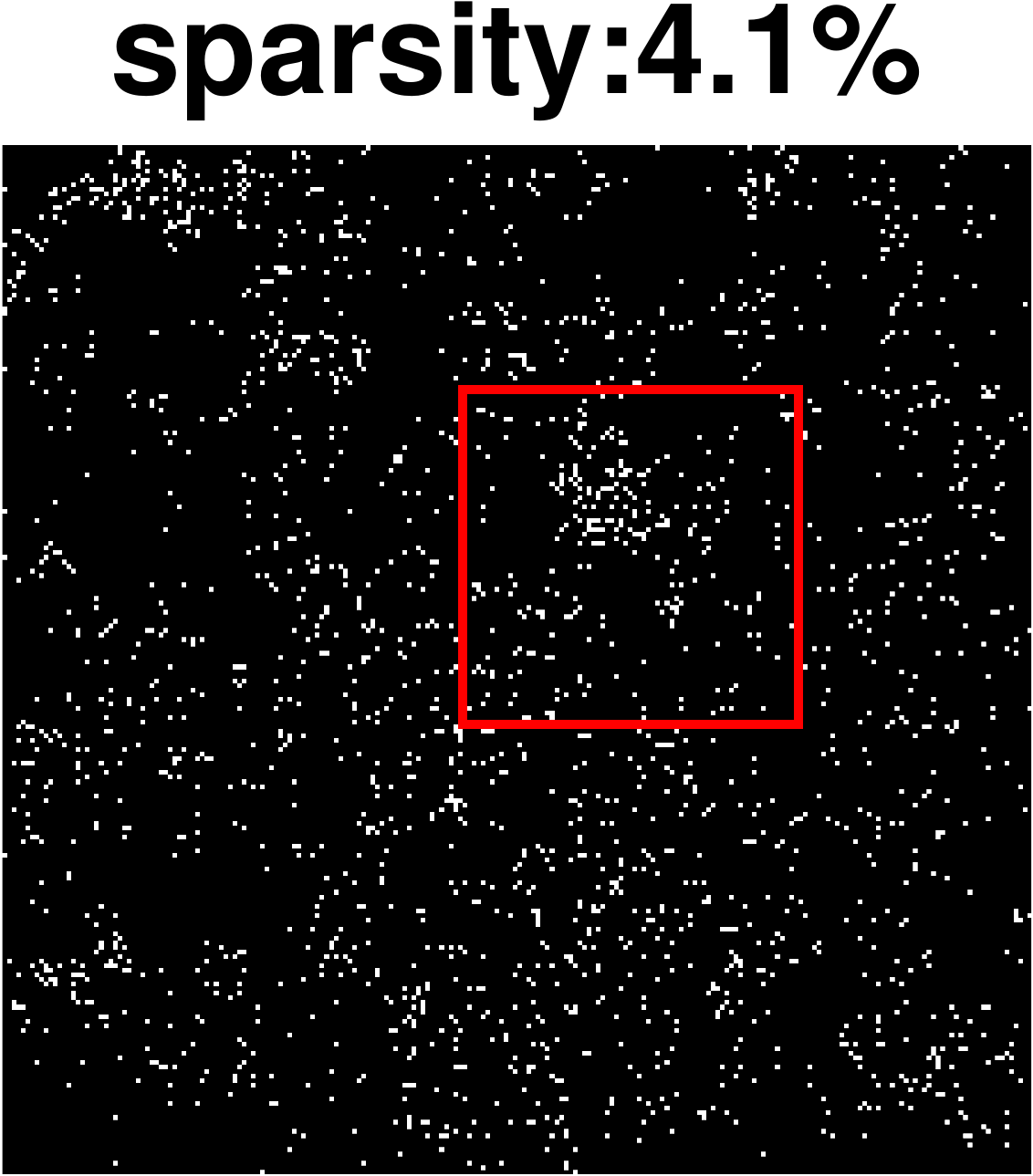}\\
		
		\includegraphics[width=0.2\columnwidth]{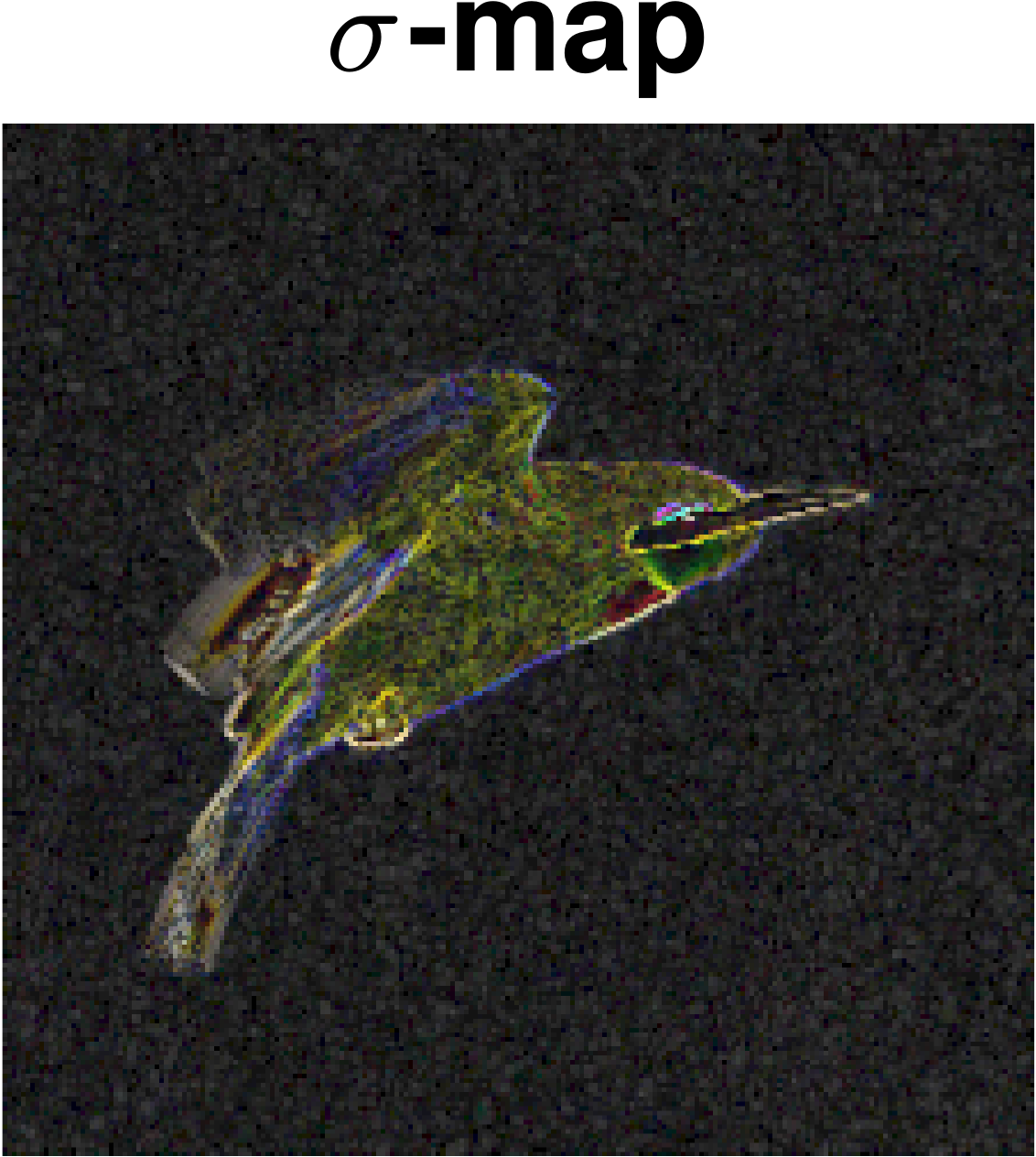}& 
		\includegraphics[width=0.2\columnwidth]{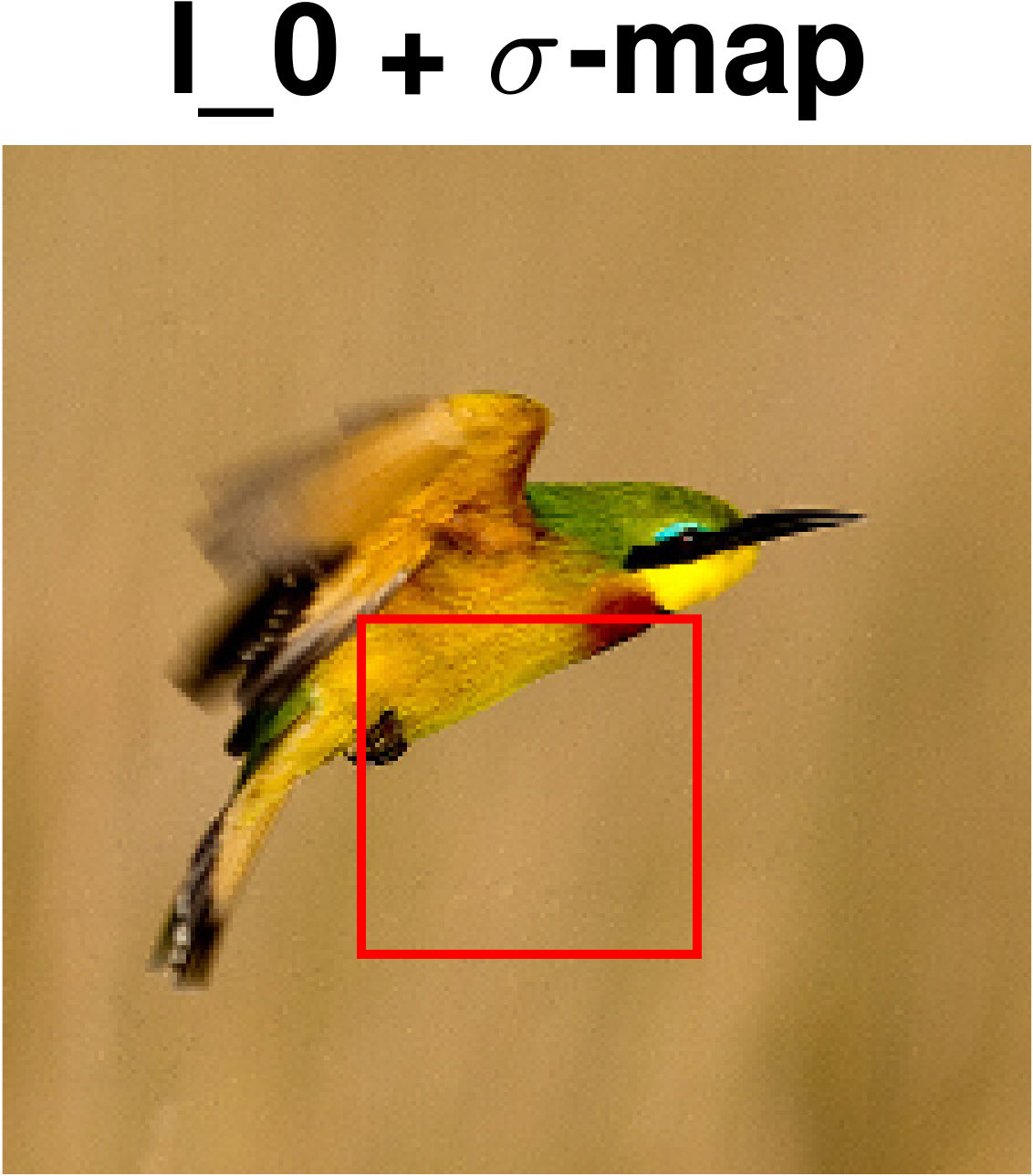}&
		\includegraphics[width=0.2\columnwidth]{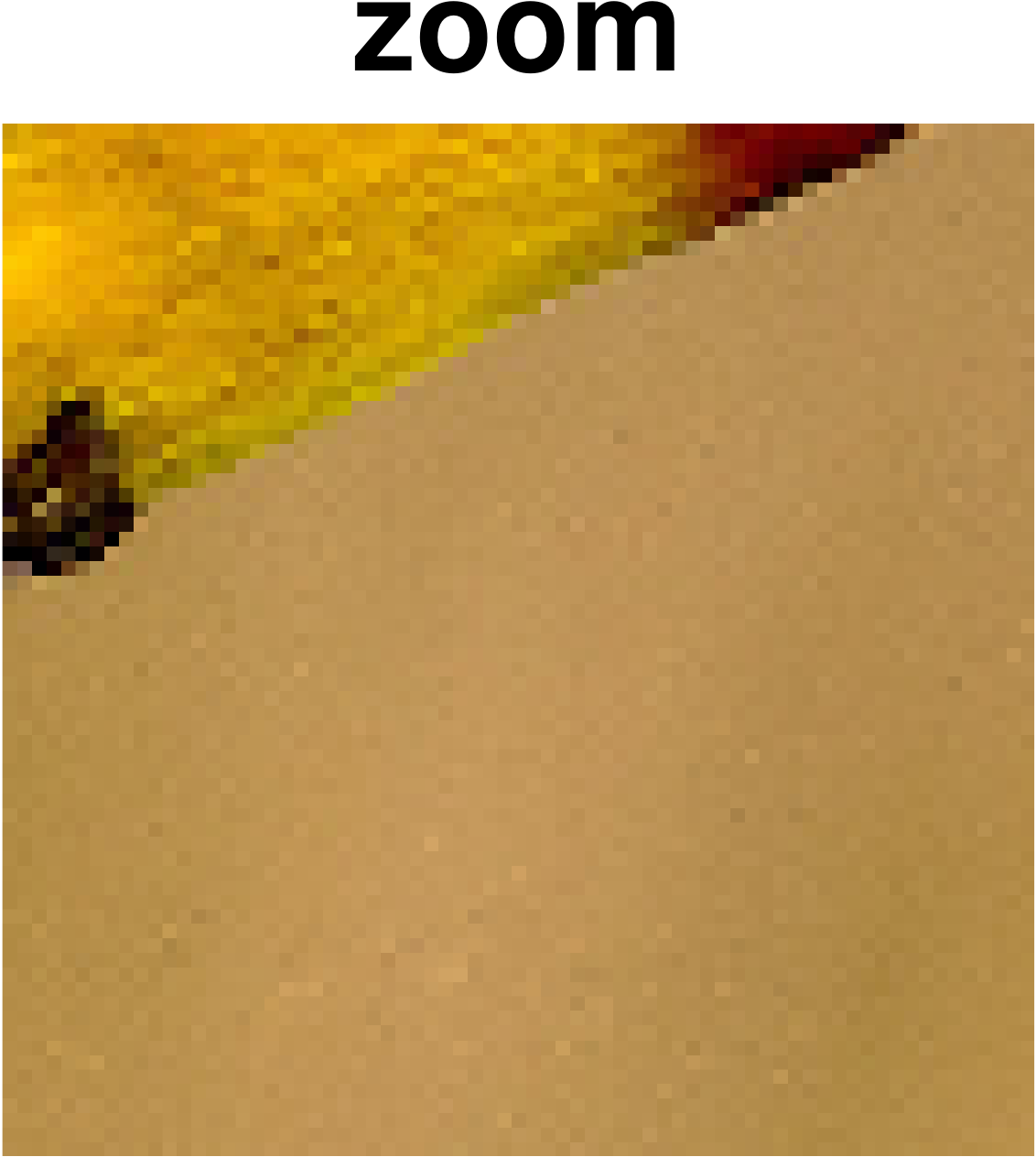}&
		\includegraphics[width=0.2\columnwidth]{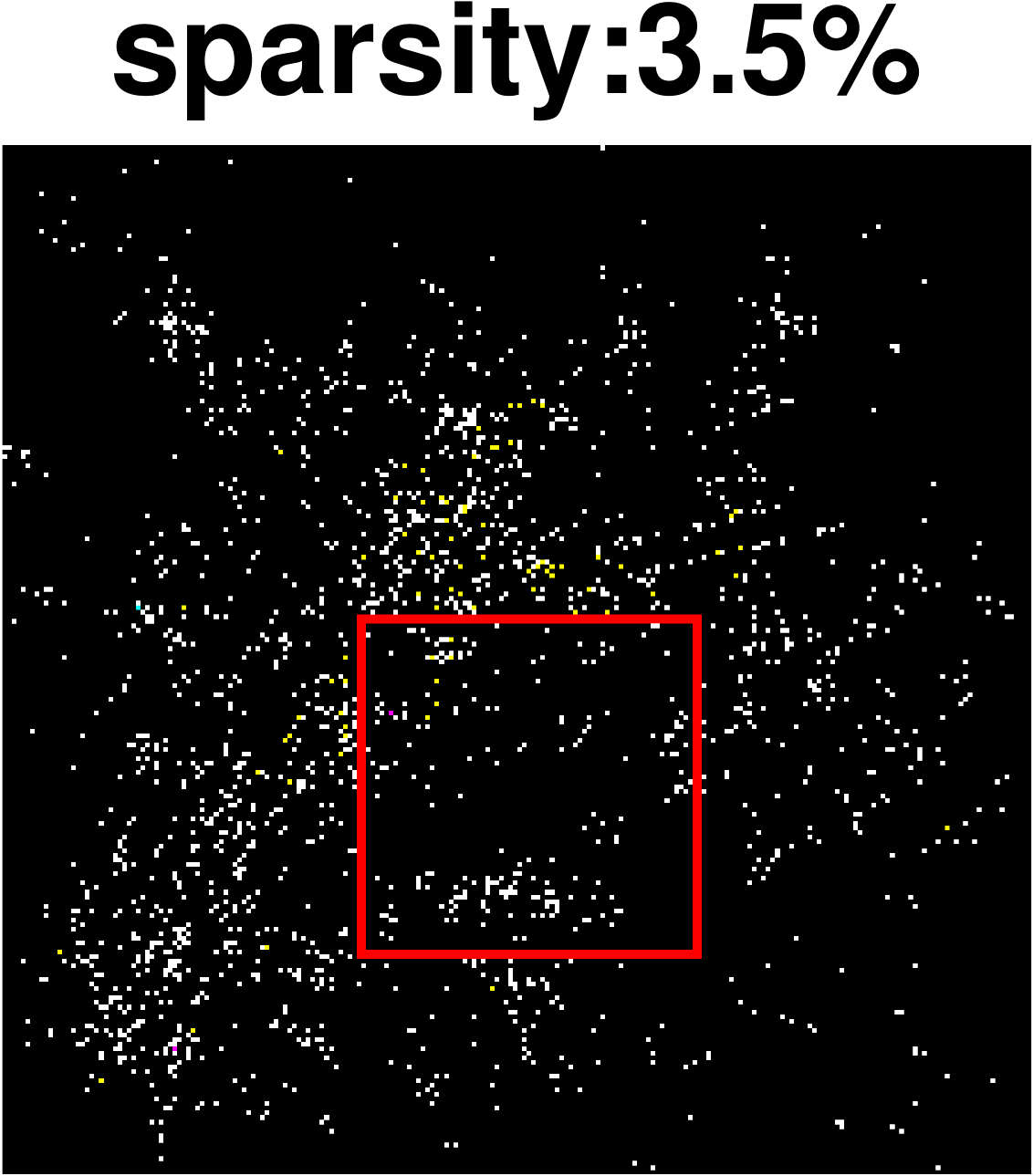}&
		
		\includegraphics[width=0.2\columnwidth]{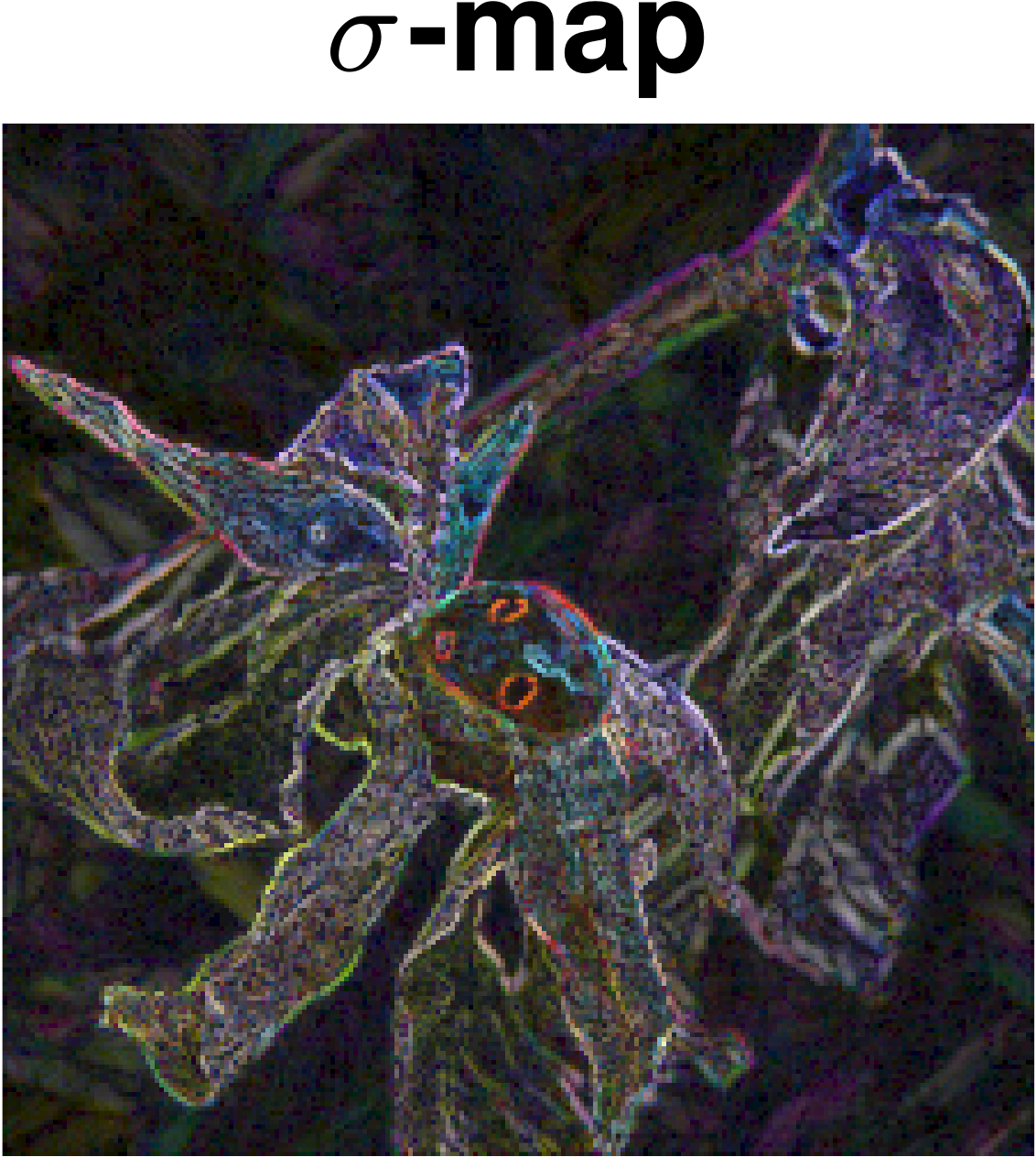}& 
		\includegraphics[width=0.2\columnwidth]{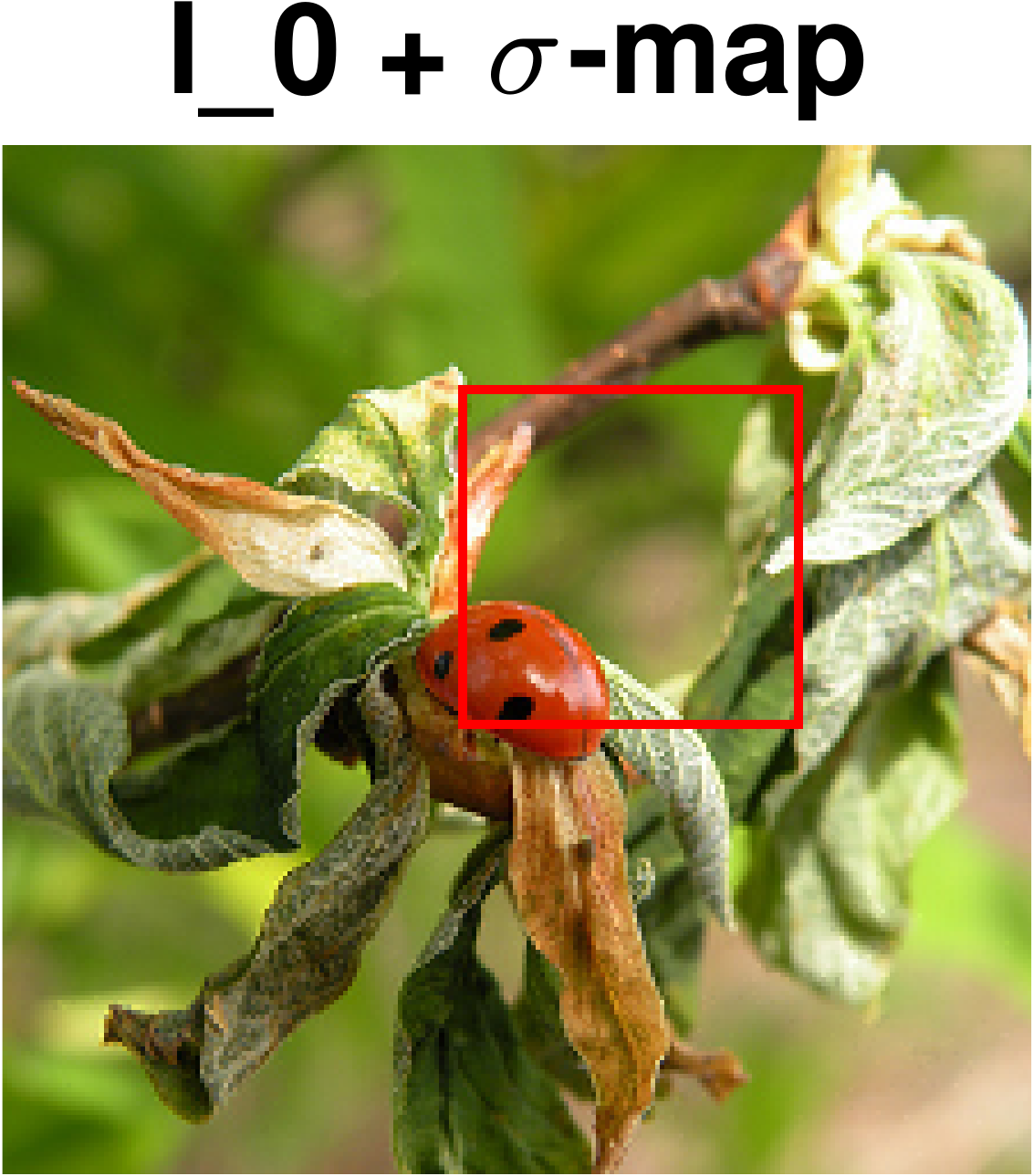}&
		\includegraphics[width=0.2\columnwidth]{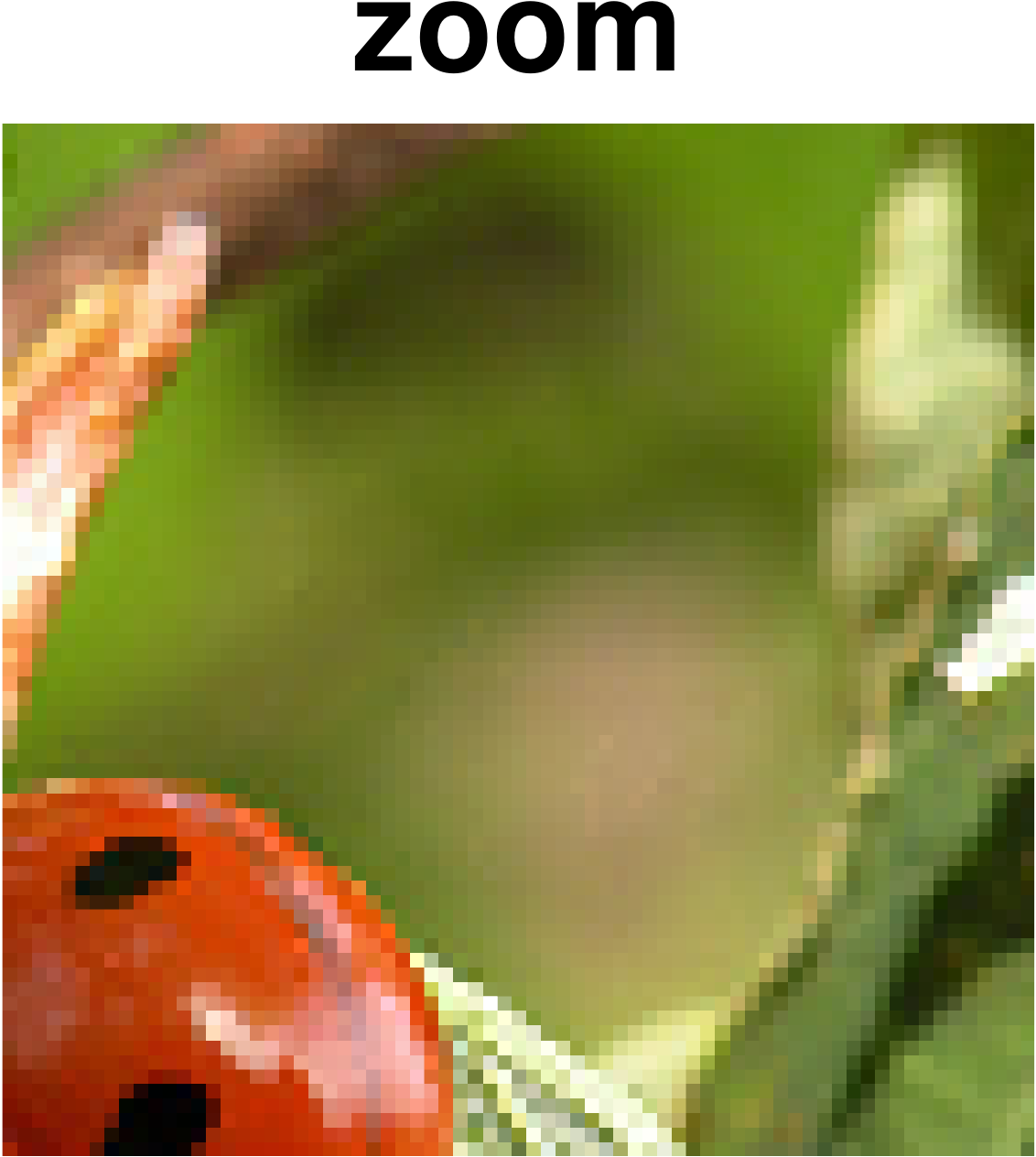}&
		\includegraphics[width=0.2\columnwidth]{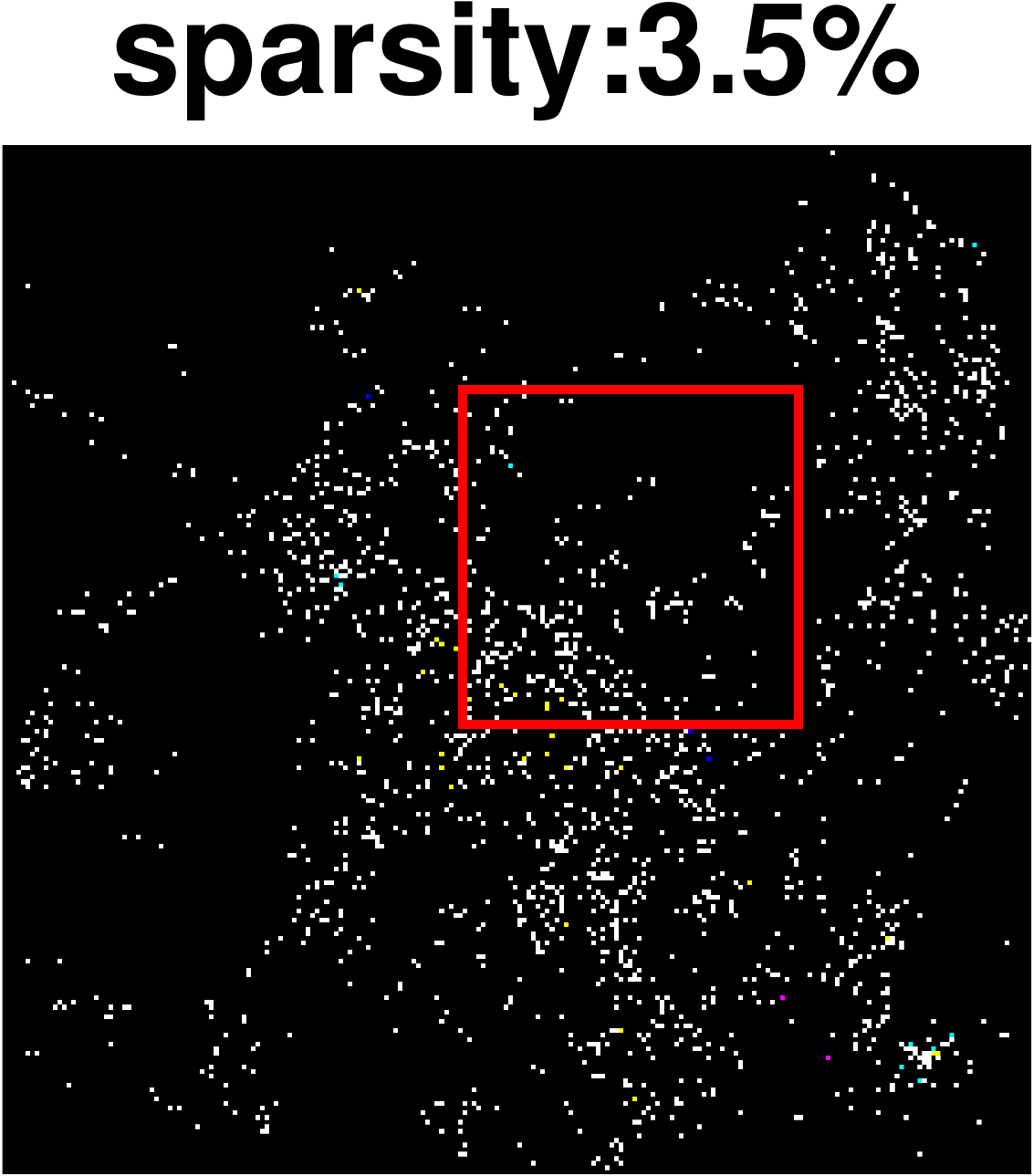} \\
	\end{tabular}
	\caption{\textbf{Different attacks on Restricted ImageNet.} We illustrate the differences of the adversarial examples (second column, zoom in third column) found by CornerSearch ($l_0$), $l_0+l_\infty$-attack and $\sigma$-CornerSearch, respectively first, second and third row. The fourth column shows the map of the modified pixels (\textit{sparsity} column). The original image is in the top left and the RGB representation of the $\sigma$-map, rescaled so that $\max_{i,j}\sigma_{ij}=1$, bottom left.} \label{fig:imp_ImageNet_app}
	\end{figure*}

\section{Adversarial examples of $\sigma$-PGD}
We want here to compare the adversarial examples generated by our two methods, $\sigma$-CornerSearch and $\sigma$-PGD. In Figures \ref{fig:pgd_MNIST} and \ref{fig:pgd_CIFAR-10} (see also \url{https://github.com/fra31/sparse-imperceivable-attacks}) we show the perturbed images crafted by the two attacks, as well as the original images and the modifications rescaled so that each component is in [0,1] and the largest one equals 1. Moreover, for $\sigma$-PGD we report the results obtained with a smaller $\kappa$. The gray images indicate unsuccessful cases.\\
It is clear that $\sigma$-CornerSearch produces sparser perturbations. Moreover, $\sigma$-PGD with the same $\kappa$ used for $\sigma$-CS gives more visible manipulations. We think that this is due to two reasons: first, the whole budget of $k$ pixels to modify is always used by $\sigma$-PGD, while this does not happen with $\sigma$-CS, and second, $\sigma$-PGD aims at maximizing the loss inside the space of the allowed perturbations. This is possibly achieved by modifying neighboring pixels, which sometimes have slightly different colors, in opposite directions (that is with $\lambda_i$ with different signs for different $i$). Conversely, $\sigma$-CornerSearch does not consider spatial relations among pixels, and thus it does not show this behaviour. However, as showed in the Figures, it is possible to recover less visible changes also for $\sigma$-PGD by decreasing $\kappa$, at the cost of a smaller success rate.
\begin{figure*}[p]
	\centering
	\begin{tabular}{c |c  c |c c |c c }
		\textbf{original}&\multicolumn{2}{c|}{\textbf{$\sigma$-CornerSearch}, $\kappa=0.8$} &\multicolumn{2}{c|}{\textbf{$\sigma$-PGD}, $\kappa=0.8$} &\multicolumn{2}{c}{\textbf{$\sigma$-PGD}, $\kappa=0.6$}\\
		\includegraphics[width=0.22\columnwidth, clip, trim=40mm 10mm 35mm 5mm]{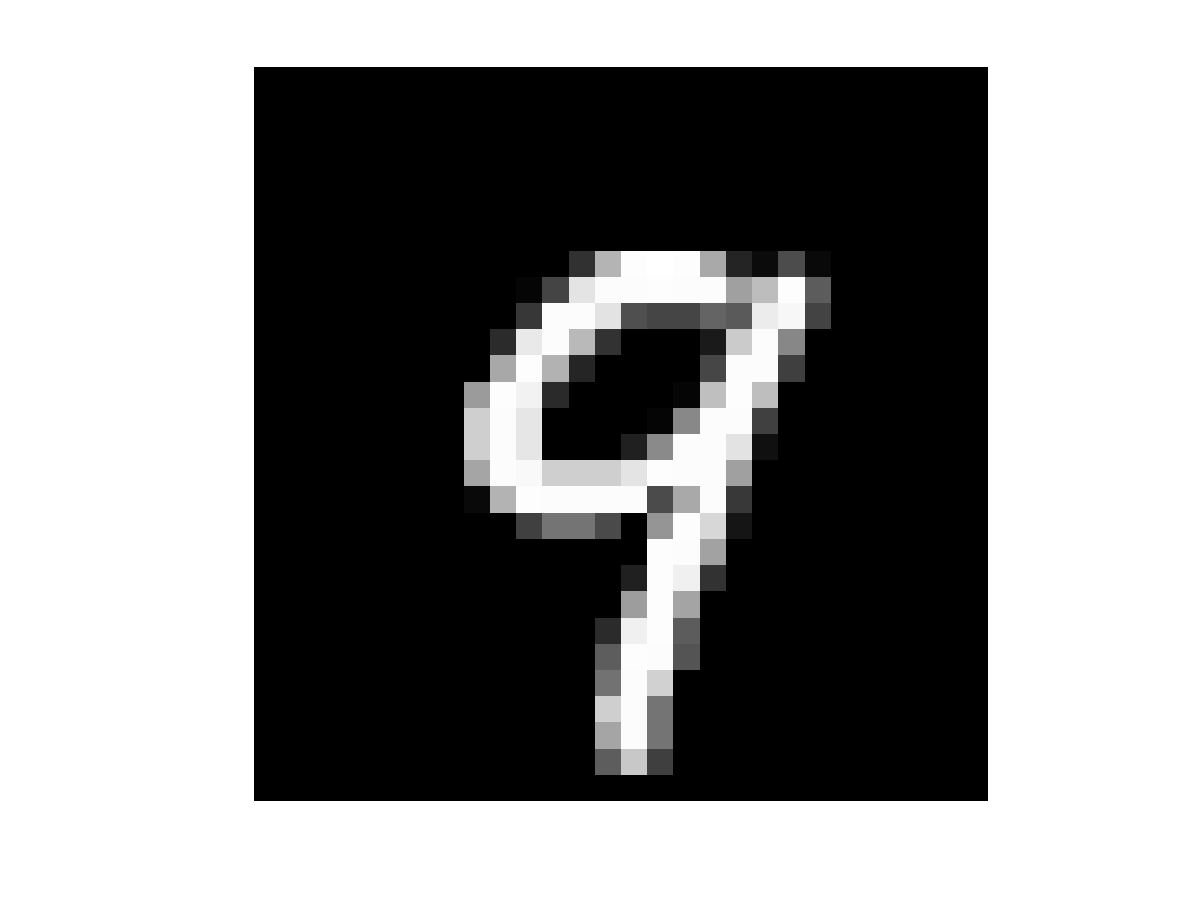}&
		\includegraphics[width=0.22\columnwidth, clip, trim=40mm 10mm 35mm 5mm]{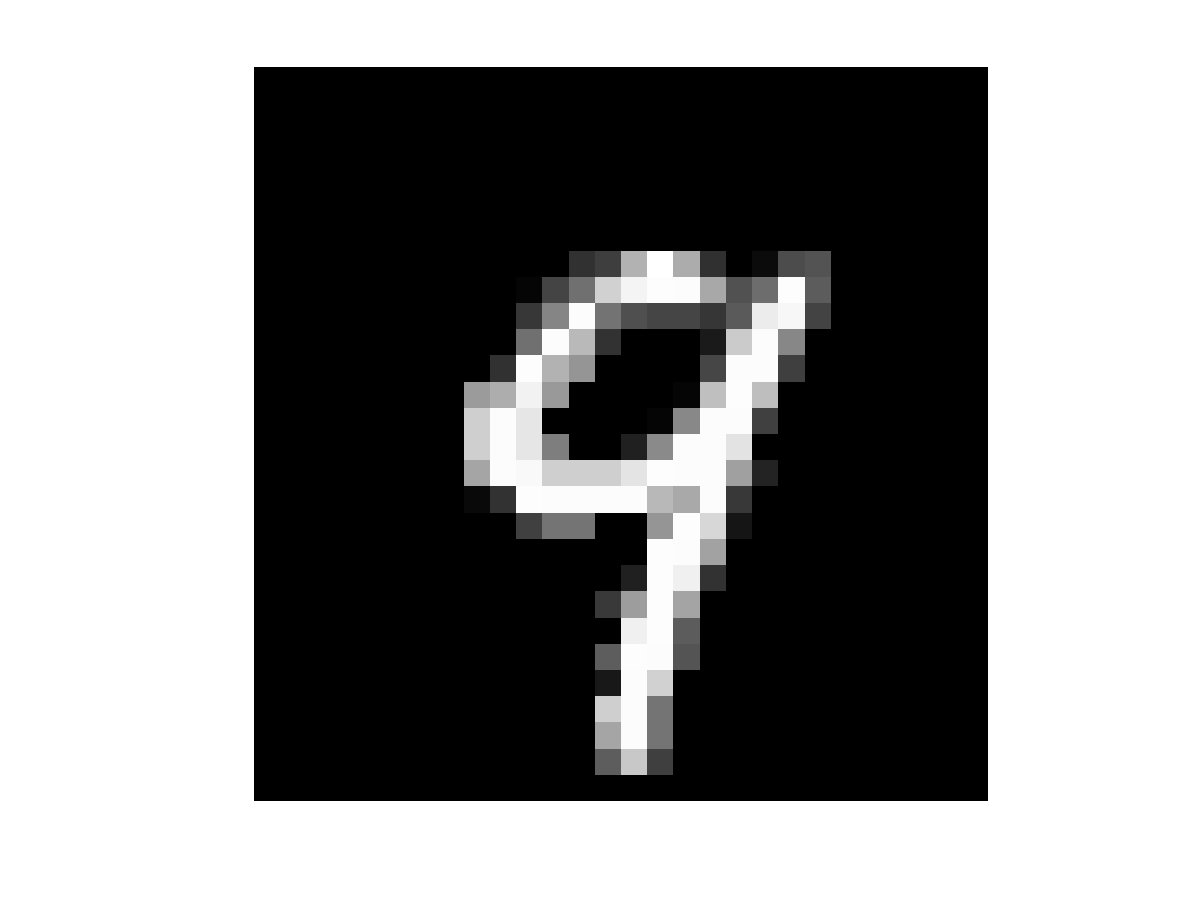}&
		\includegraphics[width=0.22\columnwidth, clip, trim=40mm 10mm 35mm 5mm]{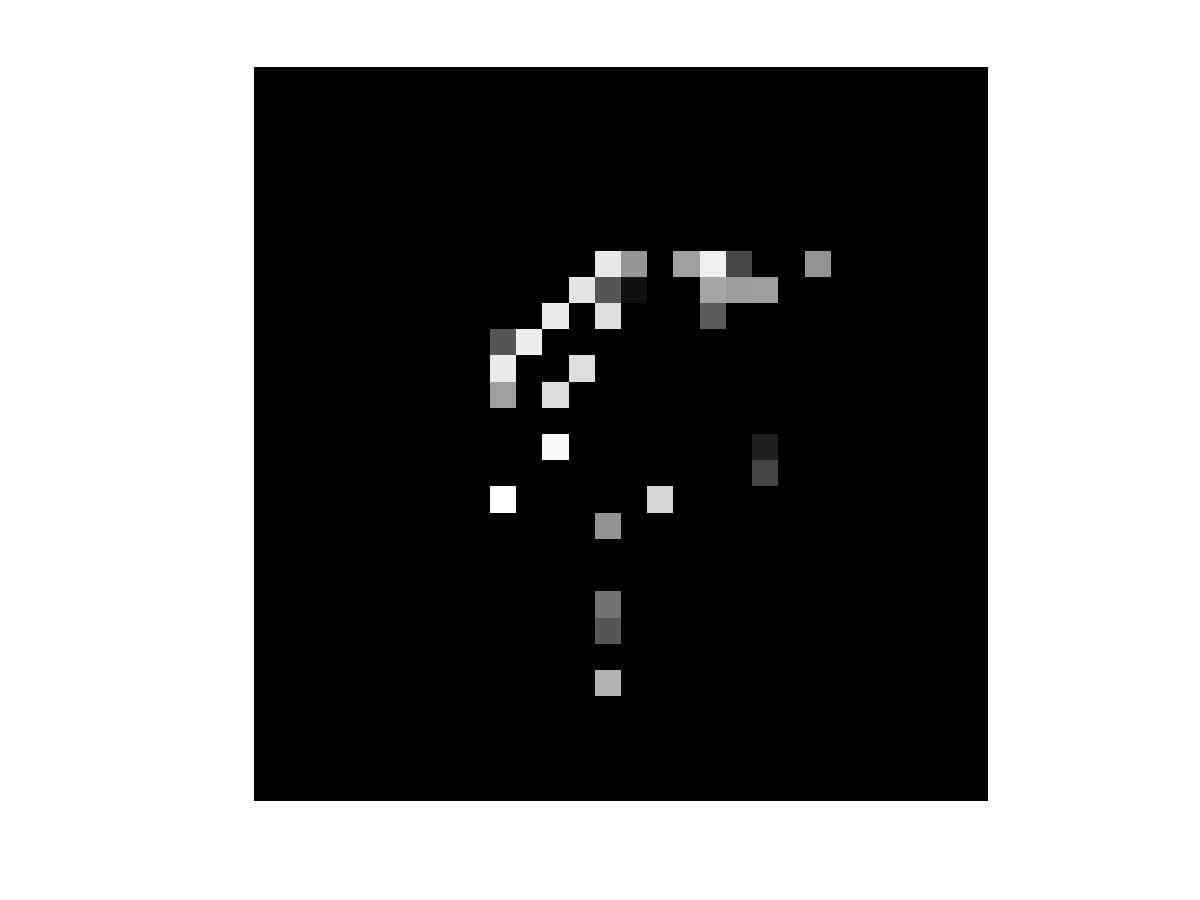}&
		\includegraphics[width=0.22\columnwidth, clip, trim=40mm 10mm 35mm 5mm]{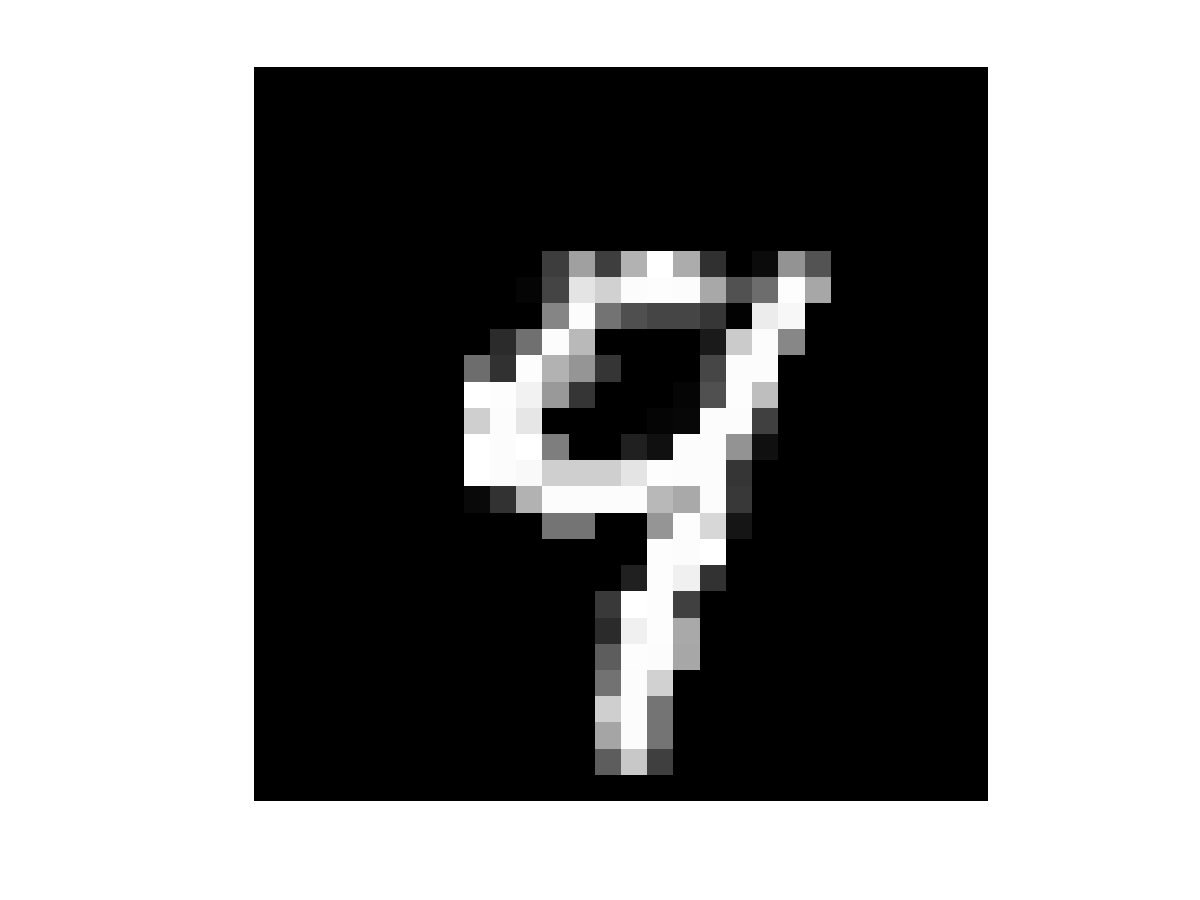}&
		\includegraphics[width=0.22\columnwidth, clip, trim=40mm 10mm 35mm 5mm]{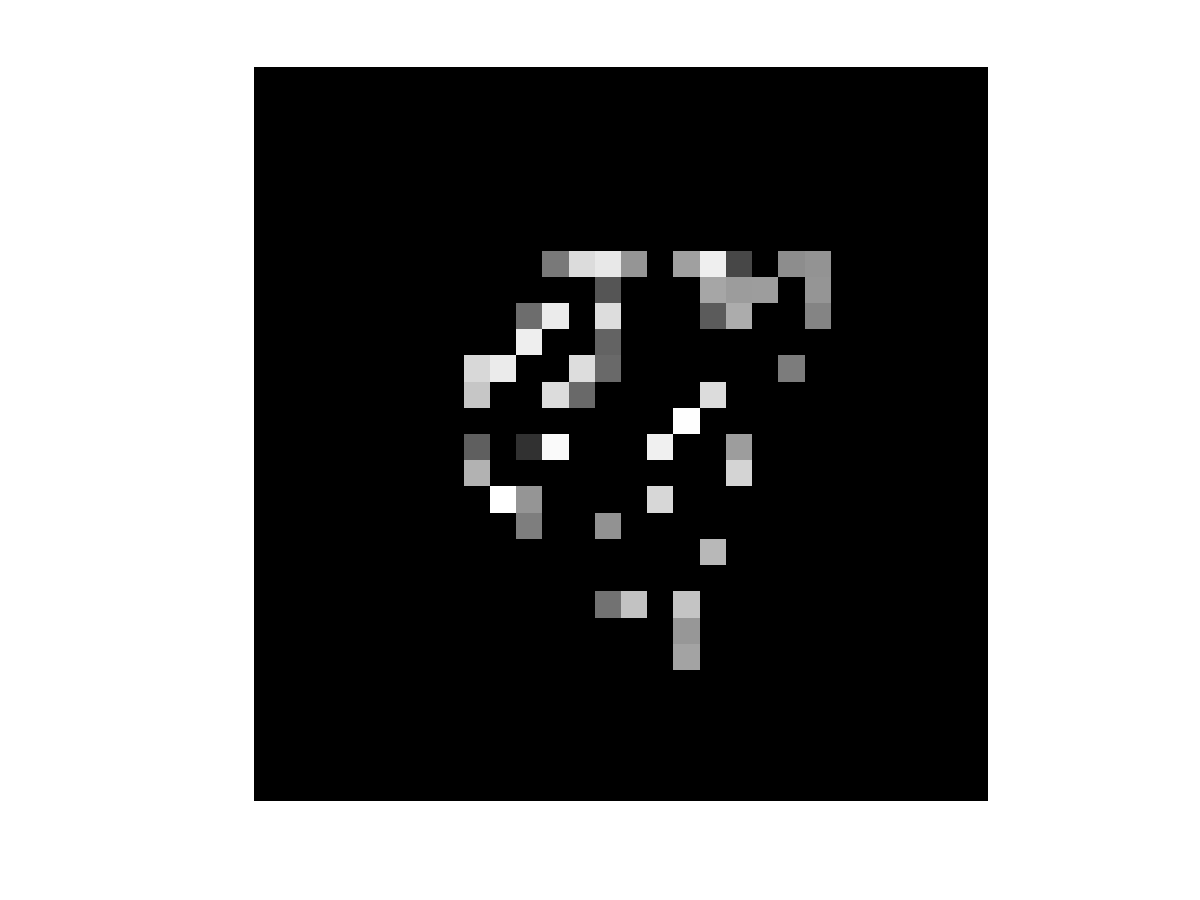}&
		\includegraphics[width=0.22\columnwidth, clip, trim=40mm 10mm 35mm 5mm]{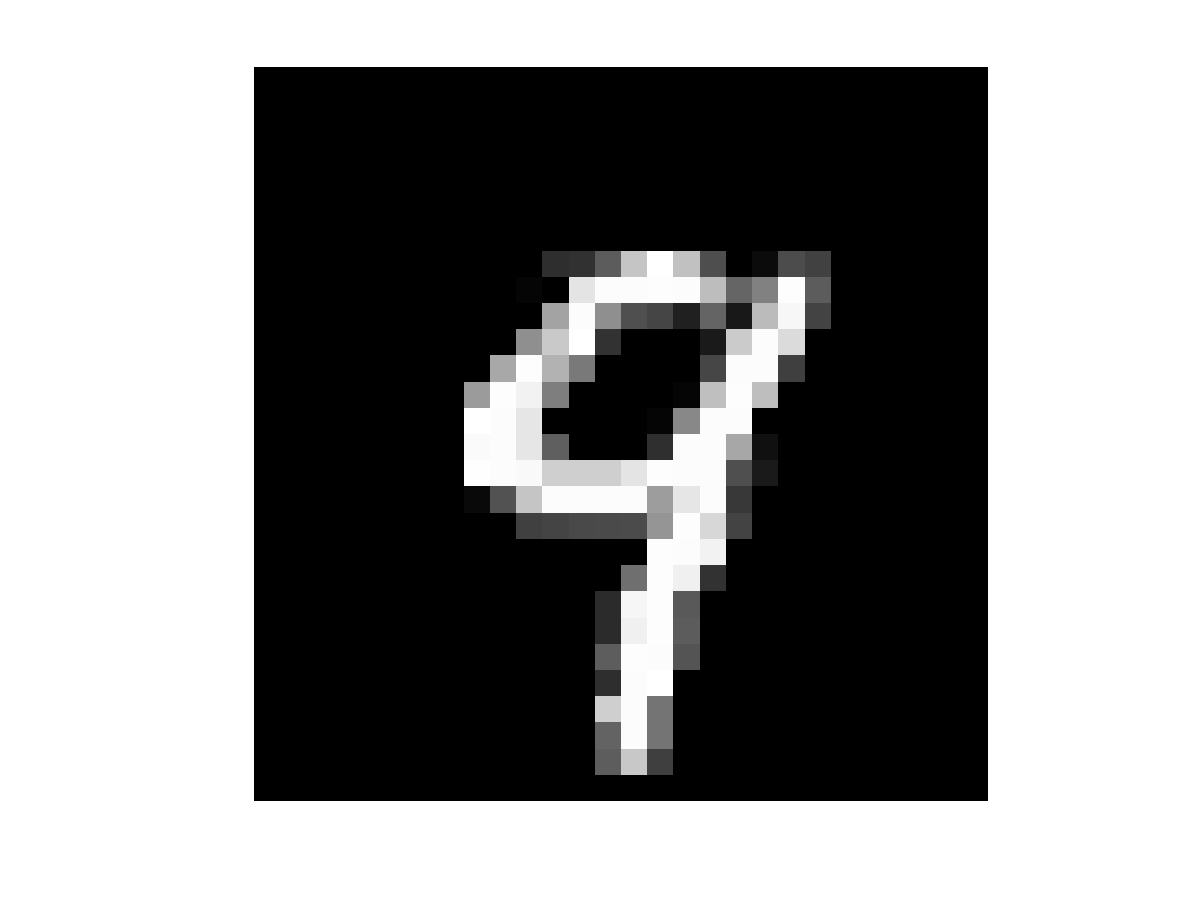}&
		\includegraphics[width=0.22\columnwidth, clip, trim=40mm 10mm 35mm 5mm]{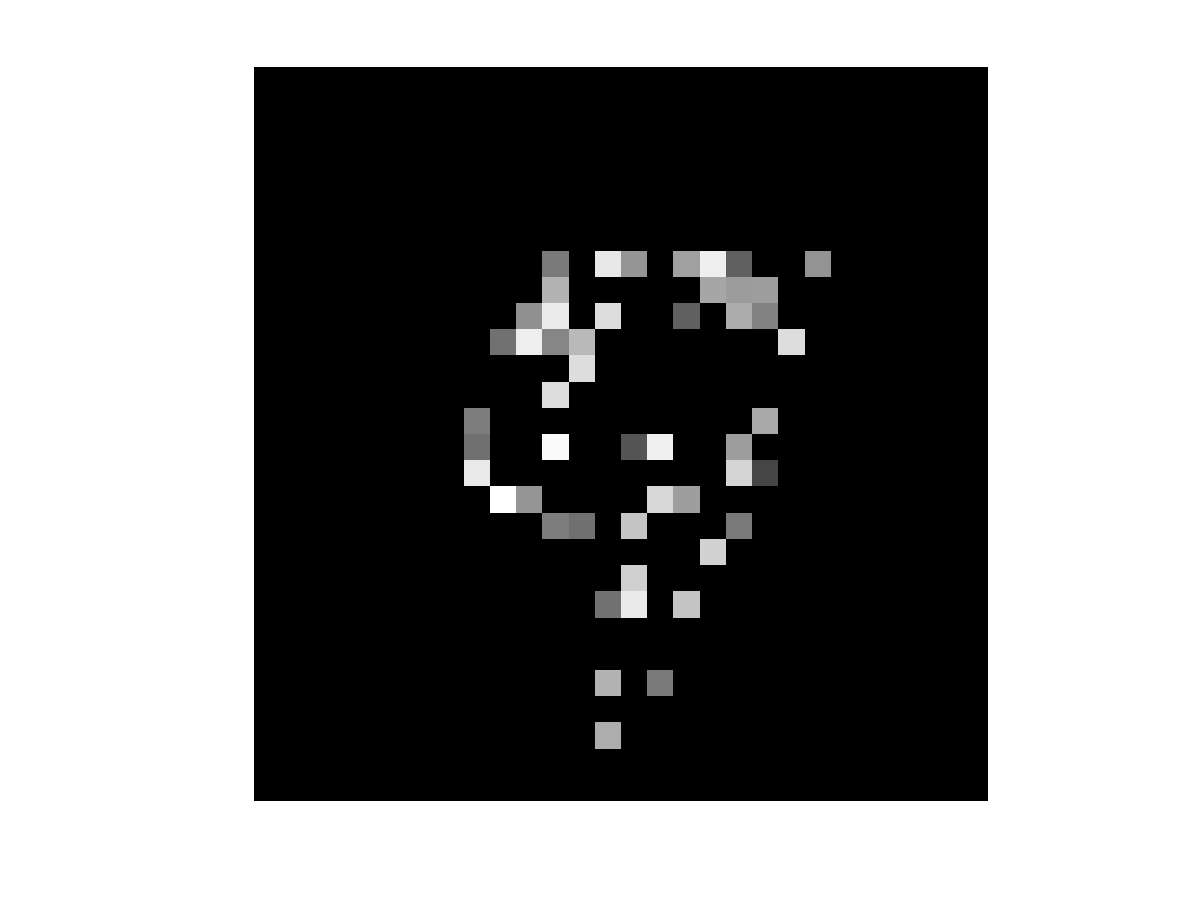}\\
		
		\includegraphics[width=0.22\columnwidth, clip, trim=40mm 10mm 35mm 5mm]{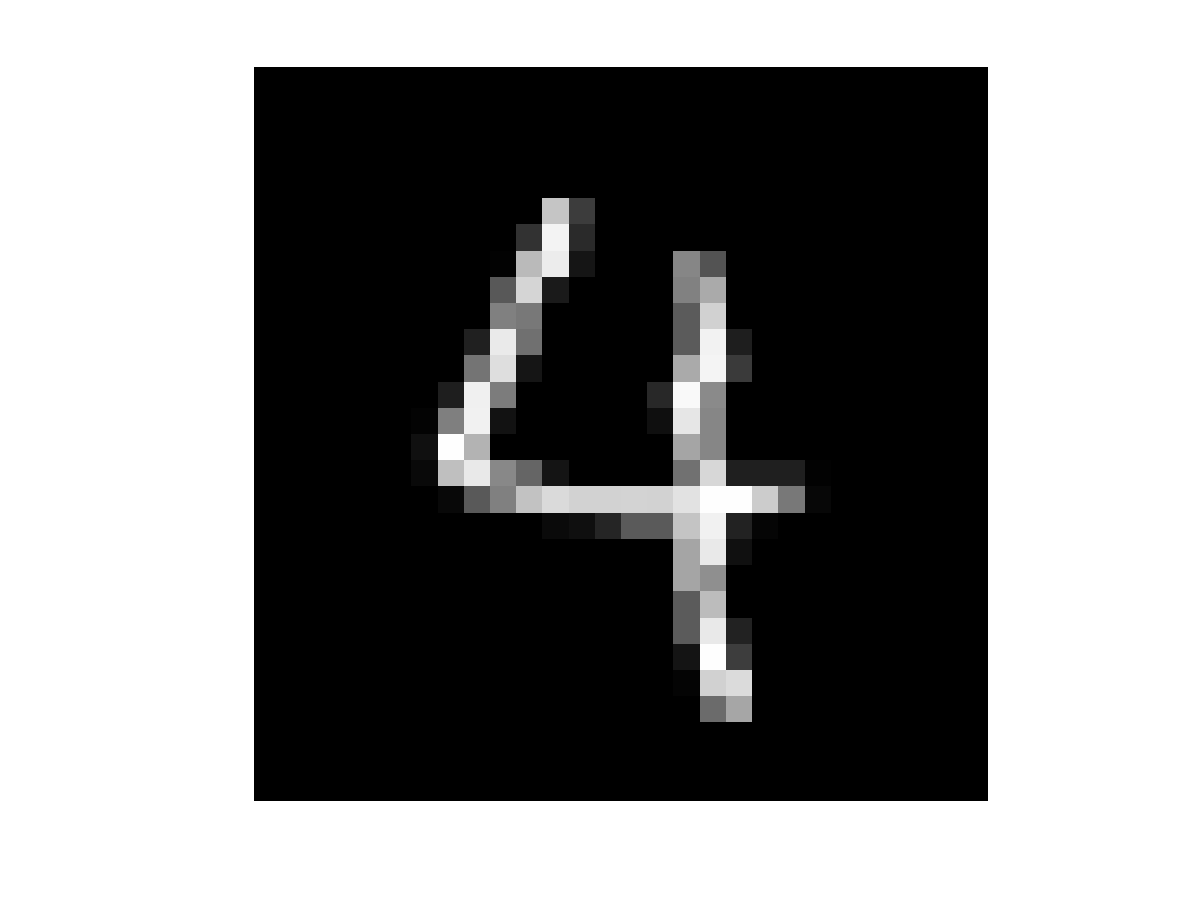}&
		\includegraphics[width=0.22\columnwidth, clip, trim=40mm 10mm 35mm 5mm]{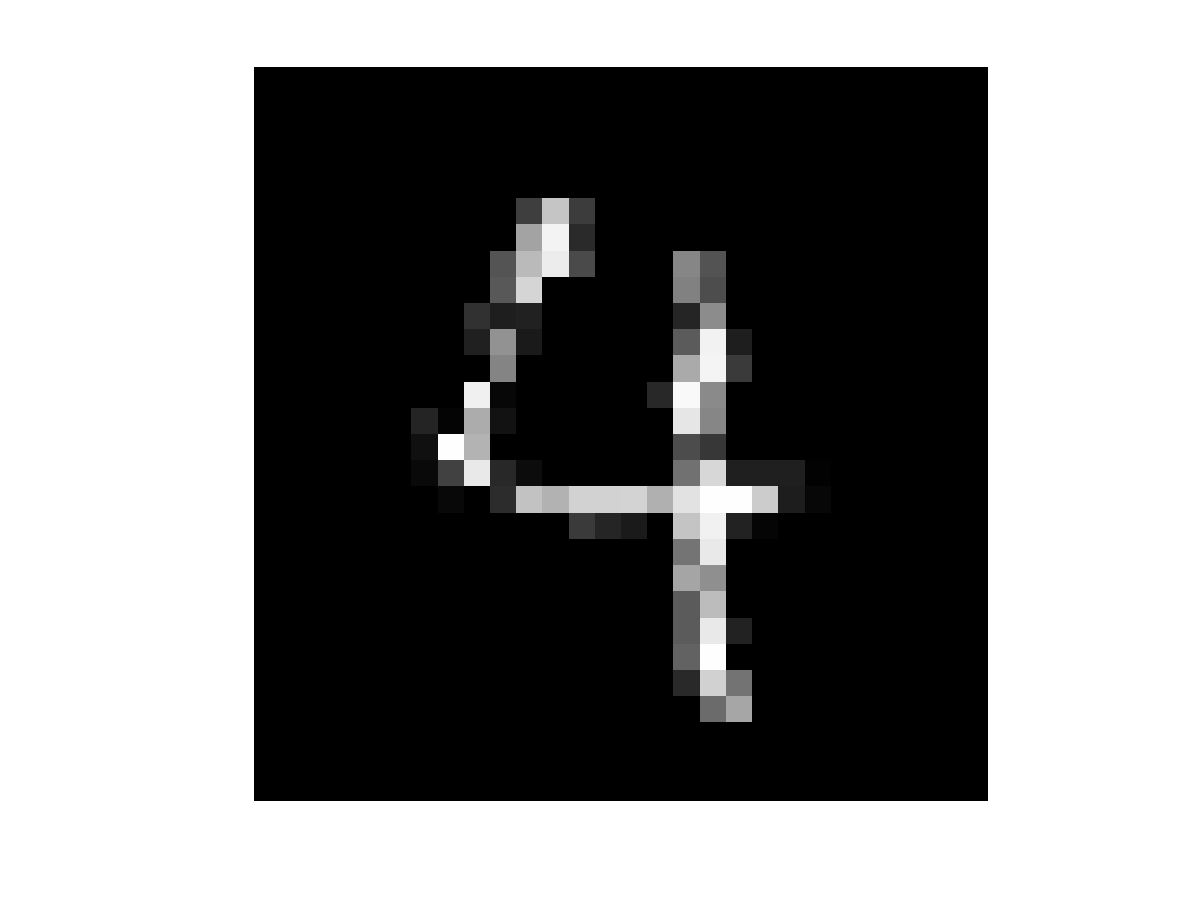}&
		\includegraphics[width=0.22\columnwidth, clip, trim=40mm 10mm 35mm 5mm]{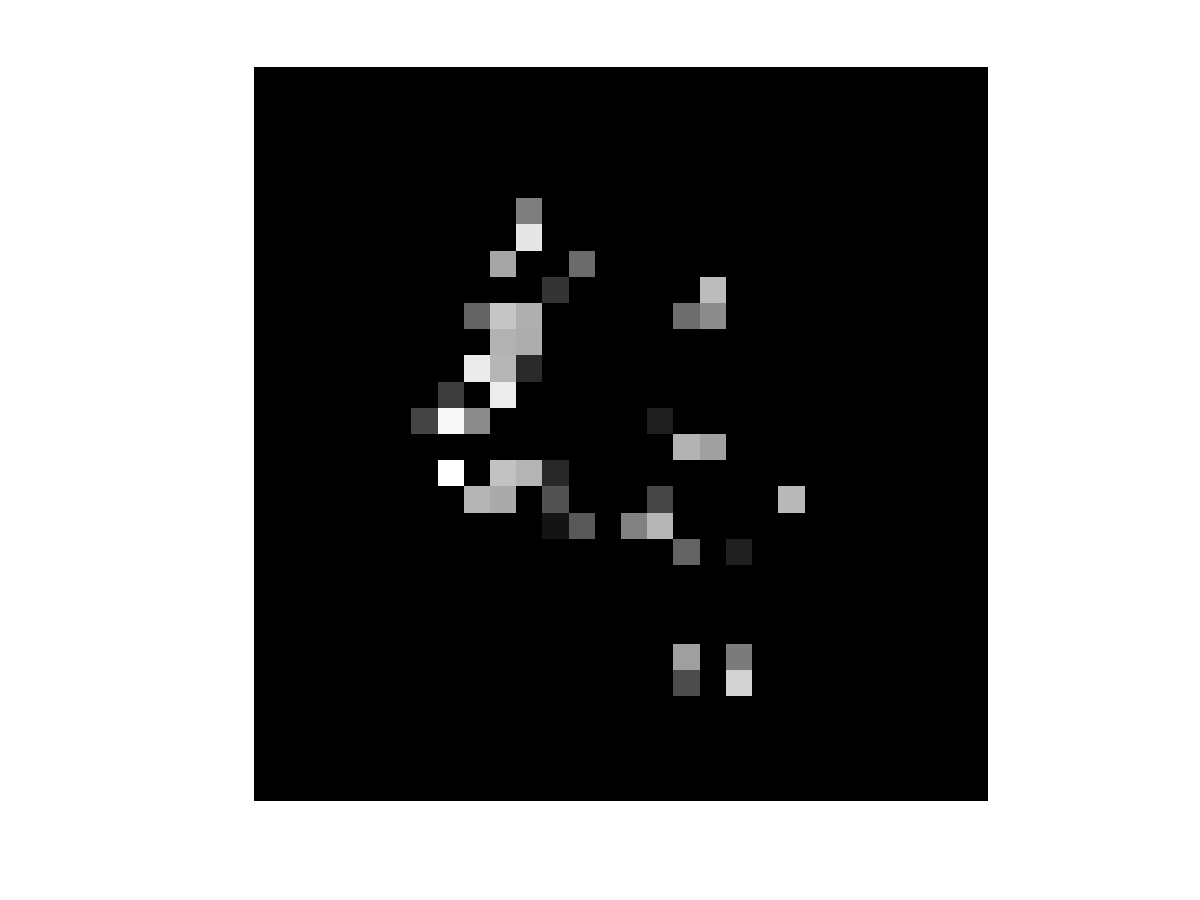}&
		\includegraphics[width=0.22\columnwidth, clip, trim=40mm 10mm 35mm 5mm]{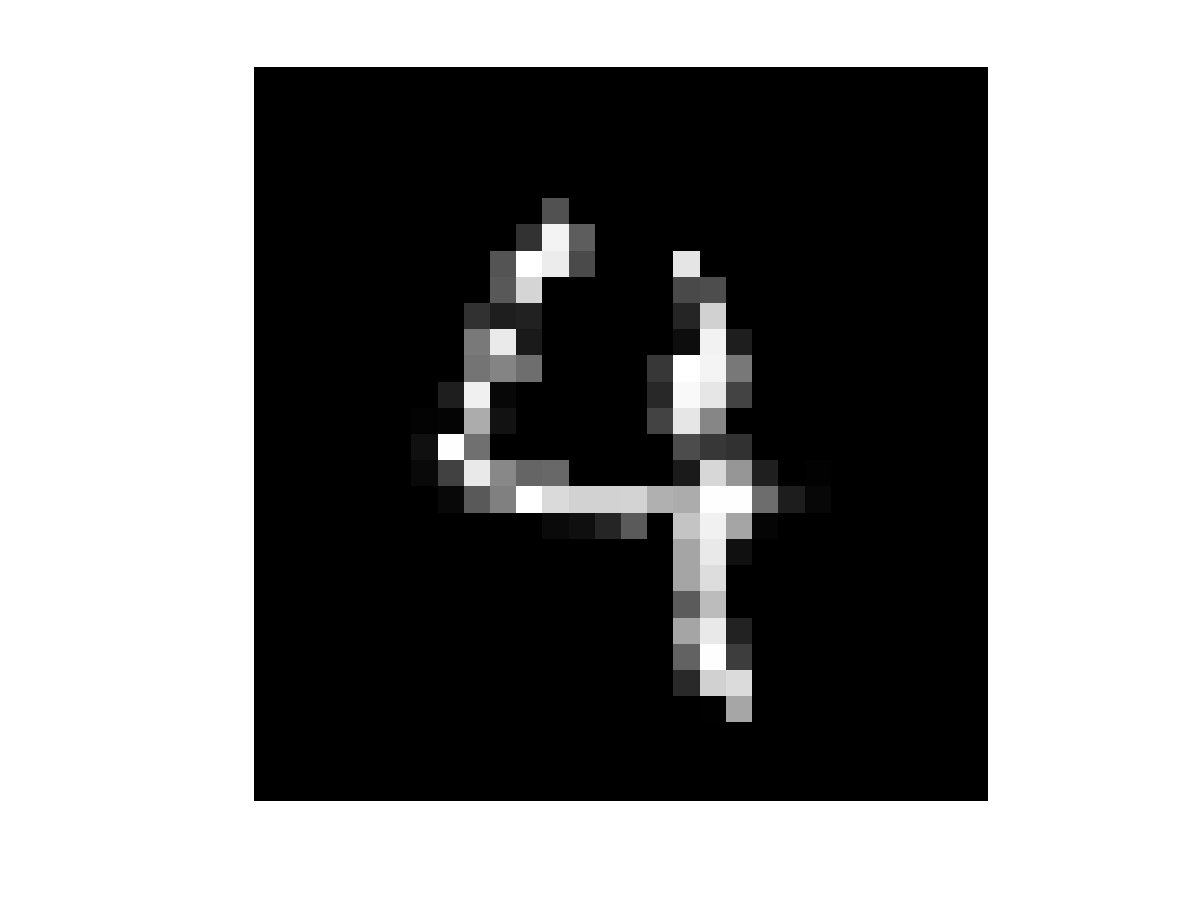}&
		\includegraphics[width=0.22\columnwidth, clip, trim=40mm 10mm 35mm 5mm]{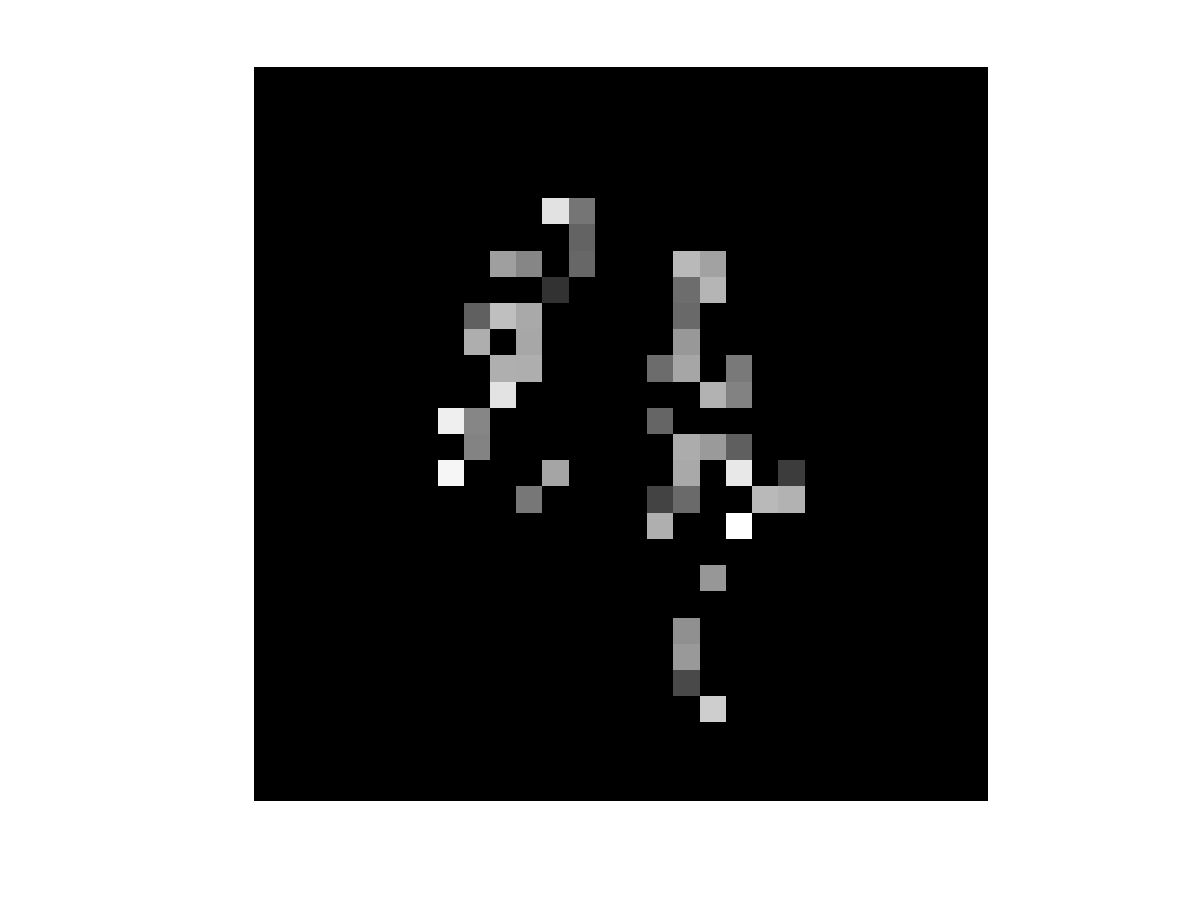}&
		\includegraphics[width=0.22\columnwidth, clip, trim=40mm 10mm 35mm 5mm]{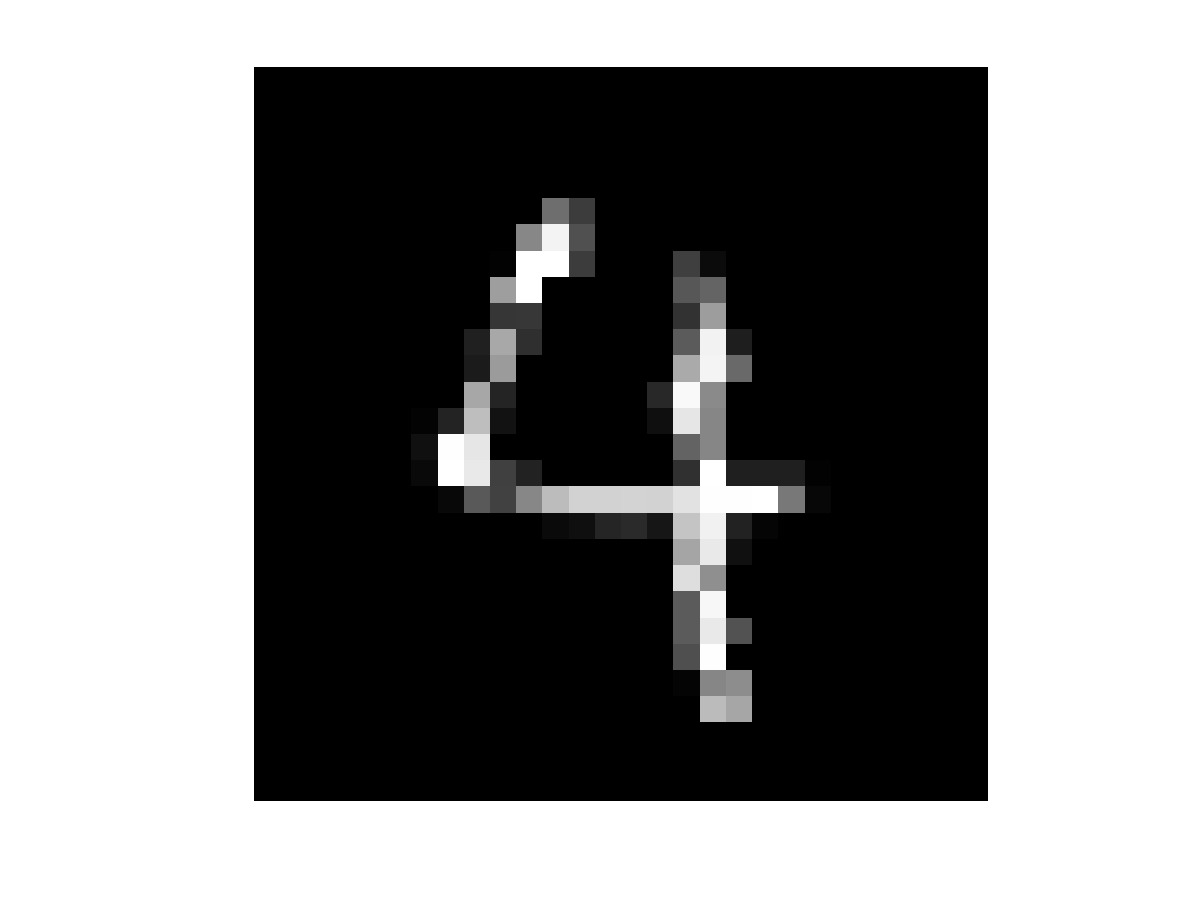}&
		\includegraphics[width=0.22\columnwidth, clip, trim=40mm 10mm 35mm 5mm]{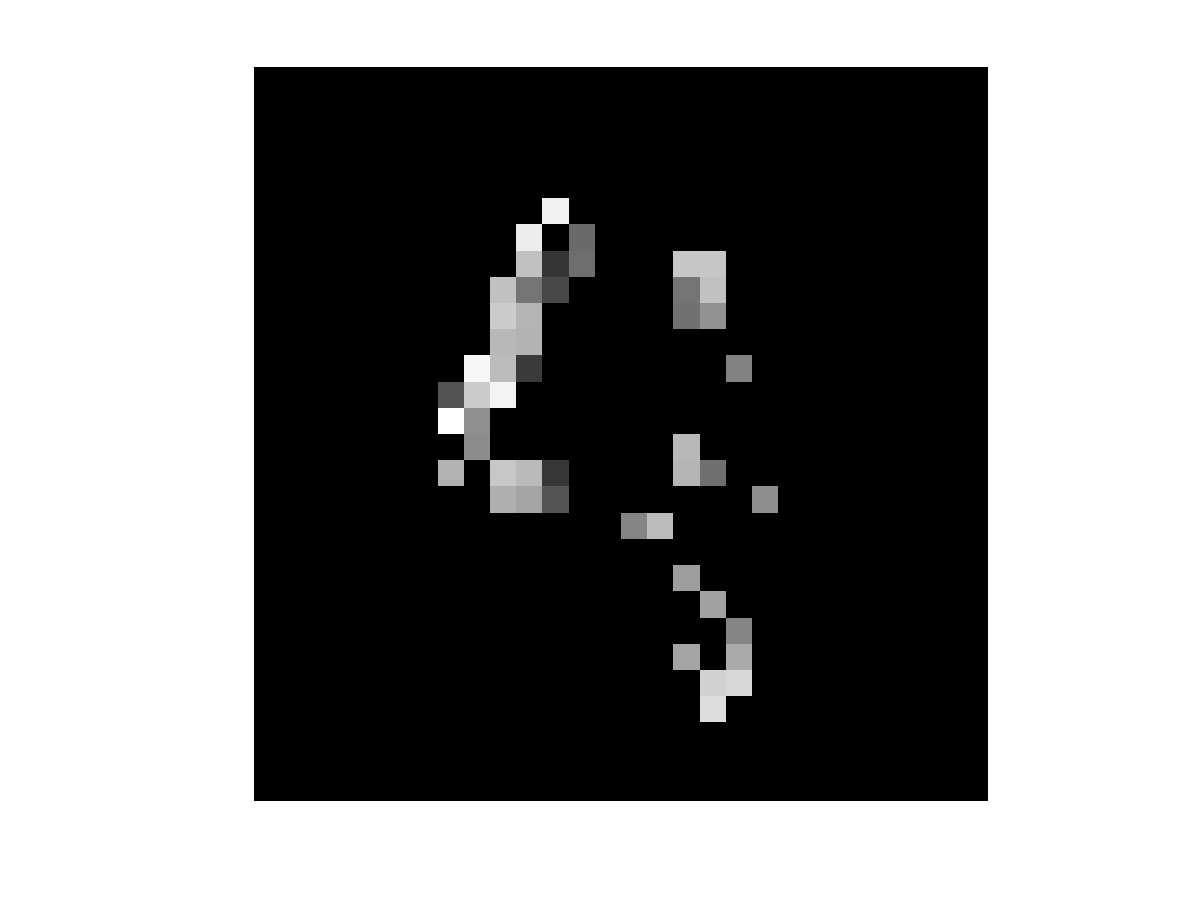}\\
		
		\includegraphics[width=0.22\columnwidth, clip, trim=40mm 10mm 35mm 5mm]{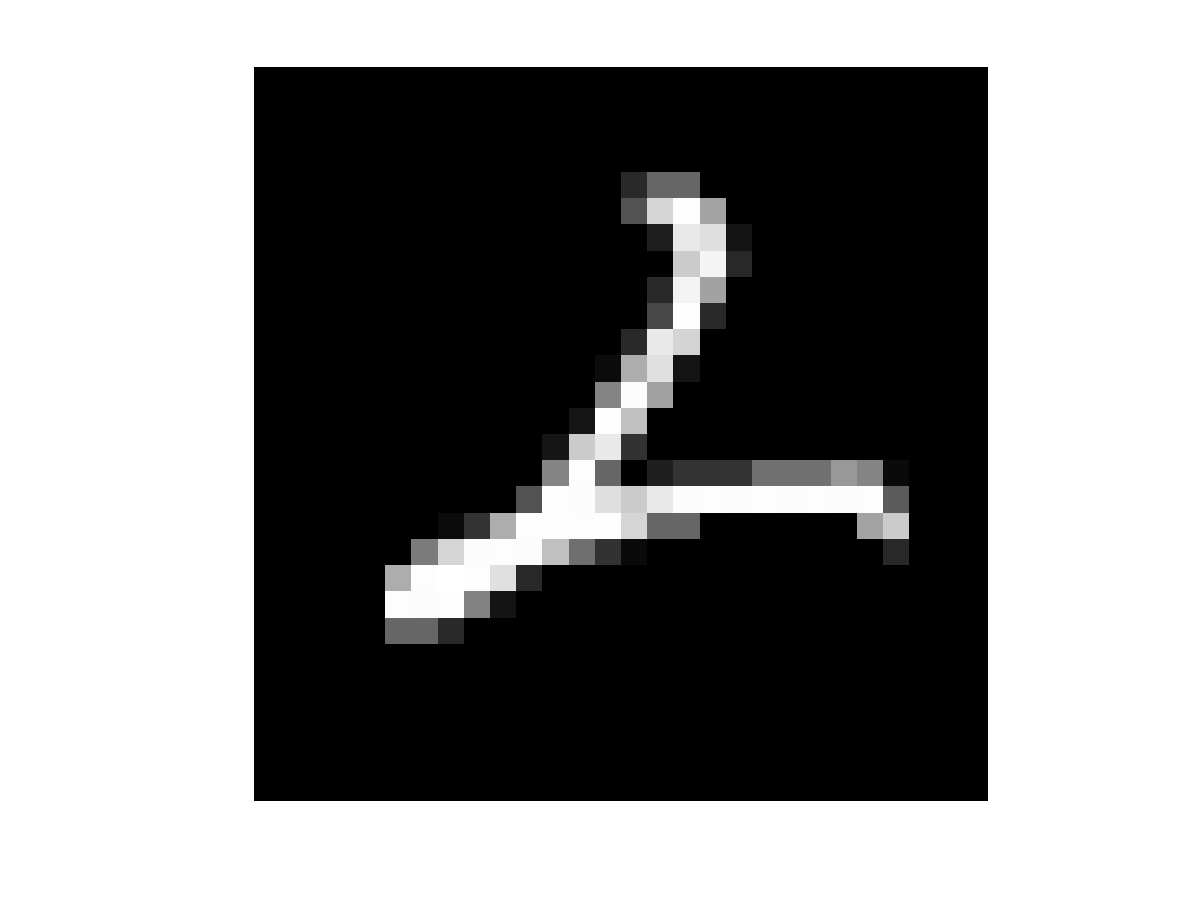}&
		\includegraphics[width=0.22\columnwidth, clip, trim=40mm 10mm 35mm 5mm]{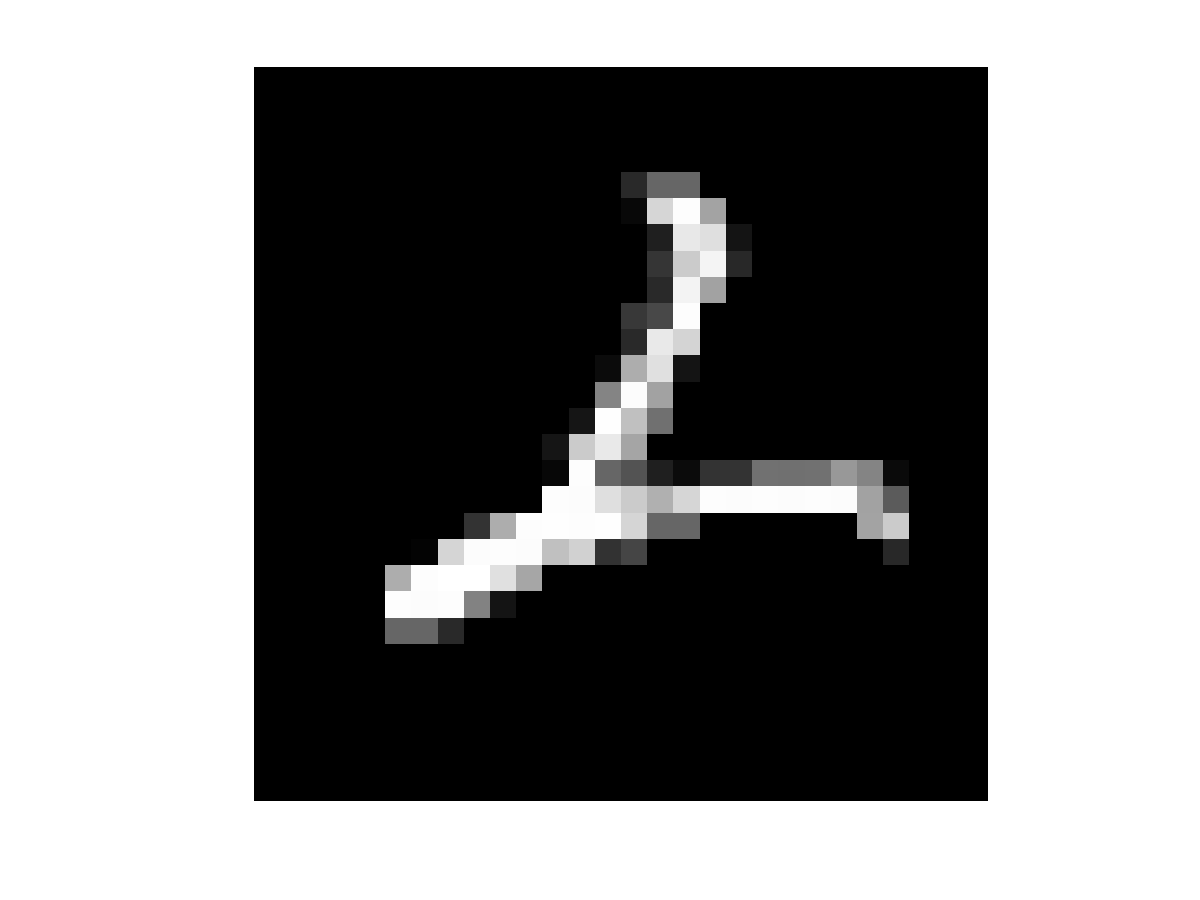}&
		\includegraphics[width=0.22\columnwidth, clip, trim=40mm 10mm 35mm 5mm]{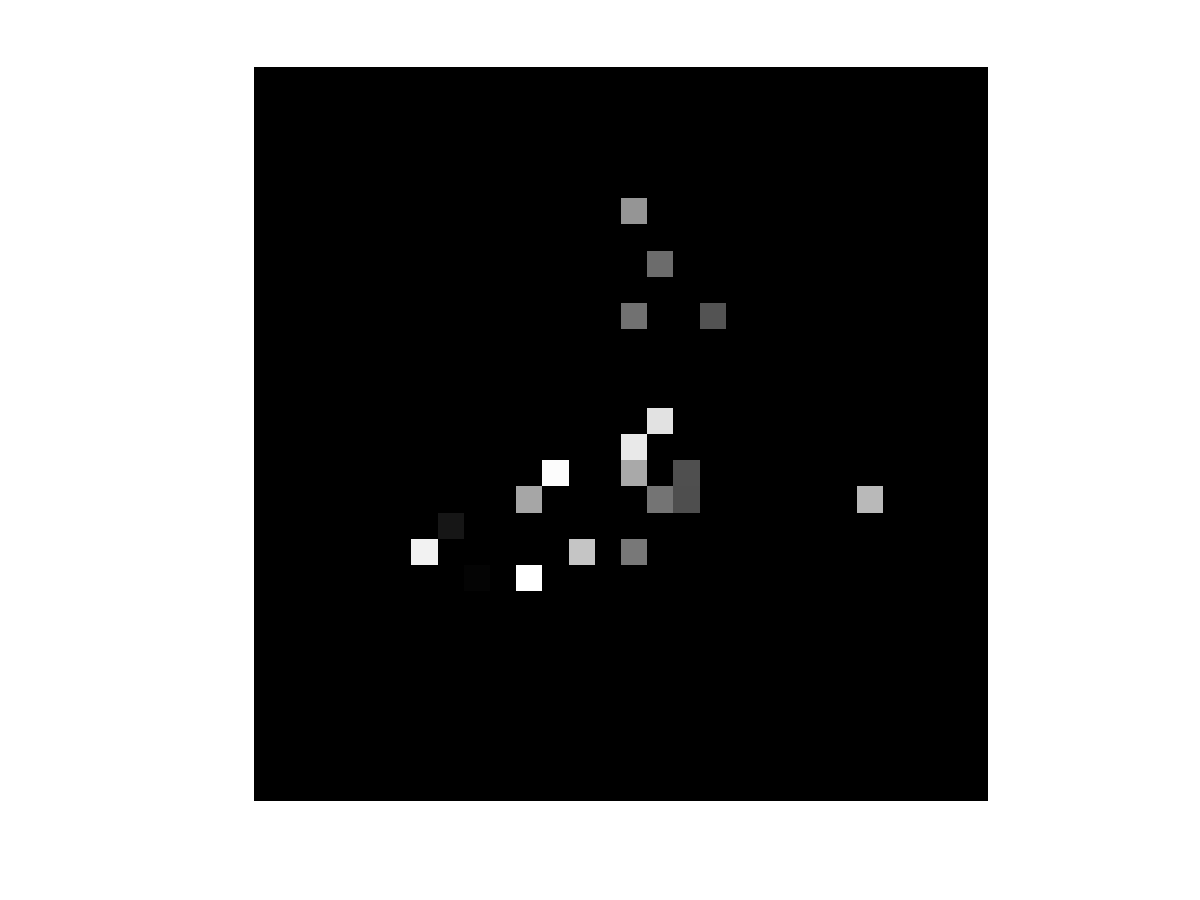}&
		\includegraphics[width=0.22\columnwidth, clip, trim=40mm 10mm 35mm 5mm]{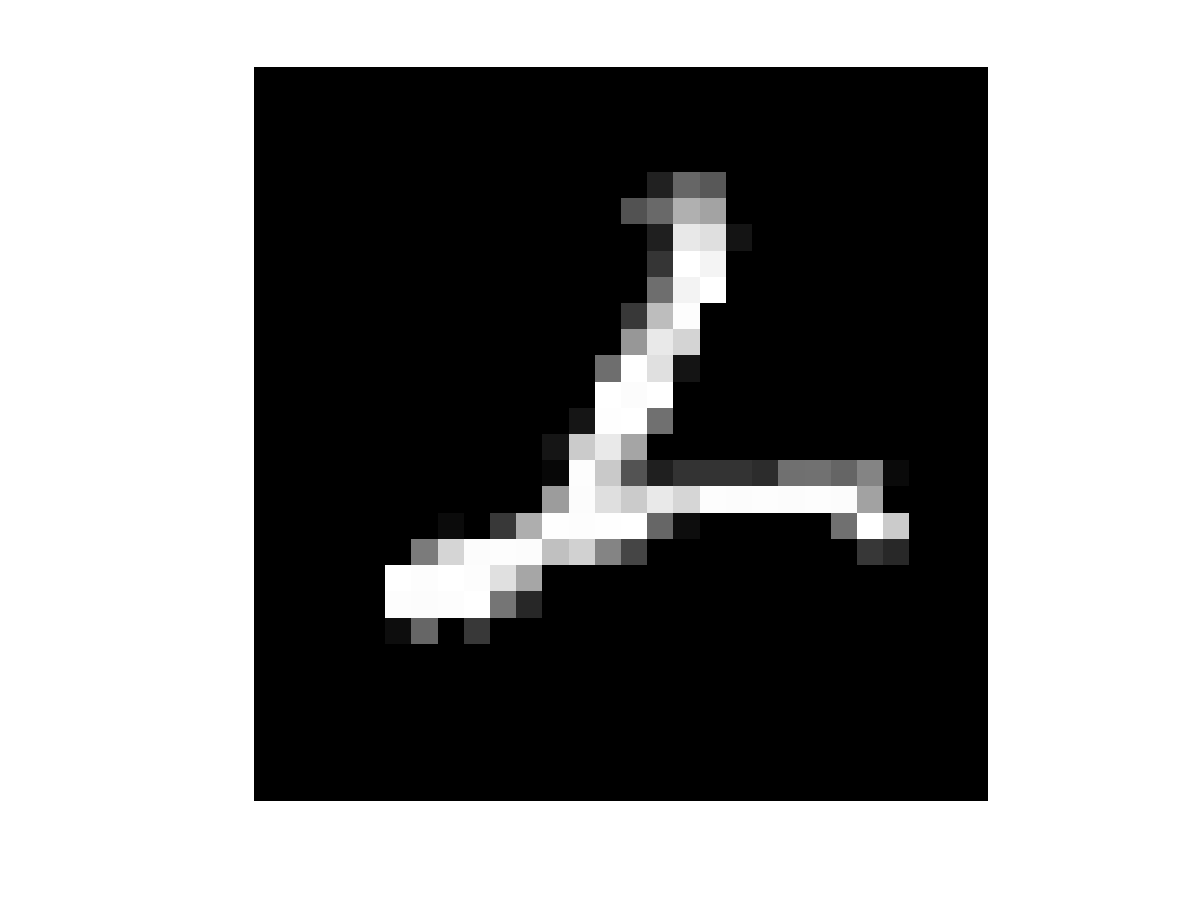}&
		\includegraphics[width=0.22\columnwidth, clip, trim=40mm 10mm 35mm 5mm]{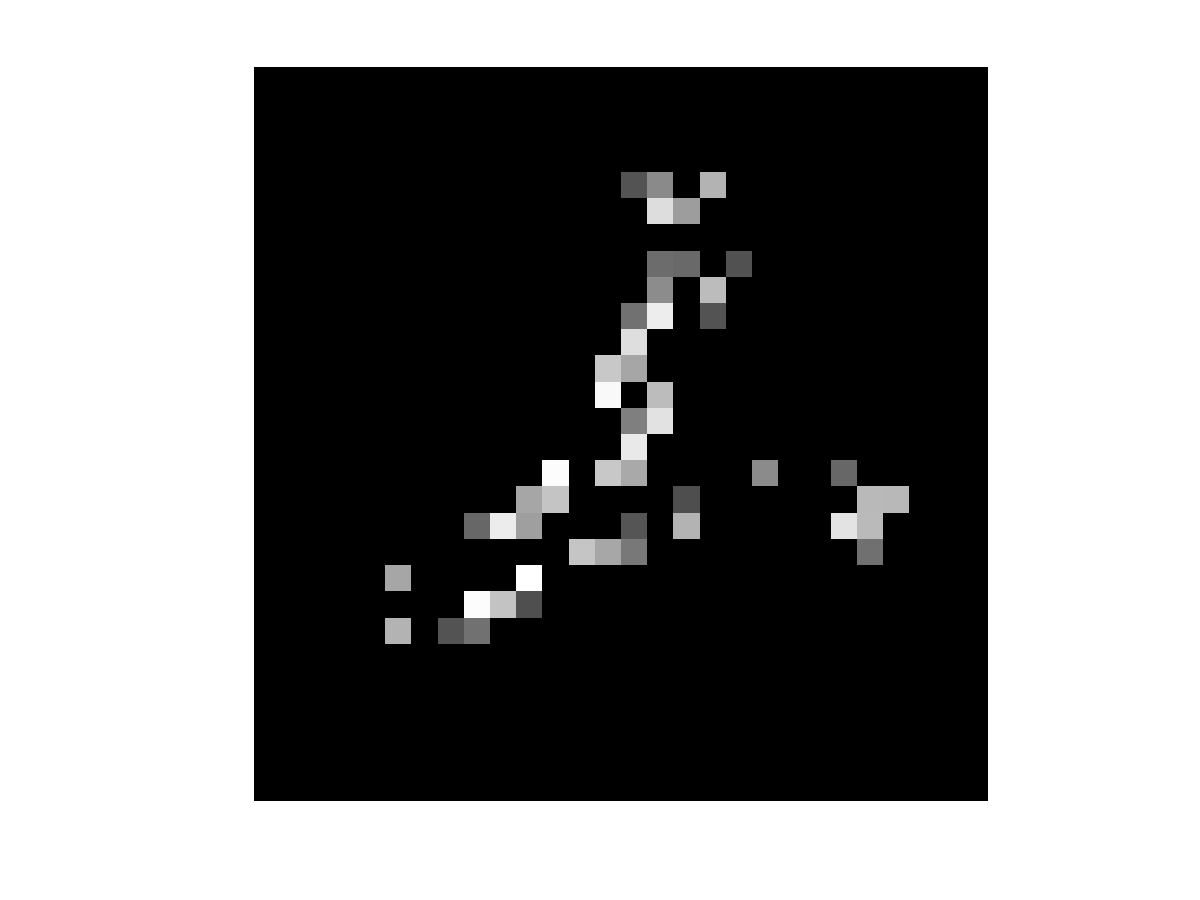}&
		\includegraphics[width=0.22\columnwidth, clip, trim=40mm 10mm 35mm 5mm]{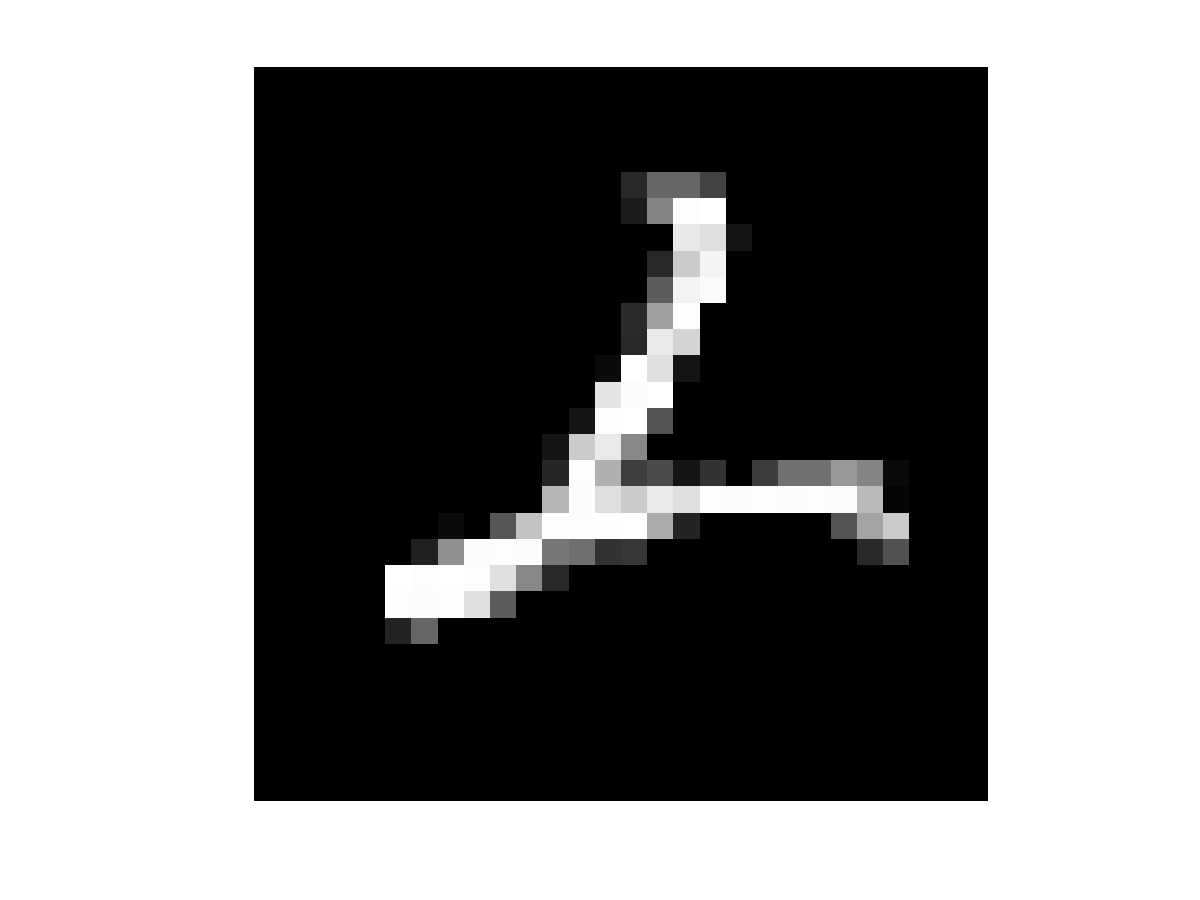}&
		\includegraphics[width=0.22\columnwidth, clip, trim=40mm 10mm 35mm 5mm]{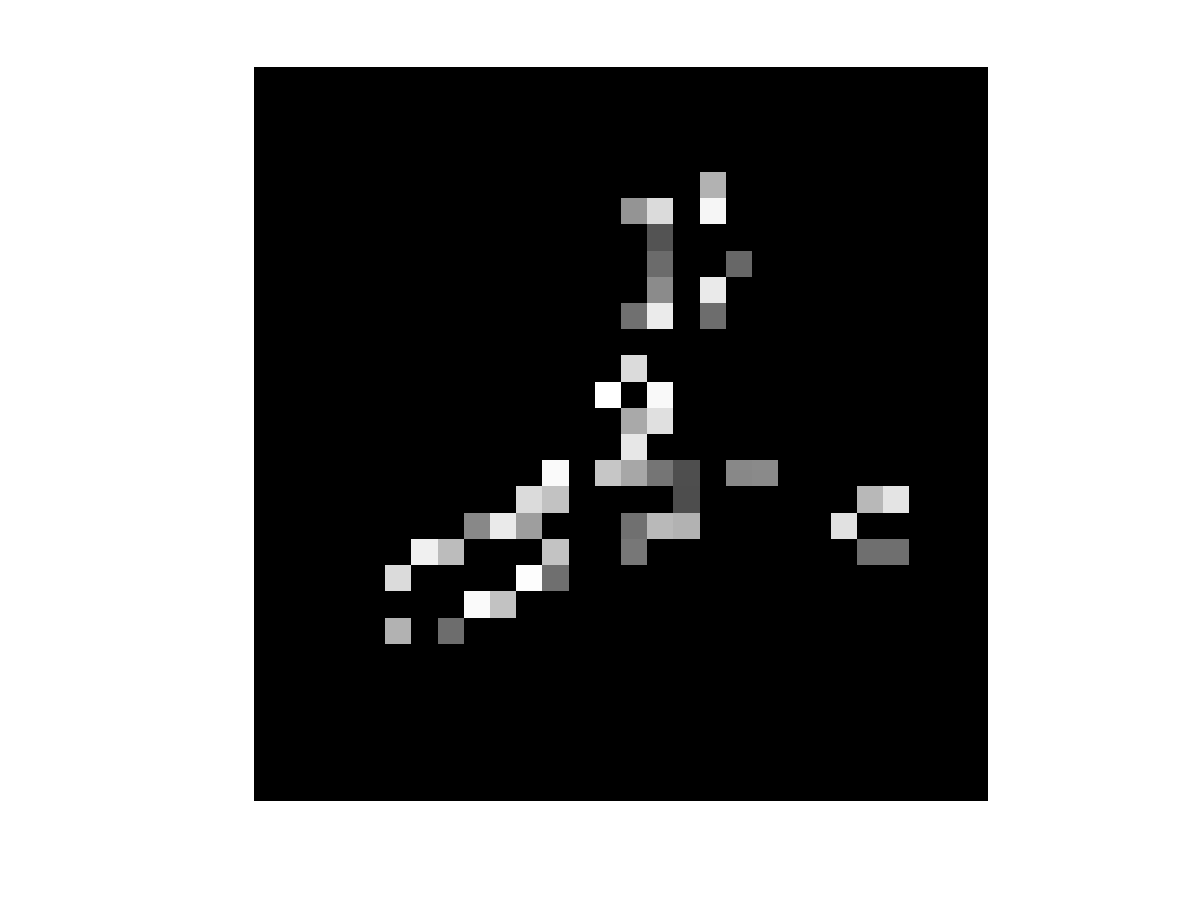}\\
		
		\includegraphics[width=0.22\columnwidth, clip, trim=40mm 10mm 35mm 5mm]{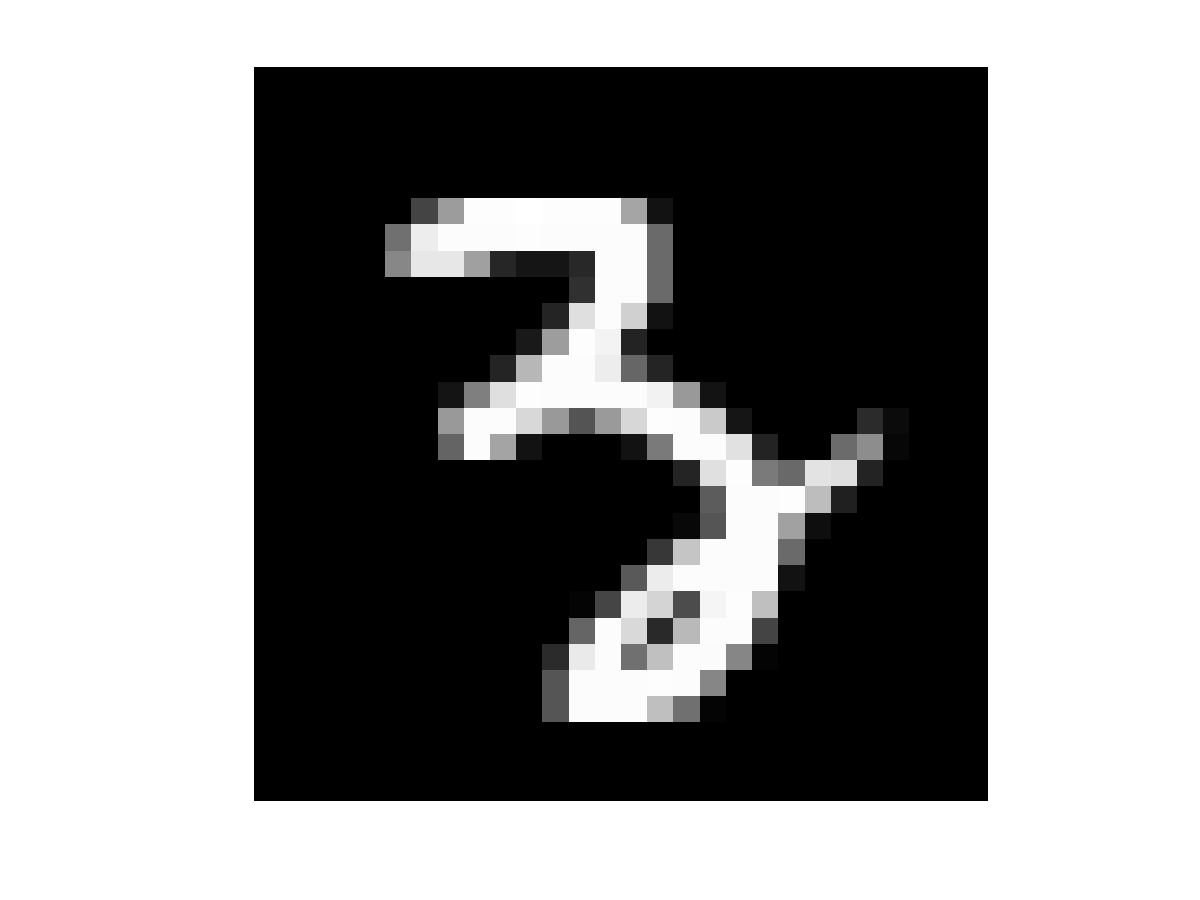}&
		\includegraphics[width=0.22\columnwidth, clip, trim=40mm 10mm 35mm 5mm]{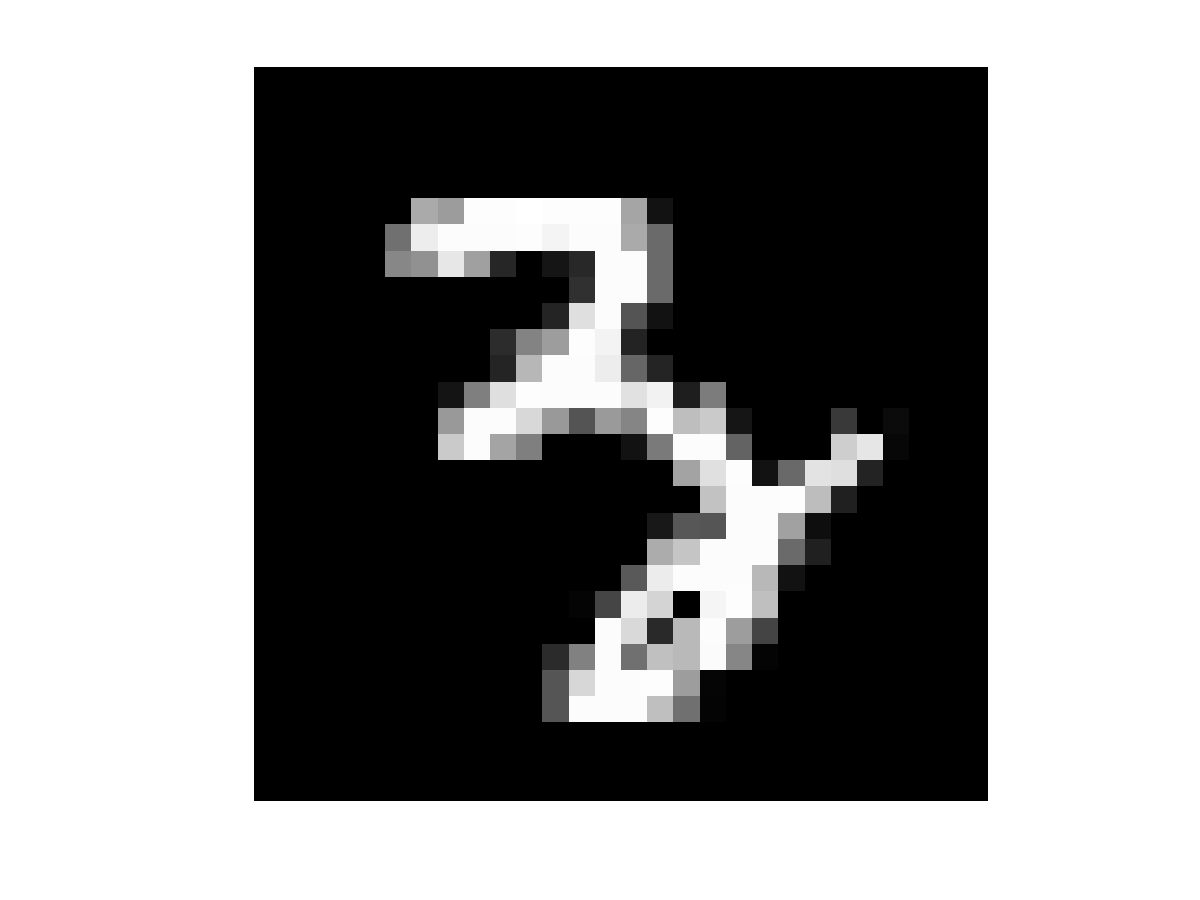}&
		\includegraphics[width=0.22\columnwidth, clip, trim=40mm 10mm 35mm 5mm]{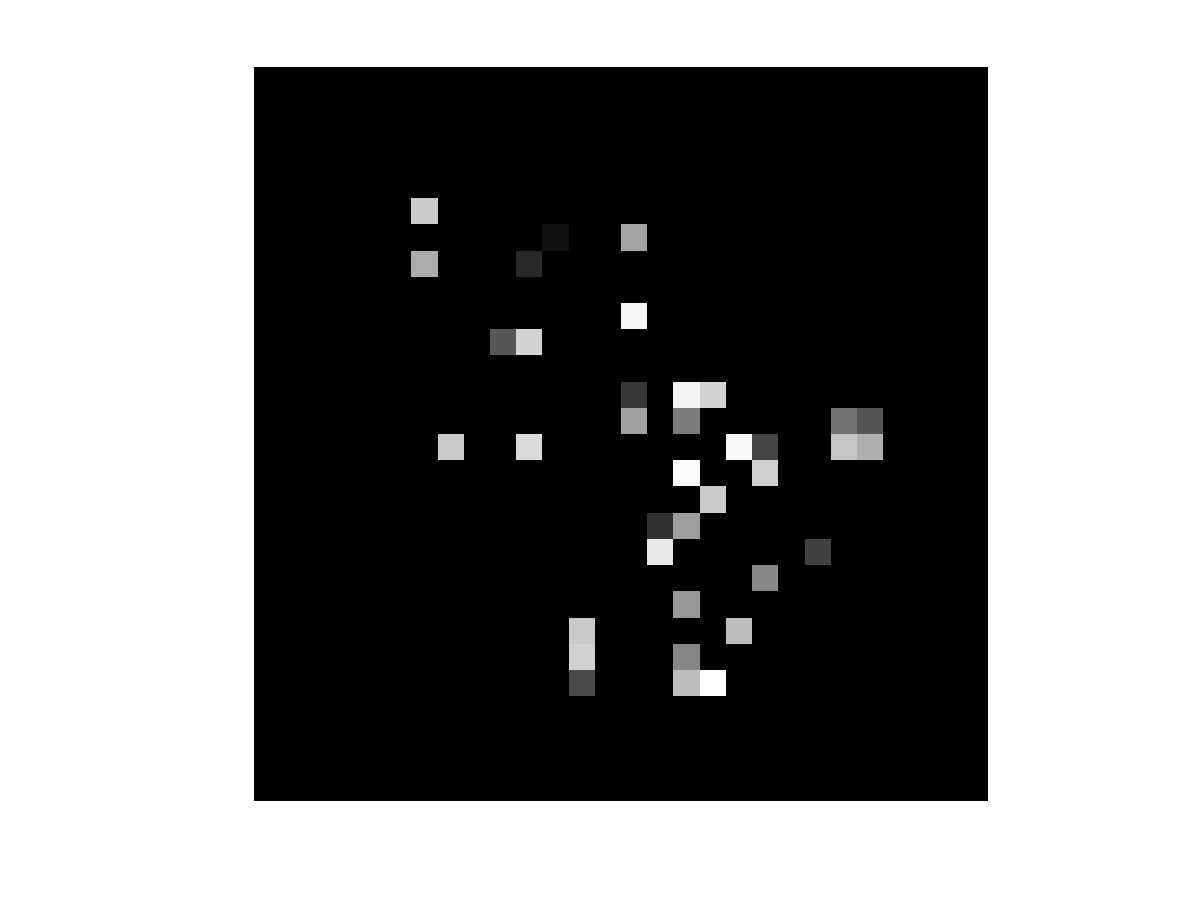}&
		\includegraphics[width=0.22\columnwidth, clip, trim=40mm 10mm 35mm 5mm]{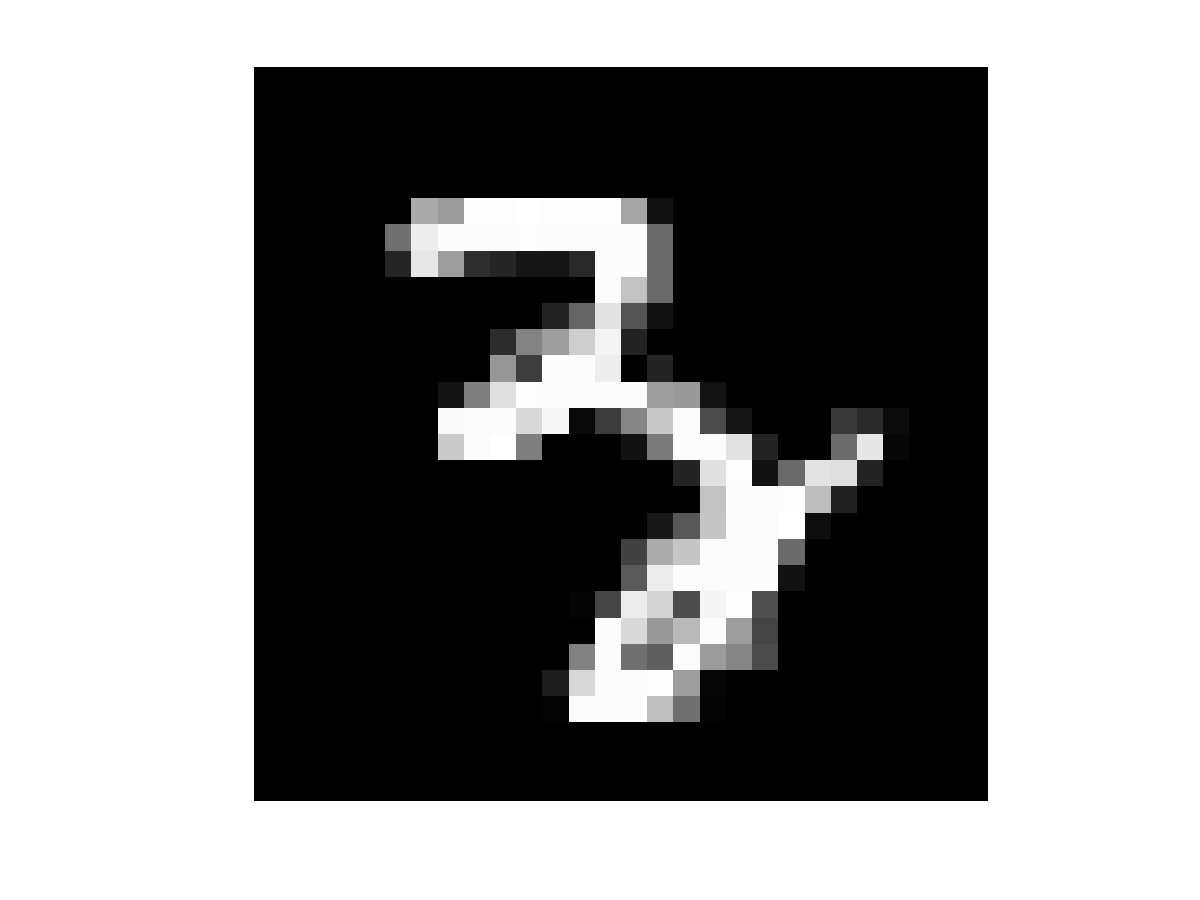}&
		\includegraphics[width=0.22\columnwidth, clip, trim=40mm 10mm 35mm 5mm]{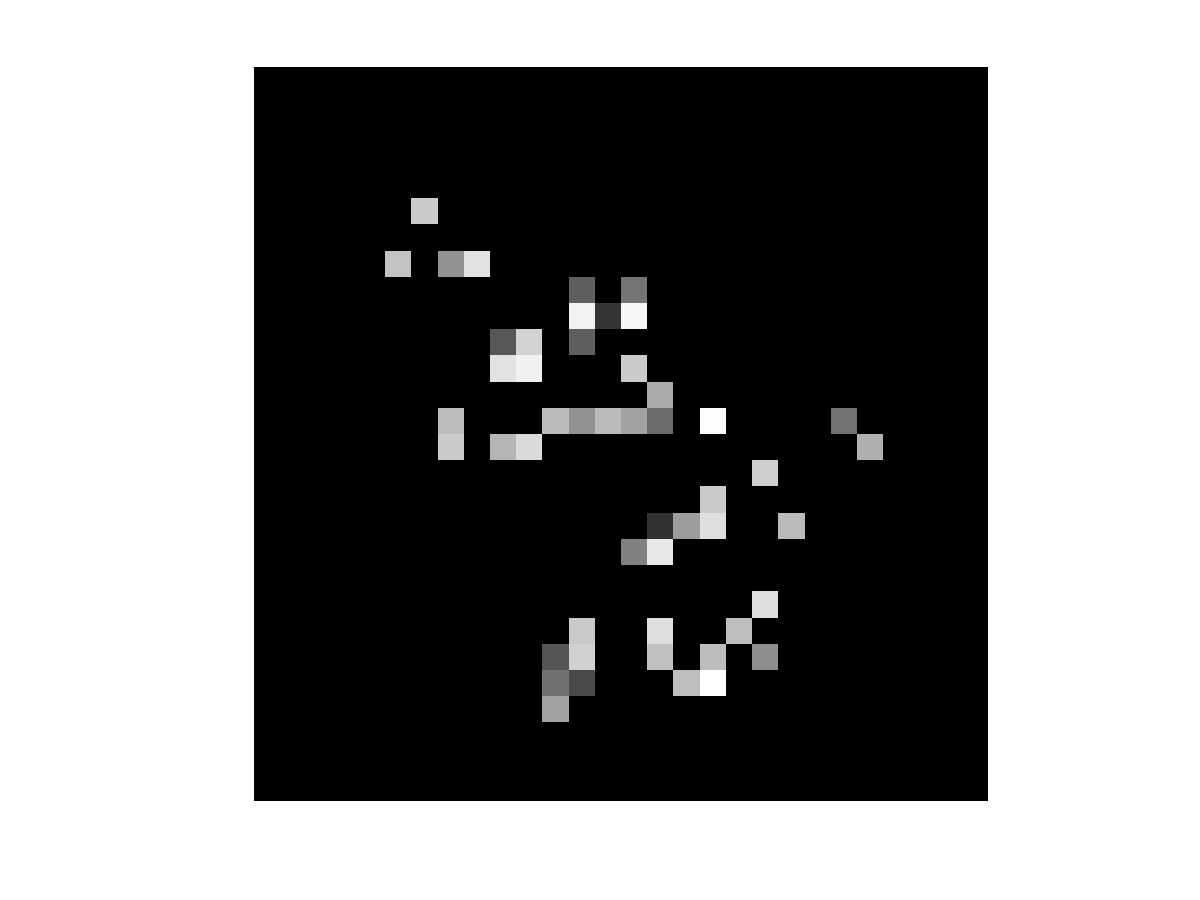}&
		\includegraphics[width=0.22\columnwidth, clip, trim=40mm 10mm 35mm 5mm]{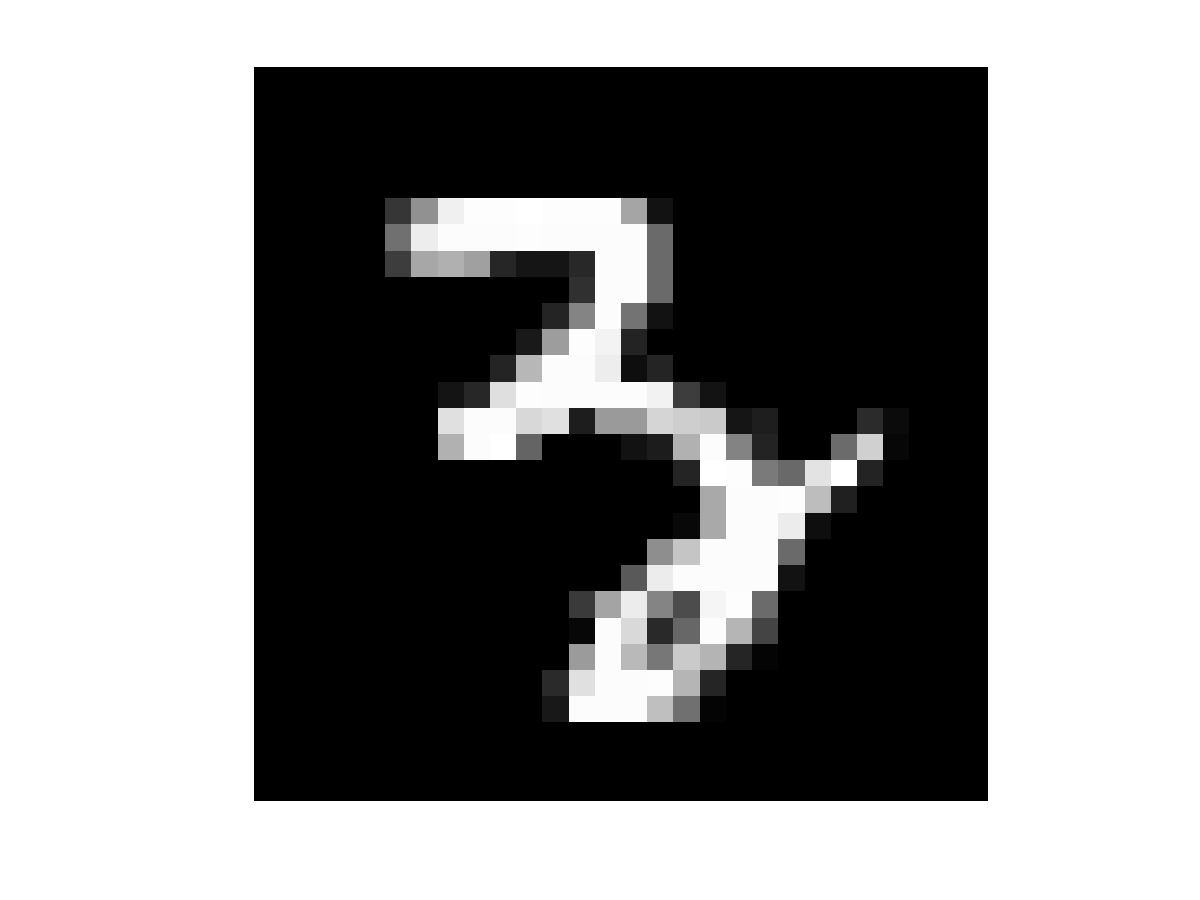}&
		\includegraphics[width=0.22\columnwidth, clip, trim=40mm 10mm 35mm 5mm]{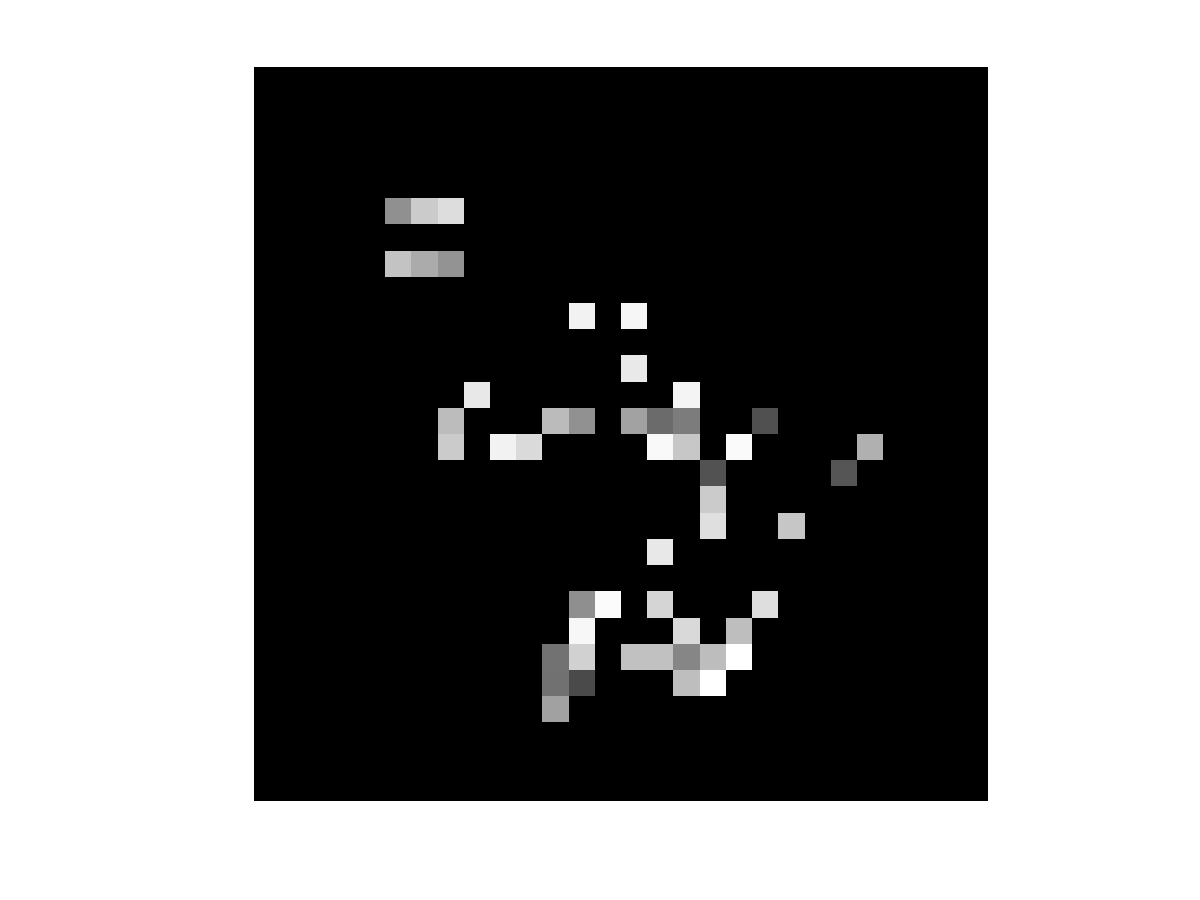}\\
	
\end{tabular}
	\caption{\textbf{Comparison $\sigma$-CornerSearch and $\sigma$-PGD on MNIST}. We show the adversarial examples generated by $\sigma$-CornerSearch with $\kappa=0.8$, $\sigma$-PGD with $\kappa=0.8$ and $\sigma$-PGD with $\kappa=0.6$, together with the respective perturbations rescaled to [0,1]. The sparsity level used for $\sigma$-PGD is $k=50$.}\label{fig:pgd_MNIST}
\end{figure*}
\begin{figure*}[p]
	\centering
	\begin{tabular}{c |c  c |c c |c c }
		\textbf{original}&\multicolumn{2}{c|}{\textbf{$\sigma$-CornerSearch}, $\kappa=0.4$} &\multicolumn{2}{c|}{\textbf{$\sigma$-PGD}, $\kappa=0.4$} &\multicolumn{2}{c}{\textbf{$\sigma$-PGD}, $\kappa=0.25$}\\
		\includegraphics[width=0.22\columnwidth, clip, trim=40mm 10mm 35mm 5mm]{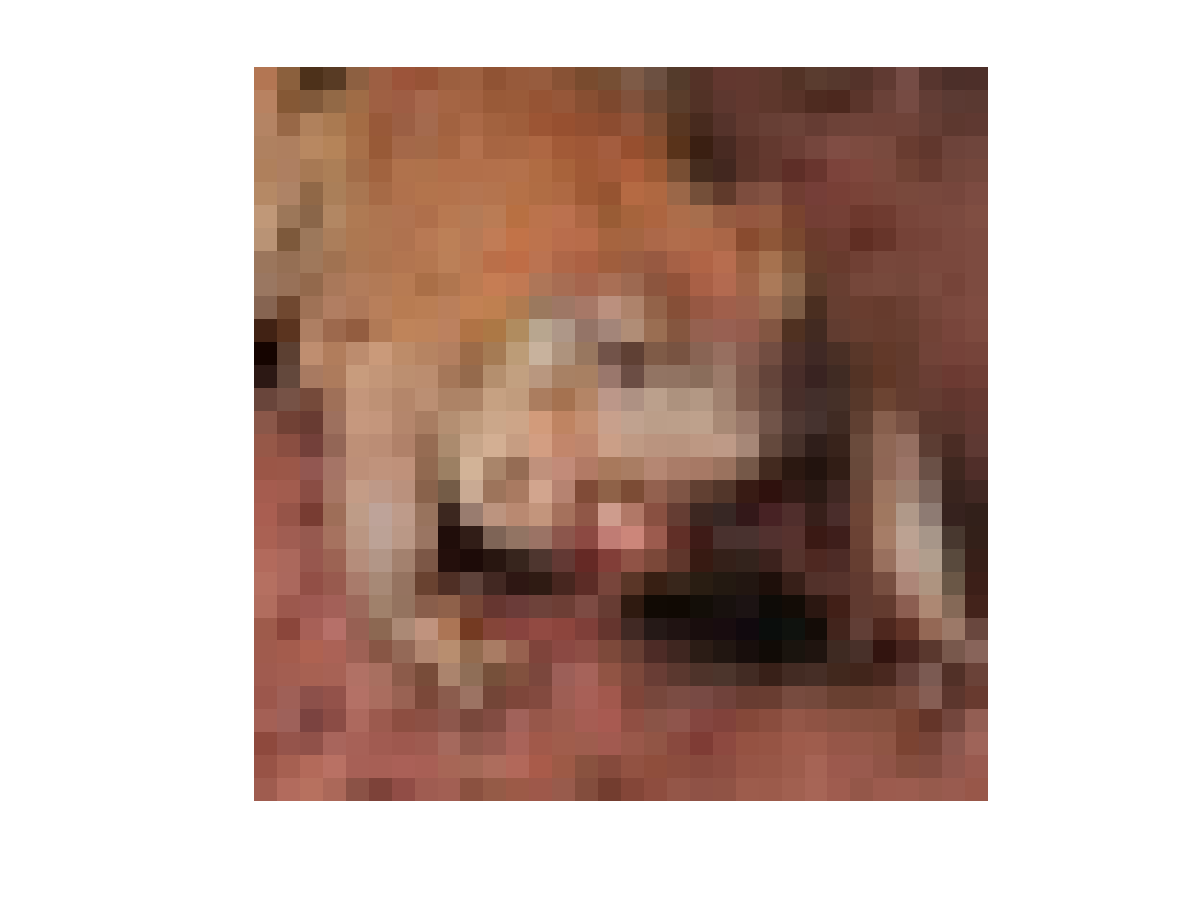}&\includegraphics[width=0.22\columnwidth, clip, trim=40mm 10mm 35mm 5mm]{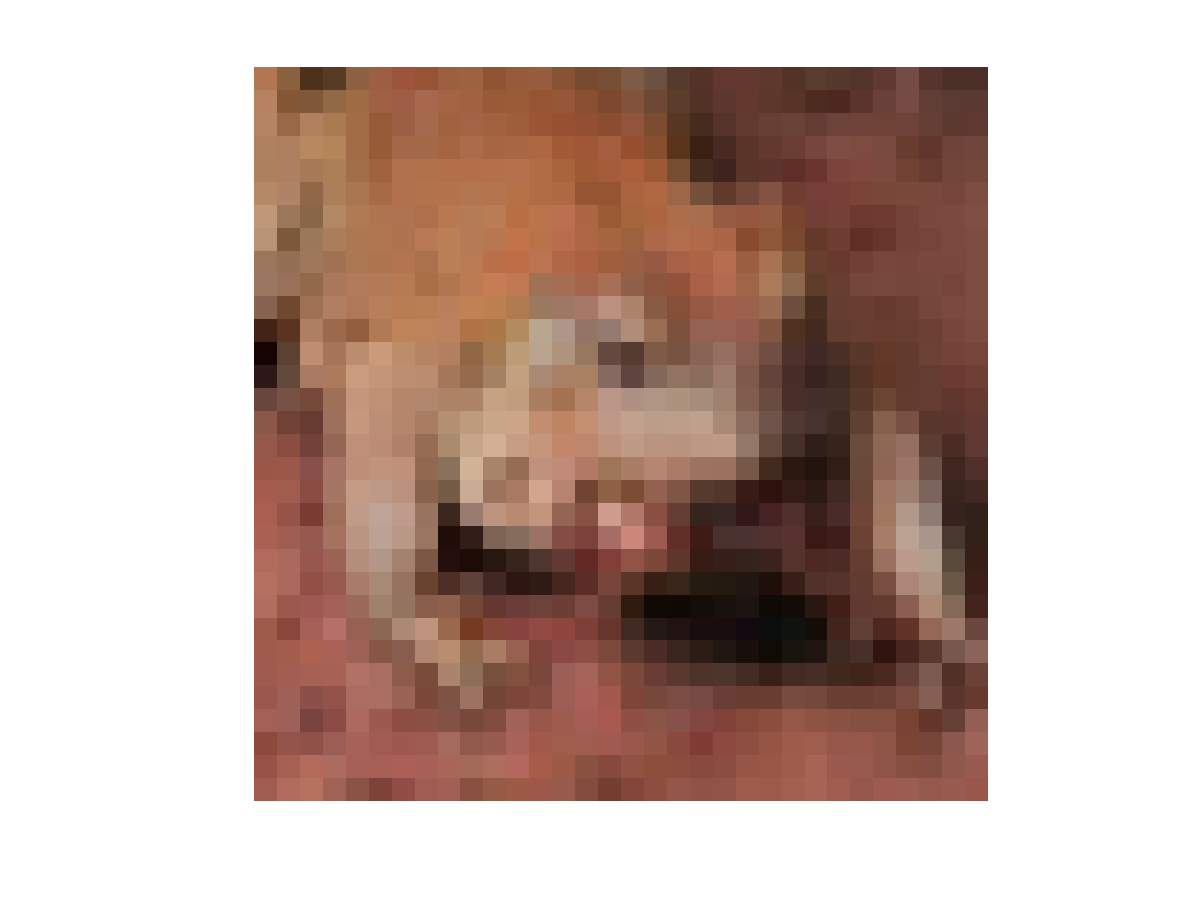}&\includegraphics[width=0.22\columnwidth, clip, trim=40mm 10mm 35mm 5mm]{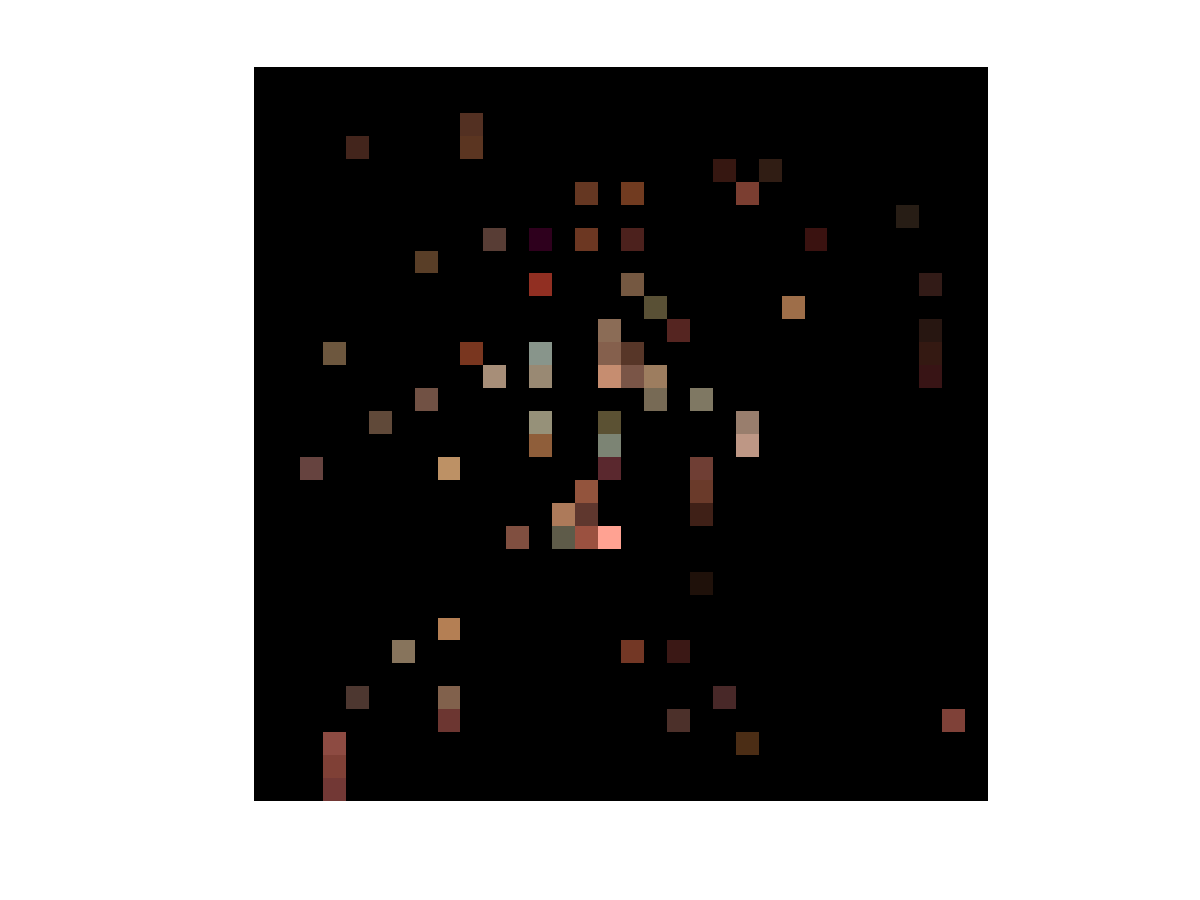}&\includegraphics[width=0.22\columnwidth, clip, trim=40mm 10mm 35mm 5mm]{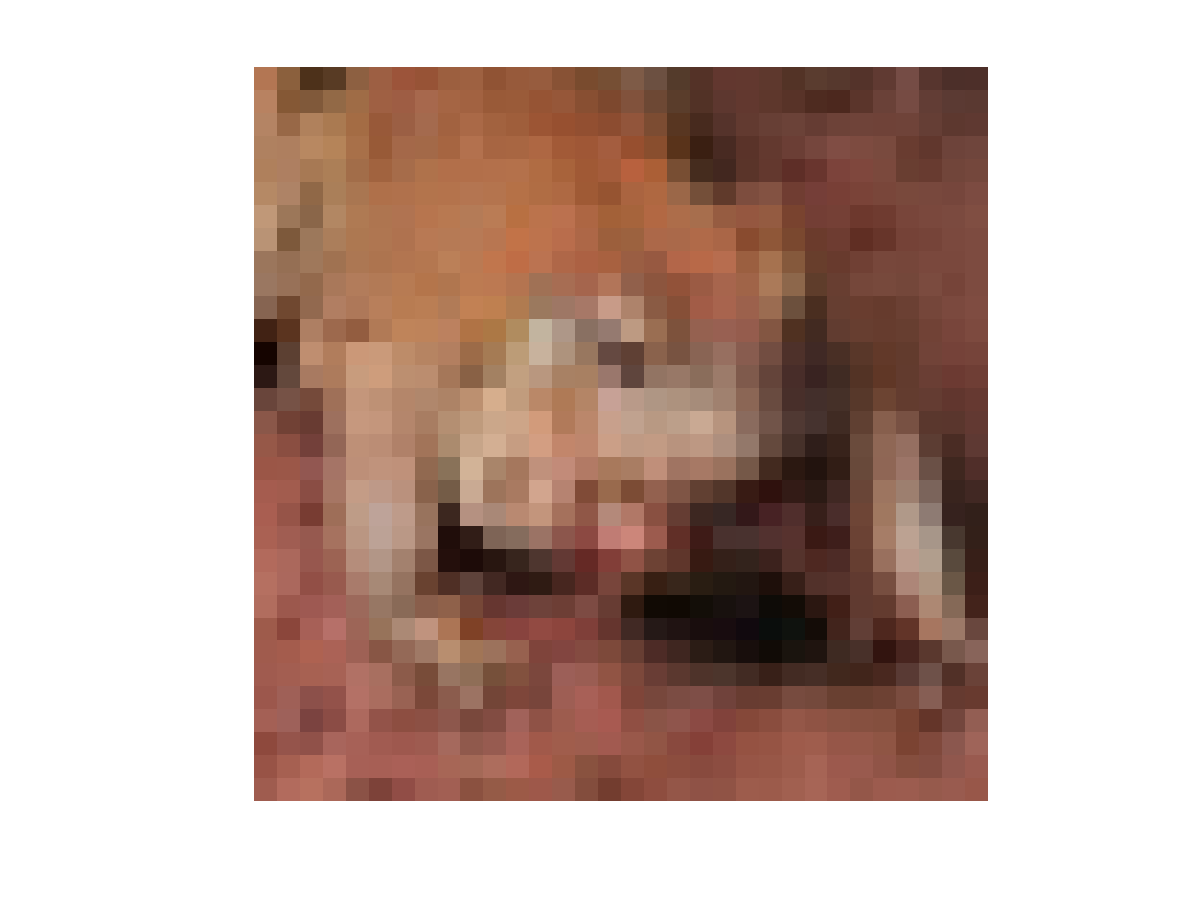}&\includegraphics[width=0.22\columnwidth, clip, trim=40mm 10mm 35mm 5mm]{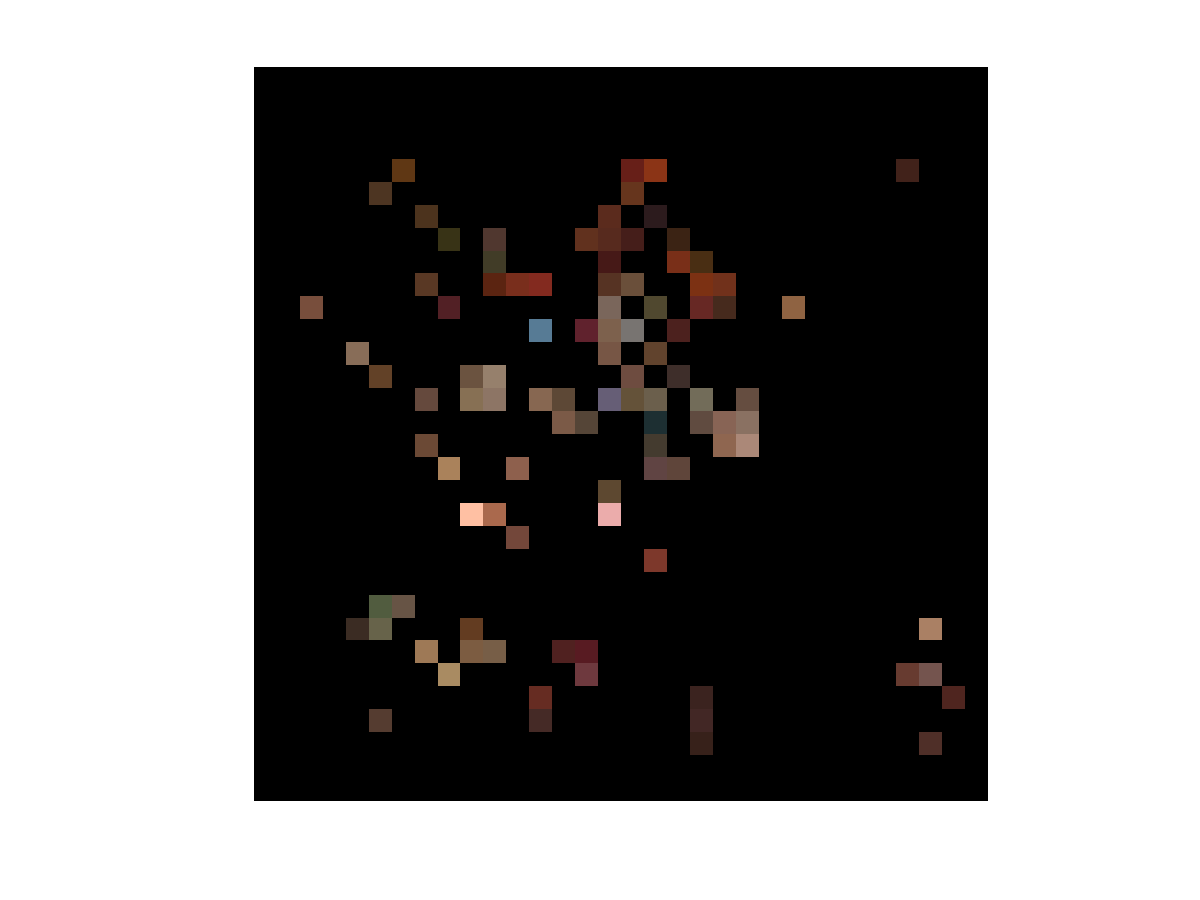}&\includegraphics[width=0.22\columnwidth, clip, trim=40mm 10mm 35mm 5mm]{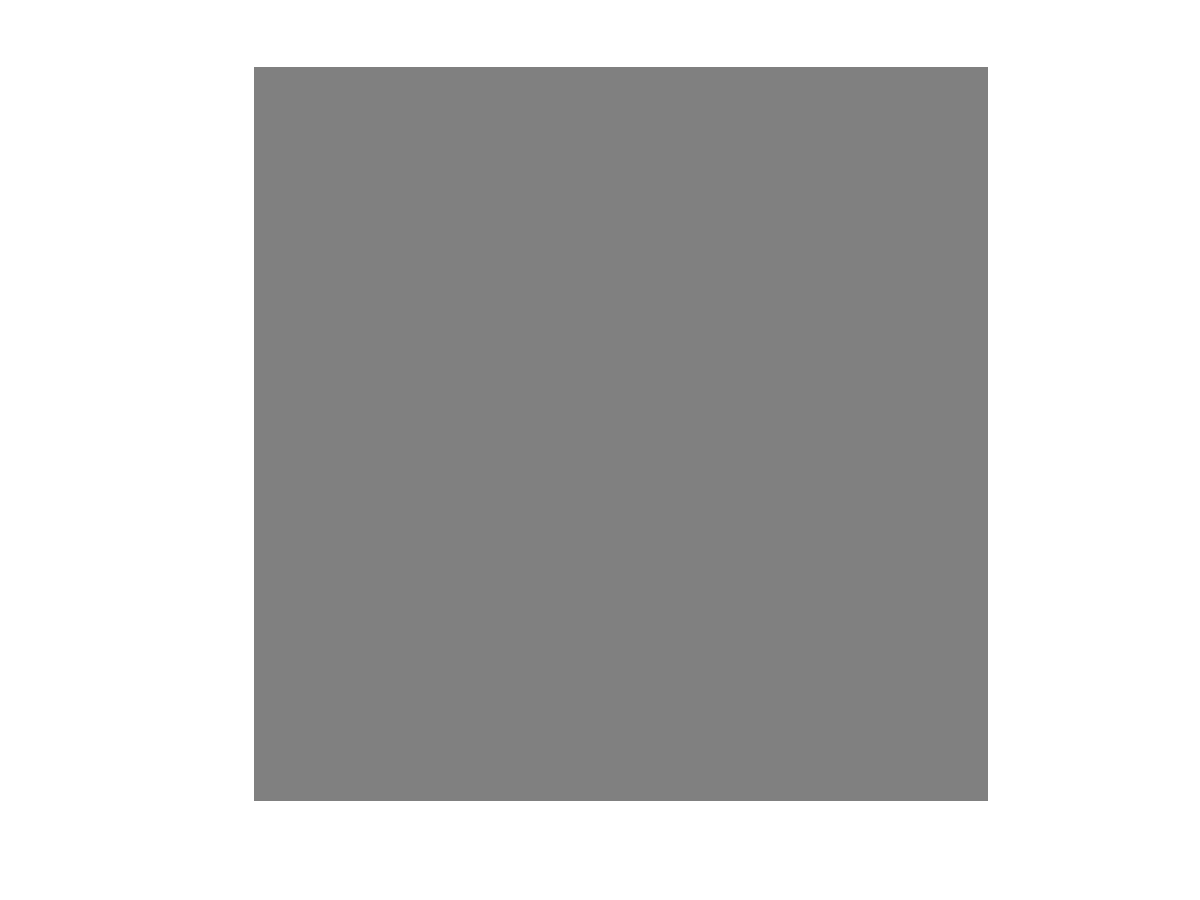}&\includegraphics[width=0.22\columnwidth, clip, trim=40mm 10mm 35mm 5mm]{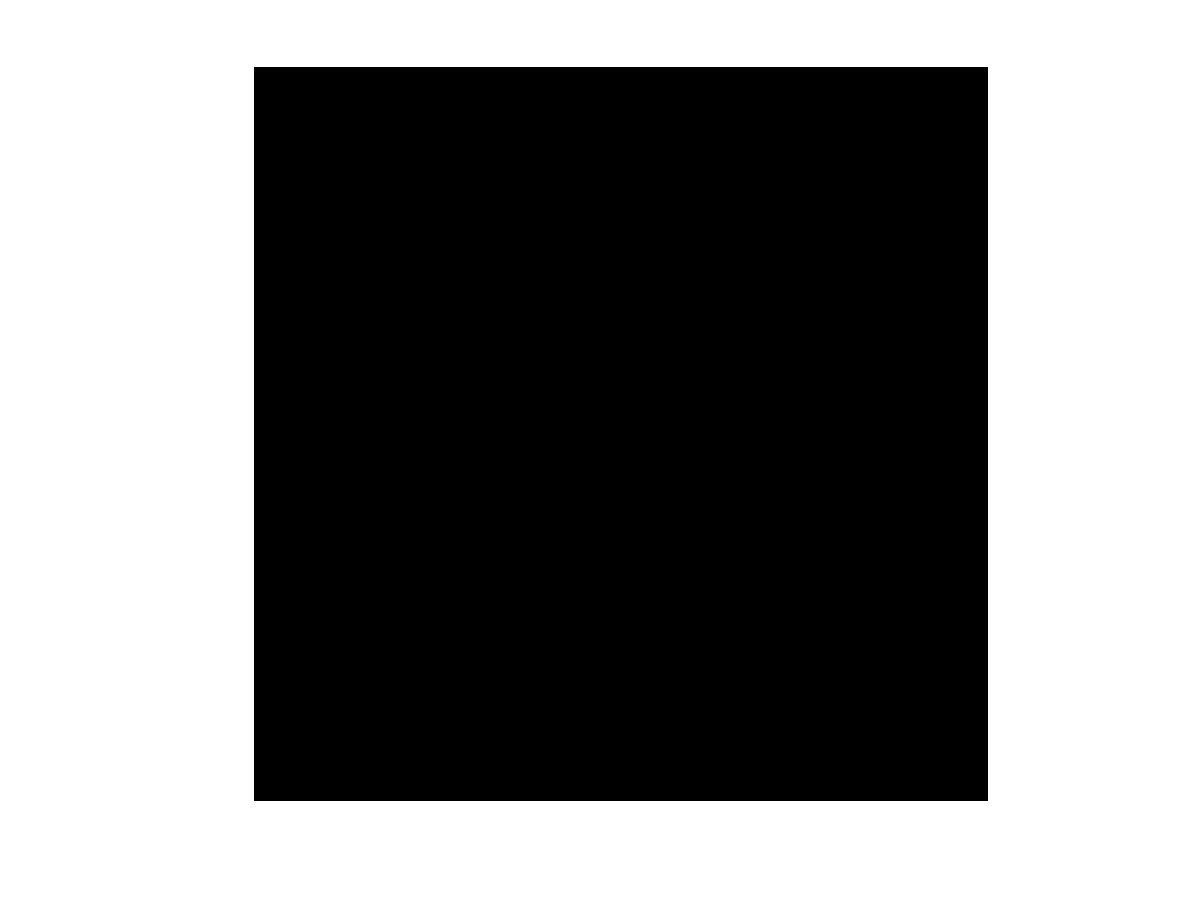}\\
		\includegraphics[width=0.22\columnwidth, clip, trim=40mm 10mm 35mm 5mm]{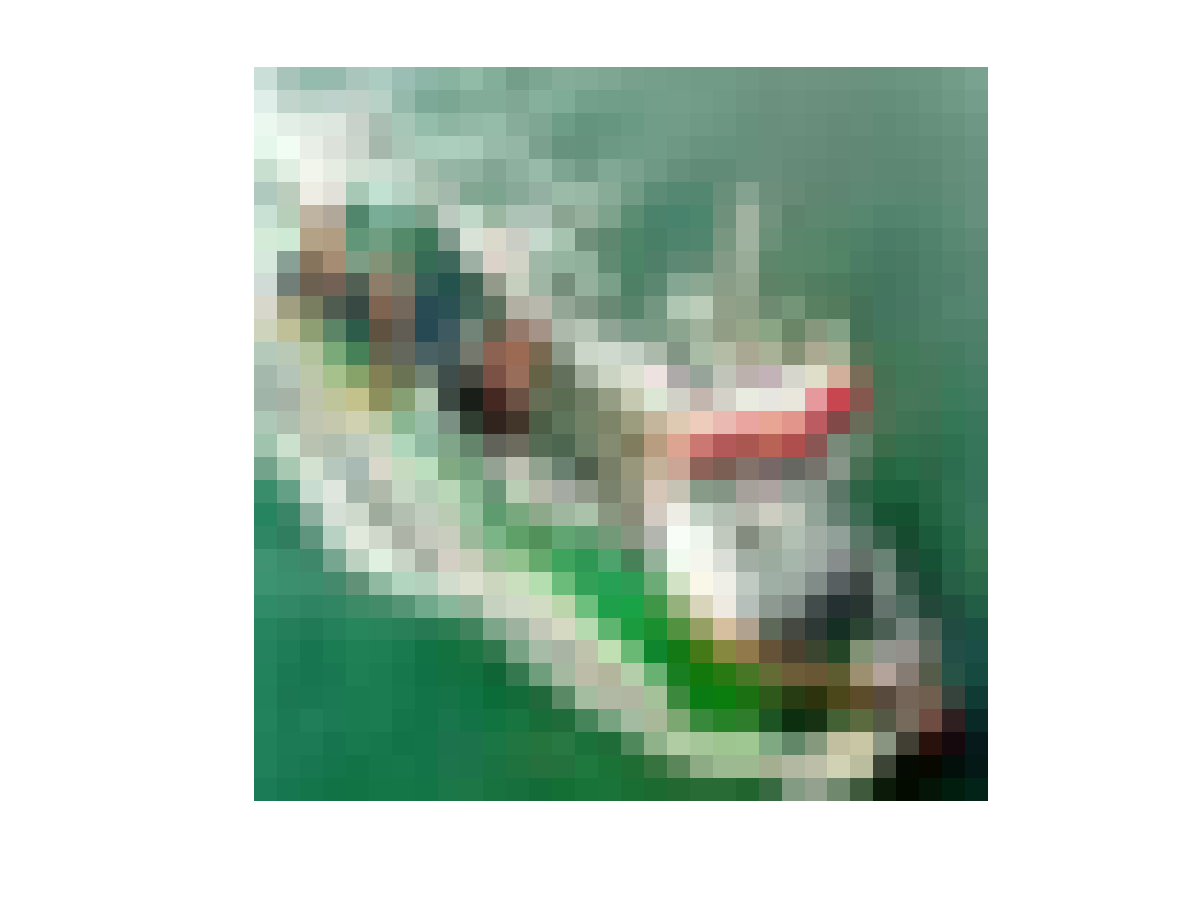}&\includegraphics[width=0.22\columnwidth, clip, trim=40mm 10mm 35mm 5mm]{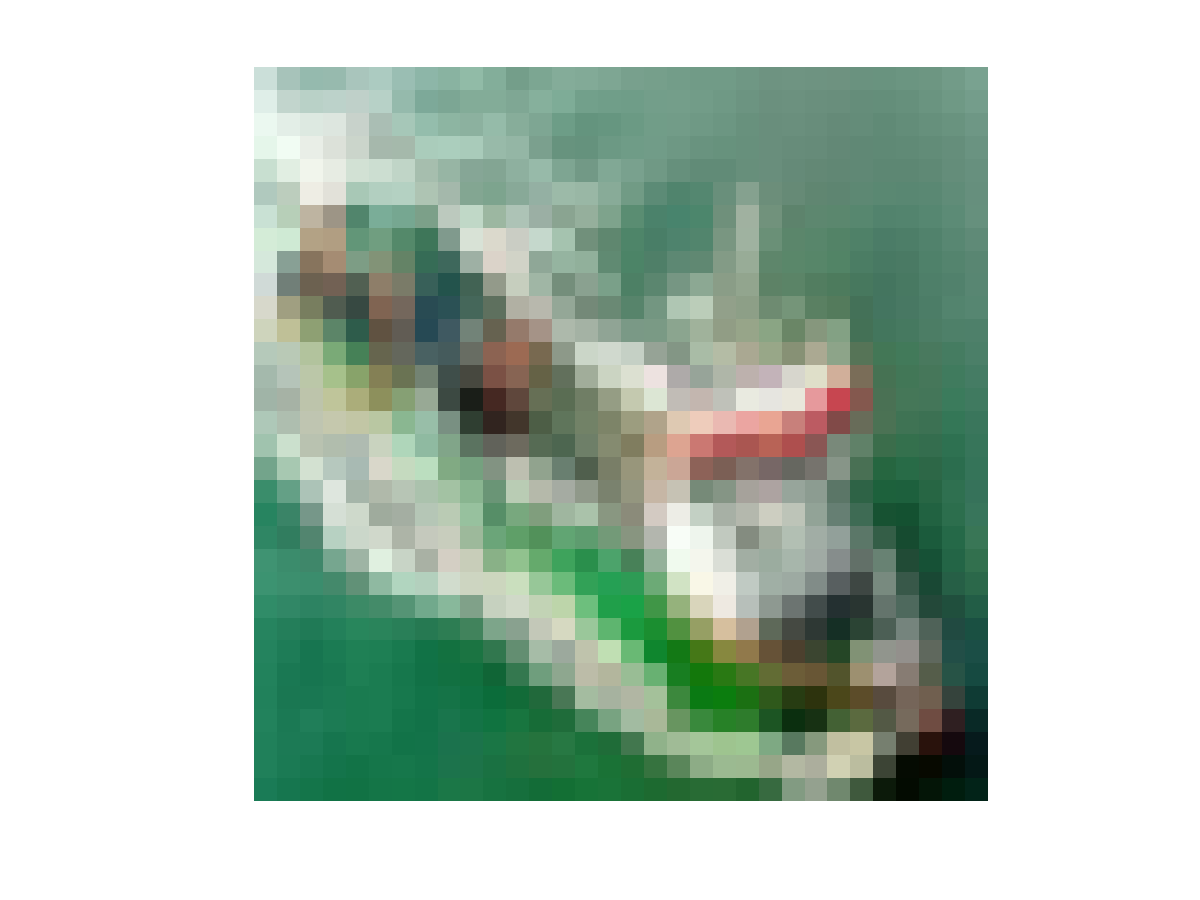}&\includegraphics[width=0.22\columnwidth, clip, trim=40mm 10mm 35mm 5mm]{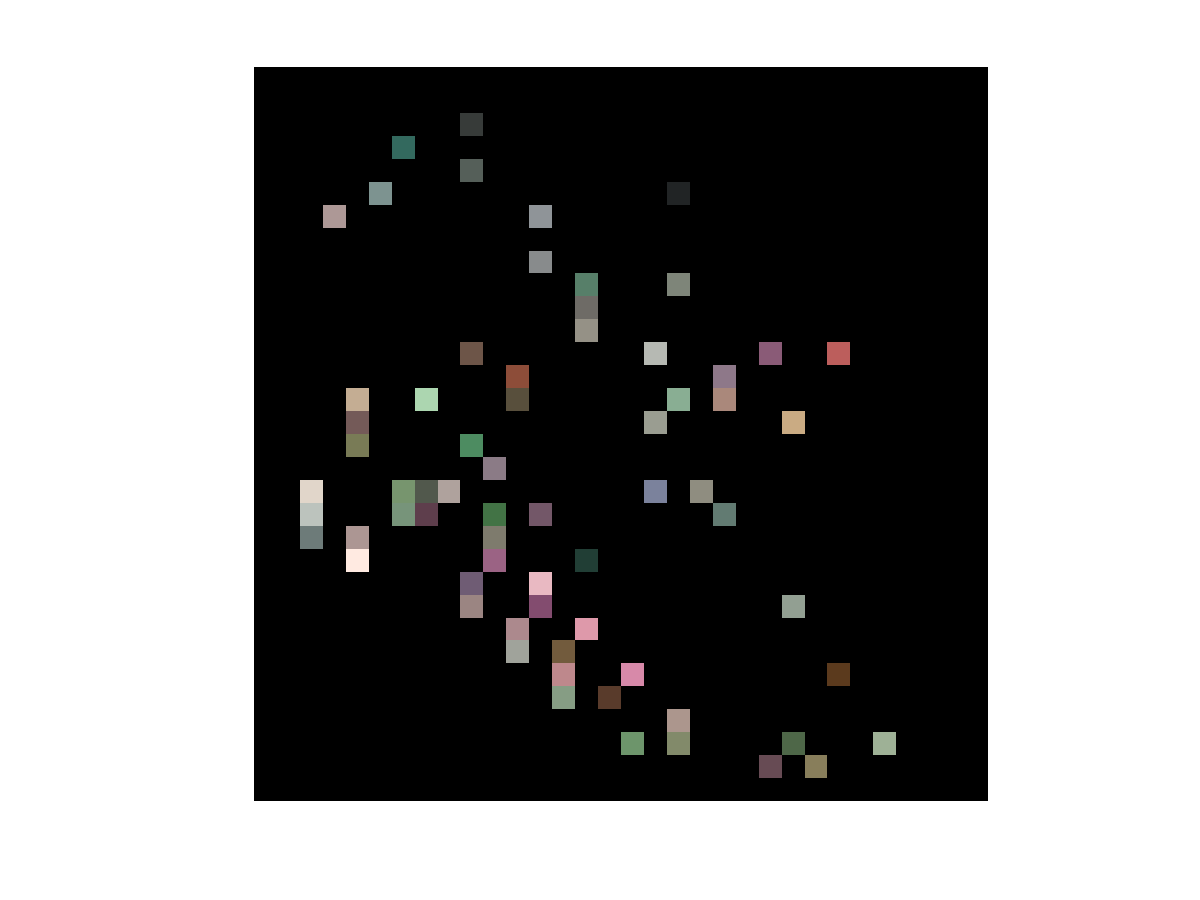}&\includegraphics[width=0.22\columnwidth, clip, trim=40mm 10mm 35mm 5mm]{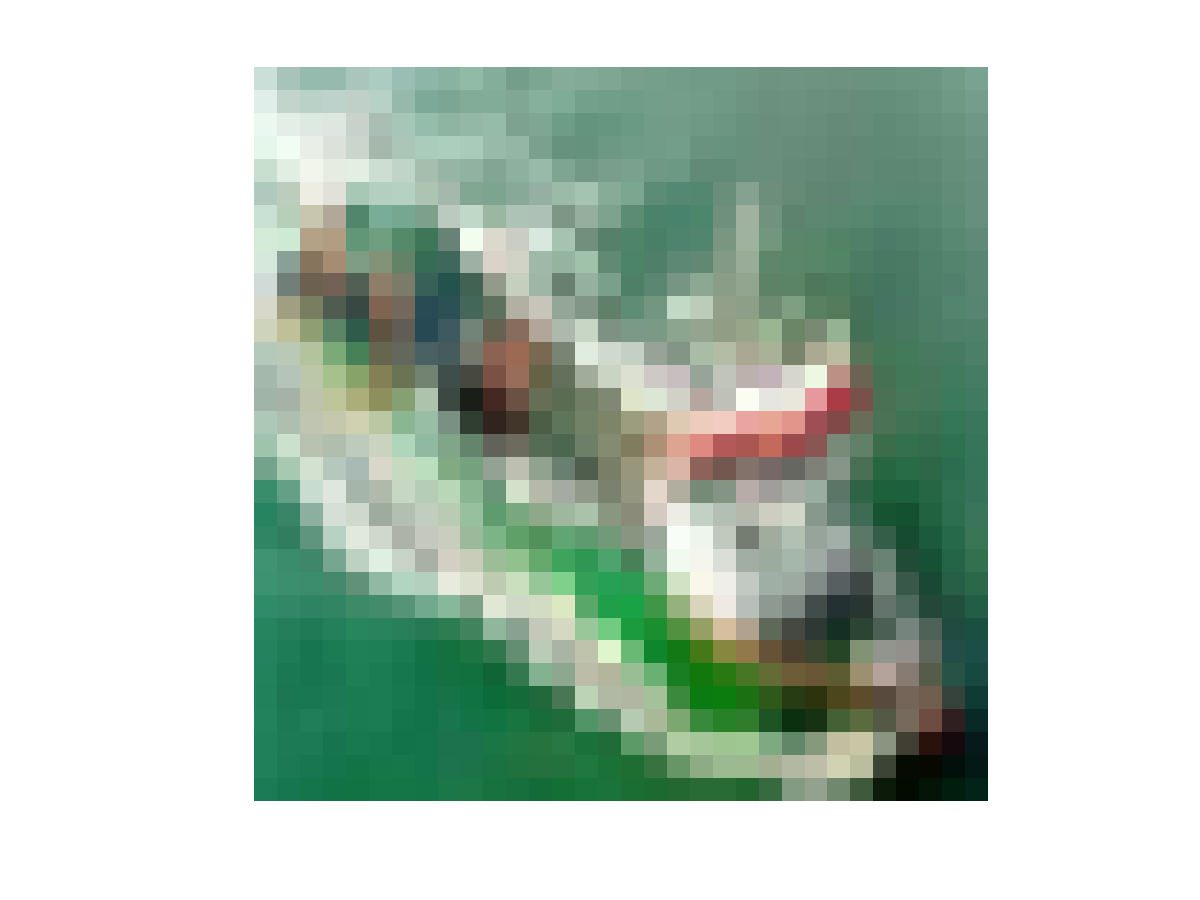}&\includegraphics[width=0.22\columnwidth, clip, trim=40mm 10mm 35mm 5mm]{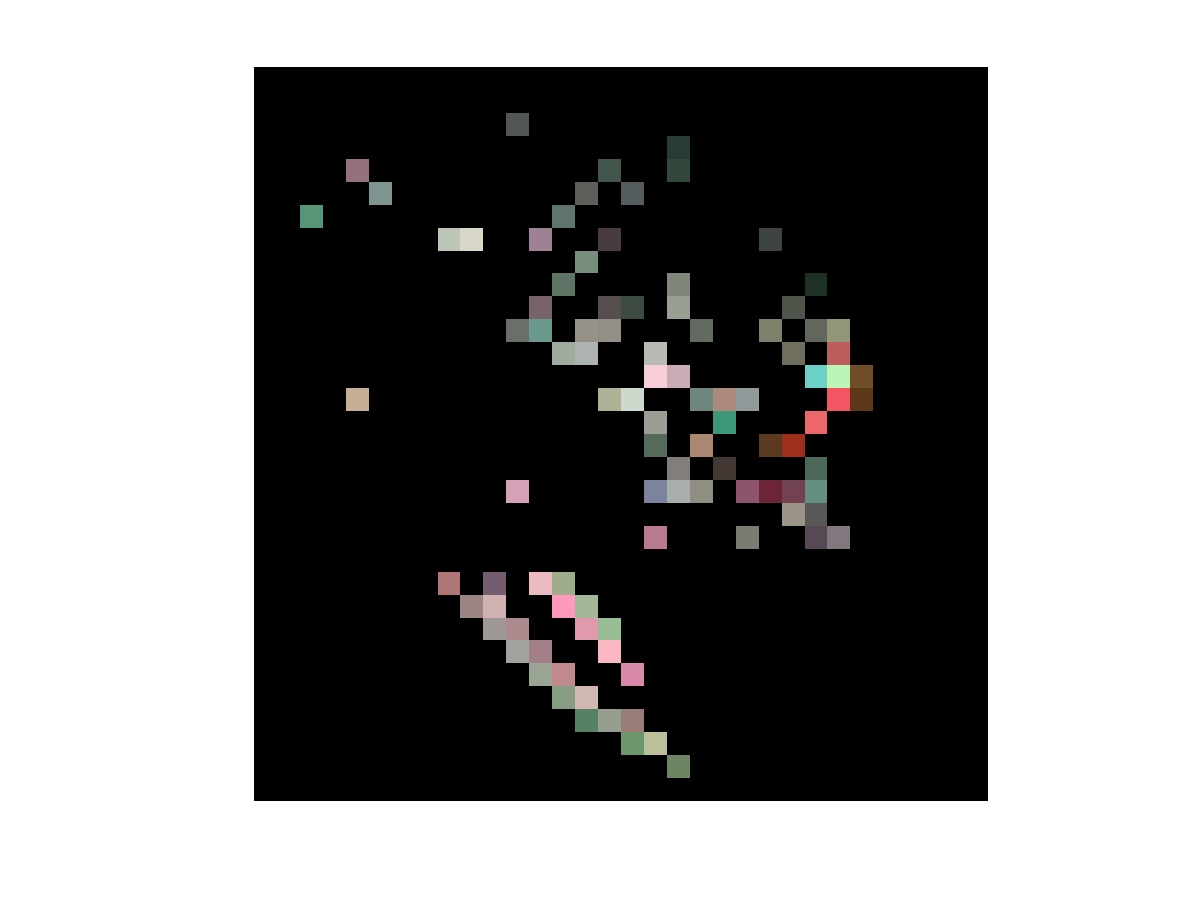}&\includegraphics[width=0.22\columnwidth, clip, trim=40mm 10mm 35mm 5mm]{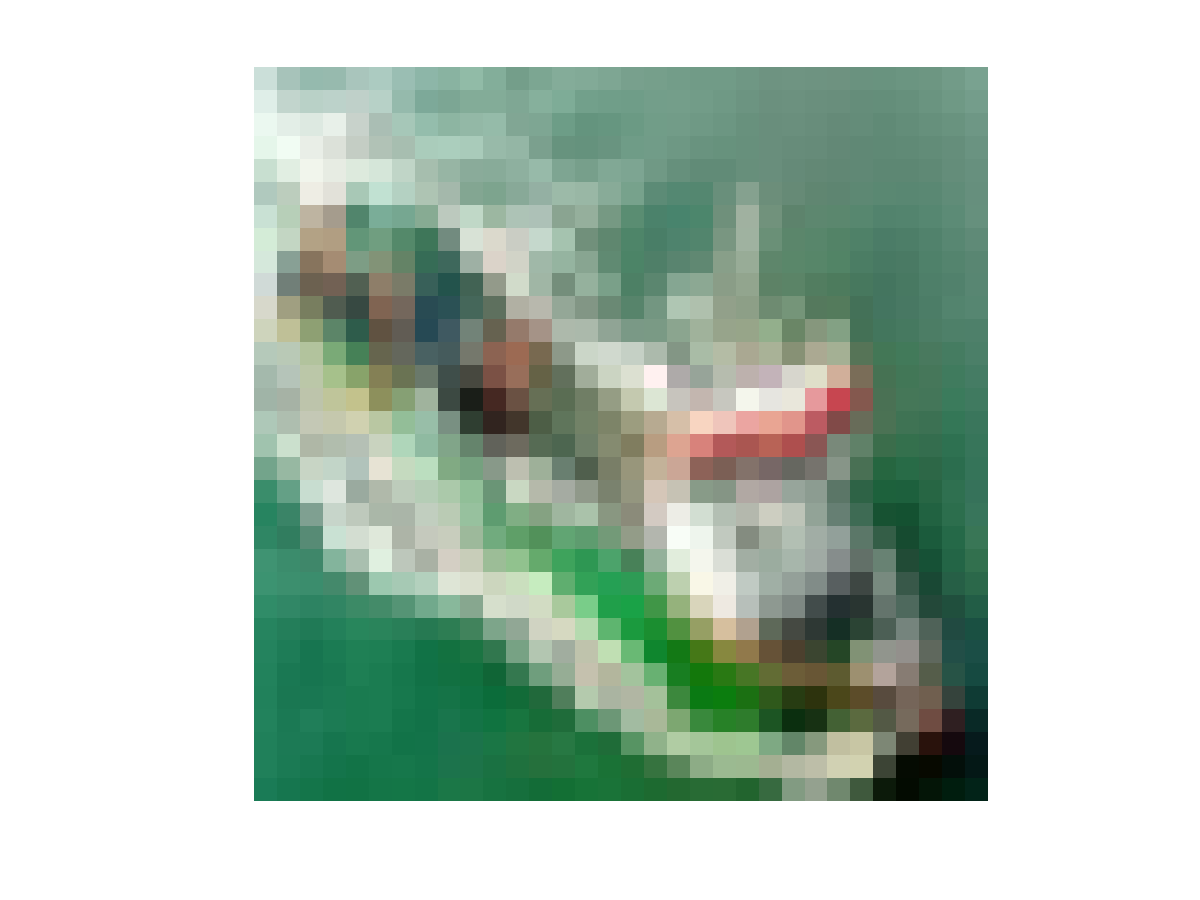}&\includegraphics[width=0.22\columnwidth, clip, trim=40mm 10mm 35mm 5mm]{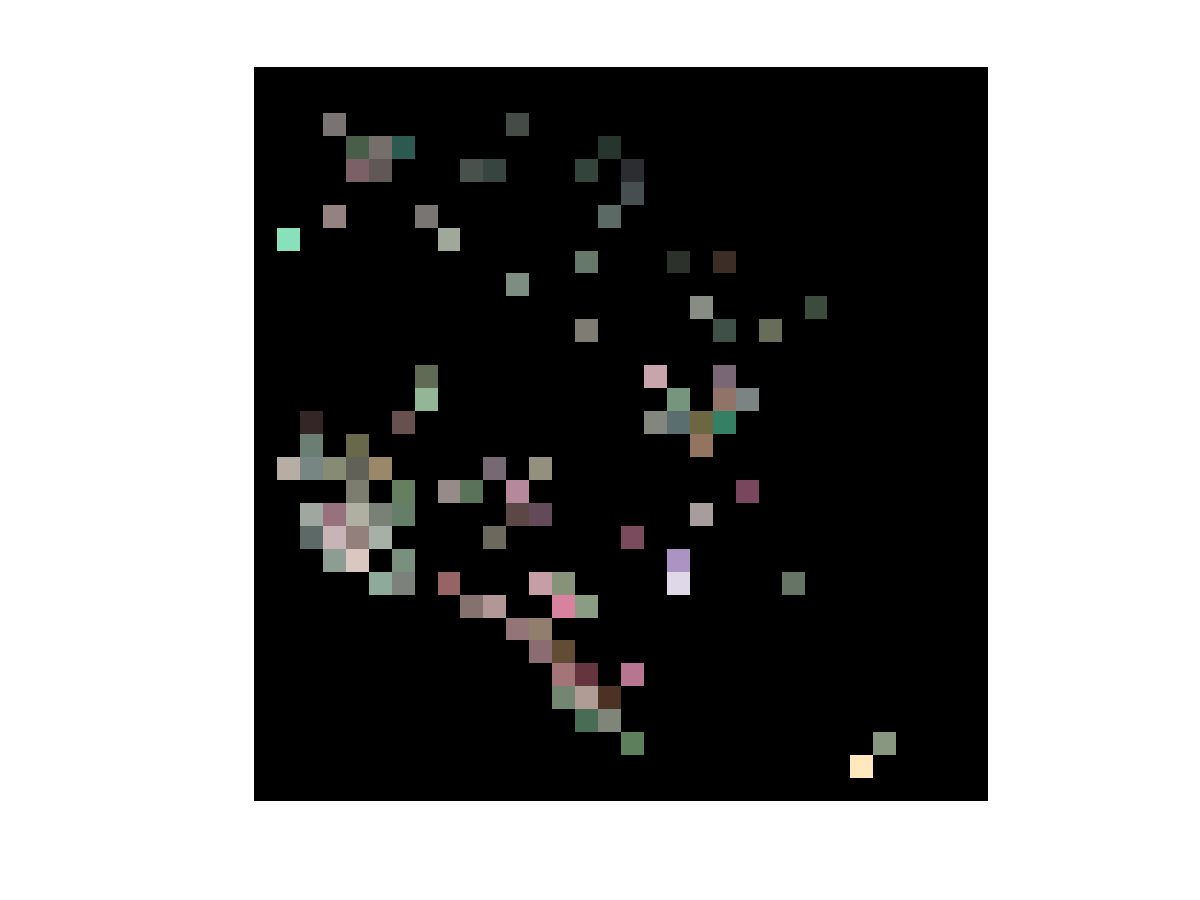}\\\includegraphics[width=0.22\columnwidth, clip, trim=40mm 10mm 35mm 5mm]{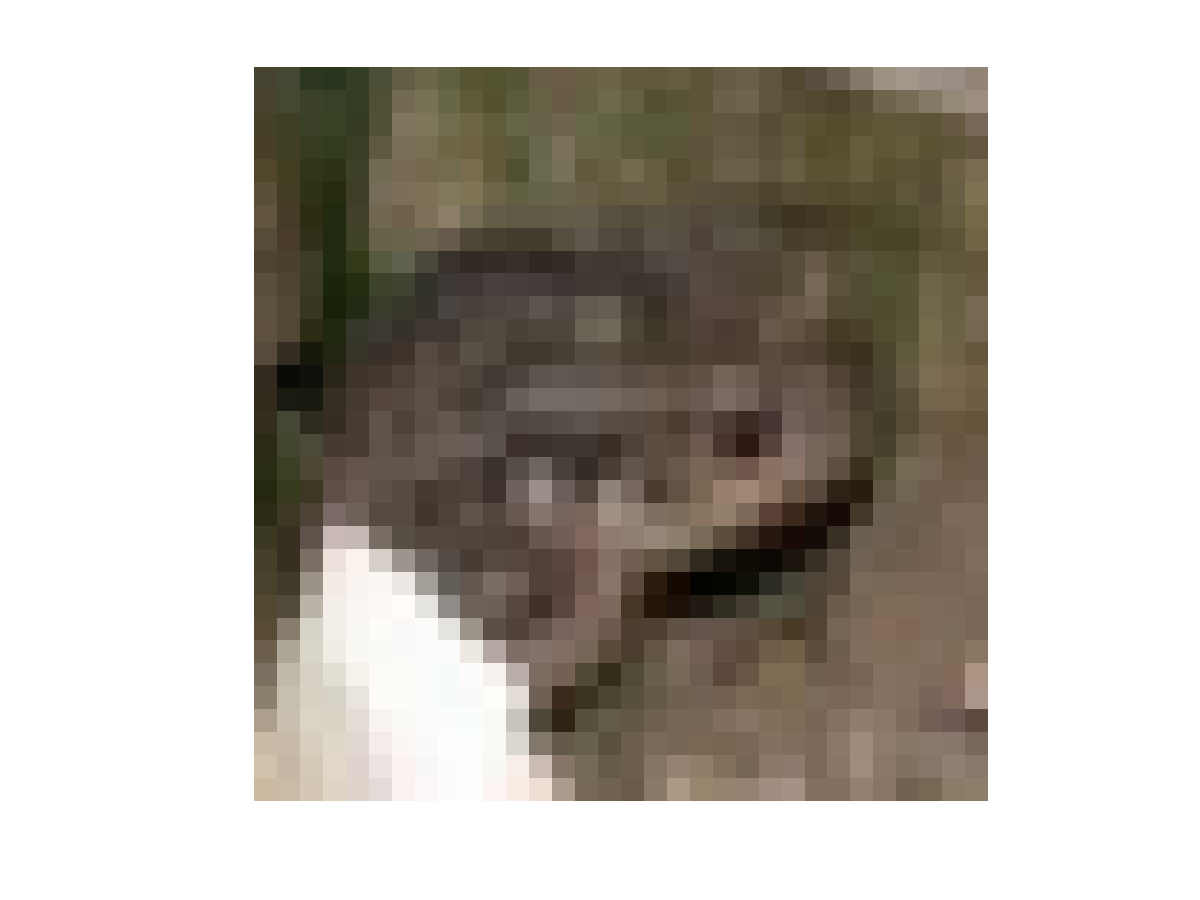}&\includegraphics[width=0.22\columnwidth, clip, trim=40mm 10mm 35mm 5mm]{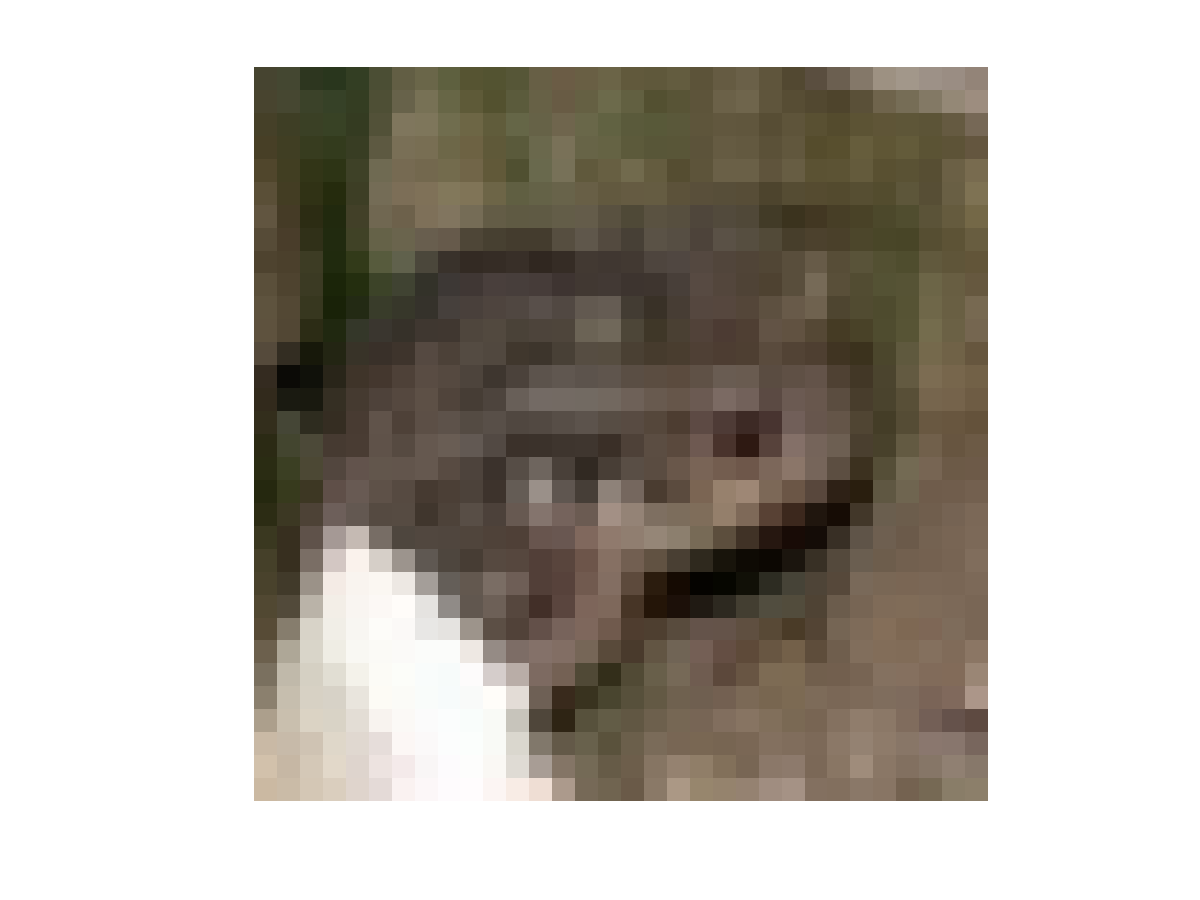}&\includegraphics[width=0.22\columnwidth, clip, trim=40mm 10mm 35mm 5mm]{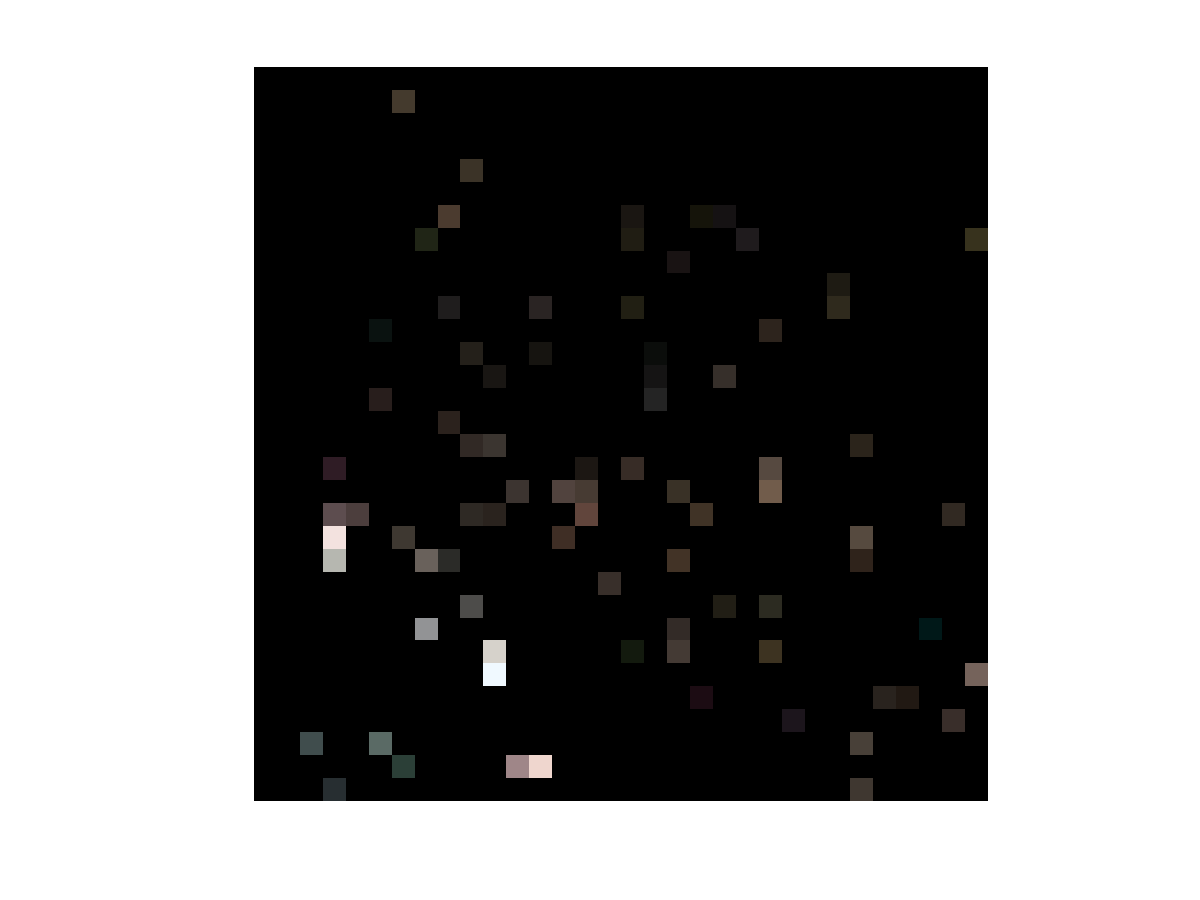}&\includegraphics[width=0.22\columnwidth, clip, trim=40mm 10mm 35mm 5mm]{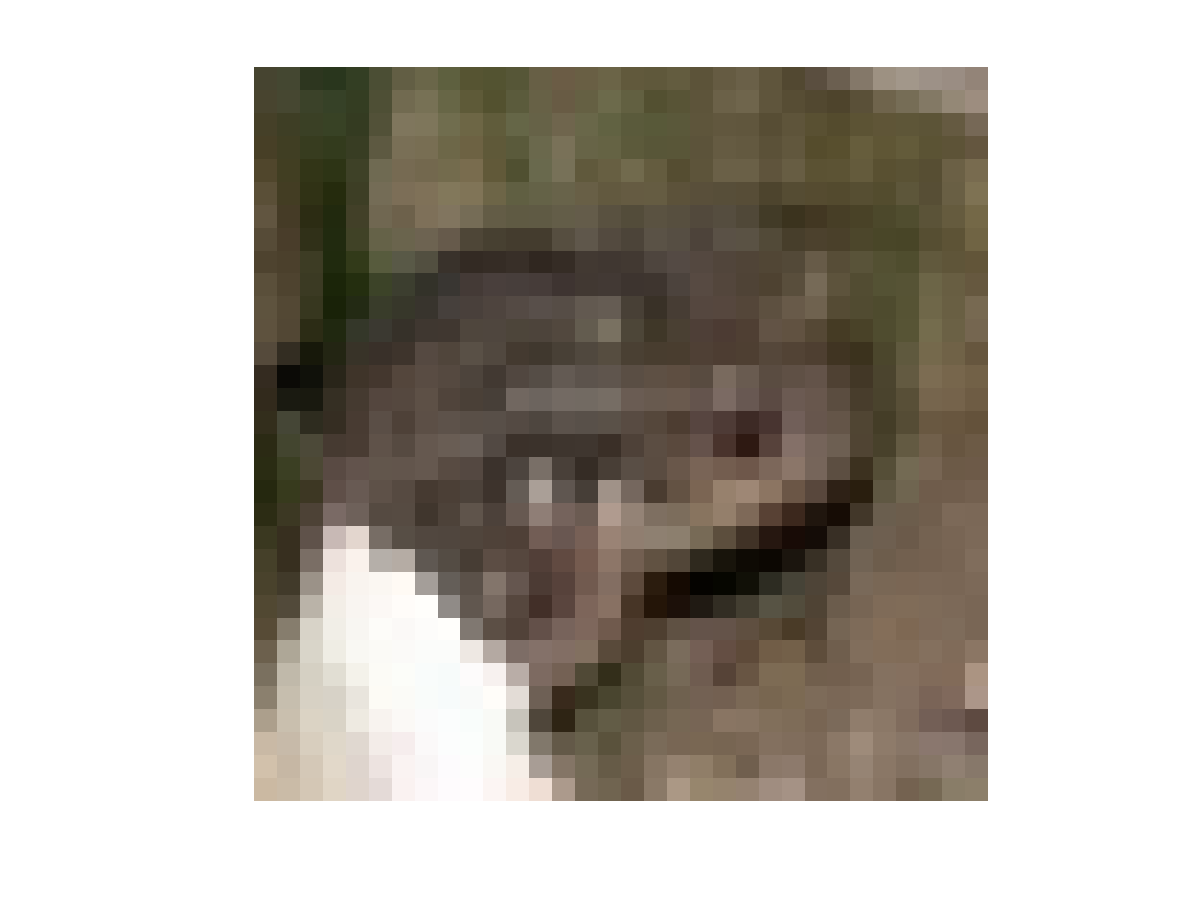}&\includegraphics[width=0.22\columnwidth, clip, trim=40mm 10mm 35mm 5mm]{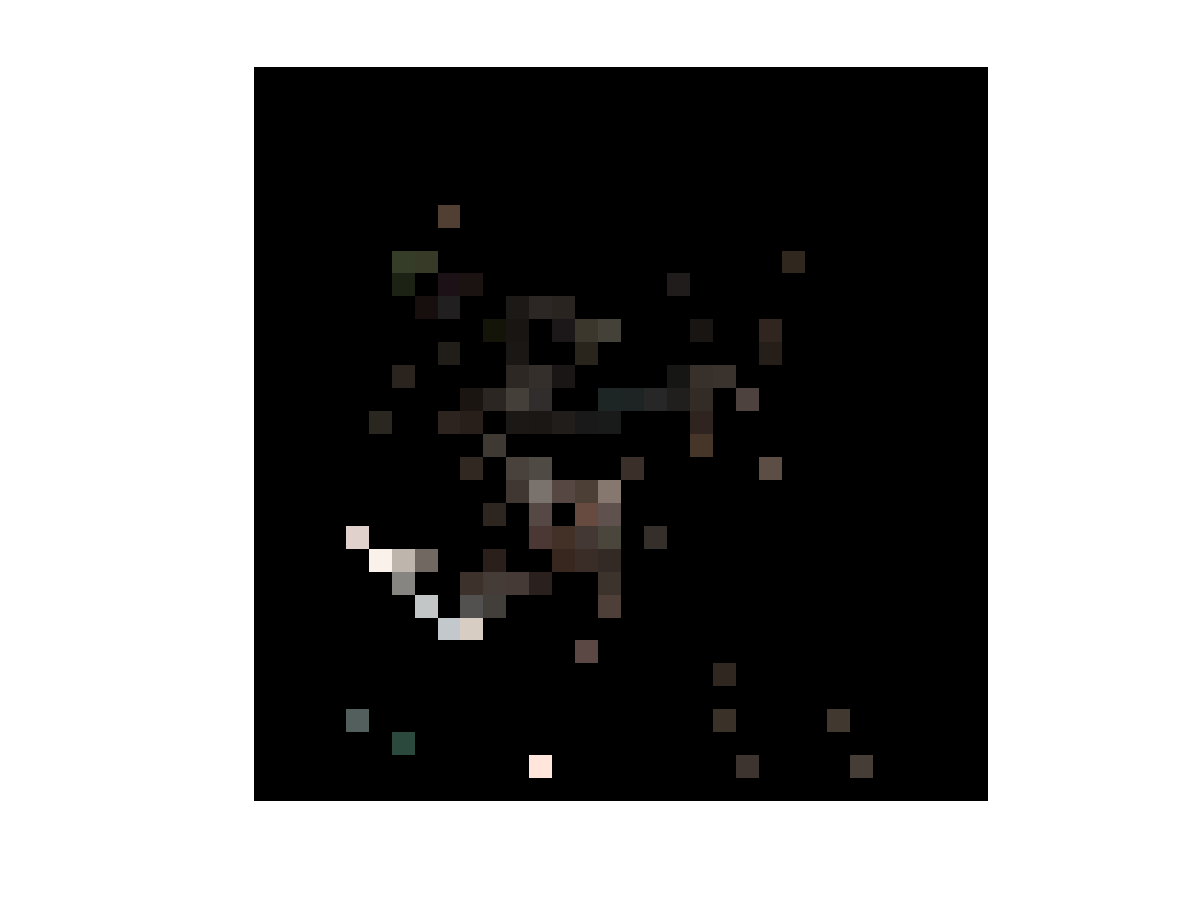}&\includegraphics[width=0.22\columnwidth, clip, trim=40mm 10mm 35mm 5mm]{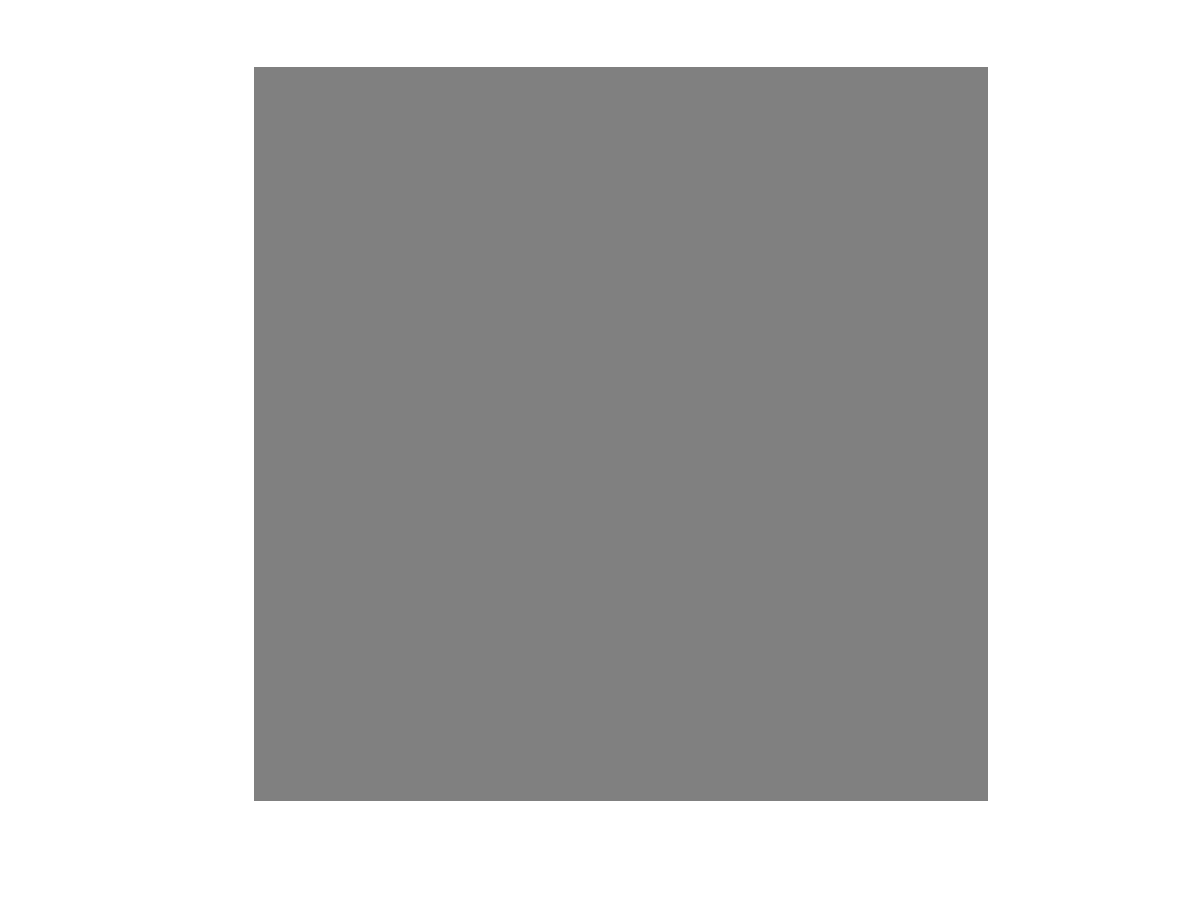}&\includegraphics[width=0.22\columnwidth, clip, trim=40mm 10mm 35mm 5mm]{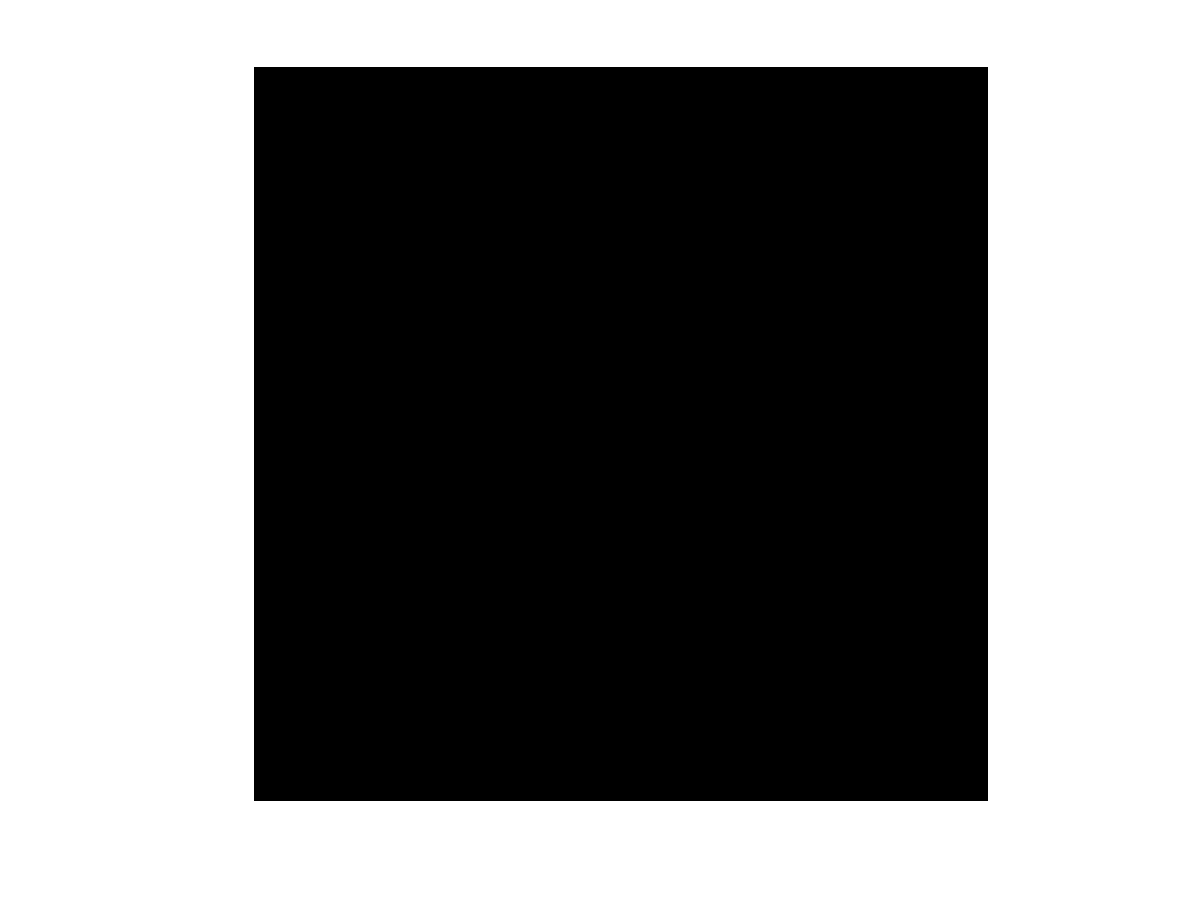}\\
		\includegraphics[width=0.22\columnwidth, clip, trim=40mm 10mm 35mm 5mm]{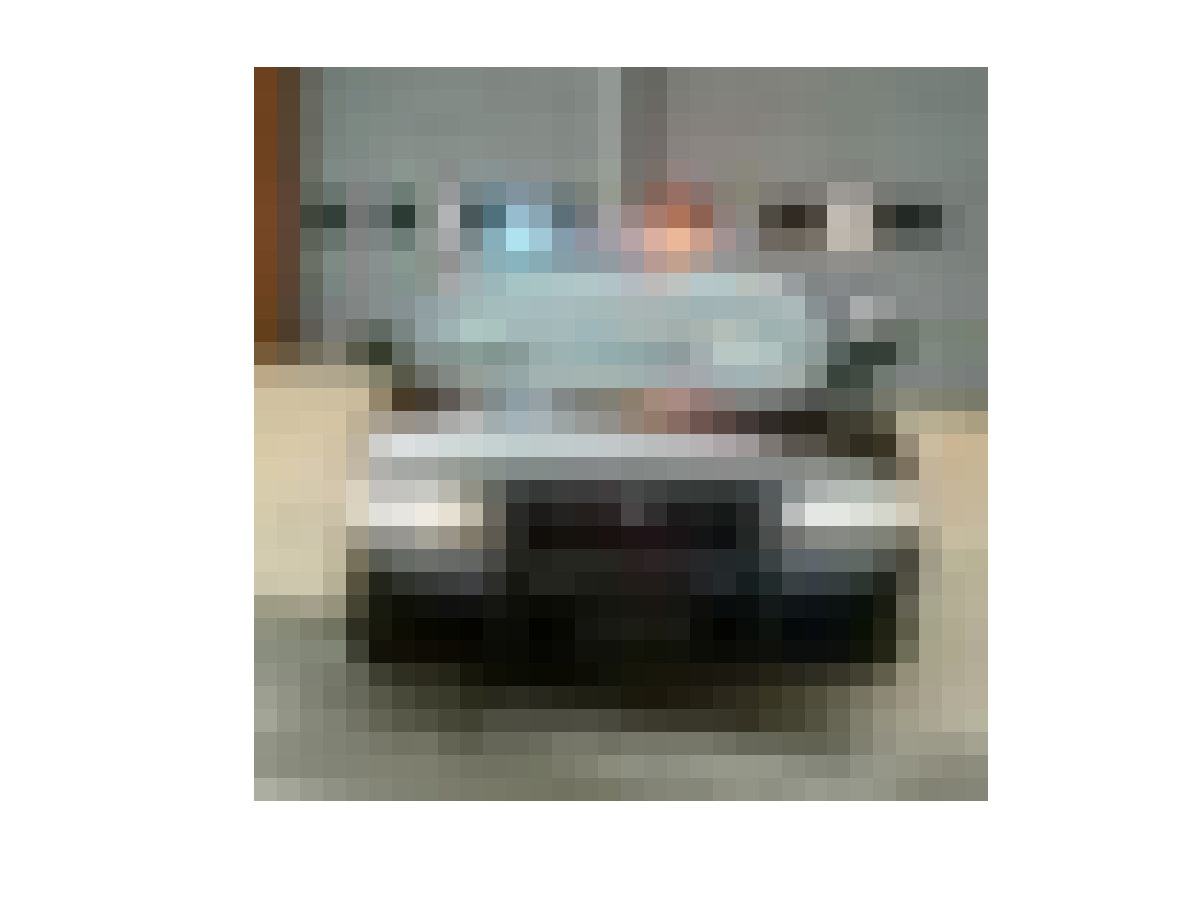}&\includegraphics[width=0.22\columnwidth, clip, trim=40mm 10mm 35mm 5mm]{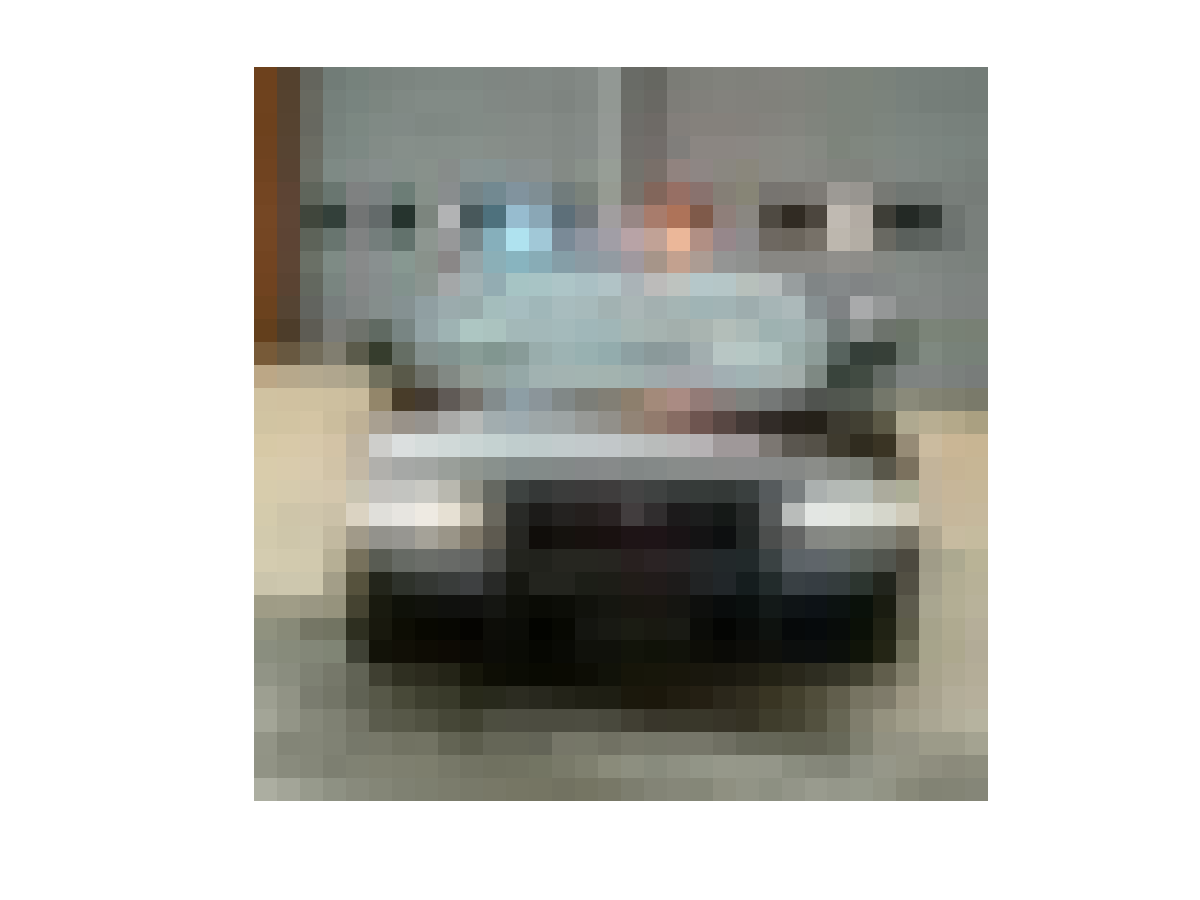}&\includegraphics[width=0.22\columnwidth, clip, trim=40mm 10mm 35mm 5mm]{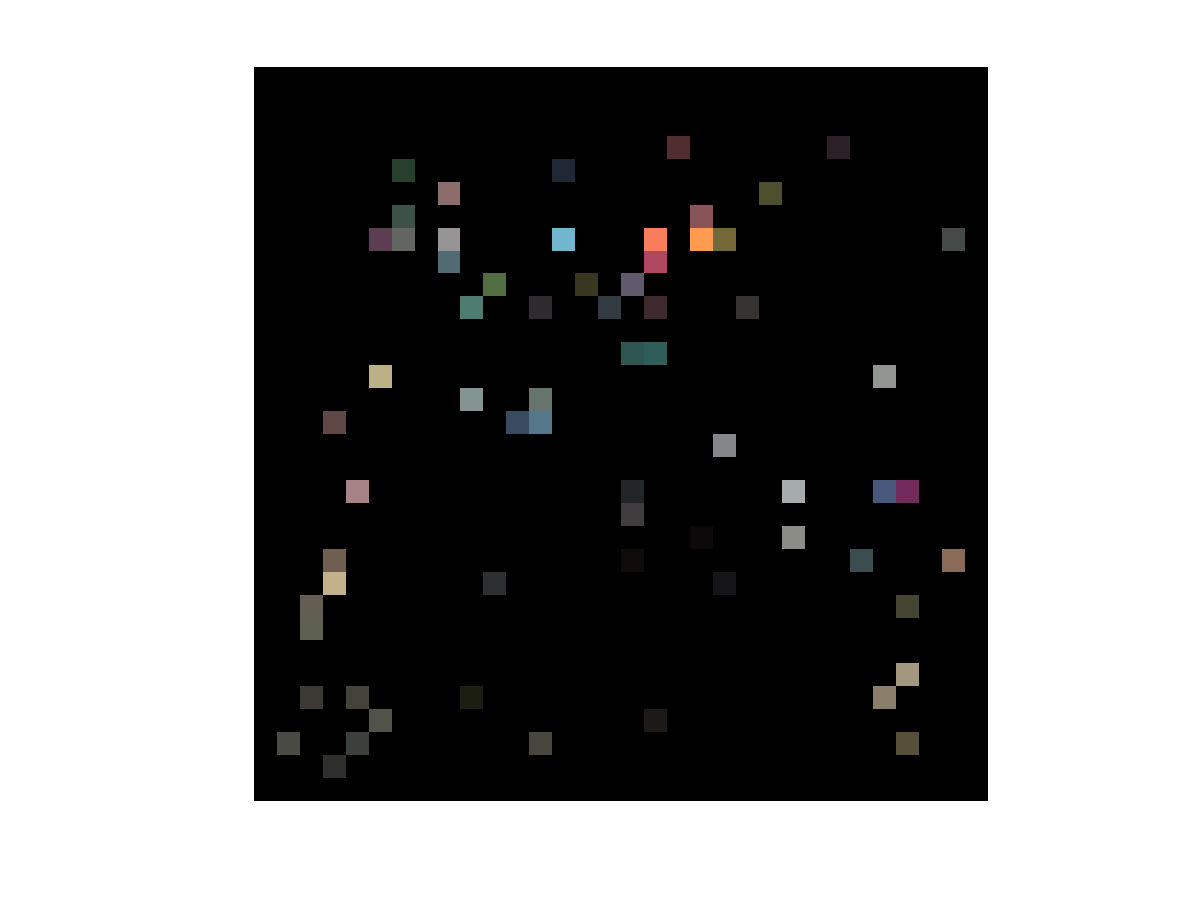}&\includegraphics[width=0.22\columnwidth, clip, trim=40mm 10mm 35mm 5mm]{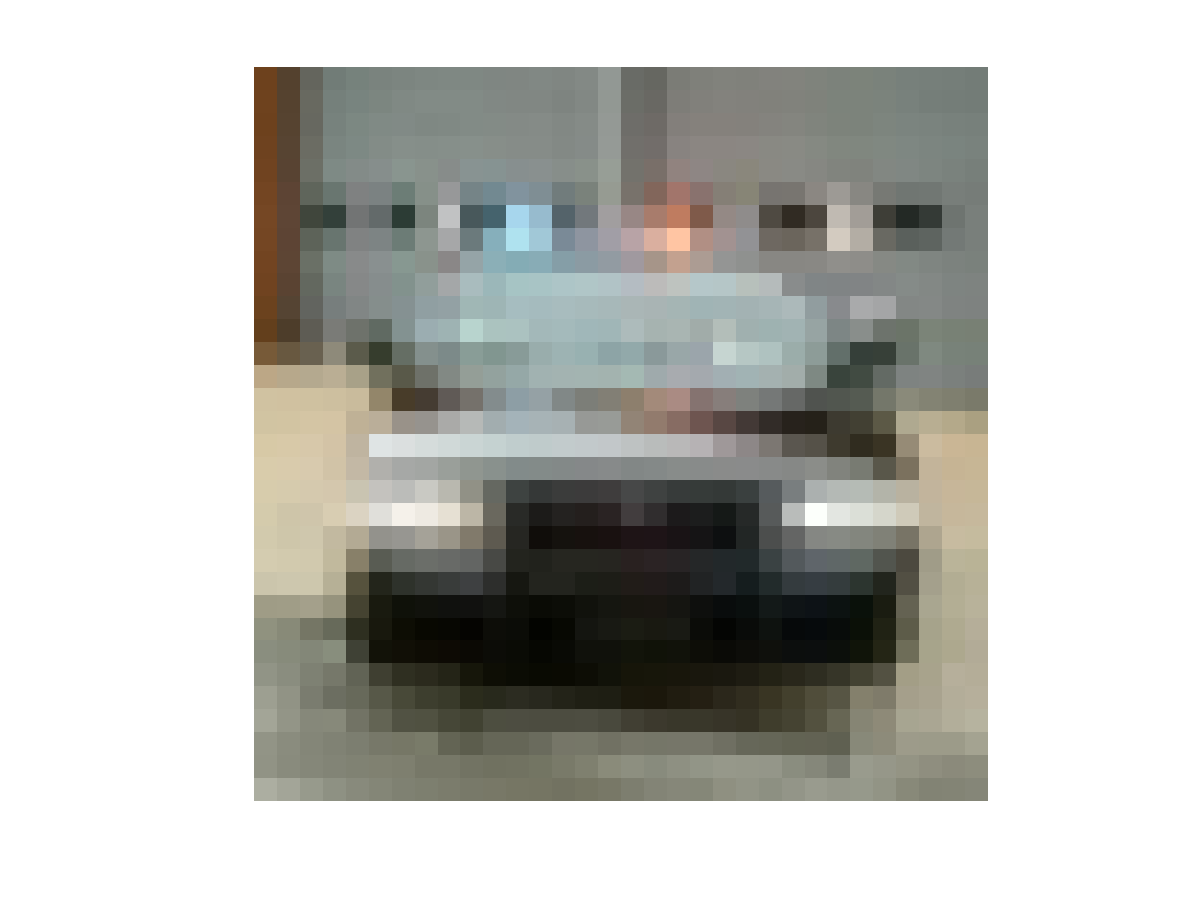}&\includegraphics[width=0.22\columnwidth, clip, trim=40mm 10mm 35mm 5mm]{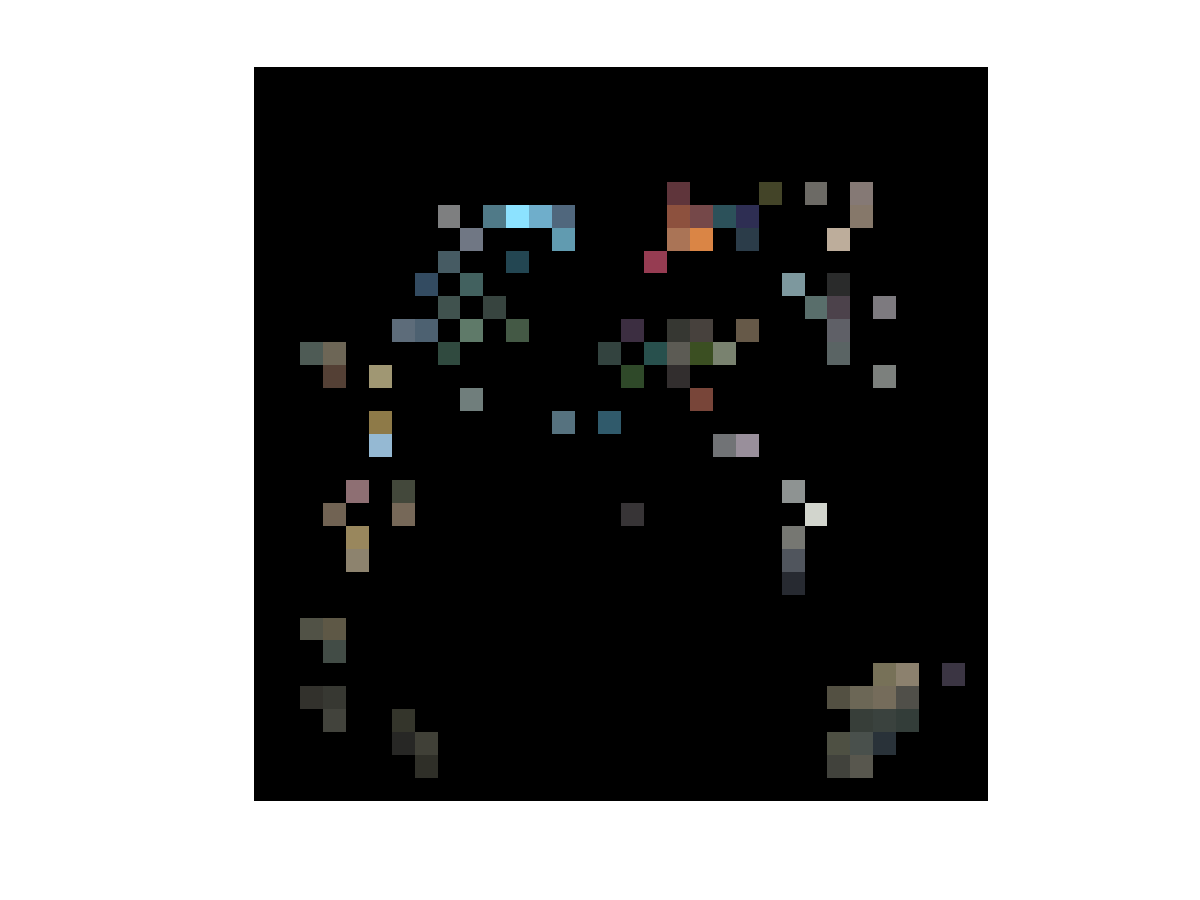}&\includegraphics[width=0.22\columnwidth, clip, trim=40mm 10mm 35mm 5mm]{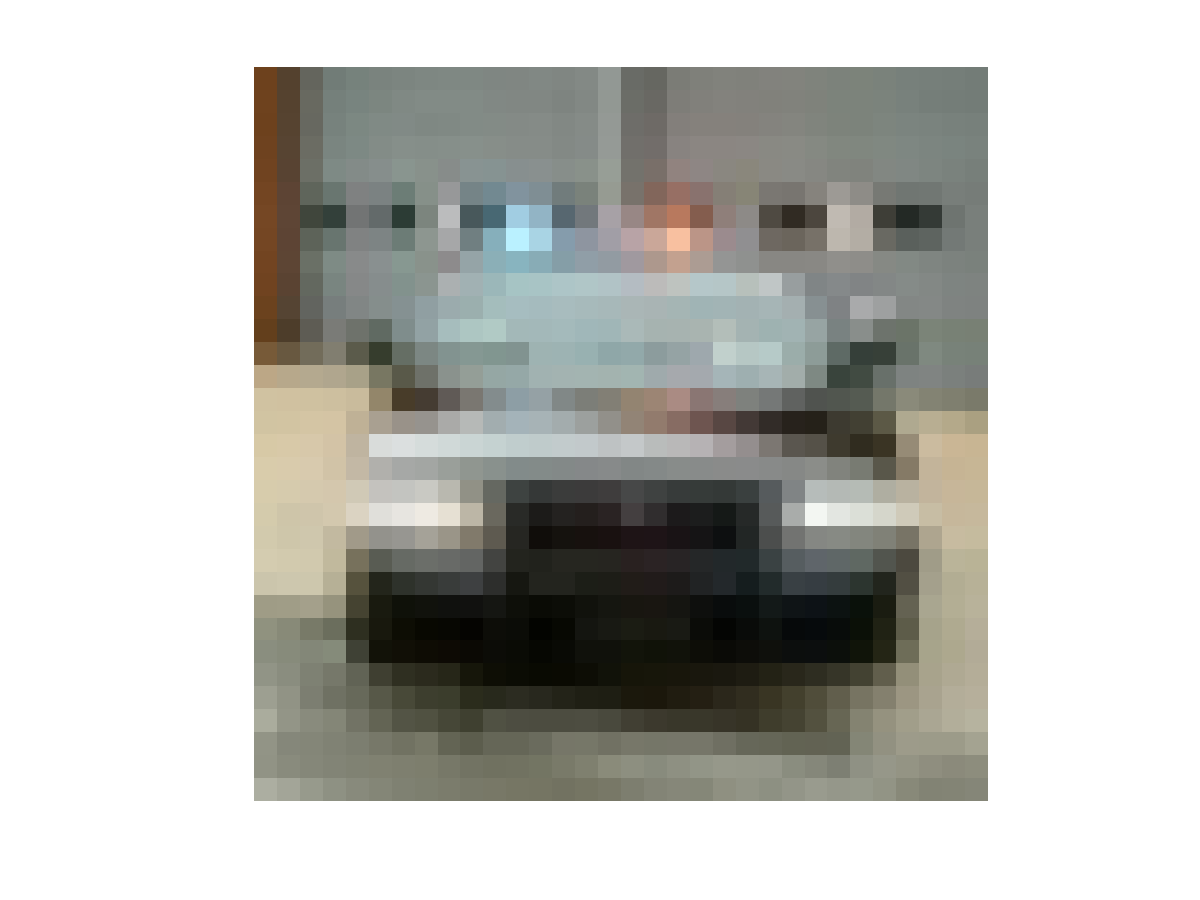}&\includegraphics[width=0.22\columnwidth, clip, trim=40mm 10mm 35mm 5mm]{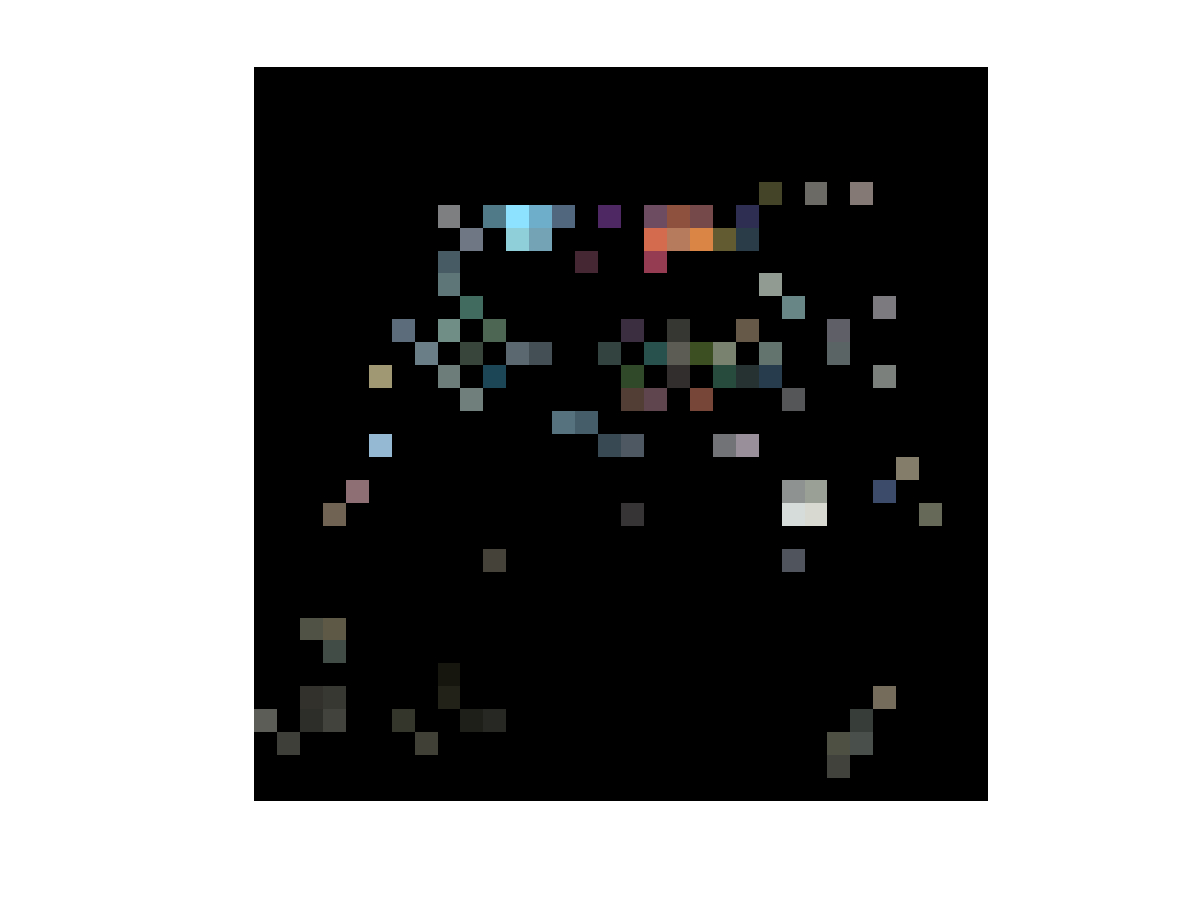}
		
\end{tabular}
	\caption{\textbf{Comparison $\sigma$-CornerSearch and $\sigma$-PGD on CIFAR-10}. We show adversarial examples generated by $\sigma$-CornerSearch with $\kappa=0.4$, $\sigma$-PGD with $\kappa=0.4$ and $\sigma$-PGD with $\kappa=0.25$, together with the respective perturbations rescaled to [0,1]. The sparsity level used is $k=100$. The gray images means the method could not find an adversarial manipulation.}\label{fig:pgd_CIFAR-10}
\end{figure*}

\section{Propagation of sparse perturbations}
\begin{figure*}
	\centering\includegraphics[scale=1.4, clip, trim= 15mm 12mm 10mm 5mm]{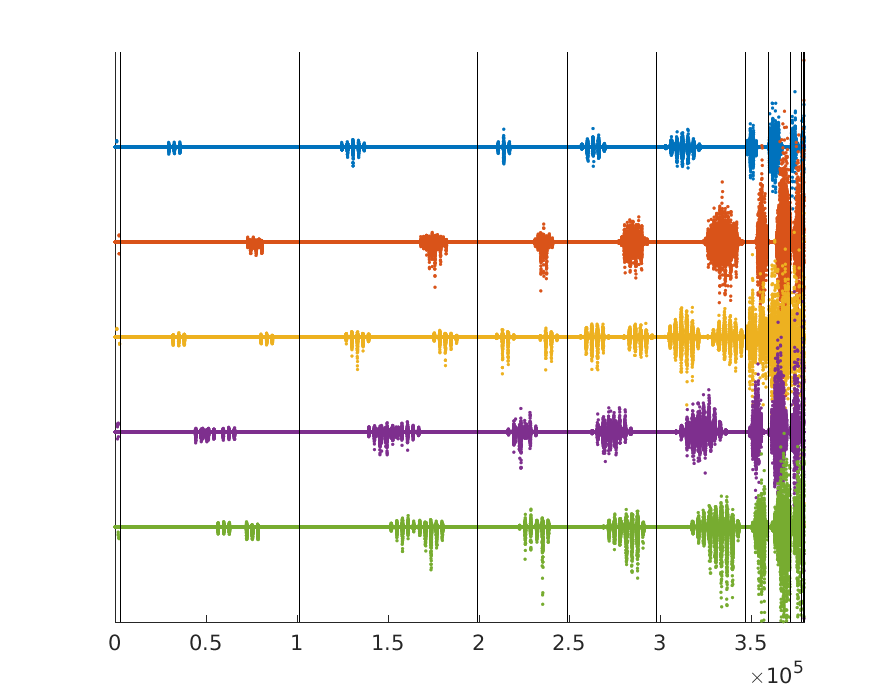}
	\caption{\textbf{Propagation of perturbations}. Difference in the values of each unit of the network obtained when propagating images of the test set and adversarial examples associated to them. The vertical segments distinguish the units of different layers, so that the input space is shown on left and the output on the right. Each color represents an image.}
	\label{fig:prop_app}
\end{figure*}
To visualize the effect of very sparse perturbations on the decision made by the classifier we can check how the output of each hidden layer is modified when an adversarial example is given as input of the network instead of the original image. We here consider the \textit{plain} model used in the comparison of the different adversarial training schemes on CIFAR-10 in Section \ref{sec:experiments} and the adversarial examples generated by CornerSearch on it.\\
We perform a forward pass first with the original images as input and then with the adversarially manipulated ones. In Figure \ref{fig:prop_app} (more examples at \url{https://github.com/fra31/sparse-imperceivable-attacks}) we plot (each color represents an image of the test set) the difference between the output values, after the activation function, of each unit of the network obtained with the two forward passes. The vertical segments separate the layers, and the leftmost section shows the difference of the inputs. The horizontal lines represent no difference in the values between the two forward passes. We can see how, going deeper into the network (towards right in Figure \ref{fig:prop_app}), the sparsity of the modifications decreases (the perturbed components of the input are on average 0.21\%, those of the last hidden layer 19.45\%) while their magnitude becomes larger, so that changing even only one pixel (that is three entries of the original image) causes a wrong classification.

\fi

\end{document}